\documentclass[plain, biber]{nowfnt}

\newif\iffinal

\finaltrue

\usepackage{comment}                    
\usepackage[T1]{fontenc}                
\usepackage{paralist}                   
\usepackage{tikz}                       
\usepackage{pgfplots}                   
\usepackage{forest}                     
\usepackage{rotating}                   
\usepackage{lscape}                  
\usepackage{longtable}                  
\usepackage{array}                      
\usepackage[binary-units=true]{siunitx} 
\usepackage{pifont}                     
\usepackage{multirow}                   
\usepackage{subcaption}                 
\usepackage{todonotes}                  
\usepackage{xspace}                     
\usepackage{nicefrac}                   
\usepackage{ifxetex}                    
\usepackage[most]{tcolorbox}           
\usepackage{xcolor}
\usepackage{tabularx}
\usepackage{rotating}
\usepackage{adjustbox}

\usepackage{float}                      
\floatstyle{plaintop}
\restylefloat{table}

\iffinal
  \usepackage{pxfonts}
  \usepackage{setspace}
  \usepackage{microtype}
\fi

\pgfplotsset{%
    discard if not/.style 2 args={
      x filter/.code={
        \edef\tempa{\thisrow{#1}}
        \edef\tempb{#2}
         \ifx\tempa\tempb
         \else
           
          \fi
        }
      }
}

\newcommand  {\st}{\textsuperscript{\textup{st}}\xspace}
\newcommand  {\nd}{\textsuperscript{\textup{nd}}\xspace}
\newcommand  {\rd}{\textsuperscript{\textup{rd}}\xspace}

\pgfplotsset{compat=1.15}

\usepgfplotslibrary{colorbrewer}
\usepgfplotslibrary{fillbetween}
\usepgfplotslibrary{groupplots}
\usepgfplotslibrary{statistics}

\usetikzlibrary{arrows}
\usetikzlibrary{external}
\usetikzlibrary{shapes.geometric}
\usetikzlibrary{positioning}
\usepgflibrary{plotmarks}
\usetikzlibrary{plotmarks}

\ifxetex
  \iffinal
  \fi
    \tikzsetexternalprefix{Figures/External/}
\fi

\newcommand{\tableyes}{\ding{51}}
\newcommand{\tableno}{\ding{55}}

\definecolor{ETH}{RGB}{192,62,28}
\definecolor{ETH}{RGB}{0,0,0}
\newcommand{\re}[1]{{\color{ETH}#1}}
\definecolor{graybox}{gray}{0.95} 

\definecolor{green1}{RGB}{102,166,30}
\definecolor{blue1}{RGB}{105,186,201}
\definecolor{yellow1}{RGB}{230,171,2}
\definecolor{brown1}{RGB}{166,118,29}
\definecolor{orange1}{RGB}{217,95,2}

\usepackage{amsmath}   
\usepackage{amssymb}   
\usepackage{amsthm}    
\usepackage{booktabs}  
\usepackage{dsfont}    
\usepackage{mathtools} 

\newcommand{\adjacency}      {\ensuremath{A}}
\newcommand{\graph}          {\ensuremath{G}}
\newcommand{\graphs}         {\ensuremath{\mathcal{G}}}
\newcommand{\edges}          {\ensuremath{E}}
\newcommand{\edge}           {\ensuremath{e}}
\newcommand{\hilbertspace}   {\ensuremath{\mathcal{H}}}
\newcommand{\innerproduct}[2]{\ensuremath{\left\langle#1,#2\right\rangle}}
\newcommand{\landau}[1]      {\ensuremath{\mathcal{O}\left(#1\right)}}
\newcommand{\vertices}       {\ensuremath{V}}
\newcommand{\vertex}         {\ensuremath{v}}
\newcommand{\Elabels}        {\ensuremath{\Sigma_{\edges}}}
\newcommand{\Vlabels}        {\ensuremath{\Sigma_{\vertices}}}
\newcommand{\graphlet}       {\ensuremath{\mathfrak{g}}}        
\newcommand{\graphlets}      {\ensuremath{\mathfrak{G}}}        
\newcommand{\paths}          {\ensuremath{\mathcal{P}}}
\newcommand{\pathlength}[1]  {\ensuremath{\left|#1\right|}}
\newcommand{\laplacian}      {\ensuremath{L}}
\newcommand{\degrees}        {\ensuremath{D}}
\newcommand{\degree}         {\mathrm{deg}}
\newcommand{\ones}           {\ensuremath{\mathds{1}}}
\newcommand{\colour}         {\ensuremath{c}}
\newcommand{\colours}        {\ensuremath{\mathcal{C}}}
\newcommand{\hash}           {\ensuremath{\mathfrak{h}}}
\newcommand{\hashfamily}     {\ensuremath{\mathfrak{H}}}

\renewcommand{\natural} {\ensuremath{\mathds{N}}}
\newcommand{\complex}   {\ensuremath{\mathds{C}}}
\newcommand{\real}      {\ensuremath{\mathds{R}}}

\DeclareMathOperator{\domain}       {dom}                      
\DeclareMathOperator{\jsd}          {JSD}                      
\DeclareMathOperator{\trace}        {tr}                       
\DeclareMathOperator{\diag}         {diag}                     
\DeclareMathOperator{\distance}     {dist}                     
\DeclareMathOperator{\gdistance}    {\distance_{\mathcal{G}}}  
\DeclareMathOperator{\fcount}       {c}                        
\DeclareMathOperator{\featurevector}{\phi}                     
\DeclareMathOperator{\flabel}       {l}                        
\DeclareMathOperator{\Vlabel}       {\flabel_{\vertices}}      
\DeclareMathOperator{\Elabel}       {\flabel_{\edges}}         
\DeclareMathOperator{\fweight}      {\mathcal{A}}              
\DeclareMathOperator{\Vweight}      {\fweight_{\vertices}}     
\DeclareMathOperator{\Eweight}      {\fweight_{\edges}}        
\DeclareMathOperator{\invariant}    {\mathcal{I}}              
\DeclareMathOperator{\relations}    {\mathcal{R}}              
\DeclareMathOperator{\match}        {\delta_{m}}               
\DeclareMathOperator{\neighbourhood}{N}                        
\DeclareMathOperator{\labelsequence}{\mathcal{L}}              
\DeclareMathOperator{\kernel}       {k}                        
\DeclareMathOperator{\basekernel}   {\kappa}                   
\DeclareMathOperator{\kernelR}      {\kernel_{\mathcal{R}}}    
\DeclareMathOperator{\kernelN}      {\kernel_{\text{node}}}    
\DeclareMathOperator{\kernelE}      {\kernel_{\text{edge}}}    
\DeclareMathOperator{\kernelEW}     {\kernel_{\text{path}}}    
\DeclareMathOperator{\kernelRW}     {\kernel_{\text{RW}}}      
\DeclareMathOperator{\kernelCP}     {\kernel_{\text{CP}}}      
\DeclareMathOperator{\kernelSP}     {\kernel_{\text{SP}}}      
\DeclareMathOperator{\kernelGL}     {\kernel_{\text{GL}}}      
\DeclareMathOperator{\kernelST}     {\kernel_{\text{ST}}}      
\DeclareMathOperator{\kernelst}     {\kernel_{\text{st}}}      
\DeclareMathOperator{\kernelNSPD}     {\kernel_{\text{NSPD}}}      
\DeclareMathOperator{\kernelkcore}     {\kernel_\text{core}}      
\DeclareMathOperator{\kernelGH}     {\kernel_{\text{GH}}}      
\DeclareMathOperator{\kernelFSL}    {\kernel_{\text{FSL}}}     
\DeclareMathOperator{\kernelMLS}    {\kernel_{\text{MLS}}}     
\DeclareMathOperator{\kernelML}     {\kernel_{\text{ML}}}      
\DeclareMathOperator{\kernelNH}     {\kernel_{\text{NH}}}      
\DeclareMathOperator{\kernelWL}     {\kernel_{\text{WL}}}      
\DeclareMathOperator{\kernelGI}     {\kernel_{\text{GI}}}      
\DeclareMathOperator{\kernelCSI}     {\kernel_{\text{CSI}}}      
\DeclareMathOperator{\kernelSM}     {\kernel_{\text{SM}}}      
\DeclareMathOperator{\kernelP}      {\kernel_{\text{Prop}}}    
\DeclareMathOperator{\kernelHC}     {\kernel_{\text{H}}}       
\DeclareMathOperator{\kernelED}     {\kernel_{\text{edit}}}    
\DeclareMathOperator{\kernelQJS}    {\kernel_{\text{QJS}}}     
\DeclareMathOperator{\kernelOA}     {\kernel_{\text{OA}}}      
\DeclareMathOperator{\kernelWD}     {\kernel_{\text{WD}}}      
\DeclareMathOperator{\kernelHGK}    {\kernel_{\text{HGK}}}     
\DeclareMathOperator{\kernelNB}     {\kernel_{\neighbourhood}} 
\DeclareMathOperator{\kernelMP}     {\kernel_{\text{MP}}}      

\newtheoremstyle{fancy-def}
                {\topsep}
                {\topsep}
                {}
                {}
                {\scshape}
                {.}
                { }
                {}

\theoremstyle{fancy-def}
\newtheorem{defn}{Definition}[chapter]

\let\originalleft\left
\let\originalright\right
\renewcommand{\left}{\mathopen{}\mathclose\bgroup\originalleft}
\renewcommand{\right}{\aftergroup\egroup\originalright}

\usepackage[%
  acronym,
  automake,
  nogroupskip,
  nopostdot,
  nonumberlist,
  toc,
  ]{glossaries}

\setglossarystyle{long}
\makeglossaries

\newcommand{\ie}{\emph{i.e.}}
\newcommand{\eg}{\emph{e.g.}}

\title{\huge Graph Kernels \\ \vspace{0.5em} \Large State-of-the-Art and Future Challenges}
\subtitle{State-of-the-Art and Future Challenges}

\maintitleauthorlist{
\\
  \textbf{Karsten Borgwardt} \\
\textbf{Elisabetta Ghisu} \\
\textbf{Felipe Llinares-L{\'o}pez} \\
\textbf{Leslie O'Bray} \\
\textbf{Bastian Rieck} \\
\\
Machine Learning and Computational Biology Lab\\
Department of Biosystems Science and Engineering\\
ETH Zurich\\
Basel, Switzerland \iffinal\\ ~ \\ \else \\ \fi
\iffinal \else \&  \fi Swiss Institute of Bioinformatics\\
}

\usepackage{mwe}

\author[1]{Borgwardt, Karsten}  
\author[1]{Ghisu, Elisabetta}  
\author[1]{Llinares-L{\'o}pez, Felipe}  
\author[1]{O'Bray, Leslie}  
\author[1]{Rieck, Bastian}  

\affil[1]{Machine Learning and Computational Biology Lab, D-BSSE, ETH Zurich\\
Swiss Institute of Bioinformatics\\
Basel, Switzerland\\
\texttt{\{karsten.borgwardt, elisabetta.ghisu, felipe.llinares, leslie.obray, bastian.rieck\}@bsse.ethz.ch}
}
\articledatabox{\nowfntstandardcitation}

\begin{document}

\setcounter{secnumdepth}{3}
\raggedbottom

\makeabstracttitle

\begin{abstract}

    Graph-structured data are an integral part of many application domains, including chemoinformatics, computational biology, neuroimaging, and social network analysis.
    Over the last two decades, numerous \emph{graph kernels}, \ie\ kernel functions between graphs, have been proposed
    to solve the problem of assessing the similarity between graphs, thereby making it possible to perform predictions in both classification and regression settings.
    This manuscript provides a review of existing graph kernels, their applications, software plus data resources, and an empirical comparison of state-of-the-art graph kernels.
\end{abstract}

\pagebreak
\tableofcontents
\pagebreak 

\chapter{Introduction}

Among the data structures commonly used in machine learning, graphs
are arguably one of the most general. Graphs allow modelling complex
objects as a collection of entities~(nodes) and of relationships between
such entities~(edges), each of which can be annotated by metadata such
as categorical or vectorial node and edge features. Many ubiquitous
data types can be understood as particular cases of graphs, including
unstructured vectorial data as well as structured data types such as
time series, images, volumetric data, point clouds or bags of entities,
to name a few. Most importantly, numerous applications benefit from the extra
flexibility that graph-based representations provide.

In \emph{chemoinformatics}, graphs have been used extensively to
represent molecular compounds~\citep{trinajstic2018chemical}, with nodes
corresponding to atoms, edges to chemical bonds, and node and edge
features encoding known chemical properties of each atom and bond in the
molecule. Machine learning approaches operating on such graph-based
representations of molecules are becoming increasingly successful in
learning to predict complex molecular properties from large annotated
data sets~\citep{duvenaud15,Gilmer17,wu2018moleculenet}, offering
a promising set of tools for drug discovery~\citep{vamathevan2019applications}.
In \emph{computational biology}, graphs have likewise risen to
prominence due to their ability to describe multi-faceted interactions
between~(biological) entities.  Examples of crucial importance include,
but are not limited to, protein-protein interaction
networks~\citep{szklarczyk2018string}, co-expression
networks~\citep{zhang2005general, lonsdale2013genotype}, metabolic
pathways~\citep{kanehisa2000kegg}, gene regulatory
networks~\citep{karlebach2008modelling}, gene-phenotype association
networks~\citep{goh2007human}, protein
structures~\citep{borgwardt2005protein} and phylogenetic
networks~\citep{huson2005application}. Graphs also play a key role in
other application domains in the life sciences, such as
\emph{neuroscience}, where they are commonly used to concisely represent
the brain connectivity patterns of different
individuals~\citep{he2010graph}, or \emph{clinical machine learning},
where they have been employed to describe and exploit relationships
between medical concepts by means of ontologies~\citep{choi2017gram} and
knowledge graphs~\citep{Rotmensch17}. In recent years,
social network analysis has become a research field on its own,
generating ever-larger graph data sets~\citep{Scott11, Wasserman94} and
spanning a wide range of applications such as viral
marketing~\citep{leskovec2007dynamics}, community
detection~\citep{du2007community}, influence
estimation~\citep{du2013scalable} or fake news
detection~\citep{tschiatschek2018fake}.

The great representational power of graph-structured data is however
 a source of important challenges for method development.
Graphs are intrinsically discrete objects, containing a combinatorial
number of substructures. As a result, even seemingly simple questions,
such as determining whether two graphs are identical~(\emph{graph
isomorphism}) or whether one graph is contained in another
graph~(\emph{subgraph isomorphism}), are remarkably hard to solve in
practice. In particular, no polynomial time algorithm is known for the
former question, while the latter question is known to be
\mbox{NP-complete}.
Machine learning methods operating on graphs must therefore grapple with
the need to balance computational tractability with the ability to
leverage as much of the information conveyed by each graph as possible. 

To this end, a popular family of early approaches, many of which were
motivated by chemoinformatics~\citep{todeschini2008handbook}, aim to
embed graphs into fixed-dimensional vectorial representations by
computing a set of hand-engineered features~(also known as \emph{topological
descriptors}). However, designing these features often proved to be
a daunting task, requiring substantial application-specific prior
knowledge and potentially depending on which statistical learning
algorithm was to be subsequently used to learn from the resulting
vectorial representations. Moreover, the amount of topological
information captured by these representations was not only limited by
the need to maintain computational tractability, but also often in
practice by the desire to obtain a parsimonious representation of
low-to-moderate dimensionality.

Crucially, the popularisation of \emph{kernel methods} in machine
learning~\citep{scholkopf2002learning} provided a principled way to
ameliorate all of the aforementioned limitations. Put briefly, kernel
methods represent objects by implicitly embedding them as elements of
a reproducing kernel Hilbert space by means of a \emph{positive definite
kernel}, which explicitly quantifies the similarity between any pair of
objects and is mathematically equivalent to the inner product between
the corresponding embeddings. This allows kernel methods to lift the
wealth of existing statistical approaches based on linear models for
vectorial data to other settings, such as non-linear modelling of
vectorial data or, as is the case for graphs, modelling of structured
data for which a vectorial representation might not be available or
might be too high-dimensional to use explicitly. Moreover, they
accomplish this while allowing to control the capacity of the underlying
model via regularisation~\citep{Hofmann08}.

%
These aspects make kernel methods a great fit for machine learning on
graph-structured data, as evidenced by the almost two decades of
fruitful research on \emph{graph kernels}\footnote{In this monograph, by
\emph{graph kernel} we refer to a kernel function between two
\emph{graphs}. Notice that the term \emph{graph kernels} sometimes is also used to refer to the different subject of kernel functions between \emph{nodes} of a single graph (\eg~\citet{kondor2002diffusion}.)} which we review in this
manuscript.
Existing graph kernels mainly differ in
\begin{inparaenum}[(i)]
  \item the type of substructures they use to define the positive definitive kernel function that measures the similarity between two
    graphs, and
  \item the underlying algorithm used to efficiently evaluate this function.  
\end{inparaenum}
In this quest to construct increasingly informative and more
computationally efficient approaches to quantify the similarity of
graphs, research on graph kernels has led to algorithms for supervised
learning \citep{Kriege2020GKsurvey}, dimensionality reduction (which can then be used to visualise graphs in
a lower dimensional space) \citep{lee2012graph}, and clustering \citep{Aggarwal2010}. 
Moreover, in doing so, the literature on graph kernels has produced
a great amount of empirical results characterising the usefulness of
different representations for graph-structured data in distinct
application domains, which we exhaustively gather, reproduce and
analyse. These experimental observations might not only pave the way to
the development of novel graph kernels, but might also be of further use
in the emerging field of graph neural networks, many of which can be
understood as natural extensions of certain graph kernels in the context
of representation learning~\citep{xu18}.

Before proceeding, we would like to mention two other recent graph kernel
surveys and highlight how our review is different. \citet{Kriege2020GKsurvey}
provide an excellent narrative overview of existing graph kernels; we
additionally provide an in-depth description of the kernels. Their review is
a good starting point for a researcher looking to understand the
landscape at a high-level or looking for a reference on which graph
kernel paper to read. \citet{nikolentzos2019graph} provide more details
about the kernels discussed; we additionally provide a conceptual
categorisation of the kernels. Unlike these two reviews, our survey
discusses trends and emerging topics in the field. Hence our review
contributes to the literature in that it provides an in-depth
description, categorisation and empirical comparison of graph kernels
and gives a detailed outlook to the future of the field.

%

This review is divided into two parts: the first part focuses on the
theoretical description of common graph kernels. After a short general
introduction to graph theory and kernels in
Chapter~\ref{chap:Background}, we provide a detailed description,
typology, and analysis of relevant graph kernels in
Chapter~\ref{chap:Kernels}. We take care to expose relations between
different kernels and briefly comment on their applicability to certain
types of data. The second part in Chapter~\ref{chap:Experiments} focuses
on a large-scale empirical evaluation of graph kernels, as well as
a description of desirable properties and requirements for benchmark
data sets. We conclude our review with an outline of future trends and
open challenges for graph kernels in Chapter~\ref{chap:Future}.
%

\chapter{Background on graph comparison and kernel methods}\label{chap:Background}

This chapter presents the required concepts and terminology from graph
theory, while also providing a brief introduction into more classical
approaches for graph comparison, such as graph isomorphism checking.

\section{A primer in graph theory}

\begin{figure}[t]

\begin{subfigure}{0.3\textwidth}
    \centering
    \includegraphics[width=0.9\linewidth]{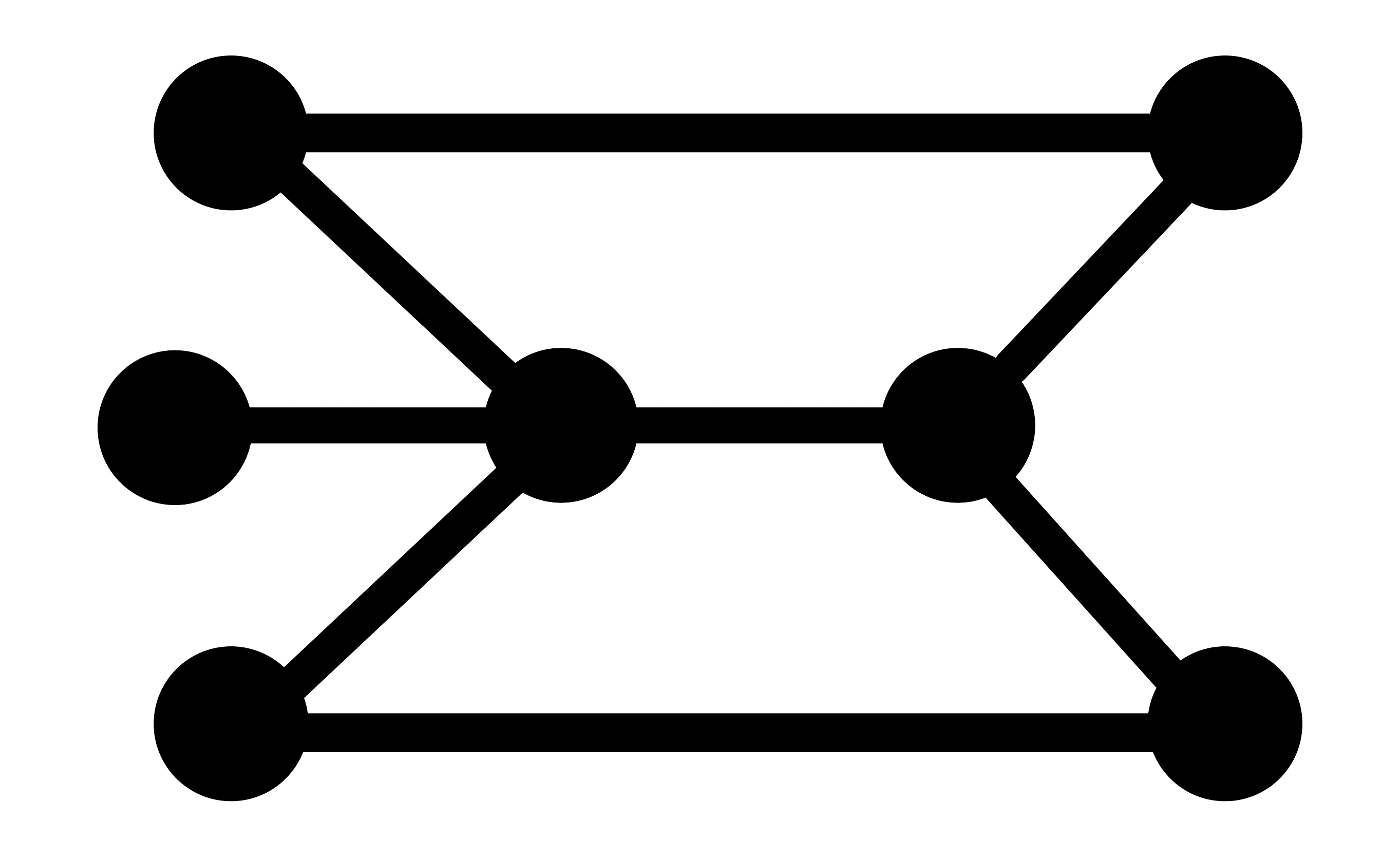} 
    \caption{An undirected graph}
    \label{fig:undirected}
\end{subfigure}
\hfill
\begin{subfigure}{0.3\textwidth}
    \centering
    \includegraphics[width=0.9\linewidth]{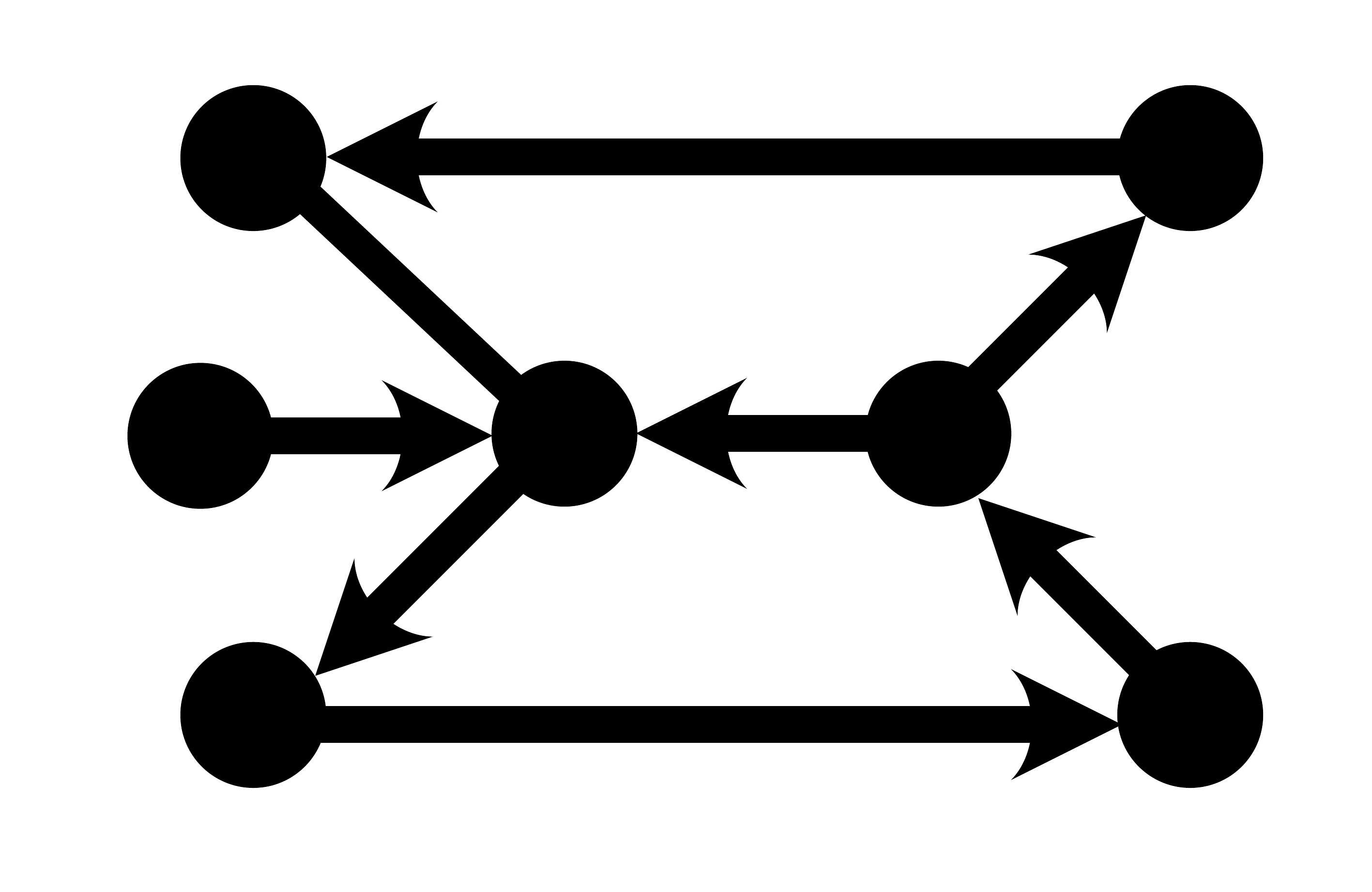}
    \caption{A directed graph}
    \label{fig:directed}
\end{subfigure}
 \hfill 
\begin{subfigure}{0.3\textwidth}
        \centering
        \includegraphics[width=0.9\linewidth]{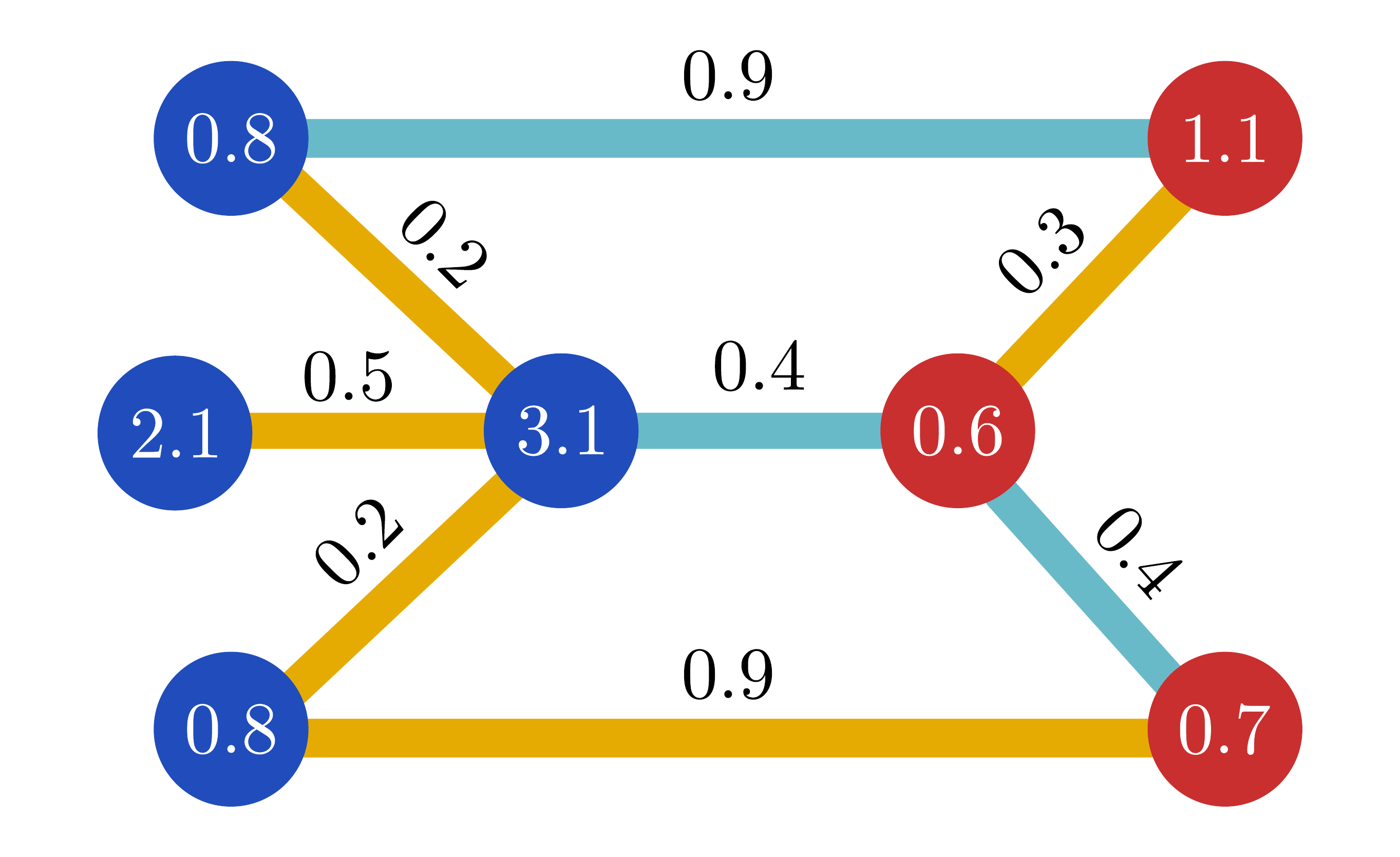}
        \caption{An attributed graph}
        \label{fig:attributed graph}
\end{subfigure}
\par\bigskip
\begin{subfigure}{0.3\textwidth}
    \centering
    \includegraphics[width=0.9\linewidth]{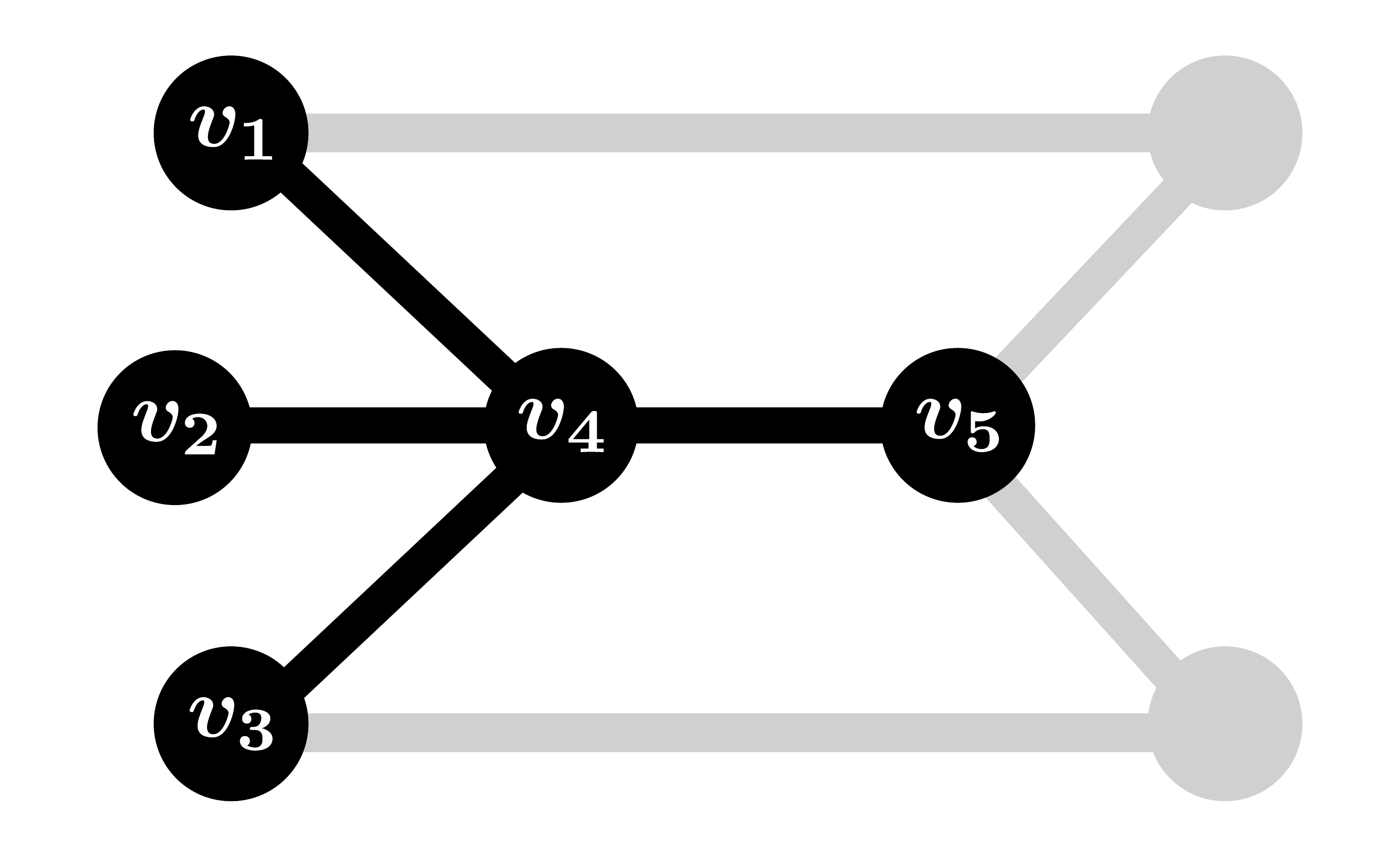} 
    \caption{$\neighbourhood^{(1)}\left(v_4\right)$}
    \label{fig:nx}
\end{subfigure}
\hfill
\begin{subfigure}{0.3\textwidth}
    \centering
    \includegraphics[width=0.9\linewidth]{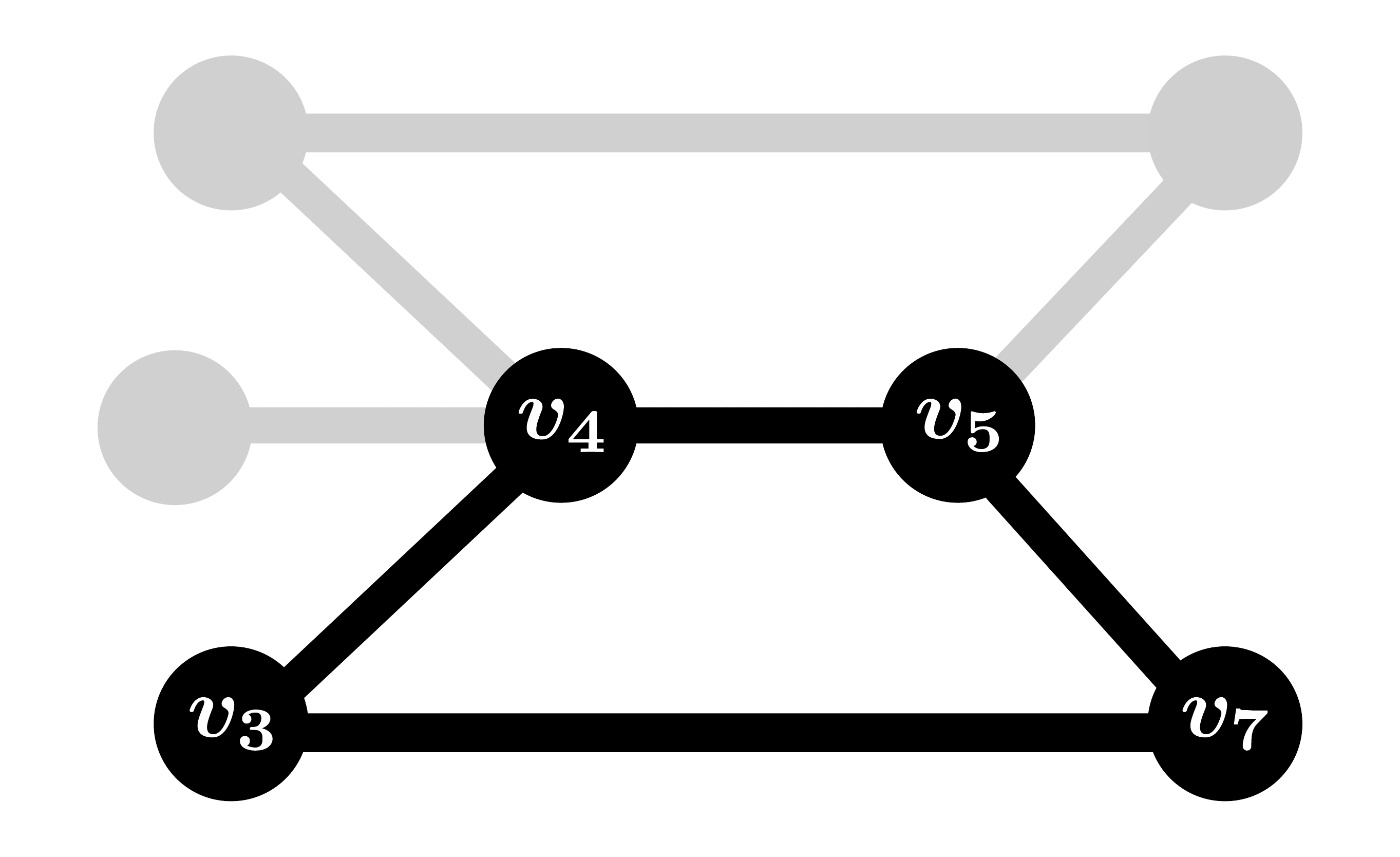}
    \caption{A cycle}
    \label{fig:cycle}
\end{subfigure}
 \hfill 
\begin{subfigure}{0.3\textwidth}
        \centering
        \includegraphics[width=0.9\linewidth]{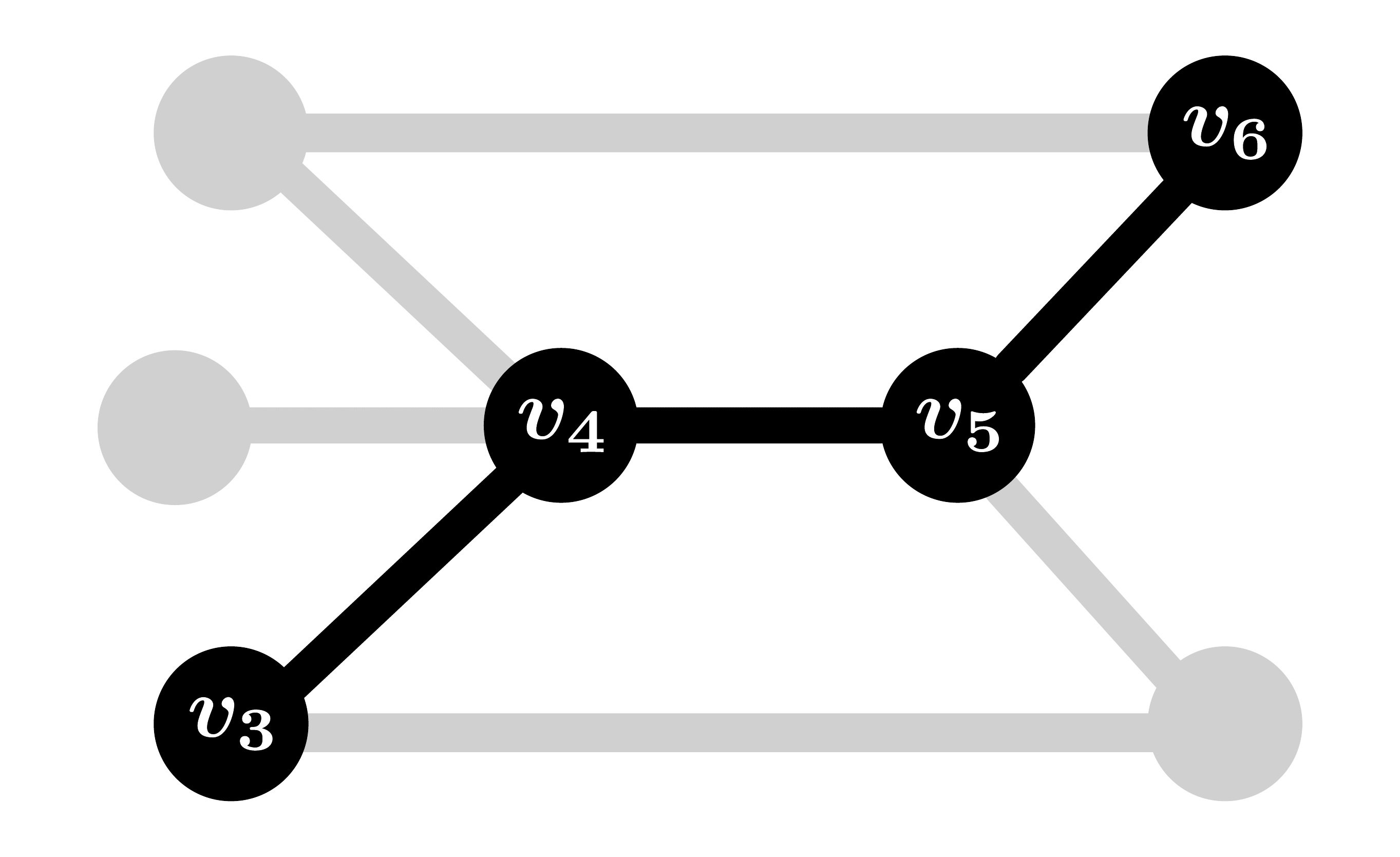}
        \caption{A shortest path}
        \label{fig:shortest_path}
\end{subfigure}

\caption{An example of an undirected graph (\subref{fig:undirected}) versus
a directed graph (\subref{fig:directed}), both of which are an example of
a \emph{fully connected} graph. (\subref{fig:attributed graph}) shows an example of an
attributed graph, where the colour of the nodes and edges represent the
node labels and edge labels respectively, and the nodes and edges also
have $1$-dimensional attributes.  We show the $1$-hop neighbourhood of
vertex $v_4$ in (\subref{fig:nx}), and also mention its degree
$\degree(v_4)=4$. In (\subref{fig:cycle}) we show a simple cycle, and in
(\subref{fig:shortest_path}) we see one of the several possible shortest paths between $v_3$ and $v_6$. Cycles and paths are examples of the more general concept of a walk.}
\label{fig:background_figure}
\end{figure}

This section presents all required concepts from graph theory. Care is
taken to define everything unambiguously so that this review is
self-contained, and we illustrate several of the following definitions
in Figure~\ref{fig:background_figure}.

\begin{defn}[Graph]
  A \emph{graph} $\graph$ is a tuple $\graph = (\vertices, \edges)$ of
  vertices $\vertices$ and edges $\edges$. For \emph{undirected} graphs,
  edges are subsets of cardinality two of the vertices, so each edge is
  of the form $(u, v)$ with $u, v \in \vertices$. For \emph{directed}
  graphs, the order of the edges in the tuple $(u, v)$ is
  relevant to indicate the direction of the edge.  If not mentioned
  otherwise, we will assume that a graph is undirected and has no
  self-loops, \ie\ edges for which $u = v$.
\end{defn}
The previous definition already gives rise to a basic graph invariant~(a
property that does not change under certain transformations such as node
renumbering), namely the \emph{degree} of a vertex.
\begin{defn}[Degree]
  The \emph{degree} of a vertex $\vertex$ of an undirected graph $\graph
  = (\vertices, \edges)$ is the number of vertices that are connected to
  $\vertex$ by means of an edge, \ie\
  $\degree\left(\vertex\right) := \left|\{ u \mid (u,
  v) \in \edges, u \neq v \}\right|$.
  For directed graphs, a vertex has an \emph{in-degree} and
  an \emph{out-degree}, depending on the direction of the edges.
  \label{def:Degree}
\end{defn}

A graph $\graph = (\vertices, \edges)$ may also contain \emph{labels} or \emph{attributes},
for example in the form of node labels, edge labels, or edge weights.
This is known as an \emph{attributed graph}.
\begin{defn}[Attributed graph]
  An \emph{attributed graph} is a graph that has either labels or
  attributes on the nodes and/or edges. When present, labels are each
  assumed to be defined over a common alphabet, $\Vlabels$ for nodes and
  $\Elabels$ for edges, with a function, $\Vlabel$ for nodes and
  $\Elabel$ for edges, to assign each entity its label. We thus have
  \begin{equation}
    \begin{split}
      \Vlabel\colon\vertices\to\,&\Vlabels\\
      \Elabel\colon\edges\to\,&\Elabels
    \end{split}
  \end{equation}
  and both of these functions need to be \emph{total}.
  A graph with additional attributes for vertices and or edges has
  attribute functions
  \begin{equation}
    \begin{split}
      \Vweight\colon\vertices\to\,&\real^{d}\\
      \Eweight\colon\edges\to\,&\real^{d}
    \end{split}
  \end{equation}
  \re{that are typically assumed} to be real-valued, \ie\ $d = 1$.
  Scalar-valued edge attributes are often also referred to as
  \emph{weights}, with the tacit assumption that the values refer
  to the strength of a specific connection.
  \label{def:Attributed graph}
\end{defn}
Generally speaking, most of the results in graph or kernel theory can be be
extended to other coefficients, such as the ring of integers, or the field of
complex numbers, but in the interest of terseness, our definitions will stay in
the field of real numbers.
In order not to clutter up the notation, we will \emph{not} mention
additional attributes in the definition of the graph~(for example by
adding more values to its tuple), but rather mention whenever we
assume or require their existence.

Since a graph defines a connectivity for each of its vertices, its edges
induce sequences for visiting them. These sequences have specific names,
depending on their properties.
\begin{defn}[Walks, paths, and cycles]
  A sequence of $k$ nodes $v_1$, \dots, $v_k$ of the vertices of a graph
  $\graph$ is called a \emph{walk} of length $k - 1$ if the edge between
  two consecutive vertices exists. More precisely, for an undirected graph,
  $(v_{i-1}, v_i) \in \edges$ needs to be satisfied for $1 < i \leq
  k$~(the case for a directed graph is analogous). Vertices of a walk
  are allowed to repeat.
  If, however, node repetition is not allowed, one typically refers to
  the node sequence as a \emph{path}, the adjective \emph{directed}
  often being added in the case of directed graphs. Special consideration
  is given to \emph{cycles}, \ie\ walks of length $k-1$ for which
  $v_1 = v_k$. If $v_1$ and $v_k$ are the only two nodes that are repeated
  in a cycle, the cycle is also called a \emph{simple cycle}.
  \label{def:Walks, paths, and cycles}
\end{defn}
Cycles are considered to be relevant descriptors of the topology of
a graph~\citep{Berger09}.
Their extraction has a high computational complexity, though,
because even detecting the presence of a single Hamiltonian cycle---a
cycle that visits all vertices exactly once, except for the start and
end vertex, which is visited twice---is known to be an NP-complete
problem~\citep{Karp72}.

The definition of walks can be extended to \emph{edge walks}~(as well as paths and cycles)
by using the neighbour relationship~(two vertices are said to be
neighbours if they are connected by an edge; we shall return to this
definition of neighbourhood shortly) between vertices: for example, an edge walk is
a sequence of edges $e_1$, \dots, $e_k$ such that exactly two vertices
of each pair $(e_{i-1}, e_i)$ coincide. The definitions for edge paths
and edge cycles are completely analogous. In an edge path, a vertex may
be visited multiple times, depending on its degree.
Certain walks are of special interest in graph theory because they
involve an added element of stochasticity.
\begin{defn}[Random walks]
  A walk in a graph is referred to as a \emph{random walk} if the next
  vertex~(or edge) is picked in a probabilistic manner. Having picked
  a start node at random, a typical choice, for example, would be to
  pick any outgoing edge of the node with uniform probability~(in case
  of unweighted graphs), or with a probability proportional to its
  weight.
\end{defn}
The notion of walks~(or, equivalently, paths and cycles), naturally
leads to a definition of connectivity in graphs.
\begin{defn}[Connected graph]
  A graph is said to be \emph{connected} if a walk between all pairs of
  nodes exists.
  Specifically, in a \emph{fully connected} graph, each pair of nodes is
  connected by an edge.
  If a graph is not connected, the set of its nodes can be partitioned
  using an equivalence relation $u \sim v$ if and only if a walk between $u$
  and $v$ exists. The equivalence classes under this relation are called
  \emph{connected components}.
\end{defn}
Likewise, paths can also be used to assess distances in a graph. This viewpoint
is often helpful when approximating high-dimensional manifolds through graphs,
and it is possible to give bounds on the dissimilarity of graph-based distances
and geodesic distances of the manifold~\citep{Bernstein00}.
\begin{defn}[Shortest paths and distances]
    Given two vertices $u$ and $v$ of a graph that are in the same connected component,
    among all the paths connecting them, there is at least one \emph{shortest path} that has the
    minimum number of vertices out of all other paths connecting the two vertices. 
    The existence of such a path is a consequence of the fact that the number of all
    paths connecting the vertices is finite.
    The distance between $u$ and $v$ is thus the number of edges of the shortest path. In case
    the graph contains edge weights, the distance is typically set to
    be the \emph{sum} of edge weights along the path.
    Although there can be multiple shortest paths between $u$ and $v$,
    the length of these shortest paths is unique.
\end{defn}
The shortest path between two vertices in a graph can be found in
polynomial time using the seminal algorithm described by
\citet{Dijkstra59}. The notion of distance in graphs is only meaningful
provided there are no negative weights along the path; in practice, this
can always be achieved by a weight transformation.

The notion of distances as defined above gives rise to the useful
concept of neighbourhoods in a graph, which extends the combinatorial
perspective~(graphs as sets and relationships) to that of a \emph{metric
space}~\citep{OSearcoid07}.
\begin{defn}[$k$-hop neighbourhood of a vertex]
  Given a vertex $\vertex$ of a graph $\graph$ and $k \in \natural_{> 0}$,
  its \emph{\mbox{$k$-hop} neighbourhood} $\neighbourhood^{(k)}\left(\vertex\right)$
  is defined as all the vertices in $\graph$ that are reachable in at
  most $k$ steps, which includes $\vertex$, assuming uniform edge weights.
  For example, $\neighbourhood^{(1)}\left(\vertex\right)$ is just the
  set of vertices that are connected to $\vertex$ by an edge and $\vertex$.
  \label{def:Neighbourhood}
\end{defn}
This definition can be connected to the idea of a ``ball'' in metric
spaces by observing that each \mbox{$k$-hop} neighbourhood of a vertex
$\vertex$ induces a subgraph of the original graph $\graph$. For
increasing values of $k$, these induced subgraphs are nested---and for
$k$ sufficiently large, the original graph $\graph$ is obtained. This
concept will play an important role later on when we define graph
kernels that operate at multiple scales, such as the Multiscale
Laplacian graph kernel in Section~\ref{sec:Multiscale Laplacian graph
kernel} on p.~\pageref{sec:Multiscale Laplacian graph kernel}.

We conclude this brief discussion of graph theory with
a description of matrices assigned to graphs.
For computational reasons, a graph is often represented through its
\emph{adjacency matrix} $\adjacency$.  A graph with $n$ vertices will thus be
represented by an $n \times n$ binary matrix whose entry $A_{ij} = 1$ if the $i$th and
$j$th vertex of the graph are connected by an edge.
The adjacency matrix is symmetrical for undirected graphs, whereas for
directed graphs $A_{ij}$ and $A_{ji}$ can be different depending on the
edge structure.
Furthermore, if edge weights are available, \ie\
$\Eweight\left(\cdot\right)$ exists and $d$ = 1, it is also possible to
derive a \emph{weighted} variant of the adjacency matrix by setting
$A_{ij}$ to the corresponding edge weight.

While the adjacency matrix $\adjacency$ can already be used to perform
random walks on graphs, another kind of matrix---the graph Laplacian---
is often employed to measure topological properties.
It is commonly defined using the adjacency matrix $\adjacency$ and the
degree matrix $\degrees$; other variants exist as well~\citep{Chung97,
Wilson08}, but they mostly differ in terms of normalisation.
%

\begin{defn}[Graph Laplacian]
  Let $\adjacency$ be the adjacency matrix of a graph $\graph = (\vertices, \edges)$.
  If weights are available, each entry $\adjacency_{ij}$ thus consists
  of the weight of the corresponding edge. Furthermore, let $\degrees$
  be the degree matrix of $\graph$. This diagonal matrix contains the
  degree of each vertex $\vertex \in \vertices$ in the unweighted case.
  For weighted graphs, each entry consists of the \emph{sum} of all edge
  weights of all edges that are incident on the corresponding vertex.
  The \emph{graph Laplacian} is then defined as
  \begin{equation}
    \laplacian := \degrees - \adjacency
  \end{equation}
  and will be a symmetric positive semi-definite matrix for undirected
  graphs. More precisely, each entry $\laplacian_{ij}$ can be described
  as
  \begin{equation}
    \laplacian_{ij} = \begin{cases}
                        -\Eweight(\edge) & \text{if an edge $\edge$ between $\vertex_i$ and $\vertex_j$ exists}\\
                        \sum_{\edge\in\neighbourhood(\vertex_i)} \Eweight(\edge) & \text{if $i = j$}\\
                        0                & \text{otherwise}
                      \end{cases}
  \end{equation}
  and in the undirected case, positive semi-definiteness is an immediate
  consequence of the~(weak) diagonal dominance.
  \label{def:Graph Laplacian}
\end{defn}

\section{Classical approaches for graph comparison}

Before diving into graph kernels, we first review earlier families of algorithms for 
graph comparison: methods based on graph isomorphism, edit distances and topological descriptors.

\subsection{The graph isomorphism problem}
\label{sec:Graph isomorphism problem}

A first family of approaches uses the most fundamental criterion for
graph comparison, namely whether an isomorphism or subgraph isomorphism
exists between the graphs. 
To formally define graph isomorphism, let $\graph = \left(\vertices,
\edges\right)$ and $\graph' = \left(\vertices', \edges'\right)$ be two
graphs.  A \emph{graph isomorphism} between $\graph$ and $\graph'$ is
a bijection between the vertex sets $\vertices$ and $\vertices'$ of
$\graph$ and $\graph'$, \ie\ $f\colon\vertices\to\vertices'$, that
preserves adjacency. Specifically, $(\vertex_i,\vertex_j) \in \edges$ if
and only if $(f(\vertex_i),f(\vertex_j)) \in \edges$ for all
$\vertex_i$, $\vertex_j \in \vertices$.
Put differently, vertices $\vertex_i$ and $\vertex_j$ are adjacent in $\graph$ if and only if $f(\vertex_i)$ and $f(\vertex_j)$ are adjacent in $\graph'$.
If an isomorphism between $\graph$ and $\graph'$ exists, the two graphs
are referred to as being \emph{isomorphic}. This gives rise to an
equivalence relation that permits partitioning a set of graphs into
different equivalence classes; graphs that belong to the same
equivalence class are indistinguishable from each other. The notion of
graph isomorphism thus gives rise to a simple similarity measure
between graphs.

In terms of computational complexity, there are currently no known
polynomial-time algorithms for solving this graph isomorphism problem,
except for some specific classes of graphs such as trees.
In fact, it is also not known to be NP-complete, making it an
interesting problem to study~\citep{Read77}. A rather recent publication~\citep{Babai15}
claims the existence of a quasi-polynomial algorithm.

In practical graph mining or graph classification, approaches based on
graph isomorphism are too restrictive, in the sense that a single
measurement error or noise in the observation of two graphs will render
them non-isomorphic. The aforementioned computational complexity also
directly results in poor scalability with the number of nodes of the
graphs. For this reason, further families of graph comparison methods
were developed and explored.

\subsection{Graph edit distances}
\label{sec:Graph edit distance}

A second of these families are so-called graph edit distances. The concept of
\emph{edit distance} refers to a general concept of how to compare two
structured objects, such as graphs. The main idea is to quantify how many
transformations are necessary to turn the first graph into the second graph.
Each transformation is measured in terms of a set of operations. For graphs,
these \emph{elementary graph operations} include
\begin{inparaenum}[(i)]
  \item vertex/edge \emph{insertions}~(subject to the creation of a new label or new attributes if the input graph is equipped with those),
  \item vertex/edge \emph{deletions}, and
  \item vertex/edge \emph{substitutions}, \ie\ the replacement of certain information, such as a label, of a given vertex/edge.
\end{inparaenum}
Each of these operations is assigned a certain \emph{cost}, and the graph
edit distance between two graphs $\graph$ and $\graph'$ is defined as the
\emph{minimum cost sum} of all operations that are required to transform
$\graph$ and $\graph'$ into each other.

The flexibility to define a cost function is both an advantage and disadvantage
of edit distances; it allows for using application-specific or domain-specific
costs, but at the same time poses the problem of careful parametrisation of
the cost function.  Once the cost function is fixed, numerous algorithms for
computing these edit distance between graphs exist; for certain classes of
graphs, such as trees, efficient polynomial-time algorithms are known, but in
general, the problem is NP-complete, as it requires to determine the maximum
common subgraph of the two given graphs (the interested reader is referred to
\citet{Riesen15} for a more extensive introduction to graph edit distance,
approximation algorithms, and applications). This rather high computational
complexity and non-trivial parametrisation of edit distances triggered
interest in yet another approach to deal with graph data, namely to map them to
a vectorial representation, as is described next.

In this manuscript, we will denote the graph edit distance between two
graphs $\graph$ and $\graph'$ as $\gdistance\left(\graph, \graph'\right)$,
with the tacit understanding that the edit distance is chosen appropriately
depending on the type of the two graphs and on their attributes.

\subsection{Invariants and topological descriptors}

A third, computationally more feasible family of approaches for graph
similarity assessment, involves computing \emph{topological descriptors}
and \emph{graph invariants}. Common to both is the idea to map a graph
to a vectorial representation: a topological descriptor is a vectorial
representation of a topological property of a graph, and a graph
invariant of a graph is a property that does not change under graph
isomorphism, such as the \emph{diameter}, \ie\ the length of the longest
shortest path, or the number of cycles.
Representing a graph by topological descriptors or graph invariants
opens the door to applying \emph{any} of the numerous machine learning
techniques for dealing with vectorial data. 

A prominent example of this family is the \emph{Wiener
Index}~\citep{Wiener47}, which is defined as the average shortest length
path in a graph.
\begin{defn}[Wiener index]
  Let $G=(V,E)$ be a graph and $\mathcal{P}$ be the set of all shortest
  paths in a graph. Then the Wiener index $W(G)$ of $G$ is defined as 
  \begin{align}
    W(G) := \frac{1}{|\mathcal{P}|} \sum_{v_i \in V} \sum_{v_j \in V} \distance(v_i,v_j),
  \end{align}
where $\distance(v_i,v_j)$ is defined as the length of the shortest path between nodes $v_i$ and $v_j$ from $G$.
\end{defn}
This index is identical for isomorphic graphs, making the Wiener
index a~(simple) graph invariant. The problem is that the
converse does not hold in general---there are graphs with identical
Wiener indices that are not isomorphic. A topological descriptor for which
the converse direction holds is called a \emph{complete graph
invariant}~\citep{Koebler08}.  However, all known complete graph invariants
require exponential runtime, as their computation is equivalent
to solving the graph isomorphism problem.

Another one of the most commonly-used descriptors is the eigenspectrum of the
graph Laplacian~(see also Definition~\ref{def:Graph Laplacian},
p.~\pageref{def:Graph Laplacian}): given the spectra of two graphs,
potentially zero-padded to ensure that they are of the same size, their
Euclidean distance can be used as a basic dissimilarity measure.
Although it is known~\citep{Wilson08} that graphs that are \emph{not}
isomorphic might still have the \emph{same} spectrum, the Laplacian
is still a useful tool in practice.
Next to interesting theoretical properties, such as stability with
respect to certain perturbations, the graph Laplacian can also be linked
to diffusion-based measures via heat kernel signatures. The reader is
referred to classical texts on spectral graph theory~\citep{Brouwer12,
Chung97} for more information.

The \emph{circular fingerprints} framework~\citep{Glem06} constitutes
a classical example of the use of topological information for specific
graphs: given graphs of molecules, where each vertex represents an atom,
it encodes information about the arrangement of other atoms~(such as
oxygen). This encoding takes into account the topological distance---the
distance defined through graph edges---to generate descriptors at
different length scales.

The central problem with topological descriptors is to find the right trade-off
between efficiency and expressivity: As stated above, all known complete graph
invariants require exponential runtime in the number of nodes of the graph,
thereby lacking efficiency. Simple topological descriptors such as the Wiener
index still lose a large amount of topological information represented by the
graph, thereby lacking expressivity. Graph kernels, the focus of this survey,
were proposed as a strategy to reach this middle ground that combines 
\emph{efficiency} and \emph{expressivity}.

\section{A brief introduction to kernel methods}\label{sec:Kernel theory}

To understand the contributions of graph kernels to the field of graph
comparison, we first have to familiarise ourselves with basic concepts
from kernel-based machine learning.

Linear models operating on inputs belonging to some vector space have
long been a staple of machine learning and statistics. Arguably
some of the most famous algorithms for both unsupervised and supervised
learning fall into this category. Examples include dimensionality
reduction methods such as principal component
analysis~\citep{pearson1901liii, hotelling1933analysis}, clustering
approaches such as \mbox{k-means}~\citep{lloyd1982least}, regression
techniques such as ridge regression~\citep{hoerl1970ridge} and
classification algorithms such as support vector
machines~\citep{boser1992training}. In a nutshell, kernel
methods provide a rich mathematical formalism to adapt this large family
of models to instead perform~(possibly non-linear) modelling of inputs
belonging to an arbitrary set $\mathcal{X}$. Intuitively, kernel-based
approaches accomplish this by embedding inputs $x \in \mathcal{X}$ as
elements of a vector space $\mathcal{H}$~(with special properties, which
we will subsequently discuss) by means of a \emph{feature map}
$\featurevector\colon\mathcal{X} \to \hilbertspace$ and applying linear models on
these transformed representations $\phi(x) \in  \hilbertspace$.
Superficially, kernel methods might resemble topological descriptors,
insofar as both of them rely on representing inputs as vectors by means
of some transformation $\phi$. However, both differ crucially in aspects
of great importance for machine learning applications.

Perhaps the most impactful distinction between both paradigms is that
while topological descriptors require specifying the mapping $\phi$
\emph{explicitly}, kernel methods typically access the \emph{feature
space}~$\hilbertspace$ only \emph{implicitly}, namely in terms of
the inner product $\langle \featurevector(x), \featurevector(x') \rangle_{\hilbertspace}$ for
any pair of inputs $x, x' \in \mathcal{X}$. A key consequence of this is
that, unlike topological descriptors, kernels can operate on a 
feature space $\hilbertspace$ of \emph{arbitrary dimensionality} without major
computational difficulties, as long as the algorithms implementing the
linear model of choice are rewritten exclusively in terms of inner
products. This observation is frequently referred to as the ``kernel
trick'' by the machine learning community~\citep{scholkopf2002learning,
Hofmann08}. Moreover, even though both designing an appropriate feature
map $\featurevector$ directly, as topological descriptors do, or indirectly, by
means of a \emph{kernel} $\kernel(x, x') := \langle \featurevector(x), \featurevector(x')
\rangle_{\hilbertspace}$, can be seen as instances of feature engineering
that require substantial domain knowledge, in many applications of
interest it is arguably more natural to use this domain knowledge to
define a notion of similarity between inputs---as captured by the
kernel---than a~(possibly high-dimensional) vectorial representation.
Kernel methods also have strong ties with statistical learning theory,
providing principled approaches to control the complexity of the
function class being used by the model. In particular, this allows
researchers to focus on designing a kernel function $\kernel(x, x')$ that
captures a meaningful notion of similarity between objects for the task
at hand, whereas other essential aspects of a learning algorithm, such
as regularisation, follow naturally and generally from the theoretical
foundations of kernel methods. These aspects make kernel methods
a particularly appealing framework to deal with structured data types,
such as graphs. 

In the remainder of this chapter, we provide a brief background on
kernel methods prior to diving into the specifics of kernel methods for
graphs in Chapter~\ref{chap:Kernels}. Interested readers can find
additional, in-depth material on the theory of kernel methods and their
applications in machine learning and statistics
in~\citet{scholkopf2002learning, shawe2004kernel, Hofmann08}.

\subsection{Fundamental concepts}

As mentioned above, kernel methods define a feature map
$\featurevector\colon\mathcal{X} \to \hilbertspace$ that represents
inputs from a set $\mathcal{X}$ as elements of a vector space
$\hilbertspace$. More precisely,  $\hilbertspace$ will be
a \emph{Hilbert space} and, as such, will be endowed with an inner product.
\begin{defn}[Real-valued Hilbert space]
  A real-valued Hilbert space $\hilbertspace$ is a vector space defined
  over $\real$, the field of real numbers, that has an \emph{inner
  product}~$\innerproduct{\cdot}{\cdot}_\hilbertspace$ and is
  \emph{complete}~(every Cauchy sequence in $\hilbertspace$ converges in
  to an element of $\hilbertspace$).
\end{defn}
The existence of an inner product for $\hilbertspace$ is instrumental in
the theory of kernel methods. In this way, the map $\featurevector$, as
well as the Hilbert space $\mathcal{H}$ in which inputs are represented,
are defined implicitly through a \emph{kernel} function $\kernel(x, x')$ that
corresponds to the inner product between the representations $\featurevector(x),
\featurevector(x')$ in $\mathcal{H}$ of any pair of elements $x, x'$ in
$\mathcal{X}$.
\begin{defn}[Kernel]
  Given a non-empty set $\mathcal{X}$, we say that a function
  $\kernel\colon\mathcal{X} \times \mathcal{X} \to \real$ is a \emph{kernel} if
  there exists a Hilbert space $\hilbertspace$ and some map
  $\featurevector\colon\mathcal{X} \to \hilbertspace$ that satisfies
  \begin{equation}
    \kernel(x, x') = \innerproduct{\featurevector\left(x\right)}{\featurevector\left(x'\right)}_\hilbertspace
  \end{equation}
  for all $x, x' \in \mathcal{X}$.
  \label{def:Kernel_gen}
\end{defn}
Hence, the kernel function $\kernel(x, x')$ plays a central role in
kernel methods, with the bulk of the research being devoted to proposing
novel kernels with favourable properties~(such as high expressivity with
low computational costs) for specific tasks. In this
regard, the field of graphs kernels is no exception. Consequently,
characterising the properties that a function $k\colon \mathcal{X}
\times \mathcal{X} \to \real$ must satisfy to be a valid kernel is of
utmost importance for theoretical and practical applications alike.

\subsection{Characterisation of kernels}

The Moore--Aronszajn theorem~\citep{Aronszajn50}, one of the seminal
results in kernel theory, fully characterises the set of functions of
the form $\kernel\colon\mathcal{X} \times \mathcal{X} \to \real$ that are kernels.
Central to this result are the notions of \emph{reproducing kernel
Hilbert space}~(RKHS) and \emph{reproducing kernel}, which we enunciate
next.
\begin{defn}[Reproducing kernel Hilbert space and reproducing kernel]
  A \emph{reproducing kernel Hilbert space} on a non-empty set
  $\mathcal{X}$ is a Hilbert space $\hilbertspace$ of \emph{functions}
  $f\colon\mathcal{X} \to \real$ with a \emph{reproducing kernel}, that
  is, a function $\kernel\colon \mathcal{X} \times \mathcal{X} \to \real$ such that
  \begin{enumerate}[(i)]
  \item $\kernel(\cdot, x) \in \hilbertspace$ for all $x \in \mathcal{X}$,
  \item $f(x) = \langle f, k(\cdot, x) \rangle_{\hilbertspace}$ for all
    $f \in \hilbertspace$ and $x \in \mathcal{X}$.
  \end{enumerate}
  As  a consequence of~(ii), we note that the reproducing kernel $k(x, x') = \langle
  k(\cdot, x), k(\cdot, x') \rangle_{\mathcal{H}}$ of a RKHS is itself
  \emph{unique} and \emph{symmetric}.
\end{defn}
Another crucial concept towards the characterisation of kernels are
\emph{symmetric positive definite} functions.
\begin{defn}[Symmetric positive definitive function]
  Let $\mathcal{X}$ be a set and $g\colon \mathcal{X} \times \mathcal{X}
  \to \real$ be a bivariate real-valued function. We say
  that $g$ is a symmetric positive definitive function if it satisfies
  the following two properties:
  \begin{enumerate}
    \item \emph{Symmetry}: for $x, x' \in \mathcal{X}$, we have
      \begin{equation}
        g(x, x') = g(x', x).
      \end{equation}
    \item \emph{Positive definiteness}: for all $\lambda_1, \dots, \lambda_k
    \in \real$ and all $x_1, \dots, x_k \in \mathcal{X}$, we have
      \begin{equation}
        \sum_{i = 1}^{k} \sum_{j = 1}^{k} \lambda_i \lambda_j g(x_i, x_j) \geq 0
      \end{equation}
  \end{enumerate}
  The second property is equivalent to saying that the matrix defined by
  the kernel function is \emph{positive definite}, \ie\ it only has
  non-negative eigenvalues.
  \label{def:Kernel}
\end{defn}%
If $\hilbertspace$ is a RKHS, its reproducing kernel $\kernel(x, x')$ \emph{is}
a kernel in the sense of Definition~\ref{def:Kernel_gen} under the map
$\featurevector\colon x \mapsto k(\cdot, x)$. Likewise, any kernel in the sense of
Definition~\ref{def:Kernel_gen} can be readily seen to be a symmetric,
positive definite function. Crucially, the Moore--Aronszajn theorem
completes the characterisation of kernels by proving that \emph{any}
symmetric positive definite function $g(x, x')$ is the reproducing
kernel $\kernel(x, x')$ of a \emph{unique} RKHS $\mathcal{H}$ and, thus, is
also a kernel as in Definition~\ref{def:Kernel_gen}. In other words, the
concepts of~(i) a kernel, (ii)~a reproducing kernel and (iii)~symmetric
positive definite functions are equivalent. From a practical
perspective, the take-away from this theoretical discussion is that
a function $\kernel\colon \mathcal{X} \times \mathcal{X} \to \real$ will be
a valid kernel if and only if it is symmetric and positive definite.

\subsection{Examples of kernels}

In this section, we will briefly mention some kernel functions defined
for Euclidean space $\mathcal{X} = \real^{n}$. Despite the fact that
$\real^{n}$ is itself a Hilbert space and, thus, inputs are already
elements of a vector space endowed with an inner product, kernel methods
still provide great practical benefits. Indeed, in this case the purpose
of the map $\featurevector\colon \real^{n} \to \hilbertspace$ implicitly defined by
the kernel is not to embed the inputs into a vector space but rather
to allow linear models to capture non-linear patterns.  

We will begin by introducing what is perhaps one of the simplest kernel functions possible, the so-called \emph{Dirac delta kernel}.
\begin{defn}[Dirac delta kernel]
  The Dirac delta kernel takes two points $x, x'$ from a set $\mathcal{X}$ and compares
  their equality, \ie\
  \begin{equation}
    \kernel_{\delta}\left(x, x'\right) :=
      \begin{cases}
        1 & \text{if $x = x'$}\\
        0 & \text{otherwise.}
      \end{cases}
  \end{equation}
  \label{def:Dirac delta}
\end{defn}
A key advantage of the Dirac delta kernel is that its simplicity makes
it generally applicable. Not only it could be applied to Euclidean
inputs, but also to categorical data, as well as structured objects such
as strings or graphs. Most importantly, as we shall see in the next
chapter, the practical importance of the Dirac delta kernel resides in
its use as a building block for more sophisticated kernels.

Next, we describe two kernel functions that are ubiquitous in the
literature, namely the \emph{polynomial kernel} and the \emph{radial
basis function~(RBF) kernel}.

\begin{defn}[Polynomial kernel]
  Given two $n$-dimensional vectors $x, x' \in \real^n$, a non-negative
  scalar $c$ and a \emph{degree} $d$, the polynomial kernel is defined
  as
  \begin{equation}
    \kernel_{\text{poly}}(x,x') = \left(\langle x, x' \rangle + c\right)^{d},
  \end{equation}
  where $\langle \cdot, \cdot\rangle$ denotes the standard inner product
  in $\real^{n}$. The polynomial kernel can be generalised, being
  applicable to inputs belonging to other inner product spaces.
  \label{def:Polynomial kernel}
\end{defn}

\begin{defn}[Radial basis function~(RBF) kernel]
  Given two $n$-dimensional vectors $x, x' \in \real^n$ and a scale
  parameter $\sigma\in\real$, the RBF kernel is defined as
  \begin{equation}
    \kernel_{\text{RBF}}(x,x') = \exp\left(- \frac{\|x - x'\|^2}{2\sigma^2}\right),
  \end{equation}
  where $\|\cdot\|$ refers to the standard Euclidean distance. The RBF
  kernel can also be defined for other inputs~(in particular, the
  metric used for the calculation can be varied); the precise definition
  will become clear from the context.
  \label{def:RBF kernel}
\end{defn}

It is also worth noting that kernels obey certain \emph{closure
properties}: for example, the sum of two kernels is another kernel, just
as the product of a kernel with a positive scalar also remains
a kernel~\citep{Vert04}, thus forming a convex cone. These and related
closure properties permit the construction of a plethora of novel
kernels from existing ones.
Finally, we conclude this chapter by discussing a general framework to
define kernels on \emph{structured} data sets such as graphs.

\subsection{$\mathcal{R}$-convolution kernels}
\label{sec:R-convolution kernels}

The $\mathcal{R}$-convolution framework was developed by
\citet{Haussler99}. It provides an algorithmic way to obtain valid
kernels for graphs based on substructure decomposition. The cornerstone
of this method is the idea of describing \emph{decompositions}.

\begin{defn}[$\mathcal{R}$-decomposition]
  Let $\graphs$ denote a family of graphs. Given a graph $\graph \in \graphs$,
  an $\mathcal{R}$-decomposition is defined as a tuple
  \begin{equation}
    \mathcal{R}\left(g_1, \dots, g_d, \graph \right),
  \end{equation}
  where $g_i \in \graphs_i$ is a ``part'' of $\graph$, such as a \emph{subgraph} or
  a subset of the vertices~(the definition of a part is purposefully
  left open in order to be as generic as possible).
  The notation is supposed to describe a \emph{relationship}, \ie\
  we can think of $\graph$ as being composed of the $g_i$. Since such
  a decomposition is not unique, it is also important to define the \emph{pre-image} or \emph{fibre}
  of the relation as
  \begin{equation}
    \mathcal{R}^{-1}\left(\graph\right) := \left\{
      \left(g_1, \dots, g_d\right)
      \mid
      \mathcal{R}\left(g_1, \dots, g_d, \graph \right)
      \right\}.
  \end{equation}
  We will use $\vec{g} := \left(g_1, \dots, g_d\right)$ to denote the tuple.
\end{defn}

If the fibre is finite, which is always the case for structural
decompositions into paths or subgraphs, as long as the graph
itself is finite, \citet{Haussler99} shows that the existence of
kernels on individual substructures, \ie\ for some $g_i$, guarantees
the existence of a kernel on~$\graphs$.

\begin{defn}[$\mathcal{R}$-convolution kernel]
  For $i \in \{1, \dots, d\}$, let $\basekernel_i$ be a base kernel on
  a subset of the parts $\graphs_i$. Then the \emph{$\mathcal{R}$-convolution kernel}
  between two graphs $\graph, \graph' \in \graphs'$ is defined as
  \begin{equation}
    \kernelR\left(\graph, \graph'\right) :=
      \sum_{\vec{g} \in \mathcal{R}^{-1}\left(\graph\right)}
      \sum_{\vec{g}' \in \mathcal{R}^{-1}\left(\graph'\right)}
      \prod_{i=1}^{d} \basekernel_i\left(g_i, g_i'\right),
      \label{eq:R-convolution kernel}
  \end{equation}
  and always constitutes a valid kernel on $\graphs$~\citep{Haussler99}.
\end{defn}

In addition to the description of the kernel in terms of relations, the
\mbox{$\mathcal{R}$-convolution} framework is also often expressed in
terms of a decomposition of a graph into sets of
substructures~$\mathcal{S}$, such as
\begin{inparaenum}[(i)]
  \item all nodes of a graph,
  \item all shortest paths of a graph.
\end{inparaenum}
In this case, Eq.~\eqref{eq:R-convolution kernel} can also be written as
\begin{equation}
    \kernelR\left(\graph, \graph'\right) :=
    \sum_{s  \in \mathcal{S}}\left(\graph\right)
    \sum_{s' \in \mathcal{S'}}\left(\graph'\right)
      \prod_{i=1}^{d} \basekernel_i\left(s, s'\right),
\end{equation}
with $\mathcal{S}$ and $\mathcal{S}'$ denoting the substructures of
$\graph$ and $\graph'$, respectively. This notation, being more
accessible, will be used throughout the subsequent chapters.

\chapter{Kernels for graph-structured data}\label{chap:Kernels}

The popularisation of kernel methods in machine learning during the
early 2000s \citep{scholkopf2002learning}, including successful applications to structured data types
such as strings or trees, has led to almost two decades of research into
designing kernels for graphs, spanning a wealth of approaches that
greatly differ in the type of substructures they consider, their
computational efficiency and the type of graphs they are applicable to.
This chapter is devoted to describing how the field has evolved during
this time, summarising the most relevant methods, discussing what
motivated their development and ultimately presenting the
state-of-the-art approaches in this domain.%

Developing practically useful kernels for graph-structured data is particularly challenging due to the fundamental nature of graphs as discrete objects with a number of substructures that grows exponentially with the size of the graph. This creates an inherent trade-off between the goals of using as much of the information contained in the graphs as possible to define the kernel and of achieving computational tractability. In particular, it is known that computing \emph{complete graph kernels}, that is, any kernel $k(G, G^{\prime})=\langle \phi(G), \phi(G^{\prime}) \rangle_{\mathcal{H}}$ such that the corresponding feature map $\phi(G)$ is injective, is at least as hard as solving the graph isomorphism problem, for which no polynomial time algorithm is known~\citep[Proposition~1]{Gaertner03}. Crucially, this result suggests that, unless a breakthrough concerning the graph isomorphism problem occurs, graph kernels which can be computed in polynomial time must forego some information such that there will always exist at least one pair of graphs, not identical to each other, which nevertheless cannot be distinguished by the kernel. However, this seemingly negative theoretical result is not at odds with the excellent empirical performance of graph kernels in both unsupervised and supervised learning applications. Statistical learning algorithms typically operate under certain regularity assumptions, such as smoothness of the target function with respect to some appropriate metric or representation for the inputs. Therefore, the use of a function class of limited capacity might actually be instrumental in being able to generalise from finite data sets rather than being a practical liability.

Much of the existing research into graph kernels can hence be understood as instances of \emph{feature engineering}, aiming to investigate which aspects of graphs are best suited to define a notion of graph similarity, quantified by the kernel function, that performs well in different statistical learning problems of interest. To this end, many graph kernels, including some of the state-of-the-art methods, have exploited the flexibility of the $\mathcal{R}$-convolution framework, exploring both the use of different types of substructures a graph can be decomposed into and of different base kernels to quantify the similarity of these substructures.

Another crucial driving force in the design of new graph kernels has been the quest for improved computational efficiency. This has motivated varied contributions ranging from the use of certain substructures that are more amenable to computation to the development of specialised algorithms to evaluate particular cases of formerly proposed graph kernels in a drastically more efficient manner. As a result of these efforts, the field has accomplished remarkable progress, with state-of-the-art approaches being orders of magnitude faster than the first ever proposed graph kernels while simultaneously performing better in many unsupervised and supervised learning tasks.

As a consequence of the emphasis on computational tractability, many
existing graph kernels make assumptions that limit the type of graphs
they can be applied to. Most often, these limitations concern the type
of attributes or labels nodes and edges are allowed to have. For
instance, some graph kernels are only applicable to graphs without
attributes while many others can handle labels but not arbitrary
continuous attributes. Overcoming these limitations by either proposing new
graph kernels applicable to graphs with continuous attributes or
providing ways to adapt previously existing graph kernels to this new
setting has been another important motivation for contributions in this
domain.

Finally, despite the great theoretical flexibility of the
$\mathcal{R}$-convolution framework, most graph kernels based on this
paradigm correspond to relatively simple particular cases. For example,
a common simplification, once again motivated by computational
considerations, is the use of Dirac kernels to compare the substructures
of choice. However, despite being convenient from a computational
viewpoint, exact matching of substructures has been known to cause
difficulties such as diagonal dominance, which could impact the
resulting generalisation performance~\citep{Green06}. Motivated by this observation,
some of the most recent developments in the field have aimed to extend
the way in which the $\mathcal{R}$-convolution framework has been
typically used to define graph kernels without compromising
computational tractability.

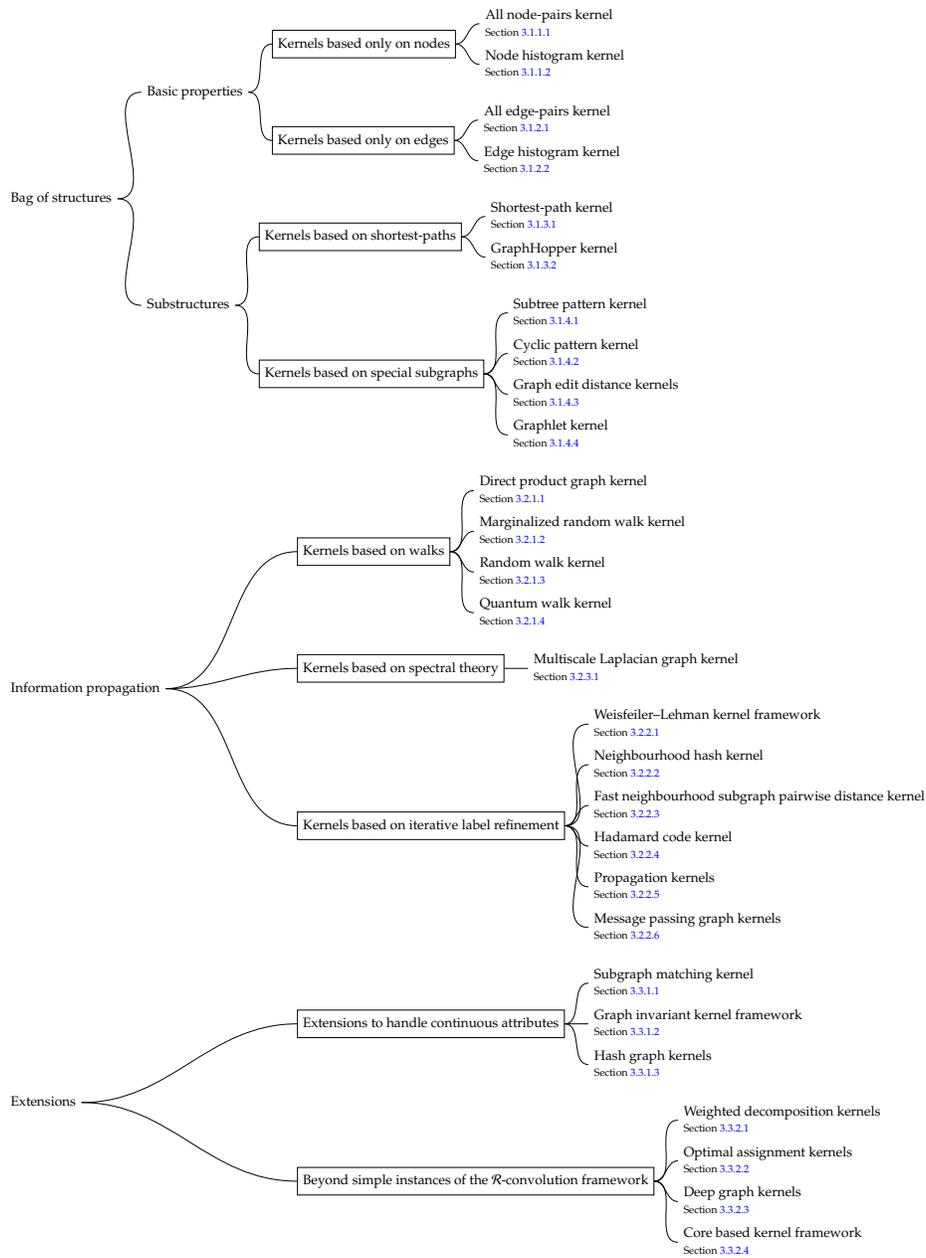
\begin{figure}
  \centering
  \scalebox{0.57}{%
    \begin{forest}
      for tree = {%
        font          = \footnotesize,
        grow'         = east,
        parent anchor = east,
        child anchor  = west,
        align         = left,
        %
        l sep+=2pt,
        edge path = {%
          \noexpand\path [draw, rounded corners=5pt, \forestoption{edge}] (!u.parent anchor) [out=0, in=180] to (.child anchor)\forestoption{edge label};
        },
        for root = {%
          ellipse,
          draw,
          parent anchor = east,
        },
        for children ={%
          /tikz/align = flush left,
          anchor      = west,
        },
      }
      [, phantom
        [Bag of structures,[Basic properties, [Kernels based only on nodes, draw, s sep-=10pt,
          [All node-pairs kernel\\\scriptsize Section~\ref{All node-pairs kernel}]
          [Node histogram kernel\\\scriptsize Section~\ref{Node histogram kernel}]
        ]
        [Kernels based only on edges, draw, s sep-=10pt,
          [All edge-pairs kernel\\\scriptsize Section~\ref{All edge-pairs kernel}]
          [Edge histogram kernel\\\scriptsize Section~\ref{Edge histogram kernel}]
        ]]
        [Substructures, [Kernels based on shortest-paths, draw, s sep-=10pt,
            [Shortest-path kernel\\\scriptsize Section~\ref{sec:Shortest-path kernel}]
            [GraphHopper kernel\\\scriptsize Section~\ref{sec:GraphHopper kernel}]
        ]
        [Kernels based on special subgraphs, draw, s sep-=10pt,
            [Subtree pattern kernel\\\scriptsize
            Section~\ref{sec:Subtree pattern kernel}]
            [Cyclic pattern kernel\\\scriptsize Section~\ref{sec:Cyclic pattern kernel}]
            [Graph edit distance kernels\\\scriptsize Section~\ref{sec:Graph edit distance kernels}]
            [Graphlet kernel\\\scriptsize Section~\ref{sec:Graphlet kernel}]
        ]]]
        [Information propagation, [Kernels based on walks, draw, s sep-=10pt, l*=6
          [Direct product graph kernel\\\scriptsize Section~\ref{sec:direct product graph kernel},]
          [Marginalized random walk kernel\\\scriptsize Section~\ref{sec:Marginalized random walk kernel}]
          [Random walk kernel\\\scriptsize Section~\ref{sec:Fast computation of walk-based kernels}]
          [Quantum walk kernel\\\scriptsize Section~\ref{sec:Quantum walk kernel}]
        ]
        [Kernels based on spectral theory, draw, s sep-=10pt,l*=6
            [Multiscale Laplacian graph kernel\\\scriptsize Section~\ref{sec:Multiscale Laplacian graph kernel}]
        ]
        [Kernels based on iterative label refinement, draw, s sep-=10pt,l*=6
          [Weisfeiler--Lehman kernel framework\\\scriptsize Section~\ref{sec:Weisfeiler--Lehman kernel}]
          [Neighbourhood hash kernel\\\scriptsize Section~\ref{sec:Neighbourhood hash kernel}]
          [Fast neighbourhood subgraph pairwise distance kernel\\\scriptsize Section~\ref{sec:Fast Neighbourhood Subgraph Pairwise Distance Kernel}]
          [Hadamard code kernel\\\scriptsize Section~\ref{sec:Hadamard code kernel}]
          [Propagation kernels\\\scriptsize Section~\ref{sec:Propagation kernels}]
          [Message passing graph kernels\\\scriptsize Section~\ref{sec:Message passing graph kernels}]
        ]]
        [Extensions,
        [Extensions to handle continuous attributes, draw, s sep-=10pt,l*=6
            [Subgraph matching kernel\\\scriptsize Section~\ref{sec:Subgraph matching kernels}]
            [Graph invariant kernel framework\\\scriptsize Section~\ref{sec:Graph invariant kernel framework}]
            [Hash graph kernels\\\scriptsize Section~\ref{sec:Hash graph kernels}]
        ]
        [Beyond simple instances of the $\mathcal{R}$-convolution framework, draw, s sep-=10pt,l*=6
          [Weighted decomposition kernels\\\scriptsize Section~\ref{sec:Weighted decomposition kernel framework}]
          [Optimal assignment kernels\\\scriptsize Section~\ref{sec:Optimal assignment kernels}]
          [Deep graph kernels\\\scriptsize Section~\ref{sec:Deep graph kernels}]
          [Core based kernel framework\\\scriptsize Section~\ref{sec:Core based kernel}]
        ]]
      ]
    \end{forest}
  }
  \caption{%
    A taxonomy of the graph kernels presented in this survey. Despite the fact that each kernel was assigned a \emph{single} category, following the structure of this chapter, some of the kernels~(in
    particular the frameworks) are highly generic and could be seen as instances of multiple categories.
    The taxonomy is not to be understood as a ``ranking'' of kernels in terms of their expressivity or any other criteria.
  }
  \label{fig:Taxonomy}
\end{figure}

We now seek to highlight a few important properties of the graph
kernels we will subsequently describe. At a high level, we group the
described graph kernels into three primary categories: 
\begin{inparaenum}[(i)]
  \item bag of structures,
  \item information propagation, and
  \item extensions of common frameworks. 
\end{inparaenum}
This hierarchy or categorisation can be seen in Figure~\ref{fig:Taxonomy}. Each of
these high-level categories can be further divided based on the type of
approach within the category. We therefore organise our descriptions
using both levels, in order to place these more
granular categories into context, to understand the relationship with
one another.
Despite kernels within a given category sharing some higher-level
principles, the specifics of what each kernel does can vary within
a category, in particular in terms of what kind of information it can
incorporate from a graph and in its computational complexity. We have
therefore added a grey box after each kernel that we describe,
summarising some of the key information about the kernel. In the
box, \emph{based on} refers to the key aspect of the graph that the
kernel uses.  \emph{Graph type} refers to the properties of the graph,
namely whether the kernel can incorporate information from undirected
graphs or directed graphs. \emph{Node type} refers to whether the kernel
can support node labels or node attributes, and \emph{edge type}
similarly details whether the kernel can incorporate edge labels
or~(continuous) edge attributes. Finally, \emph{complexity} refers to
the computational complexity of the kernel evaluation for a given pair
of graphs. We provide a sample box here to indicate how this 
information will be represented in the boxes.
\begin{figure}[h!]
    \centering
    \includegraphics[width=\textwidth]{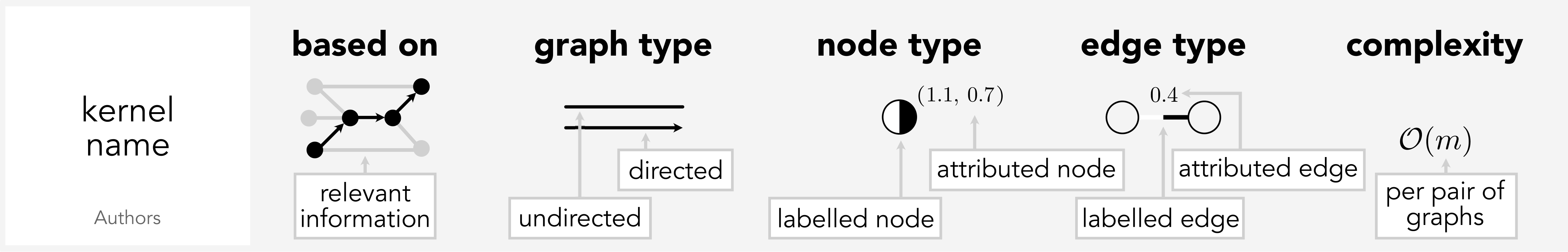}
    \label{fig:sample_gray_box}
\end{figure}

We leave a more detailed discussion on how to choose an appropriate kernel to
Section~\ref{sec:Choosing a graph kernel}, so as to incorporate also our
findings from our experiments into such a recommendation. We
note that it is also possible to choose a kernel a priori based on the
various characteristics of a kernel, whether the graph kernel can
incorporate the information in the data set, and using any relevant
domain knowledge. More specifically, the factors guiding
the choice of kernel could include:
\begin{compactenum}[(i)]
  \item Can the graph kernel handle graphs that are directed (or undirected)? 
  \item Does it include whichever node or edge labels or attributes that are present
in the graphs? 
  \item Is the graph kernel (theoretically) efficient to compute for the given data set? 
  \item Is there a particular substructure (e.g. tree patterns) that is relevant to the domain
    that would preclude the choice of a particular kernel? 
\end{compactenum}
We provide an initial reference in Figure~\ref{fig:conceptual
framework} and in Table~\ref{tab:Graph kernels} to make such a decision. However, 
we would instead recommend choosing a kernel by not
only considering what information it incorporates, but also based on its
empirical performance on benchmark data sets. We provide a more detailed
analysis of the performance of various kernels in
Chapter~\ref{chap:Experiments}, and provide the reader a more
comprehensive guide on how to choose a kernel in
Section~\ref{sec:Choosing a graph kernel}.

\begin{table}
  \resizebox{\textwidth}{!}{%
  \begin{tabular}{llccccc}
    \toprule
    Kernel                      & Graphs & N. labels & N. attr.
                                & E. labels & E. attr. & Complexity \\
    \midrule
    All node-pairs            &           & \tableyes   & \tableyes
                              &           &             & \landau{n^{2}d_v}\\
    Node histogram            &           & \tableyes   & \tableyes$^{\dagger}$   
                              &           &             & \landau{nd_v}\\
    All edge-pairs            & U, D      &             &  
                              & \tableyes & \tableyes   & \landau{m^{2}d_e}\\
    Edge histogram            & U, D      &             & 
                              & \tableyes & \tableyes$^{\dagger}$
                              & \landau{md_e}\\
    \midrule
    Shortest-path             & U, D        & \tableyes & \tableyes
                              & & 
                              & \landau{n^4d_v}\\
    GraphHopper               & U, D        & \tableyes & \tableyes
                              &  &
                              & \landau{n^4}\\
    \midrule
    Subtree pattern           & U, D & \tableyes & & & &  \landau{n^2h4^d} \\
    Cyclic pattern            & U, D & \tableyes & & \tableyes
                              & & \landau{(c+2)n + 2m}  \\
    Graph edit distance       & U, D        & \tableyes & \tableyes
                              & \tableyes & \tableyes          & \landau{n^3}\\
    Graphlet                  & U,
    D        &  & & &               & \landau{nd^{k-1}}\\
    \midrule
    Direct product graph     & U, D        & \tableyes & & \tableyes &   & \landau{n^{6}}\\
    Marginalized random walk & U, D        & \tableyes & 
                             & \tableyes &     & \landau{n^{6}}\\
    Random walk               & U, D        & \tableyes
                              &  & \tableyes
                              & \tableyes$^{\dagger}$           & \landau{n^3}\\
    Quantum walk              & U           & & & &                 & \landau{n^3}\\
    \midrule
    Weisfeiler--Lehman        & U, D        & \tableyes & & \tableyes
                              &   & \landau{hm}\\
    Neighbourhood hash        & U, D & \tableyes & & \tableyes
                              & & \landau{hm}\\
    Neighbourhood subgraph pairwise distance & U & \tableyes
                                                 & & \tableyes
                                                 & 
                                                 & \landau{nn_hm_h\log(m_h)}\\
    Hadamard code             & U, D & \tableyes & & \tableyes
                              & & \landau{hm}\\  
    Propagation framework     & U, D        & \tableyes & \tableyes
                              & \tableyes & \tableyes
                              & \landau{hm}\\
    Message passing           & U, D        & \tableyes & \tableyes
                              &  &           & \landau{n^2}\\
    \midrule
    Multiscale Laplacian      & U           & \tableyes & & & \tableyes
                              & \landau{n^2h}\\
    \midrule
    Subgraph matching         & U, D        & \tableyes & \tableyes
                              & \tableyes & \tableyes & \landau{k(n^2)^{k+1}} \\
    Graph invariant framework & U, D        &   \tableyes & \tableyes
                              & \tableyes & \tableyes          & \landau{\tau n^{2} d^{4r}}\\
    Hash graph kernels        & U, D        & \tableyes & \tableyes
                              & \tableyes & \tableyes             &\\
    \midrule
    Weighted decomposition    & U, D        & \tableyes & \tableyes
                              & \tableyes & \tableyes             & \landau{l^2}\\
    Optimal assignment        & U, D        & \tableyes & & \tableyes
                              &    & \landau{hm}\\
    Deep graph kernels        & U, D        & \tableyes & 
                              & \tableyes &              &\\
    Core based kernel framework        & U           & \tableyes & \tableyes
                              & \tableyes & \tableyes &  \\
    \bottomrule
  \end{tabular}
  }
  \caption{%
    A brief summary of graph kernel properties. $U$ and $D$ stand
      for undirected and directed graphs, respectively, with the cell
      being blank if edge information is not taken into account by the
      kernel. Node and edge labels indicate (categorical) node and edge
      labels, whereas attributes refer to (continuous) node
      attributes and edge weights. A dagger~(``$\dagger$'') implies that continuous attributes can be
      handled if the underlying kernel has an explicit feature
      representation. Note that the multiscale Laplacian requires the
      edge attributes to be $1$-dimensional.
    The complexity refers to
    the worst-case \emph{theoretical} complexity for evaluating the
    kernel between two graphs. In practice, and for certain kinds of
    graphs, some graph kernels, such as the shortest-path and
    GraphHopper kernels, can
    be evaluated much more efficiently. Missing complexity entries correspond to frameworks whose complexity would depend on the underlying base kernel.
    The table uses notation that will be reintroduced in the respective
    graph kernel description:
    $n$: number of vertices,
    $m$: number of edges,
    $d$: maximum degree,
    $d_v$: dimension of the vertex labels/attributes,
    $d_e$: dimension of the edge labels/attributes,
    $k$: size of the subgraph,
    $c$: upper bound on the number of cycles in any graph,
    $r$: diameter,
    $\tau$: maximum number of matching substructures,
    $l$: maximum number of selector-context pairs.
    }
  \label{tab:Graph kernels}
\end{table}


\begin{figure}[p]
    \centering
    \scalebox{0.9}{%
    \includegraphics[width=\textwidth]{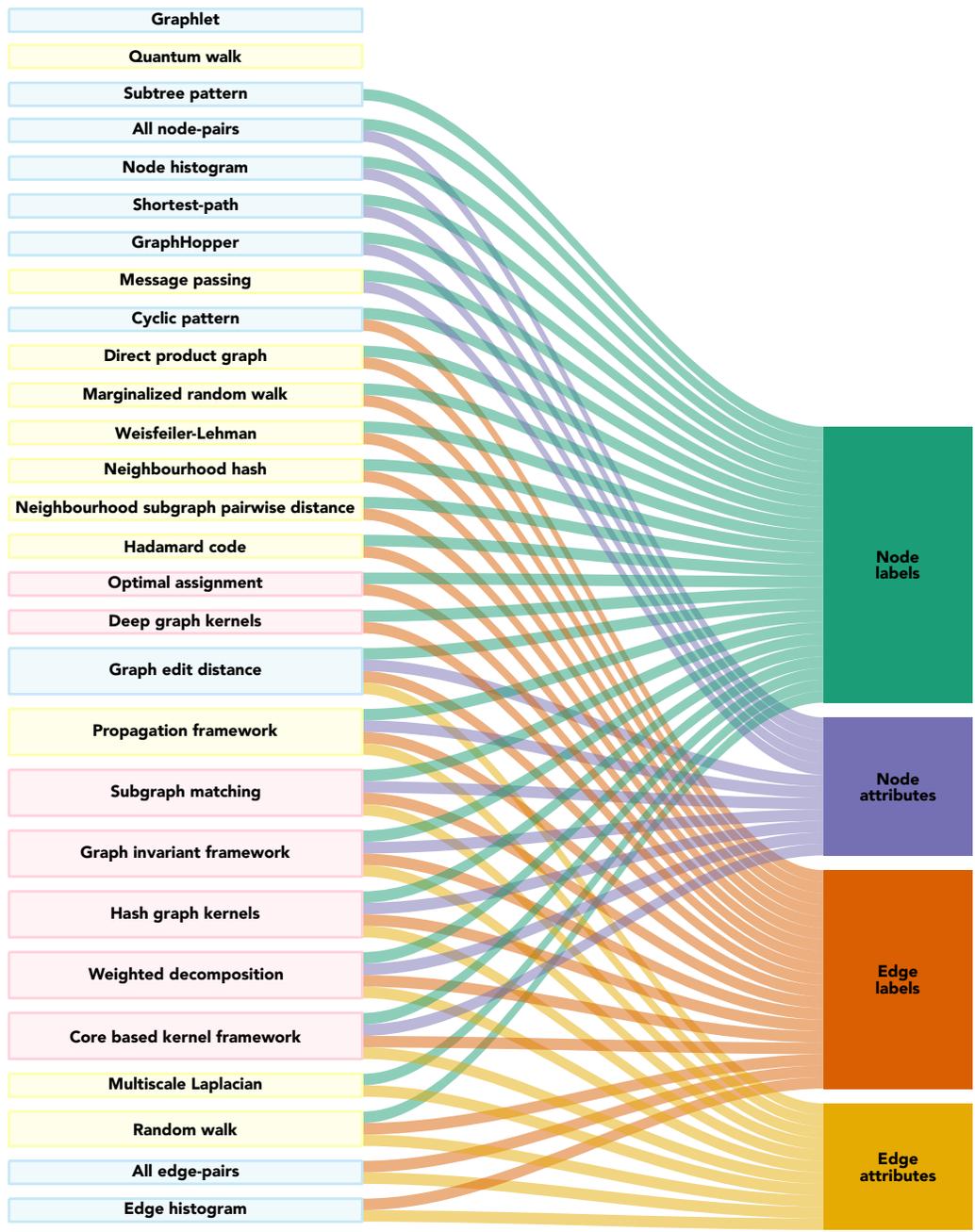}
  }
    \caption{An overview of the kernels and which node and edge
    information is used by the kernel. Labels refer to categorical
    features on the nodes or edges, whereas attributes refer to
    continuous features on the on nodes or edges. The kernels are coloured according
    to their higher level categorisation (blue: bag of structures,
    yellow: information propagation, pink: extensions), and are spaced
according to the information that is included. The graphlet kernel and
quantum walk kernel do not incorporate any node or edge labels or
attributes.}
    \label{fig:conceptual framework}
\end{figure}

The remainder of this chapter aims to present the results of almost two decades of continued progress in graph kernel research in
a concise, easy-to-follow manner. Sections have been structured to
reflect an intuitive taxonomy of the different graph kernels covered in this
review, depicted by the boxes in Figure~\ref{fig:Taxonomy}. In this way,
Sections~\ref{sec:Graph kernels based only on nodes}-\ref{sec:Graph
kernels based on spectral theory} introduce some of the most prominent
graph kernels, most of which fall under the $\mathcal{R}$-convolution
framework, categorised according to the type of substructures they are
based on. Next, Section~\ref{sec:Extending graph kernels to handle
continuous attributes} describes approaches to extend existing graph
kernels that were designed for graphs with categorical attributes to the
case where attributes might be continuous. Finally,
Section~\ref{sec:Beyond simple instances of the R-convolution framework}
discusses methods to define graph kernels that seek to alleviate some of
the limitations of simple instantiations of the
$\mathcal{R}$-convolution framework. A high-level overview of
computational complexity and supported labels and attributes for all graph kernels
under consideration is provided in Table~\ref{tab:Graph kernels} on
p.~\pageref{tab:Graph kernels}.


\section{Bag of structures}\label{sec:Bag of structures}

Many graph kernels consider the
enumeration and counting of given substructures in the graph. For
example, one can consider using basic properties about the graph, such
as the counts of node or edge labels, for use in a kernel. While these
basic graph statistics are often efficient to compute, they lack the
expressivity of more complex substructures. Another branch of this
research accordingly considers more complex structures, such as patterns
of special subgraphs or paths within a graph. The more expressive the feature, the
more computationally intense the kernel evaluation typically is, resulting in
a diverse range of time complexity within this category of kernels.

\subsection{Graph kernels based only on nodes}\label{sec:Graph kernels based only on nodes}

Graphs jointly represent a collection of entities, referred to as \emph{nodes}, as well as a set of relationships between those entities, referred to as \emph{edges}. In particular, the relational information conveyed by a graph's edges differentiates graphs from other data types, giving them great representational power, but also being responsible for most of the complexity in dealing with this type of data. As a consequence, one of the simplest ways to define a notion of graph similarity is to ignore the relational aspect of graphs altogether, effectively treating them as bags-of-nodes. Despite obvious limitations, graph kernels based exclusively on nodes are of great practical importance. Firstly, by ignoring edges, these methods provide a sensible baseline to ascertain the relative importance of graph topology for each specific task. Moreover, as we will see in Chapter~\ref{chap:Experiments}, node-only graph kernels can exhibit competitive performance in certain data sets, suggesting that modelling inputs as ``fully-fledged'' graphs might be unnecessary in some particular cases. Instead, a bag-of-nodes representation might lead to a more parsimonious and computationally efficient model for such data sets. Finally, and perhaps most importantly, graph kernels defined on nodes are used as building blocks for some of the most successful graph kernels to-date, which apply a node-only kernel to graphs that have been modified so that the attributes of each node encode information about the topology of the original graph.

\subsubsection{All node-pairs kernel}\label{All node-pairs kernel}

A fully general, node-only graph kernel can be instantiated using the $\mathcal{R}$-convolution framework by defining the $\mathcal{R}$-decomposition so that its pre-image $\mathcal{R}^{-1}(G)$ corresponds to the set of nodes of the graph $G$. This leads to the \emph{all node-pairs kernel}.

\begin{defn}[All node-pairs kernel]
  Let $G=(V, E)$ and $G^{\prime}=(V^{\prime}, E^{\prime})$ be two graphs with node attributes. The \emph{all node-pairs kernel} is defined as
  \begin{equation}
    \kernel_{\text{N}}(G, G^{\prime}) := \sum_{v \in V} \sum_{v^{\prime} \in V^{\prime}} \kernelN(v, v^{\prime})
  \end{equation}
  where $\kernelN$ stands for any p.d.~(\emph{positive definite}) kernel defined on the node attributes.
\end{defn}

The all node-pairs kernel can trivially handle both categorical and continuous node attributes by using an appropriate p.d.\ kernel $\kernelN$ between node attributes. Under the assumption that evaluating $\kernelN$ has complexity $\landau{d_v}$, the resulting graph kernel can be computed with complexity $\landau{n^{2}d_v}$ for graphs having $n$ nodes each.
\begin{figure}[h!]
    \centering
    \includegraphics[width=\textwidth]{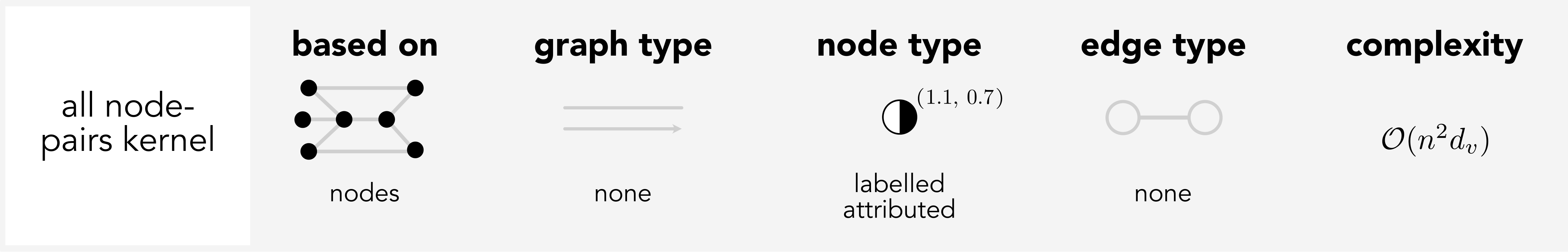}
\end{figure}

\subsubsection{Node histogram kernel}\label{Node histogram kernel}

\begin{figure}[h]
    \centering
    \includegraphics[width=0.5\linewidth]{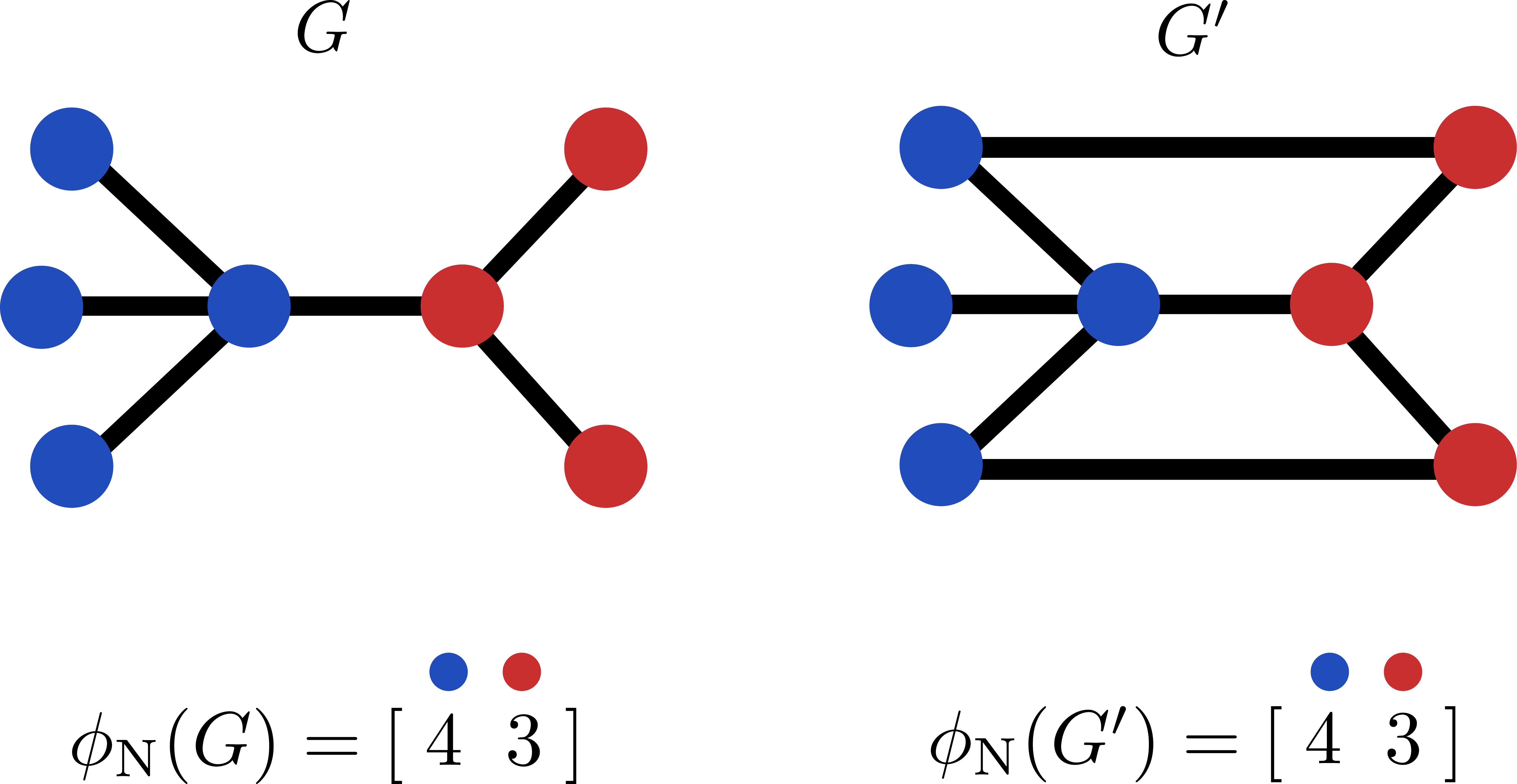}
    \caption{Given a graph $G$ and $G^{\prime}$ with node labels, the node histogram kernel can be efficiently computed using the unnormalised histogram of node labels: $\textrm{k}_\textrm{N}(G, G^{\prime}) = \langle \phi_\textrm{N}(G), \phi_\textrm{N}(G^{\prime}) \rangle_{\mathcal{H}} = 25$.}
    \label{fig:node_histogram_kernel}
\end{figure}

Denoting the feature map corresponding to the kernel on node attributes
$\kernelN$ as $\phi_{\text{node}}(\cdot)$, the all node-pairs kernel can
be expressed as $\kernel_{\text{N}}(G, G^{\prime}) = \langle
\phi_{\text{N}}(G), \phi_{\text{N}}(G^{\prime}) \rangle_{\mathcal{H}}$,
where $\phi_{\text{N}}(G) := \sum_{v \in V} \phi_{\text{node}}(v)$ can
be interpreted as the RKHS embedding of a graph $G=(V,E)$. An important
particular case arises whenever $\phi_{\text{node}}(\cdot)$ can be
computed explicitly. This occurs, for instance, when node attributes are
categorical labels over a finite alphabet $\Sigma_{\text{V}}$ and
$\kernelN$ is a Dirac kernel. Then, denoting the $i$th canonical basis
vector of $\mathbb{R}^{\vert \Sigma_{\text{V}} \vert}$ as
$\mathbf{e}_{i}$, one can define $\phi_{\text{node}}(v) :=
\mathbf{e}_{l_{\text{V}}(v)}$, where $l_{\text{V}}(v)$ stands for the
label of node $v$. The graph embedding $\phi_{\text{N}}(G)$ induced by
this kernel simply corresponds to an unnormalised histogram that counts
the occurrence of each node label in the graph.
More formally, this kernel should be referred to as a node-based kernel
with an explicit feature map, whose features are defined by label
counts.
With a slight abuse of
terminology, we will refer to this implementation of the all node-pairs
kernel as the \emph{node histogram kernel}, which is visualised in
Figure~\ref{fig:node_histogram_kernel}, even in cases where
$\phi_{\text{N}}(G)$ cannot be interpreted as a histogram. Under the
assumption that $\phi_{\text{node}}(\cdot)$ can be explicitly
represented as a $d_v$-dimensional vector, the computational complexity
of the node histogram kernel is simply $\landau{n d_{v}}$, making it one
of the most computationally efficient graph kernels for sufficiently
small values of $d_v$.
\begin{figure}[h!]
    \centering
    \includegraphics[width=\textwidth]{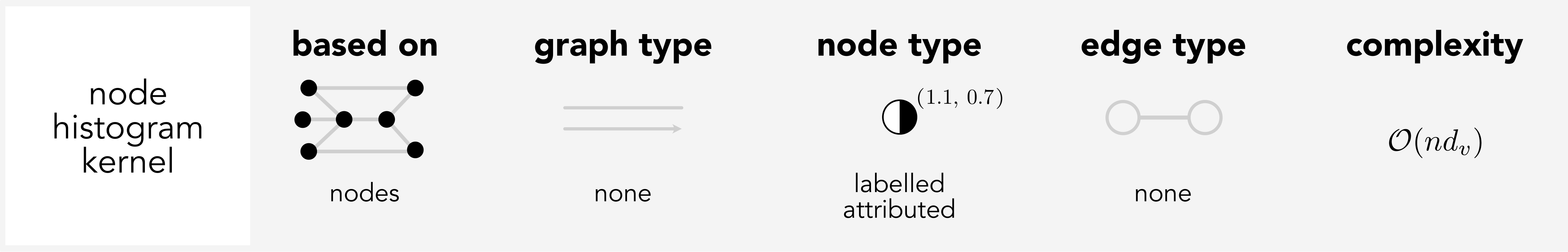}
\end{figure}

\subsection{Graph kernels based only on edges}\label{sec:Graph kernels based only on edges}

A straightforward alternative to treating graphs as
a bag-of-nodes is to model them instead as a bag-of-edges. This allows
accounting for some of the relational information contained in the
graph, though only in terms of direct relationships between entities
(nodes). Any higher-order relations defined implicitly by paths between
non-adjacent nodes are effectively ignored by this approach.
Nevertheless, much like graph kernels based only on nodes, these methods
are useful in the sense that they can serve as a basis to construct more
sophisticated graph kernels as well as provide a baseline to
characterize the relative importance of indirect relationships between
nodes for any task of interest. More generally, node-only and edge-only
graph kernels can be combined to construct a strong baseline restricted
to use only node and/or edge attributes while ignoring other aspects of
the topology of the graphs.

\subsubsection{All edge-pairs kernel}\label{All edge-pairs kernel}

Similarly to the all node-pairs kernel, defining an $\mathcal{R}$-decomposition so that its pre-image $\mathcal{R}^{-1}(G)$ corresponds to the set of edges of the graph $G$ leads to the \emph{all edge-pairs kernel}.

\begin{defn}[All edge-pairs kernel]
  Let $G=(V, E)$ and $G^{\prime}=(V^{\prime}, E^{\prime})$ be two graphs with node and/or edge attributes. The \emph{all edge-pairs kernel} is defined as
  \begin{equation}
    \kernel_{\text{E}}(G, G^{\prime}) := \sum_{e \in E} \sum_{e^{\prime} \in E^{\prime}} \kernelE(e, e^{\prime})
  \end{equation}
  where $\kernelE$ stands for any p.d.\ kernel defined on the edge attributes and/or the node attributes of the edge's endpoints.
\end{defn}

The all edge-pairs kernel can also handle both categorical and continuous attributes depending on the choice for the edge kernel $\kernelE$. If evaluating this function takes time $\landau{d_e}$, the resulting graph kernel would have complexity $\landau{m^{2}d_e}$ for graphs having $m$ edges.
\begin{figure}[h!]
    \centering
    \includegraphics[width=\textwidth]{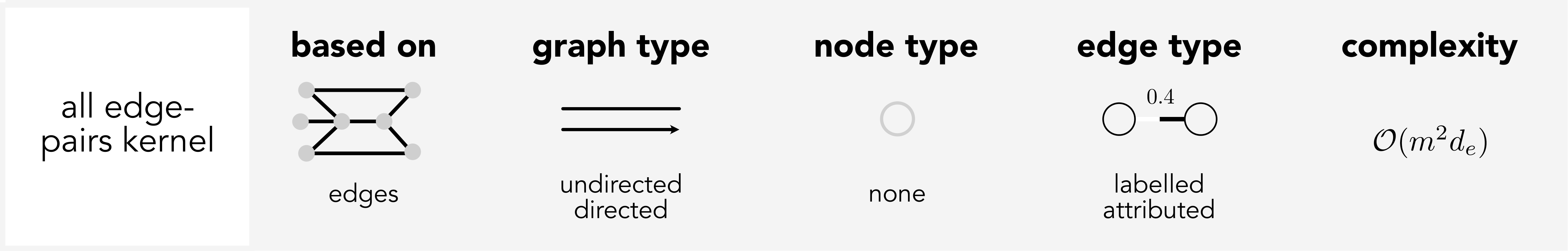}
\end{figure}
\subsubsection{Edge histogram kernel}\label{Edge histogram kernel}

\begin{figure}[h]
    \centering
    \includegraphics[width=0.5\linewidth]{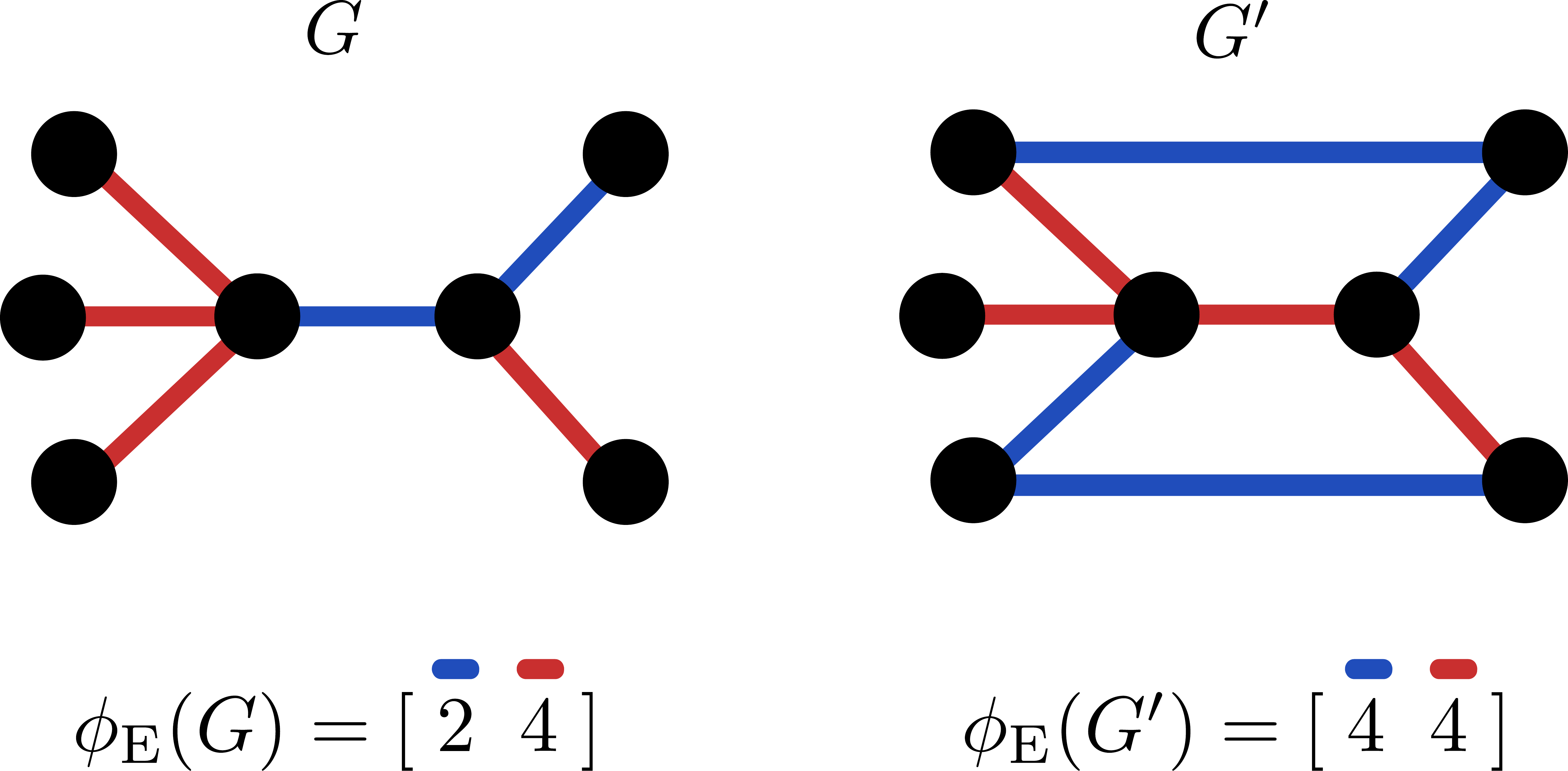}
    \caption{Given a graph $G$ and $G^{\prime}$ with edge labels, the edge histogram kernel can be efficiently computed using the unnormalised histogram of edge labels: $\textrm{k}_\textrm{E}(G, G^{\prime}) = \langle \phi_\textrm{E}(G), \phi_\textrm{E}(G^{\prime}) \rangle_{\mathcal{H}} = 24$.}
    \label{fig:edge_histogram_kernel}
\end{figure}

Analogously to the node histogram kernel, whenever the feature map
$\phi_{\text{edge}}(\cdot)$ corresponding to the edge kernel $\kernelE$
can be computed explicitly, the all edge-pairs kernel can be efficiently
calculated in terms of the induced graph embeddings $\phi_{\text{E}}(G)
:= \sum_{e \in E} \phi_{\text{edge}}(e)$, which can be interpreted as an
unnormalised histogram of edge label counts. Assuming
$\phi_{\text{edge}}(\cdot)$ admits an explicit $d_{e}$-dimensional
representation, the computational complexity of the \emph{edge histogram
kernel} is reduced to $\landau{m d_{e}}$. We present an illustration of
the edge histogram kernel in Figure~\ref{fig:edge_histogram_kernel}.
\begin{figure}[h!]
    \centering
    \includegraphics[width=\textwidth]{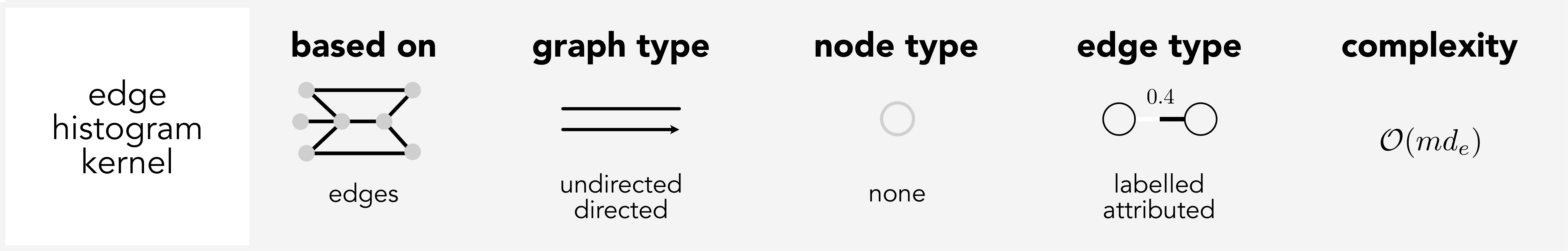}
\end{figure}

\subsection{Graph kernels based on paths}

Graph kernels built around pairwise comparisons of the node and edge
sets make limited use of the topology of graphs, failing to capture any
indirect relationships between non-adjacent nodes. Instead, representing
a graph by the \emph{paths} that are present provides a way to account
for such relations. This is the idea we will study in this subsection.
A major disadvantage, however, is that paths are less amenable to
efficient computation than comparing edges or nodes. For example, one
could propose a kernel based on comparing each path of graph $\graph$ to
each path of graph $\graph'$. However, computing this \emph{all
path-pairs kernel} has been proven to be NP-hard~\citep[Lemma
2]{Borgwardt05}. To circumvent this limitation, popular graph kernels
based on paths focus instead on shortest paths since, like walks, these
can be obtained in polynomial time.

\subsubsection{Shortest-path kernel}\label{sec:Shortest-path kernel}

\begin{figure}[h]
    \centering
    \includegraphics[width=0.75\linewidth]{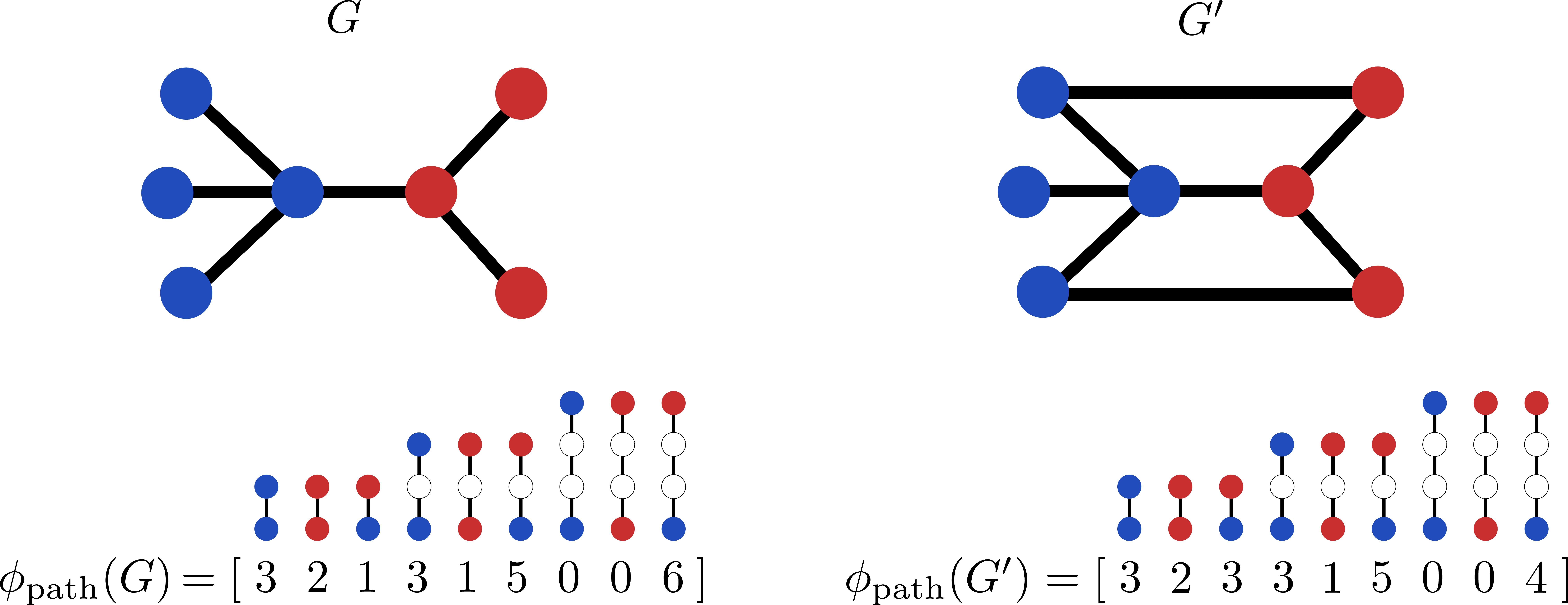}
    \caption{A possible implementation of the shortest-path kernel which
    counts the number of matching shortest paths with the same node
  labels at the end points. Here, white nodes indicate that the node can
have any label. This instance of the shortest-path kernel is the
multiplication of three Dirac delta kernels: two node kernels comparing
the labels of the source nodes and target nodes respectively, and one
edge kernel comparing the length of the shortest path. Such an instance
gives rise to an explicit feature representation (this is not always the
case), and the resulting kernel computation is
$\textrm{k}_\textrm{SP}(G, G^{\prime}) = \langle
\phi_{\textrm{path}}(G), \phi_{\textrm{path}}(G^{\prime})
\rangle_{\mathcal{H}} = 75$. While there may exist several shortest paths between two nodes, the length of the shortest path is unique.}
    \label{fig:spk}
\end{figure}

As its name suggests, the idea behind the \emph{shortest-path kernel}
(shown in Figure~\ref{fig:spk}) is
to define the similarity between two graphs in terms of the similarities
of their shortest-paths.~\citet{Borgwardt05} accomplish this by
transforming a given graph $\graph = (\vertices, \edges)$ into its
\emph{shortest-paths graph} $S = (\vertices, \edges_{S})$, which is a weighted graph.
As indicated, $S$ shares the vertices with the original graph, but its
edges $\edges_S$ are defined by the constraint that
$(v_i, v_j) \in \edges_S$ if and only if nodes $v_i$ and $v_j$ are
connected by a walk. Furthermore, the weight $w_{ij}$ of this edge
will be set to the shortest path distance of $v_i$ and $v_j$.
The transformed graph $S$ is also referred to as the \emph{Floyd-transformation}~\citep{Borgwardt05}
of $\graph$ because the original publication uses Floyd's
algorithm~\citep{Floyd62, Warshall62} to calculate shortest paths
between all pairs of nodes at the same time.
This permits us to define the shortest-path kernel.

\begin{defn}[Shortest-path graph kernel]
  Given graphs $\graph$ and $\graph'$ and their shortest-paths graphs
  $S = (\vertices, \edges_S)$ and $S' = (\vertices', \edges_S')$,
  the \emph{shortest-path graph kernel} is defined as
  \begin{equation}
    \kernelSP(\graph, \graph') := \sum_{e \in \edges_S} \sum_{e' \in
    \edges_S'} \kernelEW^{(1)}(e, e'),
  \end{equation}
  where $\kernelEW^{(1)}$ is a kernel on edge paths of length one in the
  shortest-paths graphs.
  Given two edges $e := (u, v)$ and $e' := (u', v')$,
  \citet{Borgwardt05} suggest such an edge path kernel to take the form
  of
  \begin{equation}
    \kernelEW(e, e') := \kernelN(u, u') \cdot \kernelE(e, e') \cdot
    \kernelN(v, v'),
  \end{equation}
  \ie\ a product of node kernels with an edge kernel.
\end{defn}
The definition of $\kernelEW$ allows for some flexibility in assessing
the similarity of paths.  The node kernel, for example, can be a Dirac
delta kernel that compares the labels of nodes at the beginning and end
of the path~(purposefully ignoring all other labels along the path),
while the edge kernel $\kernelE$ can be a Dirac kernel on the length
of the shortest path, and can also be easily extended to incorporate
edge features by using a measure of the difference in edge lengths of
weighted graphs~\citep{Borgwardt05}.

The computational complexity of this kernel depends on the number of
edges that have to be considered in the Floyd-transformed graphs,
leading to a worst-case runtime of \landau{n^4} for $n$ vertices. Even
though this might seem prohibitive for some applications, one of the
advantages of this kernel is its great flexibility with respect to the
node and edge kernels that can easily be adapted to make use
of arbitrary attributes. Alternatively, if the feature map $\phi_{\text{path}}(\cdot)$
corresponding to the edge path kernel $\kernelEW$ admits a $d$-dimensional
explicit representation, the computational complexity can be sharply reduced
to \landau{n^2 d}.

\begin{figure}[h!]
    \centering
    \includegraphics[width=\textwidth]{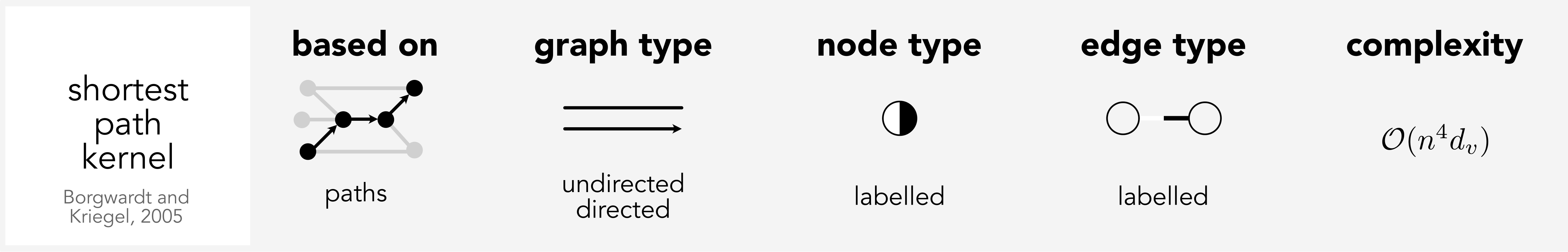}
\end{figure}

\subsubsection{GraphHopper kernel}\label{sec:GraphHopper kernel}

A drawback of the shortest-path kernel is its $\landau{n^4}$ asymptotic runtime for a graph with $n$ nodes, which can quickly become prohibitive as graphs grow in size. This motivated \citet{Feragen13} to develop the GraphHopper kernel.
Just like the shortest-path kernel, it is applicable for undirected
graphs with edge weights and optional node attributes---provided
a kernel function for comparing them is available; this is the case for
real-valued~(``continuous'') attributes, whose dissimilarity can be
assessed, for example, by means of an RBF kernel~(see Definition~\ref{def:RBF kernel}
on p.\ \pageref{def:RBF kernel}).

The central idea of the GraphHopper kernel is to compare graphs by nodes
that are encountered while the eponymous ``hopping'' along shortest
paths happens. This leads to the following general form of the kernel.
\begin{defn}[GraphHopper kernel]
  Given graphs $\graph$ and $\graph'$ and node kernel $\kernelN$, the
  GraphHopper kernel is a sum of node kernels over shortest paths, \ie\
  \begin{equation}
    \kernelGH(\graph, \graph') := \sum_{p \in \paths,\, p' \in \paths'}
    \kernel_p(p, p'),
  \end{equation}
  where $\kernel_p$ is a special path kernel that evaluates the node kernel
  $\kernelN$ along paths of \emph{equal} length $\pathlength{p}
  = \pathlength{p'}$, \ie\
  \begin{equation}
    \kernel_p(p, p') :=  \begin{cases}
                        \sum_{j = 1}^{m} \kernelN\left(p^{(j)},
                        p'^{(j)}\right) & \text{if $\pathlength{p}
                      = \pathlength{p'}$}\\
                        0                                              & \text{otherwise}
                      \end{cases},
  \end{equation}
  where $p^{(j)}$ refers to the $j$th vertex of a shortest path
  $p$.
\end{defn}
While there is a worst-case complexity of $\landau{n^4}$, it was shown~\citep{Feragen13} that the previous equation decomposes
into a weighted sum of node kernels. These weights can be calculated
efficiently, leading to an average overall worst-case complexity of $\landau{n^2 d}$
per kernel evaluation, where $n$ denotes the number of vertices and
$d$ denotes the dimension of the node attributes. This estimate assumes
that the node kernel $\kernelN$ can be calculated in time $\landau{d}$,
which is the case for most common kernels, such as the linear kernel.

\begin{figure}[h!]
    \centering
    \includegraphics[width=\textwidth]{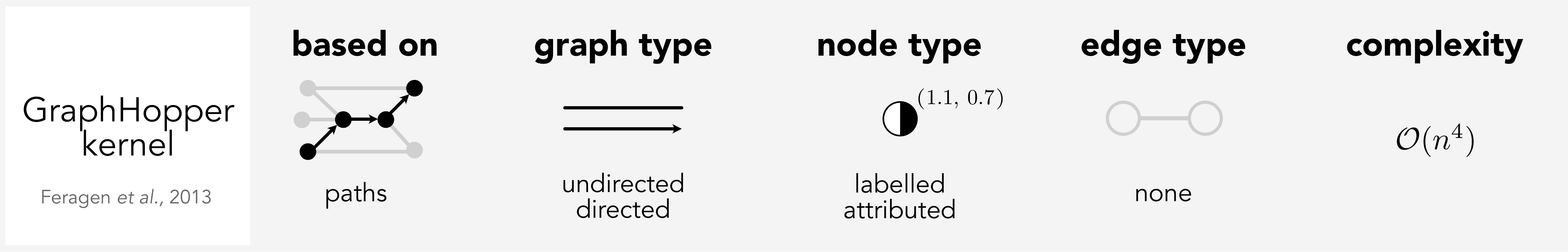}
\end{figure}

\subsection{Graph kernels based on special subgraphs}\label{sec:Graph kernels based on subgraphs}

Arguably, the most powerful representation of graphs one could obtain
would count the number of occurrences in a graph of each subgraph
occurring at least once in a given graph data set. From the perspective
of the $\mathcal{R}$-convolution framework, this corresponds to
decomposing each graph into the set of all its subgraphs and using
a Dirac kernel to quantify the similarity between these substructures.
However, is also known that computing this kernel is an NP-hard problem~\citep[Proposition 2]{Gaertner03}. As a consequence, existing graph kernels based on subgraph enumeration focus instead on counting the occurrence of special subtypes of subgraphs, as we will discuss next.

\subsubsection{Subtree pattern kernel}\label{sec:Subtree pattern kernel}

Due to the limitations mentioned above, \citet{ramon2003subtree}
proposed to limit the subgraphs considered to subtree patterns from
a root node up to a specified height $h$. For $h=1$, this amounts to
a Dirac delta kernel on the node labels, \ie\ for two vertices
$\vertex$ and $\vertex'$,  
\begin{equation}
  \kernelst^{(1)}(v,v') = 
        \begin{cases}
            1& \text{if } \Vlabel(\vertex) = \Vlabel(\vertex')\\
            0              & \text{otherwise}.
        \end{cases}
\end{equation}
For $h>1$, this considers all possible matchings $M_{\vertex,
\vertex'}$ between the nodes in the neighbourhood of the root nodes,
$\neighbourhood(\vertex)$ and $\neighbourhood(\vertex')$, and checks
whether the size of the neighbourhood is equivalent, and whether there is
a suitable match of node labels, while finally counting  how many such
matchings there are. We note that this kernel uses
\emph{subtree patterns}, as opposed to subtrees, to allow for the repetition of
nodes and edges. While Ramon and G{\"a}rtner defined the
neighbourhood to be specifically the out-degree neighbourhood, we will
use our normal neighbourhood notation since this is equivalent to the
out-degree neighbourhood in unlabelled graphs. More formally, this is to
say that $M_{\vertex, \vertex'} = \{R \subseteq \neighbourhood(\vertex)
  \times \neighbourhood(\vertex') \mid (\forall (a, a'), (b, b') \in
  R \colon a = a' \Leftrightarrow b = b' \wedge ( \forall (a, a') \in
R \colon \Vlabel(a) = \Vlabel(a')) \}$.
This leads to the following definition.
\begin{defn}
Given graphs $\graph = (\vertices, \edges)$ and $\graph' = (\vertices',
\edges')$ with node labels defined on the common alphabet $\Vlabels$,
labelling function $\Vlabel$, and vertices $\vertex \in \vertices$,
$\vertex' \in \vertices'$, the 
\emph{subtree pattern kernel} is defined as
\begin{equation}
        \kernelST^{(h)}(\graph, \graph') = \sum_{\vertex \in \vertices} \sum_{\vertex' \in \vertices'} \kernelst^{(h)}(\vertex, \vertex'),
\end{equation}
where $\kernelst^{(h)}$ is defined as
\begin{equation}
    \kernelst^{(h)} = \lambda_{\vertex}\lambda_{\vertex'} \sum_{R \in M_{\vertex,     \vertex'}} \prod_{a, a' \in R} \kernelst^{(h-1)}(\vertex, \vertex'),
\end{equation}
and $\lambda_{\vertex}$, $\lambda_{\vertex'}$ are used as a way to
give smaller weights to higher-order subtree patterns on the nodes $\vertex$ and
$\vertex'$.
\end{defn}
Due to the matching step, the subtree pattern kernel is not trivial to
compute, having a complexity of $\landau{n^2h4^d}$, where $n^2$ represents
the pairwise comparison of nodes in the two graphs, where $h$ represents
the number of iterations, and $4^d$ represents the calculation of all
matchings, where $d$ is the
maximum degree in the graph for a pair of two graphs. Nevertheless, the kernel provided the foundation for many
future kernels. For instance, \citet{mahe2009graph} extend the idea by
adding a more general parameter based on the tree patterns considered (versus at the
node level here), in order to control the effect of more complex subtree
patterns. 
\begin{figure}[h!]
    \centering
    \includegraphics[width=\textwidth]{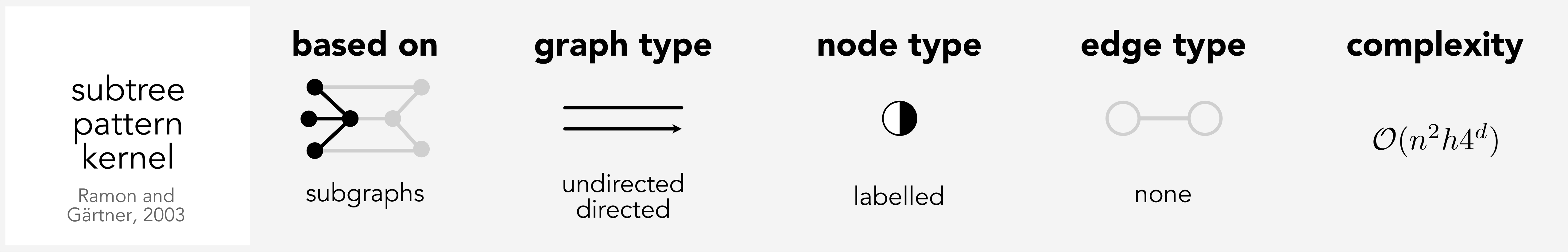}
\end{figure}

\subsubsection{Cyclic pattern kernel}\label{sec:Cyclic pattern kernel}

\citet{Horvath2004cyclic} proposed a kernel based on the patterns
of cycles and trees observed in a graph. As opposed to other
kernels that compare the frequency of given patterns, which typically
places larger importance on the patterns which are frequent, \citet{Horvath2004cyclic} instead developed a kernel to capture the diversity of
different patterns present in a given graph. 

The principal idea is to represent a graph by its set of simple cycles
$\mathcal{C}(G)$ and the set of trees $\mathcal{T}(G)$ present in
a graph $G$.  A graph $\graph$ can be decomposed into cycles and trees
by removing any \emph{cut vertices}, \ie\ vertices that will
disconnect the graph when both the vertex and its incident edges are removed. What
remains are \emph{maximal biconnected components} of $\graph$ and trees formed
by the cut vertices. These components form the sets $\mathcal{C}(G)$
and $\mathcal{T}(G)$, which are ordered using a canonical representation
that is obtained by using a function $\pi(\cdot)$, which finds the
smallest lexicographic ordering of the sequences of nodes and edges in
each cycle and tree by using the labels assigned to the nodes and edges.
For the ordering of cycles, this function is defined as
\begin{equation}
    \pi(C) := \min\{\sigma(w) \mid w \in \rho(s)\},   
\end{equation}
where $\rho(s)$ is the set of all possible orderings of a cycle $s$, and
$\sigma(w)$ assigns a value to the particular ordering $w$ using the
labels of the nodes $\Vlabel$ and edges $\Elabel$ in the sequence, \ie\
\begin{equation}
  \sigma(w) = \Vlabel(v_0)\Elabel((v_0, v_1))\Vlabel(v_1) \cdot
  \ldots \cdot \Vlabel(v_{k-1})\Elabel((v_{k-1}, v_0))
\end{equation}
for vertices $v_i$ and edges $(v_i, v_{i+1})$ in the cycle
(considering both possible directions of the sequence in a cycle).
A similar process is carried out to provide an ordering for the trees
present in the graph, and therefore ensures that identical cycles and
trees will be comparable in different graphs. The sets of cycles and
trees of a graph $\graph$ are therefore
\begin{align*}
        \mathcal{C}(\graph) &:= \{\pi(C) \mid C \in \mathcal{S}(G) \}\\
        \mathcal{T}(\graph) &:= \{\pi(T) \mid C \in \mathcal{T}(G) \}.
\end{align*}
\begin{defn}[Cyclic pattern kernel]
  Given two graphs $\graph$ and $\graph'$ with node and edge labels from
  their respective alphabets $\Vlabels$ and $\Elabels$, the \emph{cyclic
  pattern kernel} kernel is defined the cardinality of the intersection
between these two sets, \ie\
\begin{equation}
    \kernelCP(\graph, \graph') := |\mathcal{C}(\graph) \cap \mathcal{C}(\graph')| + |\mathcal{T}(\graph) \cap \mathcal{T}(\graph')|.
\end{equation}
 \end{defn}
The computation of this kernel is, as acknowledged by the
authors~\citep[Proposition~1]{Horvath2004cyclic},
NP-hard, since a graph with $n$ nodes can have more than $2^n$ cycles or
tree patterns. However, the authors propose a variation of their kernel
that works when all the graphs in a data set (or a high proportion
thereof) have a bounded number of simple cycles (graphs with more than such a bound
can be disregarded). Such a modification results in an upper bound on
the runtime of $\landau{(c+2)n + 2m}$ for a given pair of graphs, where $c$ is the bound on the number of simple cycles in any graph,
$n$ is the maximum number of nodes in any graph in the data set,
and $m$ is the maximum number of edges in any graph in the
data set.
\begin{figure}[h!]
    \centering
    \includegraphics[width=\textwidth]{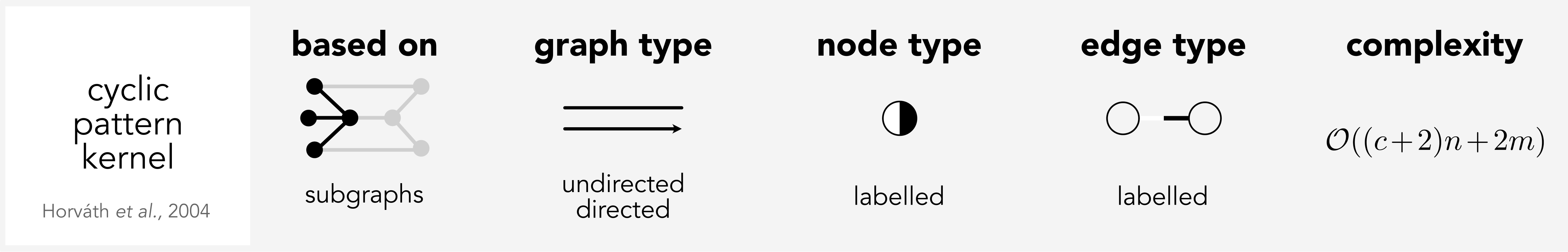}
\end{figure}

\subsubsection{Graph edit distance kernels}\label{sec:Graph edit distance kernels}

Another approach to counteract the computational bottleneck of
enumerating and counting all possible subgraphs is to select a subset of
important subgraphs, \ie\ \emph{prototypes}, a concept equivalently
known as \emph{landmarks}~\citep{Hsieh14}, and then assess how many
graph edits are necessary for a given graph to include such prototypes.
Specifically, the key ideas behind this approach are to instead~(i)
select the subgraphs on which the feature map will be based directly
from the data set of graphs we wish to represent, \ie\ data-driven
\emph{prototypes}, and~(ii) define the feature map in terms of the graph
edit distances to each of these prototypes, which can be computed for
any type of attributed graphs, rather than counting exact matches. 

This makes it possible to obtain a kernel from a metric defined on
graphs~(which is normally fraught with difficulties, such as having to
prove that the resulting kernel is p.d.; however, since this formulation
directly defines a feature map, \emph{any} p.d.\ kernel can be used to
compare the resulting feature vectors).
For example, the \emph{graph edit distance}, as briefly
described in Section~\ref{sec:Graph edit distance}, has the advantage of
being applicable to different types of graphs: by modifying the cost
functions in the appropriate fashion, it is possible to handle directed
graphs, graphs with continuous attributes, and so on. \citet{Bunke07}
thus proposed using graph edit distances to \emph{embed} graphs into
a feature space.  This permits using either feature-based classification
algorithms, \ie\ algorithms that work directly on the vector representation, or kernel-based methods.
In the following, let $\mathcal{G} := \left\{\graph, \dots, \graph_n
\right\}$ be a set of $n$ input graphs and $\gdistance\left(\cdot,
\cdot\right)$ their corresponding graph edit distance calculation
function. With a suitably-selected subset of graphs, this gives rise
to an embedding.

\begin{defn}[Graph edit distance embedding]
  Given $\mathcal{G}$ as defined above, let $\mathcal{P} := \left\{P_1,
  \dots, P_m \right\} \subseteq \mathcal{G}$ be a subset of
  \emph{prototype} graphs~(we will subsequently discuss several
  strategies for choosing them), where $m \leq n$ by definition.
  The \emph{graph edit distance embedding} of a graph
  $\graph\in\mathcal{G}$ is then defined as
  \begin{equation}
    \featurevector_{\mathcal{G}}\left(\graph\right) := \left(
      \gdistance\left(\graph, P_1\right),
      \gdistance\left(\graph, P_2\right),
      \dots
      \gdistance\left(\graph, P_m\right)
    \right),
  \end{equation}
  which creates a mapping $\featurevector_{\mathcal{G}}\colon\mathcal{G}\to\real^m$.
\end{defn}

The graph edit distance mapping can now be used either directly as the feature
vector for feature-based algorithms, or provided with an appropriate kernel on
a feature vector space.

\begin{defn}[Graph edit distance kernel]
  Let $\basekernel$ be a well-defined kernel defined for real-valued
  vector spaces of dimension $m$.
  Then the \emph{graph edit distance kernel} between two graphs $\graph$
  and $\graph'$ is defined as
  \begin{equation}
    \kernelED\left(\graph, \graph'\right) := \basekernel\left(
      \featurevector_{\mathcal{G}}\left(\graph\right),
      \featurevector_{\mathcal{G}}\left(\graph'\right)
    \right),
  \end{equation}
  which is a valid kernel between graphs by definition.
\end{defn}

\citet{Bunke07} use a variant of an RBF kernel for their experiments, \ie\
$\basekernel\left(
      \featurevector_{\mathcal{G}}\left(\graph\right),
      \featurevector_{\mathcal{G}}\left(\graph'\right)
\right) := \exp\left(-\gamma \| \featurevector_{\mathcal{G}}\left(\graph\right) -
                                \featurevector_{\mathcal{G}}\left(\graph'\right)
                             \|\right)$,
where $\|\cdot\|$ denotes the usual Euclidean norm, and $\gamma \in
\real$ is a scaling parameter. However, other choices of $\basekernel$
are possible, such as linear kernels. The choice of the \emph{prototype
set} $\mathcal{P}$ is crucial for the suitability of the embedding. \citet{Bunke07}
discuss the properties of various selection schemes, such as a \emph{spanning} selection,
which starts from randomly-selected graph and iteratively extends the selection by taking
the graph that has the maximum graph edit distance from all selected graphs.
Specifically, given a set of graphs $\graphs$, suppose that a subset
of graphs $\widetilde{\graphs} = \{\graph, \graph', \dots\}
\subseteq \graphs$ has already been selected by the prototype
selection algorithm~(the base case for $\widetilde{\graphs} = \emptyset$ is
typically solved by selecting a graph from $\graphs$ at random).
The next graph to include in $\widetilde{\graphs}$ is the graph
$\graph^{\ast} \in \graphs \setminus \widetilde{\graphs}$ that
\emph{maximises} the function
\begin{equation}
  \graph \mapsto \min\left\{
    \gdistance\left(\graph, \graph\right),
    \gdistance\left(\graph, \graph'\right),
    \dots
  \right\},
  \label{eq:Graph selection min--max}
\end{equation}
\ie\ the $\graph^{\ast}$ whose minimum distance to the set of selected
prototype graphs $\widetilde{\graphs}$ is as \emph{large} as
possible---this ensures that the most ``diverse'' set of graphs is
selected; similar strategies are very common for landmark selection
algorithms in computational geometry, for example~\citep{Silva04}.
Without loss of generality, we may assume that the graph maximising
Eq.~\ref{eq:Graph selection min--max} is \emph{unique}. If this is not
the case, $\graph^{\ast}$ can be selected at random from the set of
candidates.
For real-world applications, prototype selection can also be treated as
a hyperparameter of the algorithm, which is thus subject to
cross-validation.

In terms of complexity, the feature vector creation hinges on fast
algorithms for the graph edit distance. These algorithms depend on the
graph structure; we refer to \citet{Riesen15} for a detailed
introduction to state-of-the-art algorithms.
A recent preprint~\citep{Bai18} also deals with graph edit distance
computation through the lens of graph neural networks, the key idea
being that the network \emph{learns} how to calculate the graph edit
distance. Preliminary results indicate that the quality of the
approximation is highly dependent on the data set. Hence, there is still
a  need for other algorithmic approximations.

\begin{figure}[h!]
    \centering
    \includegraphics[width=\textwidth]{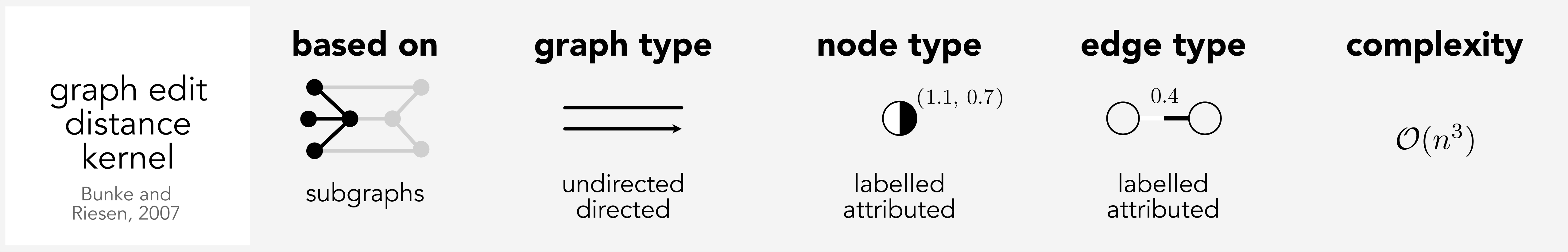}
\end{figure}

\subsubsection{Graphlet kernel}\label{sec:Graphlet kernel}

\begin{figure}[h]
\begin{center}
\begin{subfigure}{0.5\textwidth}
    \centering
    \includegraphics[width=0.9\linewidth]{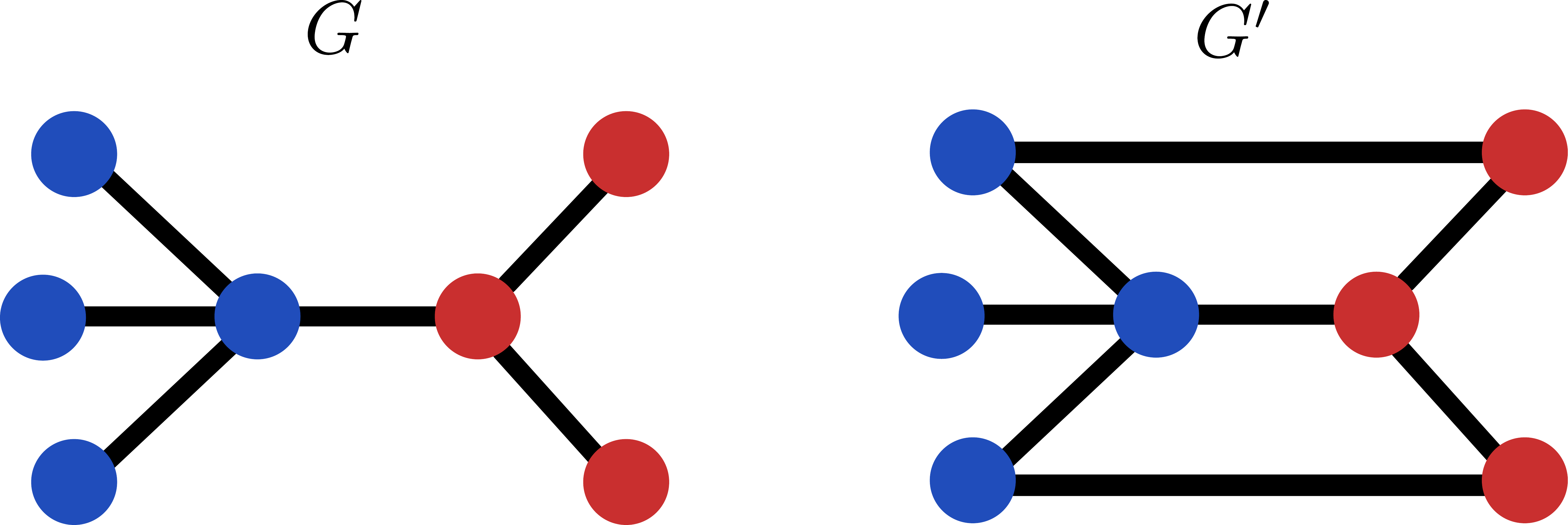} 
    \caption{Graphs $G$ and $G^{\prime}$.}
    \label{fig:graphlet_gg'}
\end{subfigure}
\end{center}

 \par\bigskip
\begin{subfigure}{\textwidth}
    \centering
    \includegraphics[width=0.9\linewidth]{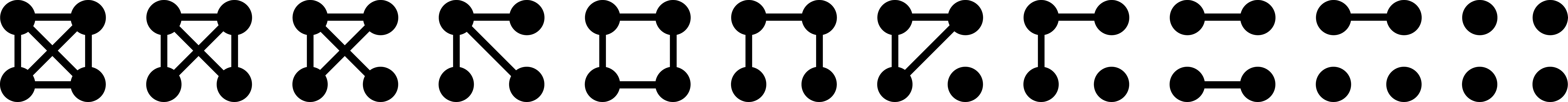}
    \caption{All subgraphs of size 4.}
    \label{fig:size_4_graphlets}
\end{subfigure}

 \par\bigskip
 \begin{center}
    \begin{subfigure}{0.8\textwidth}
        \centering
        \includegraphics[width=0.9\linewidth]{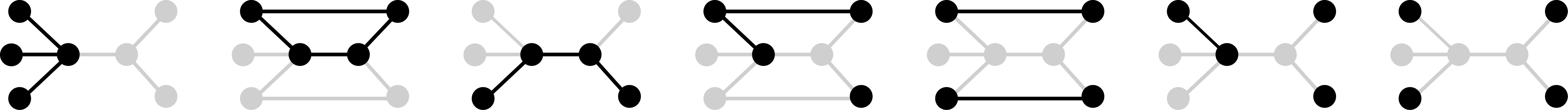}
        \caption{Examples of different size 4 graphlets found in $G$ and $G^{\prime}$.}
        \label{fig:example_graphlets}
    \end{subfigure}
\end{center}

\caption{The graphlet kernel counts the number of matching subgraphs of
size $k$ in a given graph, resulting in an explicit feature
representation for each graph. This representation, called the
$k$-spectrum, can be done using either all possible graphlets of size
$k$, as visualised in (\subref{fig:size_4_graphlets}), or using only the
connected graphlets of size $k$. The $k$-spectra are then typically
normalised, and the resulting kernel is the linear kernel between the
two spectra: $\kernelGL(\graph, \graph') :=
\featurevector_{gd}(\graph)^{\top} \featurevector_{gd}(\graph')$.}
\label{fig:graphlet_kernel}
\end{figure}

The \emph{graphlet kernel} bypasses the difficulties arising from the need to enumerate a potentially enormous number of subgraphs by restricting its feature space to counts of subgraphs of a fixed size. Specifically, we will use subgraphs with a small number of nodes, often referred to as \emph{graphlets}.
This term was introduced by \citet{Przulj04} in the context of
describing protein--protein interaction networks and subsequently
extended with a more efficient estimation procedure~\citep{Przulj06}.
\citet{Shervashidze09a} then developed a kernel, based on the notion
that two graphs should be considered to be similar if their graphlet
distributions are similar.

This results in a simple algorithm, which we will subsequently describe.
Given $k$, let $\graphlets := \{\graphlet_1, \dots, \graphlet_{N_k}\}$
refer to the set of graphlets of size $k$.
This method ignores all labels and attributes for this enumeration and
merely focuses on connectivity.
Generating different graphlets is a combinatorial problem and the set of
$k$-graphlets, even though exponential in $k$, is fully enumerable. We
show an example of enumerating graphlets and matching their occurrences
in a given graph in Figure~\ref{fig:graphlet_kernel}.
Having enumerated all graphlets, we count their occurrence in a graph
$\graph$, which yields an $N_k$-dimensional vector
$\featurevector_{gd}(\graph)$ whose $i$th entry contains the frequency
of occurrence of $\graphlet_i$ in $\graph$.  We will refer to this
vector as the \emph{k-spectrum} of a graph; it leads to a graphlet
comparison kernel.
\begin{defn}[Graphlet kernel]
  Given graphs $\graph$ and $\graph'$ and their corresponding
  k-spectra vectors for a fixed $k$, $\featurevector_{gd}(\graph)$ and
$\featurevector_{gd}(\graph')$, the graphlet kernel is defined as
  \begin{equation}
    \kernelGL(\graph, \graph') := \featurevector_{gd}(\graph)^{\top}
    \featurevector_{gd}(\graph') = \featurevector_{gd}(\graph')^{\top}
    \featurevector_{gd}(\graph),
    \label{eq:Graphlet kernel}
  \end{equation}
  \ie\ a \emph{linear kernel} between the two spectra. The $k$-spectra
  are typically normalised by dividing them by the total number of
  graphlets that occur in the graph.
\end{defn}
The computational bottleneck of the graphlet kernel is the enumeration
of all graphlets. Since the number of arbitrary graphlets is exponential
in the number of vertices in the graphlet, \citet{Shervashidze09a} suggest
that only graphlets for $k \in \{3, 4, 5\}$ be computed; the closure
properties of graph kernels make it possible to evaluate
Eq.~\ref{eq:Graphlet kernel} for different values of~$k$ and combine the
results.
Moreover, \citet{Shervashidze09a} show that their computation has
a complexity of $\landau{nd^4}$, where $d$ refers to the maximum degree
in a graph. Notably, \citet{kondor2009graphlet} proposed a follow-up approach 
that uses notions from group representation theory to extend the graphlet kernel
in order to account for the relative position of different graphlets in a graph, as well
as to incorporate node and edge attributes.

\begin{figure}[h!]
    \centering
    \includegraphics[width=\textwidth]{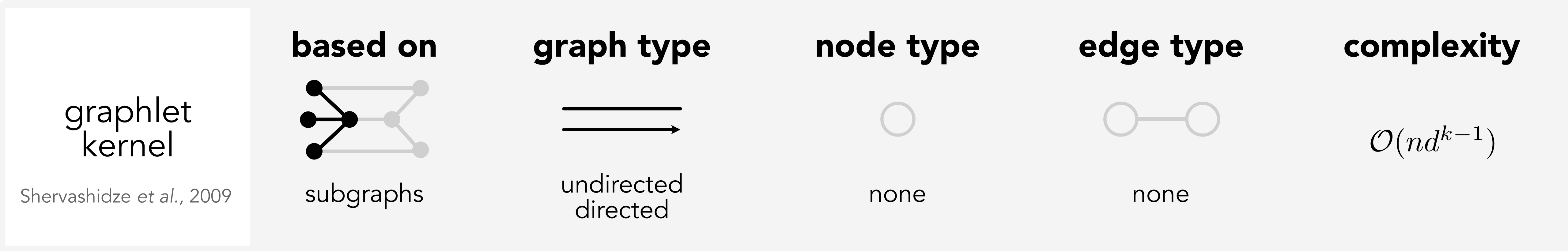}
\end{figure}
\section{Information propagation}\label{sec:Information Propagation}

Our second high-level category, information propagation, includes methods that observe how
information can be diffused throughout the graph. Walks and spectral
methods can be seen as special instances of graph-based information
diffusion processes. Similarly, iterative label refinement methods may
also be considered as iteratively propagating information between
neighbouring nodes. We now consider each of these in turn.
%
\subsection{Graph kernels based on walks}\label{sec:Graph kernels based on walks}

A crucial property of walks is that they are
amenable to efficient computation, unlike many other substructures. In
particular, letting $A$ denote the adjacency matrix of a graph
$\graph = (\vertices, \edges)$ and using the definition of matrix
multiplication, one can show by induction that $A^{k}_{i,j}$
equals the number of walks of length $k$ from node $v_{i}$ to node
$v_{j}$ in $\graph$. As a consequence, the number of walks of a certain length
between any pair of nodes in a graph can be computed in polynomial time.
This observation motivated the use of walks as the substructure of
choice for the first graph kernels proposed in the literature~\citep{Kashima03, Gaertner03}.

However, they are not without important limitations.
A well-known problem of graph kernels based on walks is the phenomenon
often referred to as \emph{tottering}~\citep{mahe2004extensions}. In
brief, tottering occurs as a consequence of walks allowing vertices to
be visited multiple times, potentially inflating the similarity of
graphs that have matching edges as these could be visited infinitely
many times. Another disadvantage of walks, investigated in-depth more
recently, is the phenomenon of \emph{halting}~\citep{Sugiyama15}.
Halting occurs because, for sufficiently small values of the
hyperparameter $\lambda$, the random walk kernel is dominated by the
constant and first-order terms of the infinite series, corresponding to
walks of length $0$ and $1$, respectively. In these cases, the random
walk kernel effectively degenerates to treating the graphs as
bags-of-nodes-and-vertices, much like the baselines introduced in
Sections~\ref{sec:Graph kernels based only on nodes} and~\ref{sec:Graph
kernels based only on edges}. However, in real-world applications, small
values of $\lambda$ might be required to guarantee convergence of the
series. One possibility to ameliorate these issues is to truncate the
series defining the random walk kernel, as discussed in~\citep[Section
4]{Sugiyama15}. 
Despite these limitations, their ability to capture higher-order
relationships between nodes while having computationally favourable
properties has made walks a critical pillar of graph kernel methods.
In this subsection, we will describe those approaches, as well as follow-up work that 
improved the computational efficiency of walk-based graph kernels and
methods that aimed to investigate alternative types of walks.
%

\subsubsection{Direct product graph kernel}\label{sec:direct product graph kernel}

\begin{figure}
    \centering
  \begin{subfigure}{0.45\textwidth}
    \centering
    \includegraphics[width=0.9\linewidth]{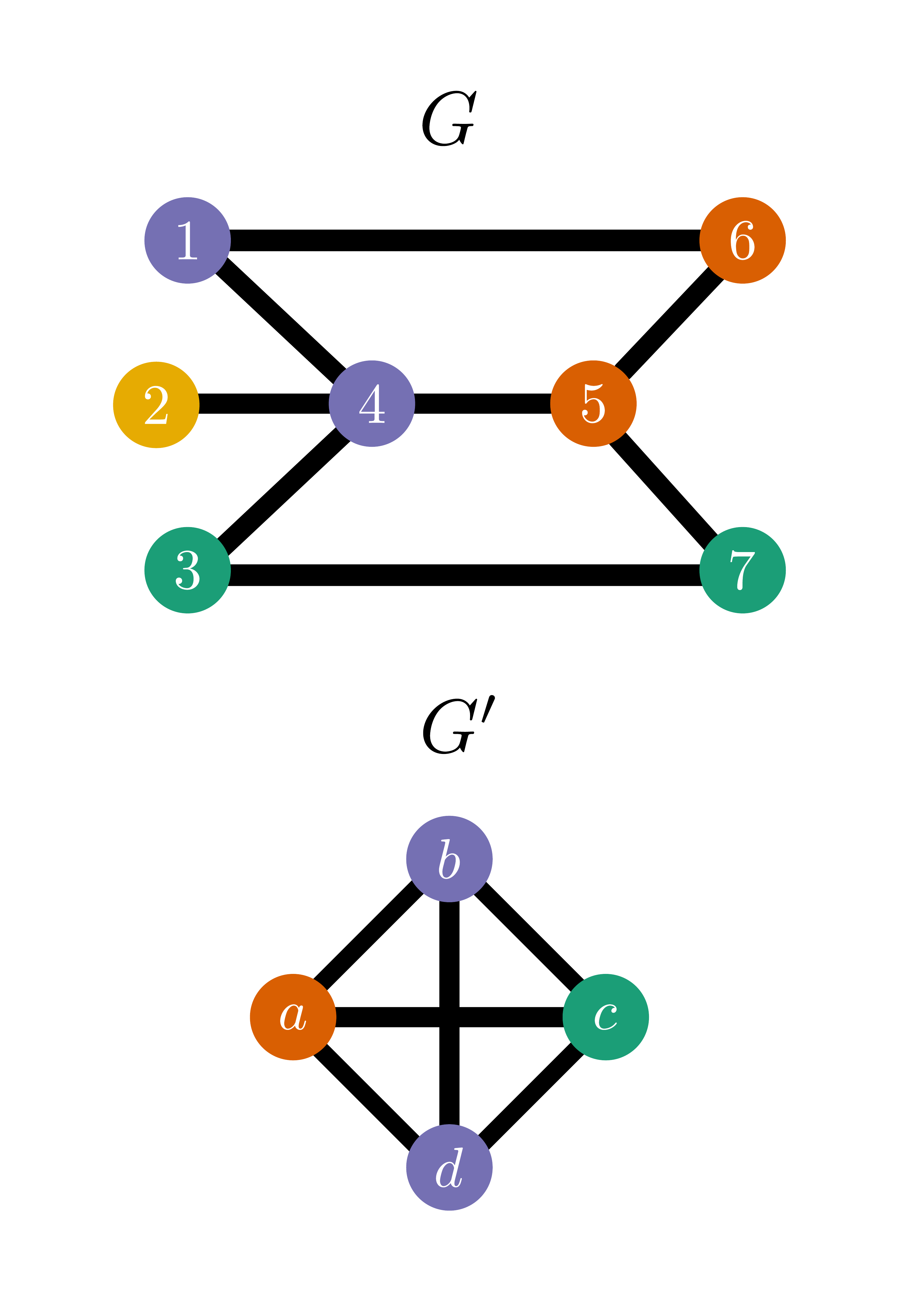} 
    \caption{Graphs $\graph$ and $\graph^{\prime}.$}
    \label{fig:dpg1}
  \end{subfigure}
  \hfill
  \begin{subfigure}{0.45\textwidth}
    \centering
    \includegraphics[width=0.9\linewidth]{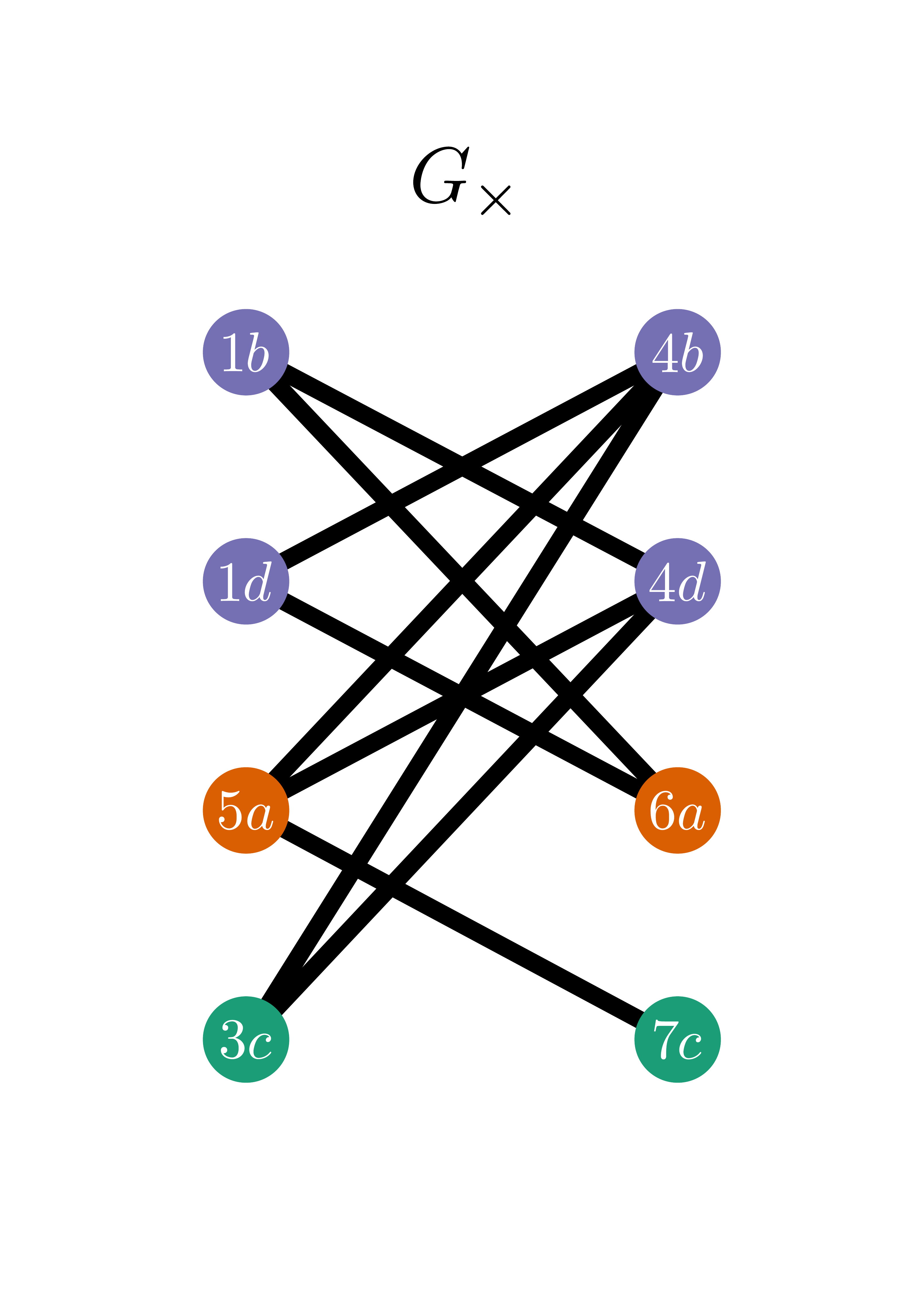} 
    \caption{The product graph $\graph_{\times}$}
    \label{fig:dpg2}
  \end{subfigure}
    \caption{The direct product graph $G_{\times}$ of graphs \graph\,
    and $\graph^{\prime}$. In $G_{\times}$, the node uses the number of
  $\graph$ and the letter from $\graph'$ as the index. The node in
$\graph_{\times}$ exists if the labels of the nodes are the same in the
two graphs, and an edge exists between two nodes in $\graph_{\times}$
if there are edges between the corresponding nodes in $\graph$ and
$\graph'$.} 
    \label{fig:direct_product_graph}
\end{figure}

Given a graph $G=(V, E)$ with categorical node and edge attributes, any walk $\omega = (v_{1}, v_{2}, \hdots, v_{k+1})$ of length $k$ in $G$ can be represented as the sequence of node and edge labels encountered along the walk, that is, $s(\omega) = \left(l_{\text{V}}(v_{1}), l_{\text{E}}( (v_{1}, v_{2})), l_{\text{V}}(v_{2}), \hdots, l_{\text{E}}( (v_{k}, v_{k+1})), l_{\text{V}}(v_{k+1})\right)$. If node and edge attributes take values in finite alphabets $\Sigma_{\text{V}}$ and $\Sigma_{\text{E}}$, then the collection of all sequences that can obtained this way forms a countable set. Conceptually, a graph could then be represented by a feature map that counts the number of occurrences of each possible label sequence in the graph. This is precisely the idea behind the \emph{direct product graph kernel} introduced by~\citet{Gaertner03}.

An indispensable tool to tractably compute a kernel based on this
feature map is the \emph{direct product graph}, from which the kernel
derives its name. We provide an example of a direct product graph in
Figure~\ref{fig:direct_product_graph}. 

\begin{defn}[Direct product graph]
  Given two graphs $\graph = (\vertices, \edges)$ and $\graph'
  = (\vertices', \edges')$, the \emph{direct product graph}
  $\graph_{\times} := (\vertices_{\times}, \edges_{\times})$ is a graph
  that captures walks that induce identical label sequences in $\graph$ and
  $\graph'$. More precisely, we have
      $\vertices_{\times} := \left\{\left(v, v'\right) \in \vertices
      \times \vertices' \mid l_{\text{V}}(v) = l_{\text{V}}(v') \right\}$
      and $\edges_{\times} := \left\{\left(\left(u, v\right), \left(u',
      v'\right)\right) \in \edges \times \edges' \mid l_{\text{E}}( (u, v)) = l_{\text{E}}( (u', v'))\right\}$.
  Thus, $\vertices_{\times}$ and $\edges_{\times}$ correspond
  to matching pairs of nodes and edges in $\graph$ and
  $\graph'$.
  \end{defn}

As shown in~\citet[Proposition 3]{Gaertner03}, the key property of the
product graph is that any walk in $\graph_{\times}$ is in one-to-one
correspondence to a pair of walks in $\graph$ and  $\graph'$ that have
the same sequence of node and edge labels. Thus, the problem of counting
matching walks between $\graph$ and  $\graph'$ can be reduced to the
problem of counting walks in $\graph_{\times}$ which, as described
above, can be solved in polynomial time via matrix multiplication. Using
this property, the direct product graph kernel can then be defined as follows.

\begin{defn}[Direct product graph kernel]
  Given two graphs $\graph$ and $\graph'$ the direct product graph kernel is calculated as
  \begin{equation}
  \kernel_{\text{CP}}(\graph, \graph') := \sum_{i=1}^{\vert \vertices_{\times} \vert} \sum_{j=1}^{\vert \vertices_{\times} \vert} \left[ \sum_{k=0}^{\infty} \lambda_{k} A_{\times}^{k} \right]_{ij},
  \end{equation}
  where $A_{\times}^{k}$ denotes the $k$-th power of the adjacency
  matrix of the direct product graph and $(\lambda_k)_{k=0}^{\infty}$ is a sequence of non-negative scalars such that $\lambda_{k}$ weights the contribution of $k$-length walks to the resulting kernel.
\end{defn}

The direct product graph kernel can also be applied to graphs without
node and/or edge labels by considering all missing attributes to be identical, in which case the conditions $l_{\text{V}}(v) = l_{\text{V}}(v')$ and/or $l_{\text{E}}( (u, v)) = l_{\text{E}}( (u', v'))$ in the definition of the direct product graph would become trivially true. Early attempts to define the sequence of weights $(\lambda_k)_{k=0}^{\infty}$ focused on computational considerations. One of the most common choices, originally proposed in~\citet{Gaertner03}, is to set $\lambda_{k} := \lambda^{k}$. This reduces the number of kernel hyperparameters to just one and guarantees convergence of the infinite series defining the kernel provided that $\vert \lambda \vert < 1 / \rho(A_{\times})$, where $\rho(A_{\times})$ stands for the spectral norm of the adjacency matrix of the direct product graph. Moreover, in this case, it can be shown that $\sum_{k=0}^{\infty} \lambda^{k} A_{\times}^{k} = \left(\mathbf{I} - \lambda A_{\times}\right)^{-1}$ and, thus, the kernel can be expressed in closed-form using matrix inversion. Since matrix inversion has complexity $\landau{p^{3}}$ for a $p \times p$ matrix and, given graphs $\graph$, $\graph'$ with $n$ nodes each, the dimension of $A_{\times}$ can be $n^{2} \times n^{2}$ in the worst case, evaluating $\kernel_{\text{CP}}(\graph, \graph')$ with $\lambda_{k} = \lambda^{k}$ in this manner results in worst-case computational complexity of the order $\landau{n^{6}}$.
\begin{figure}[h!]
    \centering
    \includegraphics[width=\textwidth]{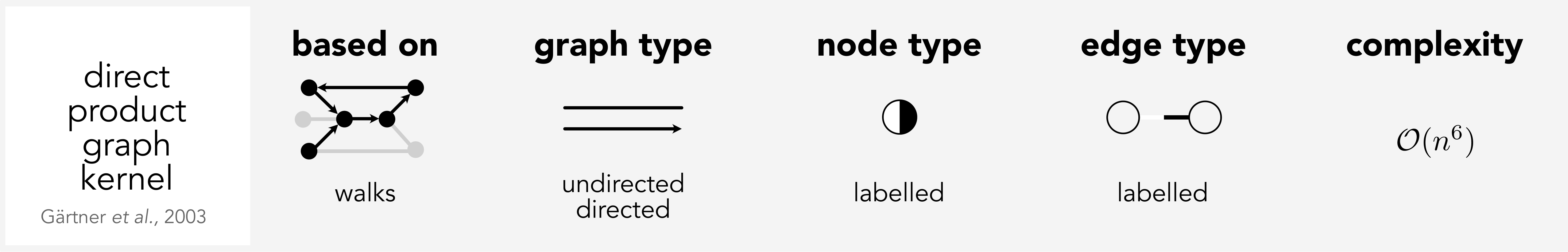}
\end{figure}
\subsubsection{Marginalized random walk kernel}\label{sec:Marginalized random walk kernel}

The direct product graph kernel represents a simple instantiation of the $\mathcal{R}$-convolution framework, decomposing a graph $G$ into the set of all its walks $\mathcal{W}$ and defining the base kernel between walks to be a Dirac kernel on the induced node and edge label sequences. By considering a more general base kernel $\kernel_{\text{walk}}$, the direct product graph kernel can be extended to handle continuous node and edge attributes. The \emph{marginalized random walk kernel}, introduced by~\citet{Kashima03}, can be understood as such a generalization.

\begin{defn}[Marginalized random walk kernel]
  Given two graphs $\graph$ and $\graph'$ the marginalized random walk kernel is obtained as
  \begin{equation}
  \kernel_{\text{MRW}}(\graph, \graph') := \sum_{\omega_{1} \in \mathcal{W}_{1}}\sum_{\omega_{2} \in \mathcal{W}_{2}} \kernel_{\text{walk}}(s(\omega_{1}), s(\omega_{2})) p(\omega_{1} \mid \graph) p(\omega_{2} \mid \graph'),
  \end{equation}
  where $\kernel_{\text{walk}}$ is a non-negative p.d.\ kernel between node and edge attribute sequences and $p(\omega \mid G)$ is the probability of $\omega$ being the outcome of a random walk in $G$.
\end{defn}

Importantly,~\citet{Kashima03} shows that, given some restrictions on $\kernel_{\text{walk}}$, the marginalized random walk kernel can also be computed by inverting a $n^{2} \times n^{2}$ matrix for graphs $\graph$, $\graph'$ with $n$ nodes each, resulting in the same worst-case computational complexity as the direct product graph kernel, that is, $\landau{n^{6}}$.
\begin{figure}[h!]
    \centering
    \includegraphics[width=\textwidth]{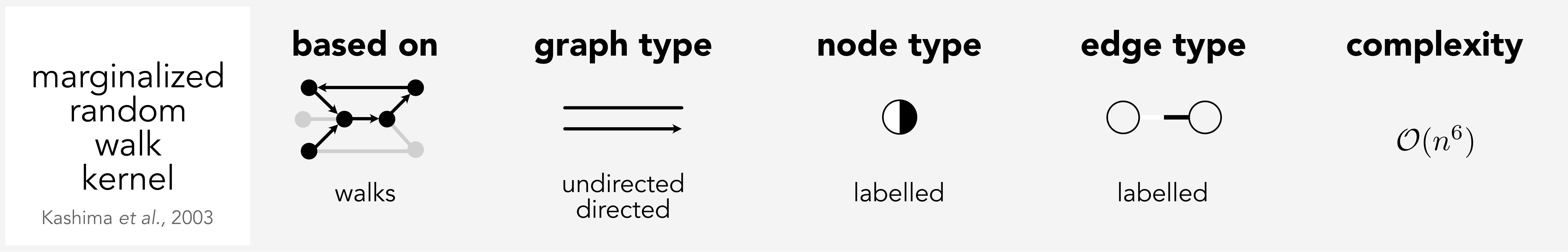}
\end{figure}

\subsubsection{Fast computation of walk-based kernels}\label{sec:Fast computation of walk-based kernels}

Both the direct product graph kernel and the marginalized random walk kernel can be computed in polynomial time. However, their asymptotic complexity, in the order of $\landau{n^{6}}$ for graphs having $n$ nodes, severely limits their practical applicability. \citet{Vishwanathan06} introduced an advance in the way graph kernels based on walks are computed, drastically reducing the computational complexity with respect to the size of the graphs to $\landau{n^{3}}$. Their method applies to a broad family of walk-based kernels, which include the direct product graph kernel as well as the marginalized random walk kernel whenever continuous node and edge attributes are compared using kernels whose feature maps can be computed explicitly. We will refer to the graph kernels that can be derived from this framework simply as \emph{random walk kernels}.

\begin{defn}[Random walk kernel]
  Given two graphs $\graph$ and $\graph'$, the random walk kernel is calculated as
  \begin{equation}
    \kernelRW(\graph, \graph') := q_{\times}^{\top} \left[\sum_{k = 0}^{\infty} \lambda_{k}  W_{\times}^{k}\right] p_{\times},
  \end{equation}
  where (i) $W_{\times} := W_{1} \otimes W_{2}$, with $W_{1}$ and $W_{2}$ being matrices whose entries are elements of a RKHS and have the same sparsity pattern as the adjacency matrices of $\graph$ and $\graph'$, respectively; (ii) $p_{\times}$ and $q_{\times}$ represent the initial and stopping probability distributions of the random walk and (iii) $(\lambda_k)_{k=0}^{\infty}$ is a sequence of non-negative scalars such that $\lambda_{k}$ weights the contribution of $k$-length walks to the resulting kernel.
\end{defn}

As previously discussed in Section~\ref{sec:direct product graph kernel}, choosing
exponentially-decaying weights $\lambda_k = \lambda^{k}$ leads to
$\kernelRW(\graph, \graph') = q_{\times}^{\top}  \left(\mathbf{I}
- \lambda W_{\times} \right)^{-1} p_{\times}$. Efficiently evaluating
the random walk kernel therefore hinges on exploiting the Kronecker
structure of $W_{\times}$ to obtain $\left(\mathbf{I} - \lambda
W_{\times} \right)^{-1} p_{\times}$ with a series of matrix-vector
products of the form $W_{\times} r$ using techniques such as fixed-point
iterations or conjugate gradient methods~\citep{Vishwanathan06}. As
a result, if the entries in $W_{1}$ and $W_{2}$ are elements of a RKHS
whose feature map can be represented by a $d$-dimensional vector, the
computational complexity of the random walk kernel is $\landau{n^{3}d}$
as opposed to $\landau{n^6}$ for a naive implementation.

Recent extensions of random walk kernels include \texttt{RetGK},
a kernel based on the return probabilities of random
walks~\citep{Zhang18}. This kernel is based on a new
descriptor of random walks that incorporates their return probabilities.
Next to being invariant under graph isomorphism, this descriptor also
permits treating attributed and non-attributed graphs within
the same framework. This is achieved by \emph{enriching} the obtained walk
representations, which do not require attribute information to be
available, with additional information about node attributes, for
instance. While \texttt{RetGK} exhibits improved expressivity and thus
improved predictive performance, its computation has a marginally higher
computational complexity of $\landau{n^3 + (k + 1)n^2} = \landau{n^3
+ kn^2}$ for random walks of at most~$k$ steps.
\begin{figure}[h!]
    \centering
    \includegraphics[width=\textwidth]{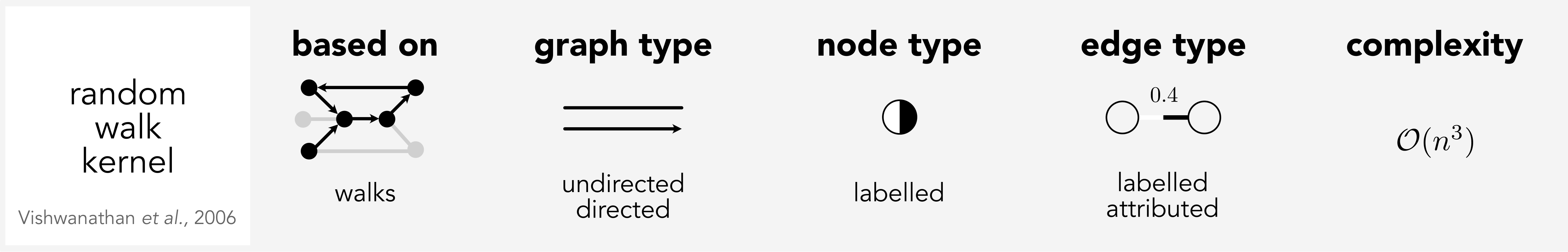}
\end{figure}
%
\subsubsection{Continuous-time quantum walk kernel}\label{sec:Quantum walk kernel}

Recent work has sought inspiration in the formalisms of quantum mechanics to make use of an alternative type of random walk in a graph such that, at any given time, the state of the walk does not correspond to a single vertex but rather to an arbitrary superposition of basis states. Quantum walks have several properties that
are not present in ``classical'' random walks, such as reversibility and non-ergodicity~\citep{Bai13}. For this subsection, we assume that we are dealing with unlabelled, undirected graphs.
Given such a graph $\graph = \left(\vertices, \edges\right)$, we first
define the evolution of a general quantum walk in close analogy to
random walks.
%
\begin{defn}[Continuous-time quantum walk]
  Let $n := \left|\vertices\right|$ be the number of vertices in the
  graph and $U := \{u_1, \dots, u_n\}$ be an orthonormal basis of a
  complex Hilbert space $\mathcal{H}$. Given a set of time-varying amplitude
  vectors $\alpha\left(u_j, t\right) \in \complex$, the state of a
  \emph{continuous-time quantum walk} is defined as
  \begin{equation}
    \psi\left(t\right) = \sum_{i=j}^{n} \alpha\left(u_j, t\right) u_j.
  \end{equation}
  The amplitude vectors can be defined using the Laplacian matrix
  $\laplacian$ of $\graph$, leading to
  \begin{equation}
    \psi\left(t\right) = \exp(-i \laplacian t) \psi\left(0\right),
  \end{equation}
  which can be seen as a solution of the Schr{\"o}dinger equation for this graph~\citep{Bai13}.
  The initial state $\psi\left(0\right)$ in the previous equation is defined using
  the \emph{degree} of a given vertex, \ie\
  \begin{equation}
    \psi\left(0\right) := \sum_{j=1}^{n} \frac{d_{u_j}}{\sqrt{\sum_{k=1}^{n} d_{u_k}^2}} u_j,
  \end{equation}
  which is equal to the steady state of a classical random walk on the graph.
\end{defn}
%
To obtain a kernel for comparing different quantum walks, a notion of
\emph{entropy} is introduced~\citep{Bai13}. The underlying idea is to
define a density matrix over the individual states of the graph. This
is achieved by rephrasing the continuous formulation from above into
a discrete form~(as proposed by \citet{Bai13}, we use the same
terminology as before to make the link clearer).
Specifically, we define a maximum number of time steps $s \in \natural$
and using a uniform probability $p := 1 / s$ for assuming each of the
states $\psi(1), \dots, \psi(s)$. This leads to a \emph{density
operator} $\rho_\graph$ of the graph $\graph$ as
\begin{equation}
  \rho_\graph = \sum_{t=1}^{s} p \cdot \psi(t) \psi(t)^\top,
\end{equation}
\ie\ the weighted sum of an \emph{outer product} of two state vectors.
Finally, the \emph{von Neumann entropy} of this density operator is
defined as
\begin{equation}
  H_N(\rho_\graph) := -\trace\left(\rho_\graph \log \rho_\graph\right),
  \label{eq:von Neumann entropy}
\end{equation}
where $\trace$ refers to the trace of the resulting matrix. In practice,
\citet{Bai13} note that the von Neumann entropy is calculated from
a spectral decomposition of the density matrix, but for notational
simplicity, we refrain from doing so.
This leads to the definition of the quantum Jensen--Shannon kernel for graphs.
%
\begin{defn}[Quantum Jensen--Shannon kernel]
  Let $\graph$ and $\graph'$ be two undirected graphs, and $s\in\natural$
  be the maximum number of time steps. With the respective density operators
  $\rho_{\graph}$ and $\rho_{\graph'}$, the \emph{Jensen--Shannon divergence}
  between $\graph$ and $\graph'$ is defined as
  \begin{equation}
    \jsd(\graph, \graph') := H_N\left(\frac{\rho_{\graph} + \rho_{\graph'}}{2}\right)
                               - \frac{1}{2} H_N\left(\rho_{\graph}\right)
                               - \frac{1}{2} H_N\left(\rho_{\graph'}\right),
  \end{equation}
  where $H_N(\cdot)$ refers to the von~Neumann entropy as defined in
  Equation~\ref{eq:von Neumann entropy}.
  From this, the \emph{quantum Jensen--Shannon kernel} is defined by an
  appropriately-scaled exponential expression, \ie\
  \begin{equation}
    \kernelQJS(\graph, \graph') := \exp\left(-\lambda \jsd\left(\graph, \graph'\right)\right),
  \end{equation}
  where $0 < \lambda < 1$ is a decay factor that ensures that large values do not tend
  to dominate the kernel value.
\end{defn}
%
The computational complexity of the kernel is dominated by the
calculation of the eigendecomposition of the Laplacian, which has
a complexity of the order $\landau{n^3}$, where $n$ is the maximum number
of vertices in the two graphs for which the kernel is calculated.

Several types or variants of this kernel exist~\citep{Bai14, Bai15}.
For example, an extension~\citep{Bai15} employs an additional matching
or assignment procedure~(see also Section~\ref{sec:Optimal assignment
kernels} for more details). In a more general setting, quantum walks
based on graphs have also demonstrated favourable performance in other
application domains, such as edge detection in images~\citep{Curado15}.
\begin{figure}[h!]
    \centering
    \includegraphics[width=\textwidth]{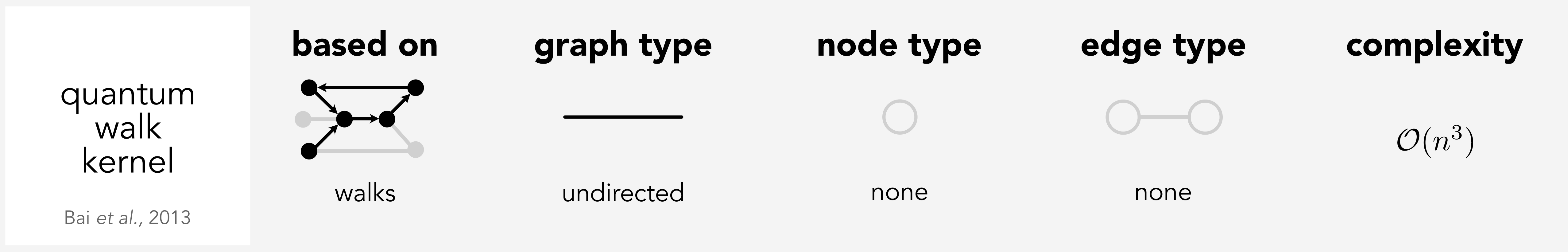}
\end{figure}

\subsection{Graph kernels based on iterative label refinement}\label{sec:Graph kernels based on iterative label refinement}

An important advance in the field of graph kernels occurred in 2009, when~\citep{Shervashidze09b, Hido09} concurrently introduced two graph kernels based on the same underlying idea. Most previously existing approaches defined graph similarity directly in terms of pairwise comparisons between a large number of graph substructures. Instead, these methods proposed to first substitute the original node attributes in each graph by a new set of node attributes that also incorporate topological information about the $k$-hop neighbourhood of each node. Then, one can subsequently apply a simple graph kernel based only on nodes to the modified graphs to obtain a computationally efficient graph kernel that nonetheless can make use of fine-grained information about graph topology. 

More precisely, these approaches recursively \emph{refine the node labels} by applying local transformations of the form
\[
 l_{\text{V}}^{(\text{new})}(v) = f\left(l_{\text{V}}^{(\text{old})}(v), g\left(\left\{ l_{\text{V}}^{(\text{old})}(v') \mid v' \in \mathcal{N}(v)\right\}\right)\right),
 \] 
 where $\mathcal{N}(v)$ denotes the set of nodes adjacent to $v$, $g$ a permutation-invariant function and $f$ an arbitrary function. As we shall see, these operations can be defined to be very efficient, such that computing $l_{\text{V}}^{(\text{new})}(v)$ for all nodes in a graph can be done in only $\landau{m}$ time for a graph with $m$ edges. Applying these transformations in succession results in a sequence of modified graphs, each of which has node attributes that aggregate information about increasingly large $k$-hop neighbourhoods. This general idea can give rise to a multitude of distinct graph kernels, depending on (i) the specific form of the functions $f$ and $g$; (ii) which kernels are used to compare the resulting modified graphs and (iii) how the graph similarities at multiple scales ($k$-hop neighbourhoods), captured by the different modified graphs obtained during the sequence of label refinement operations, are aggregated into a single similarity value.

The success of this family of methods has been three-fold. Firstly, they
lead to extremely efficient graph kernels, often orders of magnitude
faster than previously existing methods for sufficiently large graphs.
Secondly, as will be seen in Chapter~\ref{chap:Experiments}, approaches
based on such iterative label refinement schemes achieve
state-of-the-art performance in many different supervised learning
tasks, often outperforming kernels based on other substructures by
a significant margin. Finally, as will be discussed in
Chapter~\ref{chap:Future}, these graph kernels have strong ties to more
recent approaches, being \emph{de facto} the precursors of most modern
\emph{graph neural networks}. 

In this section, we will study this family of graph kernels by first introducing the two original methods, as proposed by~\citet{Shervashidze09b} and~\citet{Hido09}. Next, we will describe follow-up approaches that aim to further speed-up these kernels as well as to extend them to handle continuous node and edge attributes.

\subsubsection{The Weisfeiler--Lehman kernel framework}\label{sec:Weisfeiler--Lehman kernel}

\begin{figure}[h]

\begin{subfigure}{0.3\textwidth}
    \centering
    \includegraphics[width=.9\linewidth]{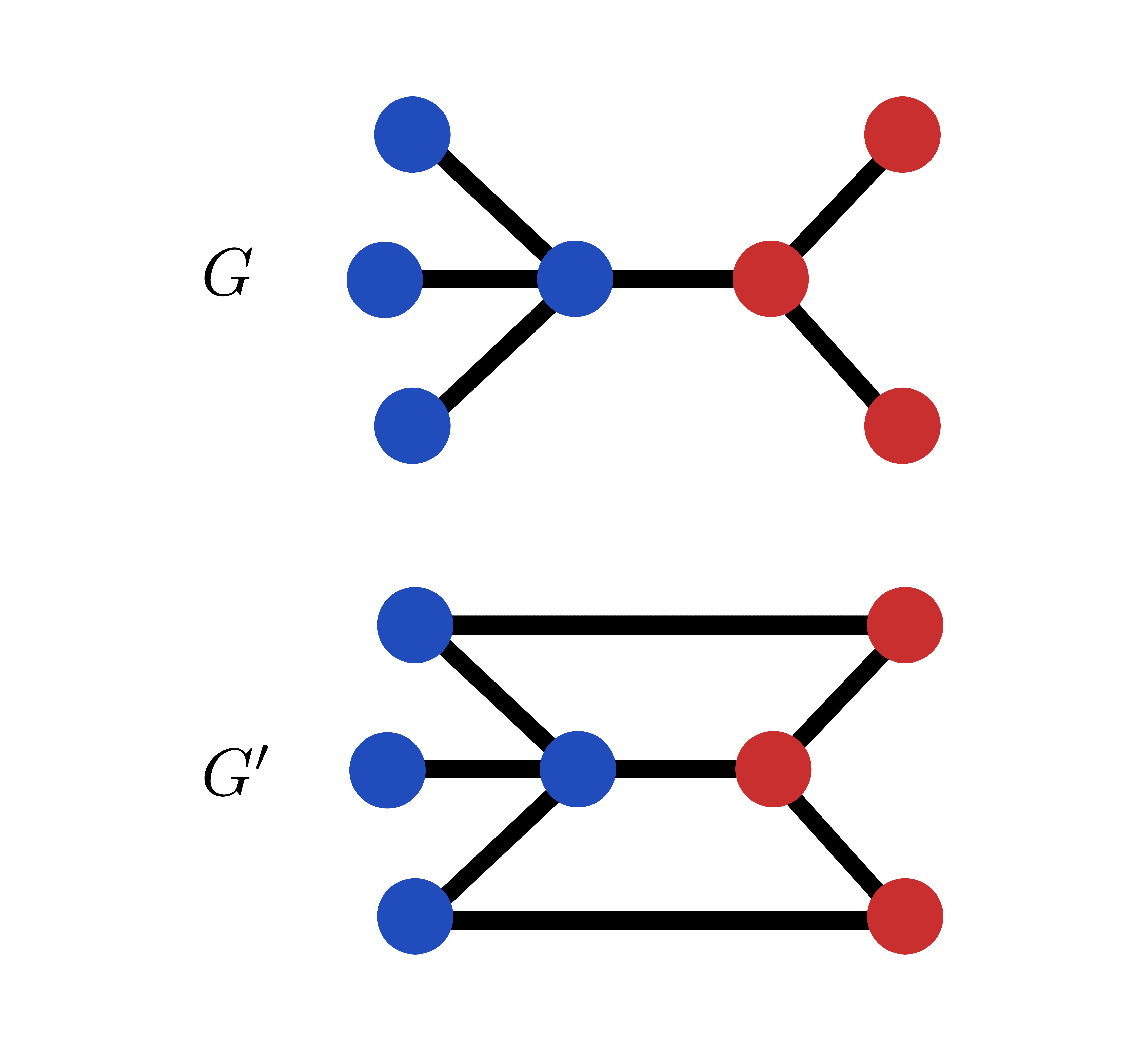} 
    \caption{$G$ and $G^{\prime}$, \ie\ $h=0$.}
    \label{fig:wl0}
\end{subfigure}
\hfill
\begin{subfigure}{0.3\textwidth}
    \centering
    \includegraphics[width=.9\linewidth]{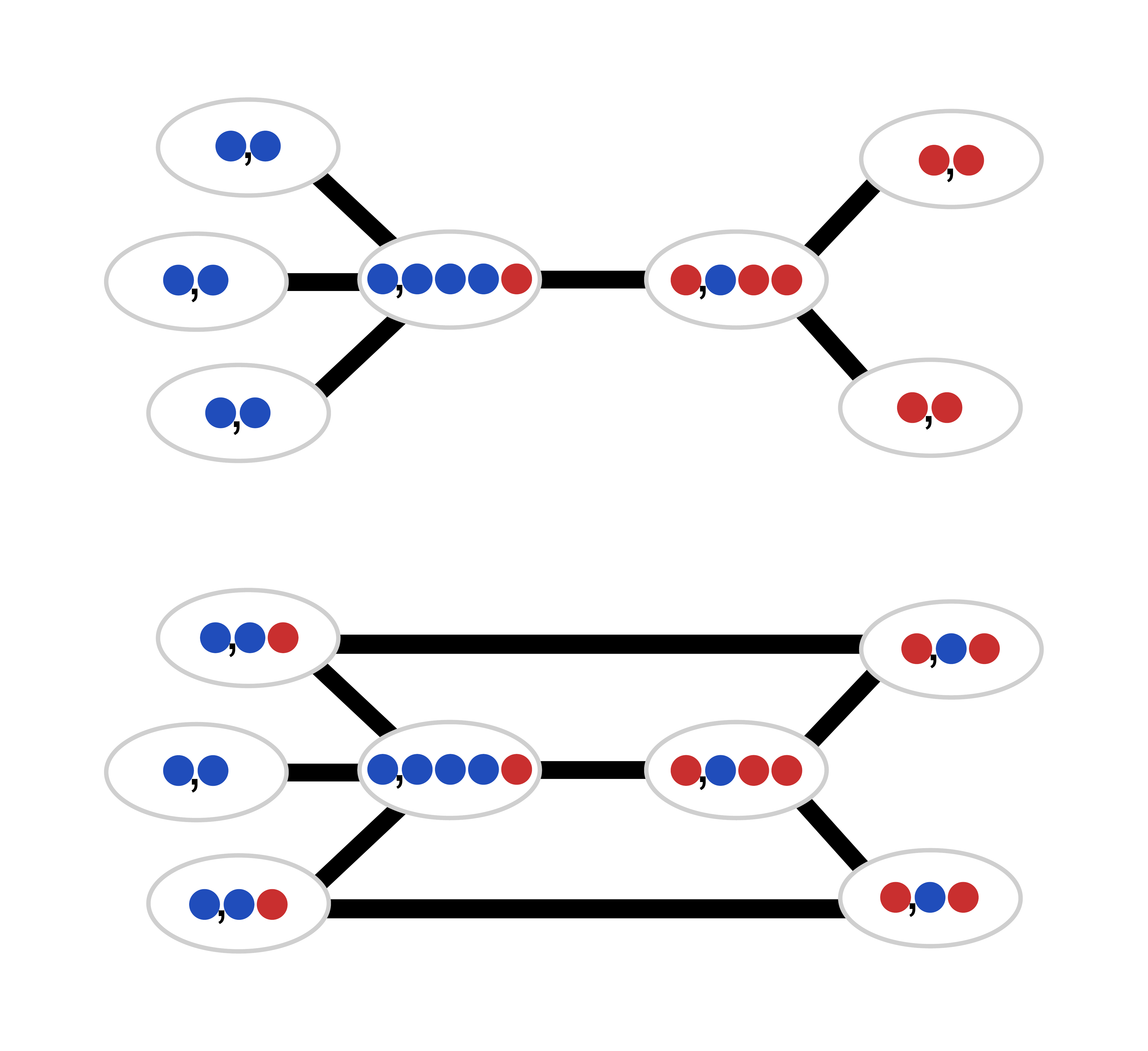}
    \caption{The sorted multisets.}
    \label{fig:wl_multiset}
\end{subfigure}
\hfill
\begin{subfigure}{0.3\textwidth}
    \centering
    \includegraphics[width=.9\linewidth
    ]{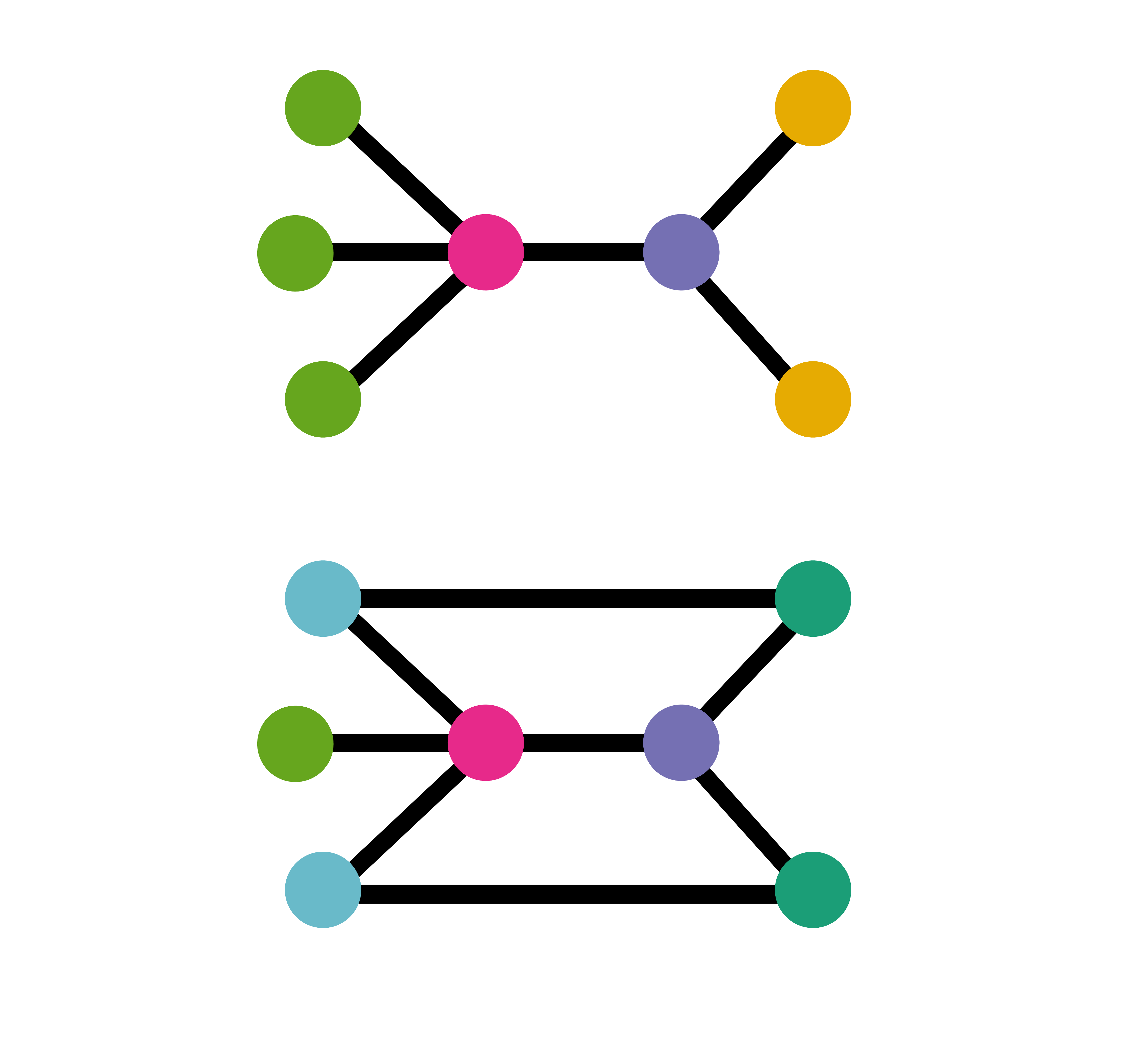}
    \caption{$h=1$.}
    \label{fig:wl1}
\end{subfigure}

 \par\bigskip
\begin{subfigure}{0.35\textwidth}
    \centering
    \includegraphics[height=1.5cm]{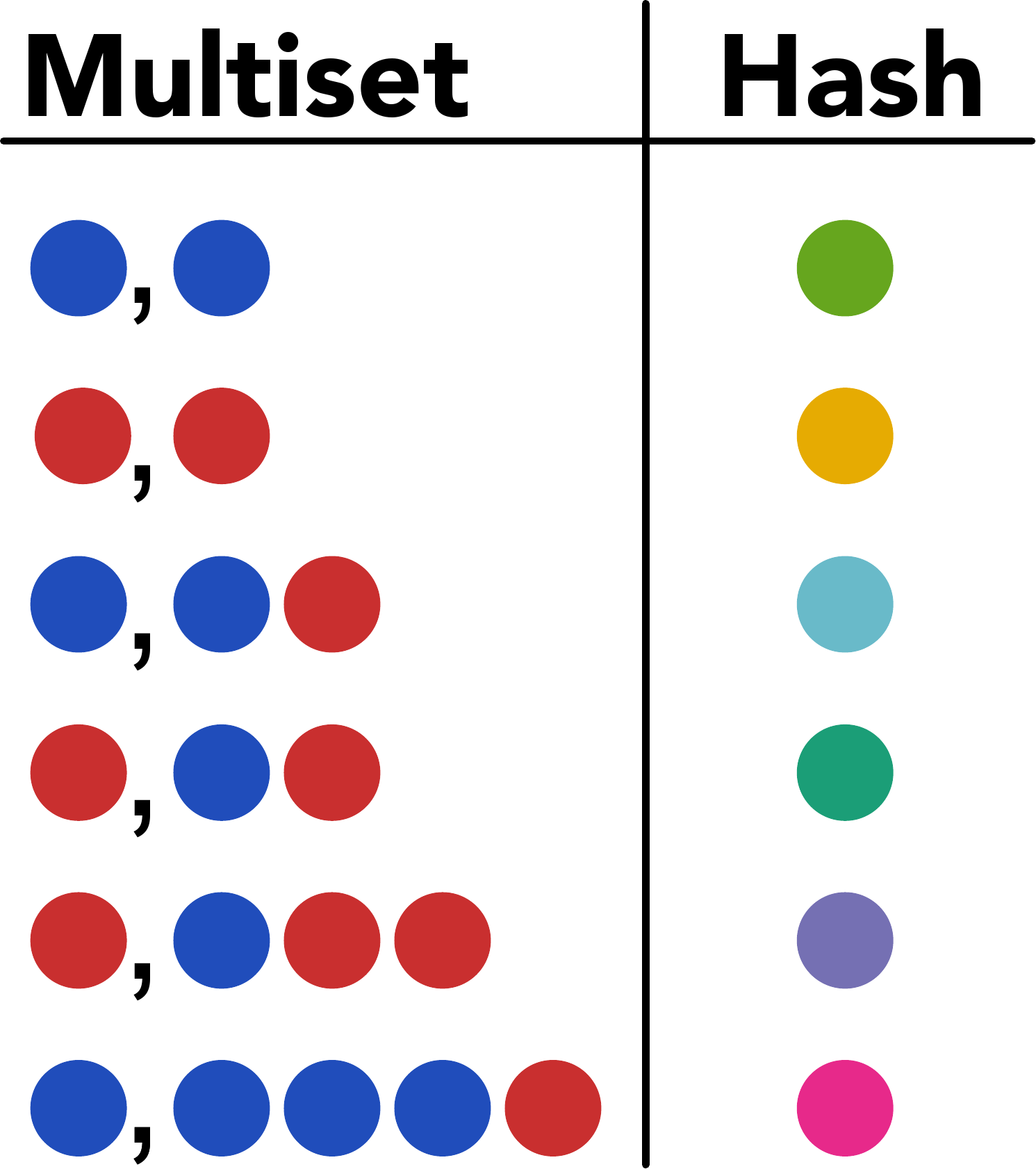}
    \caption{The hash function.}
    \label{fig:wl_hash}
\end{subfigure}
\hfill
\begin{subfigure}{0.65\textwidth}
    \centering
    \includegraphics[width=0.9\linewidth]{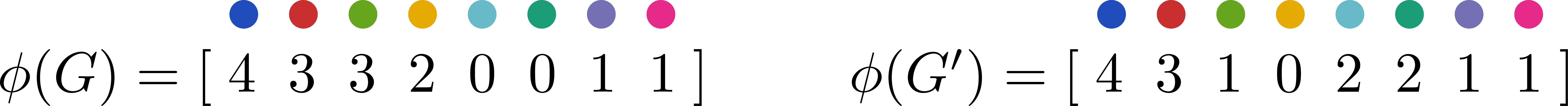}
    \caption{The feature vector representations of $G$ and $G^{\prime}$.}
    \label{fig:wlk_map}
\end{subfigure}

\caption{An example of the Weisfeiler--Lehman kernel where $h=1$. The
  nodes in graphs $G$ and $G^{\prime}$ are relabeled using a hash
  function of the multiset of the given node's label and the sorted labels of its neighbours. The expanded alphabet from this hashing procedure $\Sigma_{WL}^{(1)}$ gives rise to a feature vector representation of the graphs which counts the instances of each label. We obtain our kernel value as $\textrm{k}_\textrm{WL}^{(1)}(G, G^{\prime}) = \langle \phi(G), \phi(G^{\prime}) \rangle_{\mathcal{H}} = 30$.
}
\label{fig:wlkernel}
\end{figure}

The first of the two original approaches that pioneered the use of iterative label refinement operations to derive graph kernels~\citep{Shervashidze09b, Shervashidze11} was inspired by the Weisfeiler--Lehman test for isomorphism~\citep{Weisfeiler68}, which gave it its name.
A central component of this test
is the concept of a \emph{multiset}. Informally put, a multiset is
a generalization of a set that permits the same element to be added
multiple times. By this definition, a set can be considered as a multiset, in which all elements have a count of $1$. For a formal definition, please
see~\citet{Blizard88}.
Briefly put, the Weisfeiler--Lehman test now uses two undirected graphs $\graph
= \left(\vertices, \edges\right)$ and  $\graph' = \left(\vertices',
\edges'\right)$ with a set of node labels from the same alphabet and
repeatedly augments node labels by the sorted multiset of labels of the
neighbours of a vertex. The augmented label is subsequently compressed,
and the process is repeated until the label multisets of the two graphs are
different~(indicating that the graphs cannot be isomorphic), or until
the maximum number of iterations $h_{\text{max}} := \max\left(\left|\vertices\right|, \left|\vertices'\right|\right)$
has been reached. This procedure is guaranteed to produce identical
sequences for isomorphic graphs.  While it remains possible for two
non-isomorphic graphs to also have identical sequences, if the generated
label sequences are equal, the two graphs are isomorphic with high
probability~\citep{Babai79}.

It was observed by \citet{Shervashidze09b} that the Weisfeiler--Lehman
test can be seen to give rise to an iteration that creates subsequent refinements of
vertex labels $\Vlabel\left(\vertex\right)_{\text{WL}}^{(h)}$ for
a vertex $\vertex$ and $h \in \natural$. The base case for $h = 0$ of
this iteration uses the original labels of the graph, so that
$\Vlabel\left(\vertex\right)_{\text{WL}}^{(0)} :=
\Vlabel\left(\vertex\right)$.
For $h > 0$, each vertex is assigned a new label that uniquely identifies the tuple formed by the current Weisfeiler--Lehman label of the vertex, $\Vlabel\left(\vertex\right)_{\text{WL}}^{(h)}$, and the multiset of the current Weisfeiler--Lehman labels of its neighbours, $\left\{
    \Vlabel\left(\vertex'\right)_{\text{WL}}^{(h)} \mid \vertex' \in \neighbourhood\left(v\right)
  \right\}$, \ie\, the updates take the form
\begin{equation}
  \Vlabel\left(\vertex\right)_{\text{WL}}^{(h+1)} := f\left( \left(\Vlabel\left(\vertex\right)_{\text{WL}}^{(h)}, \left\{
    \Vlabel\left(\vertex'\right)_{\text{WL}}^{(h)} \mid \vertex' \in \neighbourhood\left(v\right)
  \right\} \right)\right),
  \label{eq:WL relabelling}
\end{equation}
where $f(\cdot)$ is a hashing function that compresses the tuple into a \emph{single} integer-valued label. Crucially, as we shall see, what sets this method apart from the approach concurrently proposed by~\citet{Hido09} is the use of \emph{perfect hashing} for $f(\cdot)$, following the Weisfeiler--Lehman test for graph isomorphism. This leads to a highly expressive representation of topological information, able to differentiate neighbourhoods differing by a single node. To accomplish this, \citet{Shervashidze11} proposed an approach based on Counting Sort and Radix Sort, that manages to keep the time complexity linear with respect to the number of edges in the graph.
Applying this relabelling scheme recursively gives rise to a sequence of \emph{Weisfeiler--Lehman graphs}.

\begin{defn}[Weisfeiler--Lehman sequence]
  Given an undirected graph $\graph = \left(\vertices, \edges\right)$ with a
  label function $\Vlabel\left(\cdot\right)$ and $h\in\natural$, the Weisfeiler--Lehman
  relabelling operation as described in Eq.~\ref{eq:WL relabelling} results in a
  sequence of graphs $(G_0, G_1, \dots, G_h)$, where
  \begin{equation}
    G_h := \left(\vertices, \edges, \Vlabel\left(\cdot\right)_{\text{WL}}^{(h)}\right)
  \end{equation}
  and each graph only differs in terms of its labels. This sequence is referred
  to as the \emph{Weisfeiler--Lehman sequence}.
\end{defn}
%
The sequence of graphs can be seen as a multiscale description of its
neighbourhoods. By comparing them with a suitable kernel function, it
is possible to obtain a kernel for the graph itself~\citep{Shervashidze11}.
%
\begin{defn}[Weisfeiler--Lehman kernel]
  Let $\graph$ and $\graph'$ be two undirected graphs with node labels
  defined over the same alphabet.
  Given a well-defined base kernel $\basekernel$ for graphs and
  $h\in\natural$, the \emph{Weisfeiler--Lehman kernel} is defined as
  \begin{equation}
    \kernelWL^{(h)}\left(\graph, \graph'\right) := \sum_{i=0}^{h} \basekernel\left(\graph_i, \graph_i'\right),
  \end{equation}
  where $\graph_i$ and $\graph_i'$ refer to the $i$th graph of the
  Weisfeiler--Lehman sequences of $\graph$ and $\graph'$, respectively.
  \label{def:Weisfeiler--Lehman kernel}
\end{defn}
%
This framework gives rise to a multitude of kernels~\citep{Shervashidze11}; we only
discuss the \emph{subtree kernel} here, which makes direct use of the label sequence.
Letting $\Sigma_{\text{WL}}^{(h)} := \{\sigma_1^{(h)}, \sigma_2^{(h)}, \dots\}$
refer to the alphabet of \emph{all} compressed labels in step $h$ of the Weisfeiler--Lehman
relabelling operation, the subtree kernel uses a \emph{count} function $\fcount\left(\cdot\right)$
such that $\fcount\left(\graph, \sigma_i^{(h)}\right)$ is the number of occurrences of the label
$\sigma_i^{(h)}$ in $\graph$. Thus, a graph $\graph$ is assigned a feature vector
\begin{equation}
  \phi\left(\graph\right) = \left(
    \fcount\left(\graph, \sigma_1^{(0)}\right),
    \dots
    \fcount\left(\graph, \sigma_i^{(1)}\right),
    \dots,
    \fcount\left(\graph, \sigma_j^{(h)}\right),
    \dots
  \right)
\end{equation}
for a total of $h$ steps of the Weisfeiler--Lehman iteration, and the
\emph{Weisfeiler--Lehman subtree kernel} is defined as the \emph{inner
product} of these features vectors, \ie\,
\begin{equation}
  \basekernel_{\text{subtree}}\left(\graph, \graph'\right) :=
  \left\langle
    \phi\left(\graph\right), \phi\left(\graph'\right)
  \right\rangle,
\end{equation}
where $\graph$ and $\graph'$ are undirected labelled graphs as described above.
This sort of propagation scheme is simple but extremely powerful, as it
automatically makes it possible to represent the graph at coarse
scales~(small values of $h$) or fine scales~(large values of $h$).
Multiple variants of this kernel exist. For instance, it is simple to
extend this to incorporate edge labels as well as node labels by using
the triple of ($\Vlabel(\vertex)$, $\Vlabel(\vertex_i)$,
$\Elabel((\vertex, \vertex_i))$) to represent a given neighbour $v_i$
of node $v$, rather than using only $\Vlabel(\vertex_i)$, as is done in
the base case implementation of the kernel. Nevertheless, the
aforementioned subtree kernel is the most common one, and is depicted in
Figure~\ref{fig:wlkernel}.

The complexity of the Weisfeiler--Lehman kernel computation depends on the
selected base kernel. For the subtree kernel, \citet{Shervashidze11}
show that, given a perfect hashing function, the computation of a full
kernel matrix for $N$ graphs and $h$ steps of the Weisfeiler--Lehman iteration
has a complexity of $\landau{Nhm + N^2hn}$,
where $n$ is the maximum number of vertices of a graph and $m$ is the
maximum number of edges. More generally, the complexity of $h$
Weisfeiler--Lehman relabelling iterations~(not accounting for any kernel
calculations) is $\landau{hm}$. 
Compared to previously existing, popular graph kernels, such as the random walk kernel, the shortest-paths kernel or the graphlet kernel, the availability of an approach whose runtime scales only linearly with the number of nodes and edges while simultaneously achieving state-of-the-art performance in a variety of statistical learning tasks constituted an important step forward in the field.
\begin{figure}[h!]
    \centering
    \includegraphics[width=\textwidth]{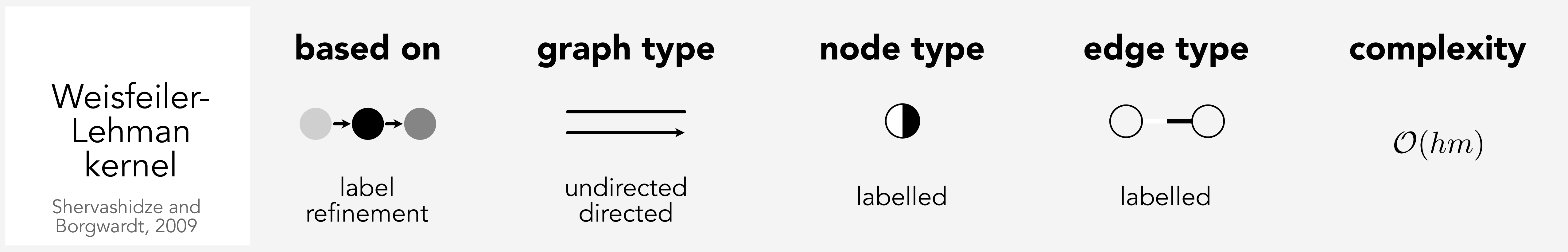}
\end{figure}
\subsubsection{Neighbourhood hash kernel}\label{sec:Neighbourhood hash kernel}

\begin{figure}[h]
\begin{subfigure}{0.3\textwidth}
    \centering
    \includegraphics[width=.9\linewidth]{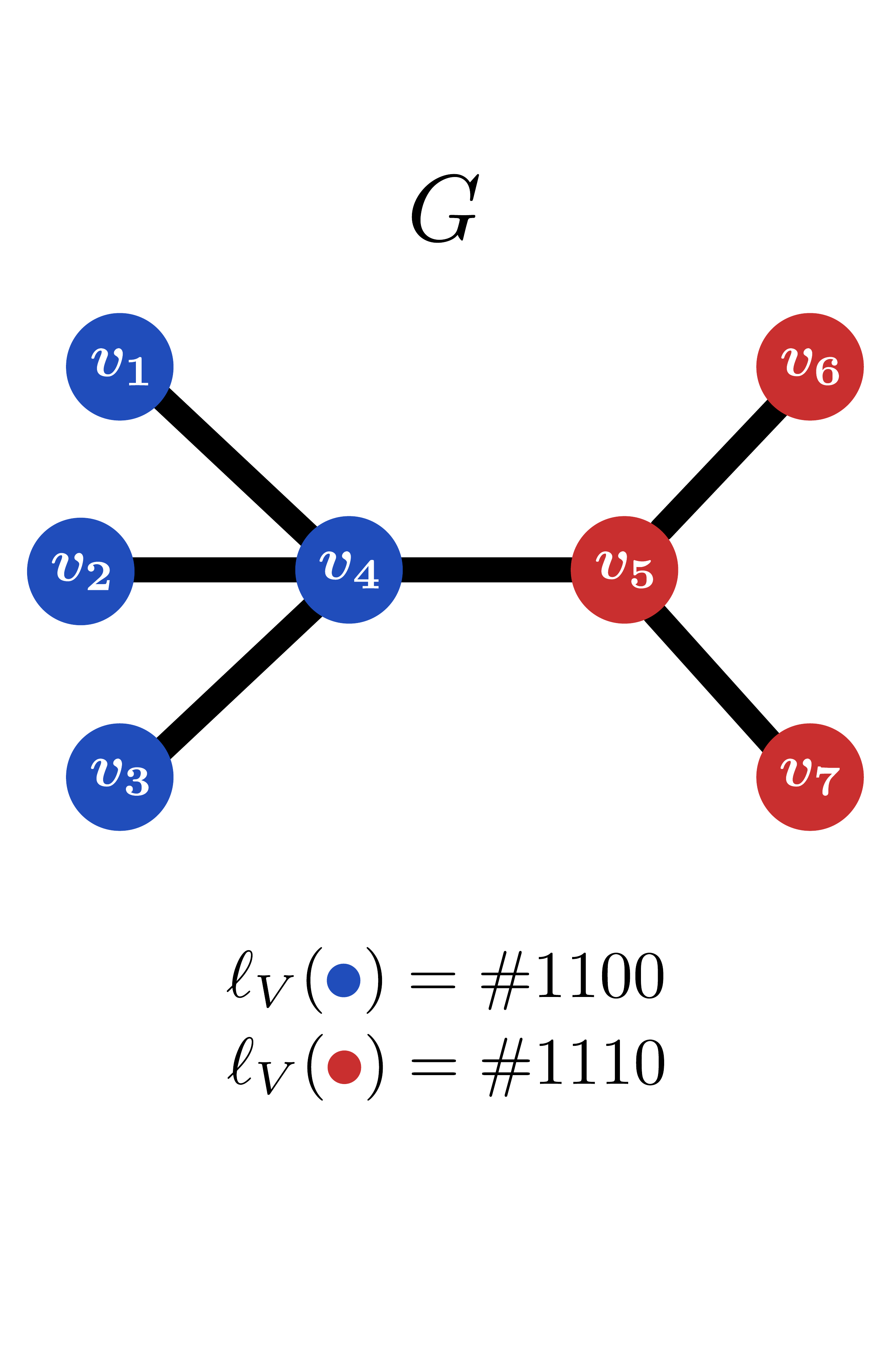} 
    \caption{$\graph$ and initial node label hash values.}
    \label{fig:nh0}
\end{subfigure}
\hfill
\begin{subfigure}{0.6\textwidth}
    \centering
    \includegraphics[width=.9\linewidth]{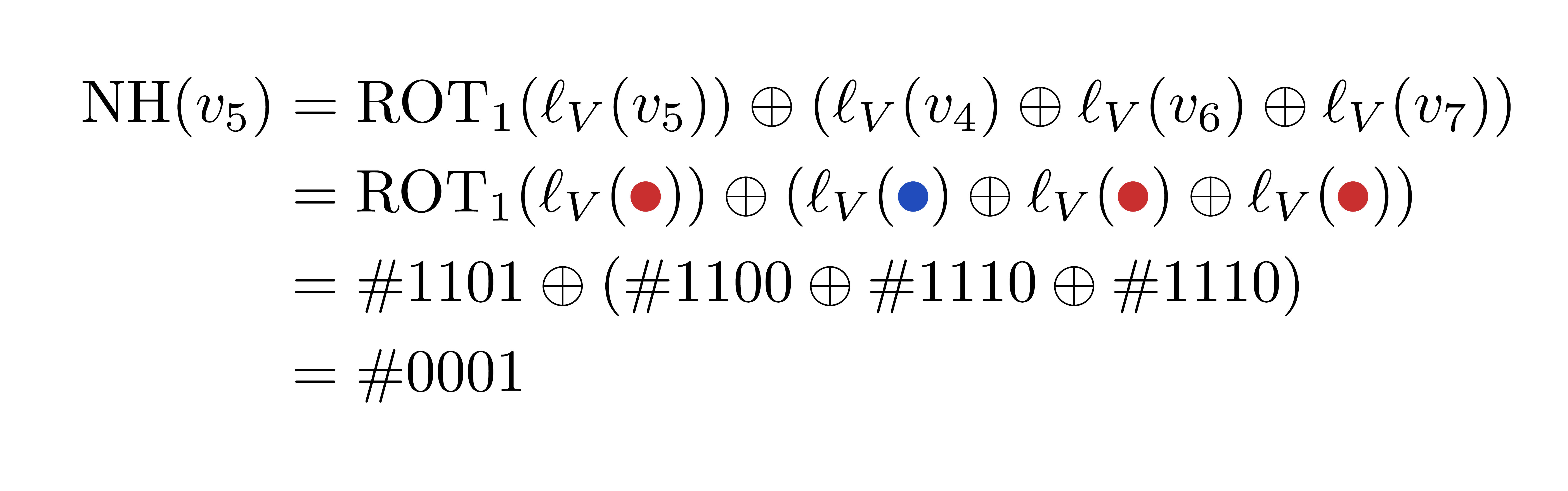}
    \caption{The simple neighbourhood hash label for $v_5$.}
    \label{fig:nh_nh}
    \vspace{2ex}

    \includegraphics[width=.9\linewidth
    ]{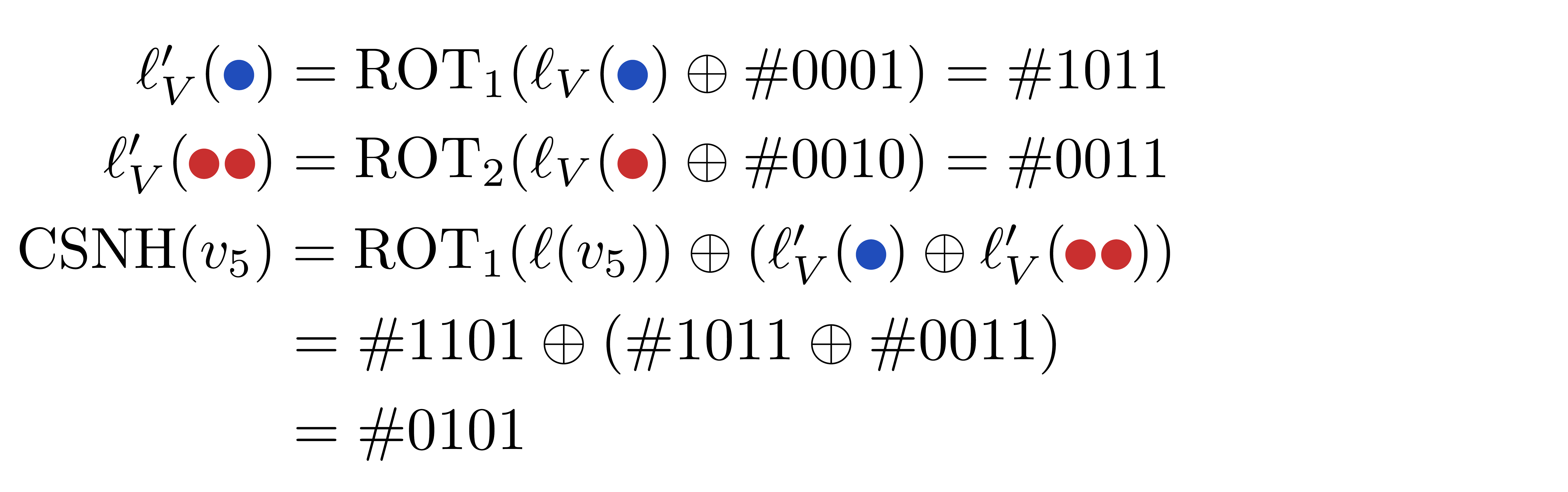}
    \caption{The count-sensitive hash function for $v_5$.}
    \label{fig:nh_csnh}
\end{subfigure}
\caption{An example relabeling of the node $v_5$ in a graph \graph\ with
  initial node label function (\subref{fig:nh0}),
  with nodes indexed as $v_1,\ldots,v_7$, using the simple
  neighbourhood hash function (\subref{fig:nh_nh}) and the count-sensitive hash
function (\subref{fig:nh_csnh}). After each node has been hashed, the
vector representation is a count vector of each unique hash label. The
neighbourhood hash kernel is then calculated according to Eq.~\ref{eq:neighbourhood hash kernel}.}
\label{fig:nh_image}
\end{figure}

In parallel, \citet{Hido09} developed the \emph{neighbourhood hash kernel} with the same goal: to obtain a highly computationally efficient graph kernel that accounts for information about the graph topology and achieves good predictive performance in real-world problems.

Conceptually, the neighbourhood hash kernel, shown in
Figure~\ref{fig:nh_image}, and the Weisfeiler--Lehman
framework converged to similar ideas. Both are based on iteratively
refining node labels, combining information about the current node label
of a vertex and those of its neighbours, to create a sequence of graphs
which can then be compared by means of simple criteria. Moreover, both
approaches use hashing-based schemes to implement this refinement step.
However, while the Weisfeiler-\-Lehman framework relies on perfect
hashing, \citet{Hido09} instead use simpler hashing techniques based on
binary arithmetic. While both approaches have the same asymptotic
scaling with respect to the size of the graphs, the neighbourhood hash
kernel has better constant factors thanks to its simpler hashing
function, being slightly faster in practice, and is also more
memory-efficient. However, this comes at the cost of the possibility of
having accidental hashing \emph{collisions}, which can limit the
expressivity of the resulting feature map if nodes that are
rather different get hashed to the same value.
Notice that collisions are not problematic per se. In fact, the primary
lesson of the Weisfeiler--Lehman iteration is that graph isomorphism is
a perspective that is too restrictive---for most applications, graph
\emph{similarity} is much more relevant. Collisions that result in
similar graphs being hashed together are therefore less problematic
than collisions that result in highly dissimilar graphs being assigned
the same hash.

The neighbourhood hash kernel assumes that node labels can be embedded
into binary strings~(``bit arrays'') of a pre-determined length $s$.
Thus, for the remainder of this section, we assume that each node label
$l$ is represented by a sequence of bits $l = \{b_1, b_2, \dots, b_s\}$,
where $b_i \in \{0, 1\}$.  Shorter strings (small values for $s$) will
lead to faster hashing but increase the probability of accidental collisions occurring. The hashing schemes in \citet{Hido09} are based on this binary representation of node labels, making use of \texttt{XOR} operations and bit rotations~(also known as \emph{shifts}). 

The \texttt{XOR} operation between
two bits $b_i, b_j$ is defined as
\begin{equation}
  b_i \oplus b_j = \begin{cases}
                     1 & \text{if $b_i \neq b_j$}\\
                     0 & \text{otherwise}
                   \end{cases}
\end{equation}
and can be extended to bit strings of the same lengths. A key aspect of the \texttt{XOR} operation is that it is associative and commutative, making the output permutation invariant. Consequently, it is a suitable function to hash multisets of labels of neighbouring nodes.

The bit rotation function circularly shifts all bits by a pre-defined amount, \ie\
\begin{equation}
  \mathrm{ROT}_k(b_1, \dots, b_l) := \{b_{k+1}, b_{k+2}, \dots, b_l, b_1, \dots, b_{k}\},
\end{equation}
which does not change the size of the bit string. In this context, bit-rotations are useful to treat certain elements \emph{asymmetrically} during the hashing process. This allows, for example, hashing the tuple formed by the current node label of a vertex and the multiset of labels of neighbouring nodes.

Building on these ideas, the first of the hashing schemes in \citet{Hido09} is defined as follows.

\begin{defn}[Simple neighbourhood hash function]
  Given a vertex $\vertex \in \vertices$ and its neighbours $\vertex_1,
  \dots, \vertex_k$, the \emph{simple neighbourhood} hash is calculated
  as
  \begin{equation}
    \mathrm{NH}\left(\vertex\right) := \mathrm{ROT}_1\left(\Vlabel(\vertex)\right) \oplus \left( \Vlabel(\vertex_1) \oplus \dots \oplus \Vlabel(\vertex_k) \right).
    \label{eq:NH}
  \end{equation}
  The label of the initial vertex is shifted by one bit in order to make
  it distinct from the other vertices. This hash function can now be applied multiple
  times, and each iteration will thus use information from higher-order
  neighbourhoods, since every vertex receives propagated information
  from its direct neighbours.
\end{defn}
\citet{Hido09} note that this hashing scheme is severely limited in its ability to handle repeated occurrences of the same node label in the multisets. Indeed, the \texttt{XOR} operation can be seen to compute the bitwise \emph{parity} of the bit strings being hashed. Thus, the only information retained about the number of occurrences of each node label in the multisets is whether the count is even or odd. To alleviate this limitation, \citet{Hido09} proposed a second hashing scheme that makes explicit use of the label counts using bit rotations.

\begin{defn}[Count-sensitive neighbourhood hash function]
  If a specific label $\Vlabel\left(\vertex_j\right)$ occurs $l$ times in
  the neighbourhood, let
  $\Vlabel'\left(\vertex_j\right) := \mathrm{ROT}_l\left(\Vlabel\left(\vertex_j\right) \oplus l\right)$
  denote its transformed version: by shifting a label by $l$ bits after
  calculating its $\mathrm{XOR}$, hash values are unique and only depend
  on the number of occurrences of a label. This leads to the
  \emph{count-sensitive neighbourhood hash function}
  \begin{equation}
    \mathrm{CSNH}\left(\vertex\right) := \mathrm{ROT}_1\left(\Vlabel(\vertex)\right) \oplus \left( \Vlabel'(\vertex_1) \oplus \dots \oplus \Vlabel'(\vertex_k) \right).
    \label{eq:CSNH}
  \end{equation}
\end{defn}

The \emph{count-sensitive neighbourhood hash function}, while still prone to collisions, manages to avoid some of the main pitfalls of the previous hashing scheme. However, to obtain the label counts in each multiset, a sorting operation must be applied, which reduces the computational advantage with respect to the Weisfeiler--Lehman framework.

Based on either of the two hash functions, the neighbourhood hash kernel can now
be defined as the overlap between the label sets of two graphs.
\begin{defn}[Neighbourhood hash kernel]
  Given two labelled \linebreak graphs $\graph$ and $\graph'$, select a hash
  function as described above and calculate its hashed labels according
  to Eq.~\ref{eq:NH} or Eq.~\ref{eq:CSNH} above. This results in two
  sets of labels $\mathcal{L}$ and $\mathcal{L}'$. The \emph{neighbourhood hash kernel}
  is the \emph{Tanimoto coefficient}~\citep[Chapter~2]{Tan19} of
  $\mathcal{L}$ and $\mathcal{L}'$, \ie\
  \begin{equation}
    \kernelNH(\graph, \graph') := \frac{c}{n + n' - c},
    \label{eq:neighbourhood hash kernel}
  \end{equation}
  where $c$ is the number of matching labels between $\mathcal{L}$ and
  $\mathcal{L}'$, and
  $n$, $n'$ denotes the number of vertices in $\graph$ and $\graph'$,
  respectively.
\end{defn}
Ignoring the fact that the quantities involved are bit strings, the
previous equation can also be seen as an instance of the
Weisfeiler--Lehman framework, which takes the linear kernel between two
count vectors instead of an overlap measure.

The asymptotic complexity of the neighbourhood hash kernel with respect
to the size of the graphs is comparable to that of the
Weisfeiler--Lehman framework. However, as discussed above, it offers
certain advantages in terms of runtime; in the best case, hash function
calculations and comparisons can be performed in a single CPU
instruction, respectively. 
%
\begin{figure}[h!]
    \centering
    \includegraphics[width=\textwidth]{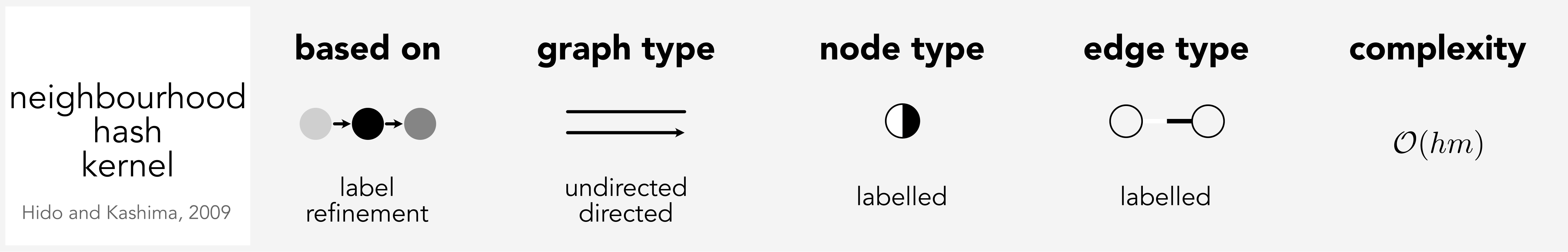}
\end{figure}
\subsubsection{Fast Neighbourhood Subgraph Pairwise Distance Kernel}\label{sec:Fast Neighbourhood Subgraph Pairwise Distance Kernel}

\citet{costa2010fast} present another method that also uses a node's
neighbourhood as a mechanism to propagate information to a node. In
graphs without edge weights, this neighbourhood corresponds to the $k$-hop
neighbourhood of a node $\vertex$ defined in
Chapter~\ref{chap:Background}, where~$k$ is now referred to as the
radius $r$, \ie\ $\neighbourhood^{(r)}(\vertex)$, and accordingly
includes any node $u$ that is reachable from $\vertex$ in at most $r$
hops. In weighted graphs, we can extend this definition using the sum
of edge weights to define the shortest distance to $\vertex$, and
in both cases, require that the distance~$d$ between the two nodes is
less than or equal to the radius~$r$. For a root node $v$ in a graph
$\graph = (\vertices, \edges)$, with a given radius $r$, we can
therefore generalise the definition of its radius $r$ neighbourhood
$\neighbourhood^{(r)}(\vertex)$ as $\{ u \in \vertices \mid d(u, \vertex) \leq r \}$.
The subgraph we will use is the corresponding induced subgraph of $\graph$, denoted by $\graph^{(r)}_{\vertex}$. 

The neighbourhood subgraph pairwise distance kernel does a pairwise
comparison using pairs of subgraphs in each graph, where the two root
nodes $\vertex_1$ and $\vertex_2$ have a shortest path distance to one
another less than or equal to $d$. It then evaluates this for increasing
values of $r$ and $d$. While the original formulation wants to test the
subgraphs for isomorphism, we have mentioned this is not yet
computationally tractable, and so the authors instead settle for a proxy
test.
%
\begin{defn}[Neighbourhood subgraph pairwise distance kernels]
  Let $\graph=(\vertices, \edges)$ and $\graph'=(\vertices', \edges')$
  be two graphs with node and edge labels defined on alphabets
  $\Vlabels$ and $\Elabels$ and labelling functions $\Vlabel$ and
  $\Elabel$ respectively. A subgraph of root node $\vertex$ of radius
  $r$, $\graph^{(r)}_\vertex$ is represented as the lexicographically
  sorted list of updated edge labels in the subgraph,
  $\Elabel^{(r)}(e_{ij})$, which contains the updated node labels
  incident to the edge, as well as its original edge label. That is,
  \begin{equation}
    \Elabel^{(r)}(e_{ij}) = \left( \Vlabel^{(r)}(i), \Vlabel^{(r)}(j),
    \Elabel(e_{ij}) \right), 
  \end{equation}
  and where the updated node labels are again a sorted representation of
  \begin{equation}
    \Vlabel^{(r)}(i) = \left( (d(i, u), \Vlabel(u)) \, \forall u \in
    \vertices(\graph^{(r)}_{\vertex}), d(i, \vertex) \right)
  \end{equation}
  \ie\ the distance of node $i$ to all other nodes in the subgraph
  combined with the label of the other node, and the distance of $i$ to the root node $v$. This
  representation is then hashed to an integer to be used in the Dirac
  delta kernel. For a given radius $r$, the neighbourhood subgraph
  pairwise distance kernel is calculated as
  \begin{equation}
      \kernelNSPD(\graph, \graph') = \sum_{r} \sum_{d} \basekernel_{r,d}(\graph, \graph'),
  \end{equation}
  where $\basekernel_{r,d}(\graph, \graph')$ assesses the similarity of
  the pairs of subgraphs induced by pairs of root nodes whose distance
  is less than or equal to $d$. If we call $\mathfrak{R}$ the set of
  tuples of root nodes ($\vertex_1$, $\vertex_2) \in \vertices$ where
  $d(\vertex_1, \vertex_2) \leq d$, and $\mathfrak{R}'$ is the set of
  tuples ($\vertex'_1$, $\vertex'_2) \in \vertices'$ where
  $d(\vertex'_1, \vertex'_2) \leq d$, and $\text{k}_{\delta}$ is the
  Dirac delta kernel, this leads to the final formulation of
 \begin{equation}
    \basekernel_{r,d}(\graph, \graph') = \sum_{(\vertex_1, \vertex_2) \in \mathfrak{R}} \sum_{(\vertex'_1, \vertex'_2) \in \mathfrak{R}'} \text{k}_{\delta}(\graph^{(r)}_{\vertex_1}, \graph'^{(r)}_{\vertex'_1}) \text{k}_{\delta}(\graph^{(r)}_{\vertex_2}, \graph'^{(r)}_{\vertex'_2}).
 \end{equation}
\end{defn}
%
The complexity of the neighbourhood subgraph pairwise distance kernel is
determined primarily from the repeated process of relabelling the nodes
and edges in the induced subgraphs, resulting in an overall complexity
of $\landau{nn_hm_h\log(m_h)}$, where $n$ is the maximum number of nodes in
a graph, $n_h$ and $m_h$ are the maximum number of nodes and edges
respectively in the induced subgraphs from the root nodes. Since $d$ and
$r$ are typically small, this is reduces to
being linear in the number of nodes in a graph. 
\begin{figure}[h!]
    \centering
    \includegraphics[width=\textwidth]{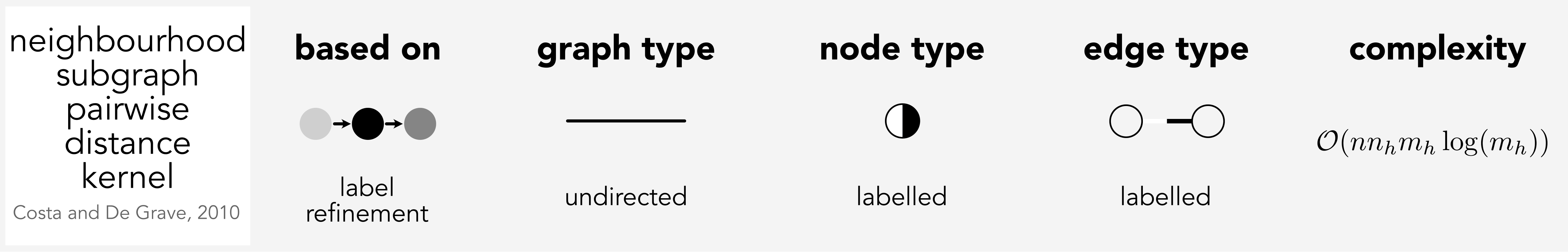}
\end{figure}

\subsubsection{Hadamard code kernel}\label{sec:Hadamard code kernel}

The two hashing schemes proposed by~\citet{Hido09} incur a trade-off
between computational efficiency and expressivity. The simple
neighbourhood hash function is highly efficient, but can also result in
accidental collisions. In contrast, the count-sensitive neighbourhood
hash function circumvents some of the most obvious cases of collisions
in the previous scheme, such as those due to node label duplications,
but at the price of requiring a sorting step that might slow down the
resulting algorithm. 

Motivated by this observation, \citet{Katoka16} proposed a different hashing scheme, aiming to keep the computational efficiency of the simple neighbourhood hash function while being more robust to collisions. The key element of their construct are Hadamard matrices, which we introduce next.

\begin{defn}[Hadamard matrix and Hadamard code]
  A square matrix $H$ of dimension $n$ whose entries are from $\{-1, 1\}$
  is called a \emph{Hadamard matrix} if its rows are mutually orthogonal.
  Hence, given $i \neq j$, for rows $h_i$ and $h_j$ we have $\langle
  h_i, h_j\rangle = 0$, where $\langle \cdot, \cdot \rangle$ denotes the
  usual real-valued dot product. Any row $h_i$ of $H$ is called
  a \emph{Hadamard codeword}~(there is also the notion of \emph{Hadamard
  codebook}, which refers to the union of rows of $H$ and $-H$, but we
  do not require it here).
\end{defn}
%
Hadamard codes can be shown to have interesting mathematical properties,
such as the ability to self-correct; we refer the reader to
\citet[Chapter~17.5]{Arora09} for more details. 

The key idea behind the Hadamard code kernel is to represent each categorical node label in a finite alphabet $\Sigma_{\text{V}}$ by a different Hadamard codeword rather than a bit string. Since distinct codewords are mutually orthogonal, this can be understood as an alternative feature map for a Dirac kernel defined on the set of node labels. However, compared to the more commonly used embeddings given by the canonical basis vectors of $\mathbb{R}^{\vert \Sigma_{\text{V}} \vert}$, Hadamard codewords have favourable computational properties when many such embeddings are to be summed together. In particular, since entries of Hadamard codewords take values $-1$ or $+1$ with equal probability, the expected value of most entries of a sum of Hadamard codewords is $0$. \citet{Katoka16} argue that, due to this property, a small number of bits should suffice to represent them faithfully. In contrast, if one were to use the canonical basis instead, entries would explode as the number of embeddings being summed grows, requiring many bits to be represented accurately. Building on this idea, the relabelling step of the Hadamard code kernel is defined as a simple aggregation of individual Hadamard codewords.

%
\begin{defn}[Hadamard code relabelling]
  Let $\graph = (\vertices, \edges)$ be an undirected graph with a set
  of vertex labels from a common label alphabet $\Vlabels$.
  Let $s := 2^{\left\lceil\log_2\left|\Vlabels\right|\right\rceil}$
  and $H$ be a Hadamard matrix of dimension $s$, such that there is at least one different codeword in $H$ per label in $\Vlabels$.
  For a vertex $\vertex$ with original label $\Vlabel(\vertex) = \sigma_i$, set its initial Hadamard label $\Vlabel(\vertex)_H^{(0)}$ to be $h_i$, the $i$th row of $H$.
  Given $h \in \natural$, the Hadamard code label of $\vertex$ at stage
  $h + 1$ is calculated recursively as
  \begin{equation}
    \Vlabel(\vertex)_H^{(h+1)} = \Vlabel(\vertex)_H^{(h)} + \sum_{\vertex'\in\neighbourhood\left(\vertex\right)}\Vlabel(\vertex')_H^{(h)},
  \end{equation}
  where $\neighbourhood\left(\vertex\right)$ are the vertices adjacent to $\vertex$.
  \label{def:Hadamard code relabelling}
\end{defn}
%

Due to Hadamard codewords being mutually orthogonal,  $\Vlabel(\vertex)_H^{(h)}$ can be understood as an unnormalised, rotated histogram of node label counts. Indeed, the dot product of $\Vlabel(\vertex)_H^{(h)}$ with $h_i$ yields a scaled count of occurrences of node label $\sigma_i$ in a $k$-hop neighbourhood around $\vertex$, whose size depends on $h$. However, as previously mentioned, the entries of these rotated histograms fluctuate symmetrically around zero, rather than growing monotonically as $h$ increases, hence requiring fewer bits to store. In this way, as noted by~\citet{Katoka16}, the Hadamard relabelling step bears some resemblance to the Weisfeiler--Lehman relabelling step, but has some key differences, such as not treating the central vertex of each $k$-hop neighbourhood asymmetrically when computing the new node labels.
The Hadamard code kernel is finally defined as an iterative summation of
Dirac delta kernels on the sequence of relabelled graphs.

\begin{defn}[Hadamard code kernel]
  Given two graphs $\graph = \left(\vertices, \edges\right)$ and
  $\graph' = \left(\vertices', \edges'\right)$ with node labels
  that are defined over the same node label alphabet $\Vlabels$
  and $h \in \natural$, the Hadamard code base kernel at step $h$
  is defined as
  \begin{equation}
    \kernelHC^{(h)}\left(\graph, \graph'\right) := \sum_{\vertex\in\vertices}\sum_{\vertex'\in\vertices'}
      \delta\left(\Vlabel(\vertex)_H^{(h)}, \Vlabel(\vertex')_H^{(h)}\right),
  \end{equation}
  where $\Vlabel(\vertex)_H^{(h)}$ refers to Hadamard code label of
  $\vertex$ at stage $h$ according to Definition~\ref{def:Hadamard code
  relabelling}, and $\delta$ denotes a Dirac delta kernel.
  Building on this, the \emph{Hadamard code kernel} is defined as
  \begin{equation}
    \kernelHC\left(\graph, \graph'\right) := \sum_{h}^{h_{\text{max}}} \kernelHC^{(h)}\left(\graph, \graph'\right),
  \end{equation}
  where $h_{\text{max}}$ denotes the maximum number of iterations of the
  relabelling step.
\end{defn}
%
In terms of computational efficiency, \citet{Katoka16} exploit that the Hadamard
codewords can be stored as bit strings of a fixed length. While collisions,
as for the neighbourhood hash kernel, still occur with a non-zero probability,
the kernel is able to perform well in practice without requiring any sorting
operations. Hence, even though its asymptotic computational complexity is the same 
as that of the neighbourhood hash kernel and the Weisfeiler-Lehman framework, it is 
faster than the version of the neighbourhood hash kernel based on the 
count-sensitive neighbourhood hash function while matching its predictive 
performance.
\begin{figure}[h!]
    \centering
    \includegraphics[width=\textwidth]{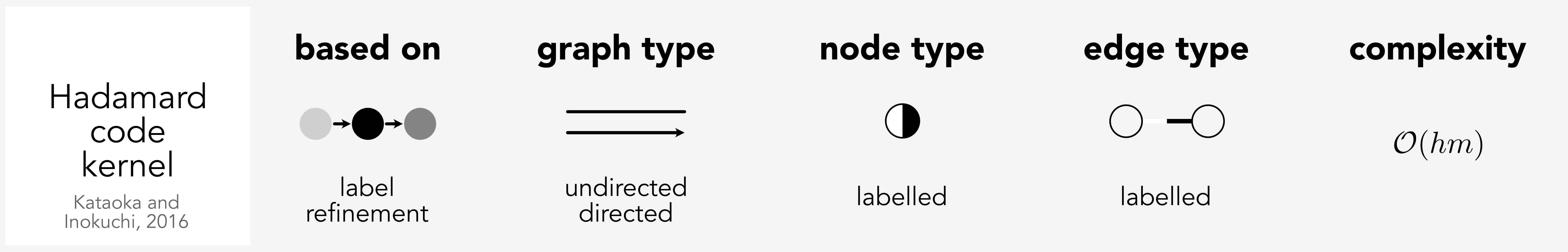}
\end{figure}

\subsubsection{Propagation kernels}\label{sec:Propagation kernels}

\begin{figure}[ht!]

\begin{subfigure}{0.25\textwidth}
    \centering
    \includegraphics[width=.9\linewidth]{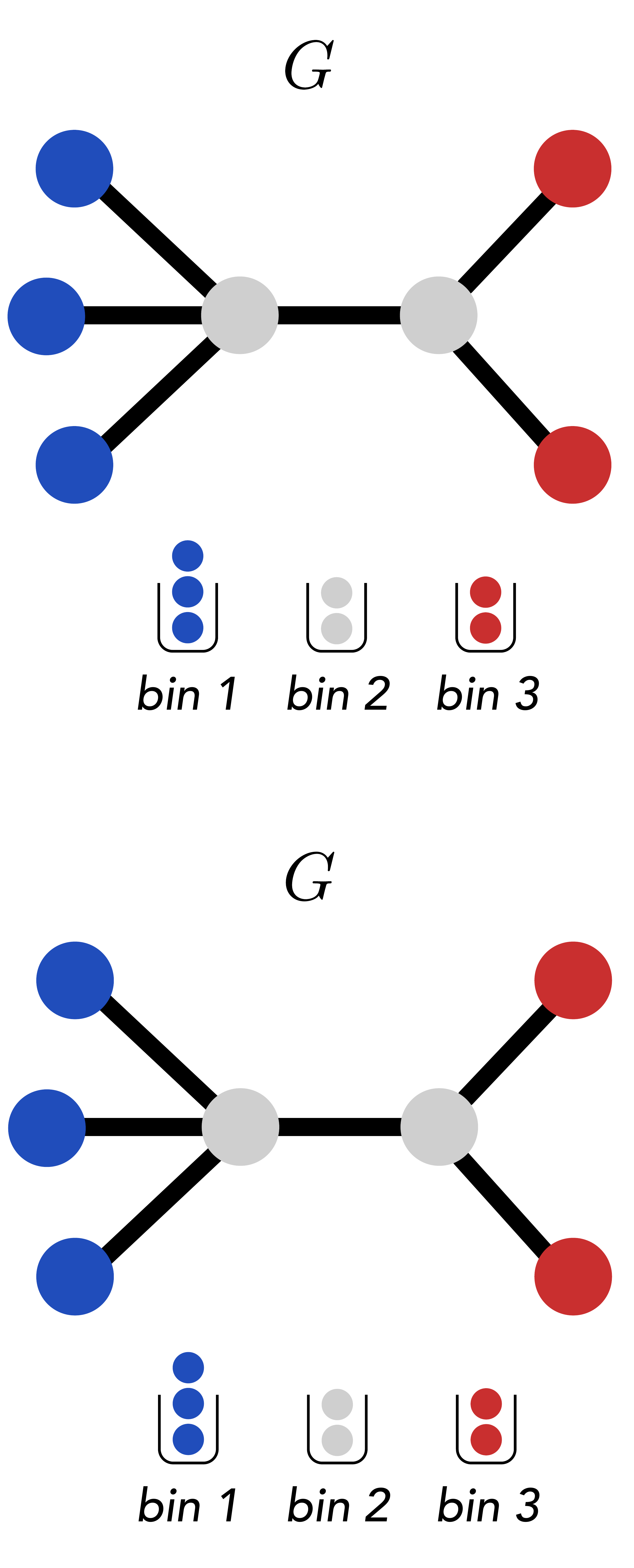} 
    \caption{$G$ and $G^{\prime}$}
    \label{fig:pk1}
\end{subfigure}
\hfill
\begin{subfigure}{0.25\textwidth}
    \centering
    \includegraphics[width=.9\linewidth]{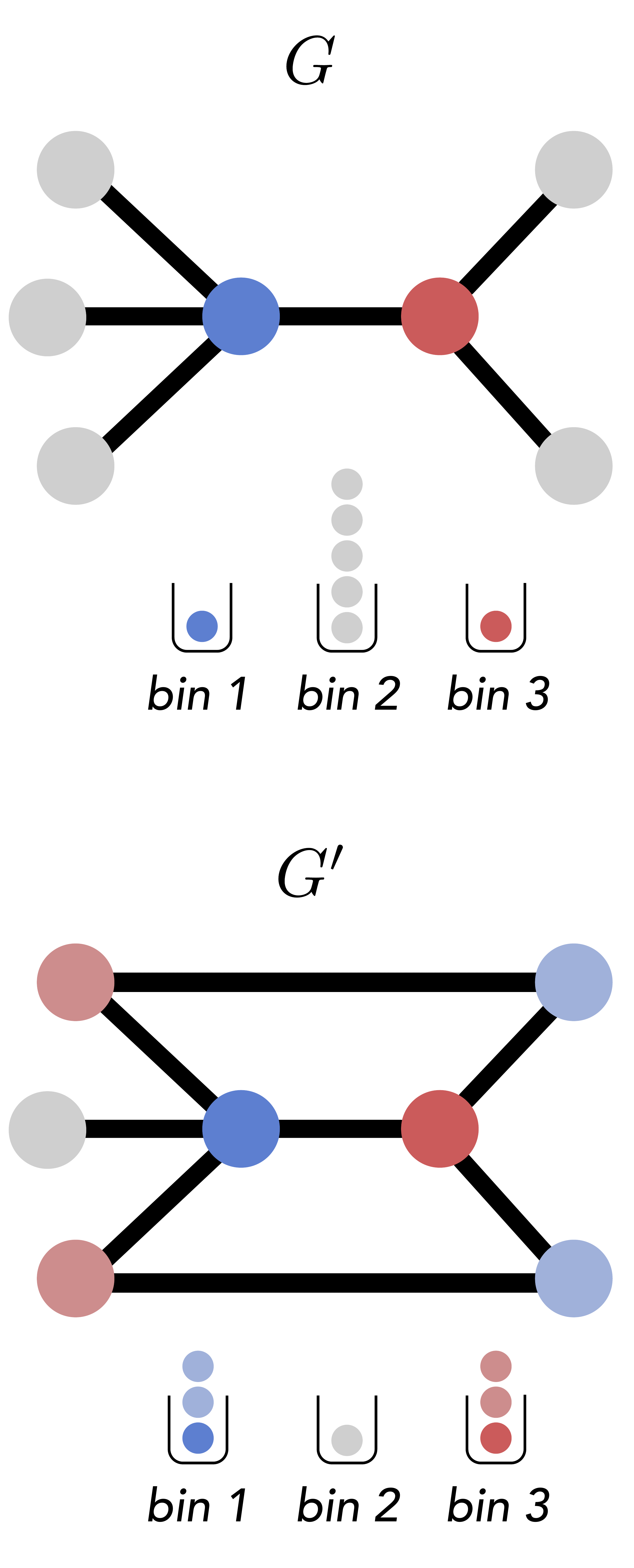}
    \caption{$G$, $G^{\prime}$ at iteration 1}
    \label{fig:pk2}
\end{subfigure}
\hfill 
\begin{subfigure}{0.25\textwidth}
    \centering
    \includegraphics[width=0.9\linewidth]{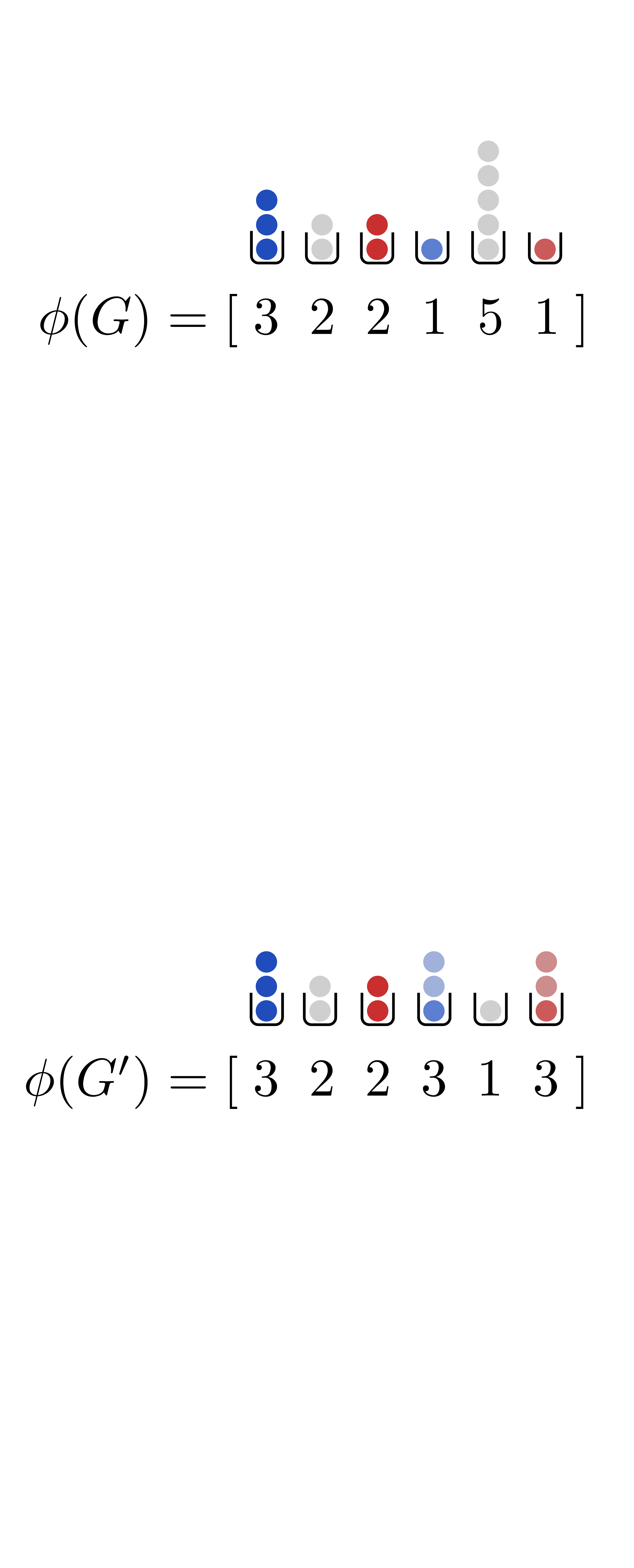}
    \caption{$\phi(G)$ and $\phi(G^{\prime})$}
    \label{fig:pk3}
\end{subfigure}
\par\bigskip
\begin{center}
    
\begin{subfigure}{0.75\textwidth}
    \centering
    \includegraphics[width=0.9\linewidth]{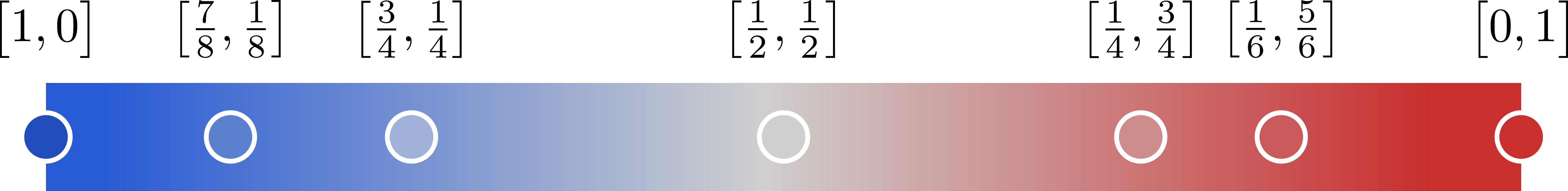}
    \caption{The distribution of node attributes, with the values appearing in this figure identified.}
    \label{fig:pk4}
\end{subfigure}
\end{center}
\caption{A visualisation of a one iteration propagation kernel. The node
attribute distributions at iteration $t$ is $P_t \in \mathbb{R}^{n
\times k}$, where $k$ is the dimension of the node attributes.  Node
attributes are denoted by colour, and their corresponding values in
(\subref{fig:pk4}). $P_0$ thus contains $[1, 0]$ in the indices
corresponding to blue nodes in the original graphs, $[\frac{1}{2},
\frac{1}{2}]$ for the gray nodes corresponding to gray nodes, and $[0,
1]$ to the red nodes. The node attribute distributions are updated via
the following: $P_{t+1} = TP_t$, where $T = D^{-1}A$, \ie\ the row-normalised adjacency matrix. In each propagation, the node label distributions are binned, the feature vector of a graph is the count of nodes in each bin throughout all propagation steps.}
\label{fig:propagation_kernel}
\end{figure}

An important limitation of many of the first graph kernels based on
iterative label refinement is their reliance on categorical node and/or
edge labels. Indeed, all approaches described so far exploit the fact that node labels are discrete to construct efficient hashing schemes that lie at the core of their respective relabelling steps. Motivated by this, \citet{Neumann16} introduced the propagation kernels, a \emph{family} of graph kernels that, much like the Weisfeiler--Lehman framework, are flexible with respect to the choice of base kernel, but can be applied to a more general class of graphs, including those having continuous node and/or edge attributes or even graphs with partially-missing attributes.

Their central idea is similar to other approaches already introduced in this section. Information, such as label sequences, is \emph{propagated} through a series of local transformations to assign a sequence of relabelled graphs to each graph of the input data set. The similarity of two graphs may then be assessed by using any valid kernel that compares their individual sequences.

\begin{defn}[Propagation kernels]
  Let $\graph$ and $\graph'$ be two graphs (with or without attributes). We
  assume that there is a \emph{propagation scheme}, which we will
  discuss later, that assigns a \emph{sequence}~(of equal length $t$) of
  graphs to each graph.
  We thus have a sequence $\graph_0, \graph, \dots, $ \linebreak $\graph_{t-1}$ for
  $\graph$ and a sequence $\graph'_0, \graph'_1, \dots, \graph'_{t-1}$
  for $\graph'$. Letting $\vertices_i$ and $\vertices'_i$ refer to the
  vertex sets of the $i$th graph in the propagation scheme of $\graph$
  and $\graph'$, respectively, we can use any node kernel $\kernelN(\cdot)$
  to define a propagation kernel as
  \begin{equation}
    \kernelP\left(\graph, \graph'\right) := \sum_{i=0}^{t-1} \sum_{\vertex\in\graph_i} \sum_{\vertex'\in\graph'_i} \kernelN(\vertex, \vertex'),
  \end{equation}
  which amounts to the sum of individual node kernels over the
  propagation process.
\end{defn}

The node kernel in the previous equation should be chosen according to
the problem domain. Typical examples include the linear kernel~(for
real-valued attributes) or a Dirac kernel~(for labels). The most important
property of this kernel is that the propagation scheme can be adapted easily
to deal with missing information in the graph. \citet{Neumann16} present
multiple suitable schemes for this purpose: for labelled and unlabelled
graphs, there is a simple \emph{diffusion scheme} based on iterative
updates of the labels~(or degrees, for unlabelled graphs) of the
vertices in a one-hop neighbourhood around each vertex.
%
For partially labelled graphs, this scheme can be slightly
adapted~\citep{Neumann16}, whereas for graphs with continuous
attributes, the simplest propagation scheme assumes that attributes can
be modelled according to a mixture of Gaussian distributions, whose
parameters are then adjusted in every step. We provide an example of
a propagation kernel in Figure~\ref{fig:propagation_kernel}.

The computational complexity of this kernel clearly depends on the calculation of
the kernel for each propagation scheme. \citet{Neumann16} note that in many cases,
it is possible to use a \emph{hashing} function~(or, similarly, \emph{binning} of
node features) to perform the evaluation of the kernel on two graphs
$\graph_i$ and $\graph'_i$ in \emph{linear}~(linear in the number of
bins) time. Such a binning process is straightforward in the case of
discrete attributes; see the discussion of hashing functions in
Section~\ref{sec:Weisfeiler--Lehman kernel} for more details.
In the case of continuous node attributes, \citet{Neumann16} propose
using \emph{locality-sensitive hashing}~\citep{Datar04}, \ie\ a family
of hashing functions, for this purpose.
Using these speed-up techniques, propagation kernels have
a computational complexity of order $\landau{t N m + t N^2 n}$ for computing
features based on counts, where $t$ is the number of iterations of the kernel,
 $N$ is the number of graphs, and $n$ and $m$ are the maximum number of
 vertices and edges, respectively. Thus, its asymptotic complexity is comparable
 to the Weisfeiler--Lehman framework or the neighbourhood hash kernel.
\begin{figure}[h!]
    \centering
    \includegraphics[width=\textwidth]{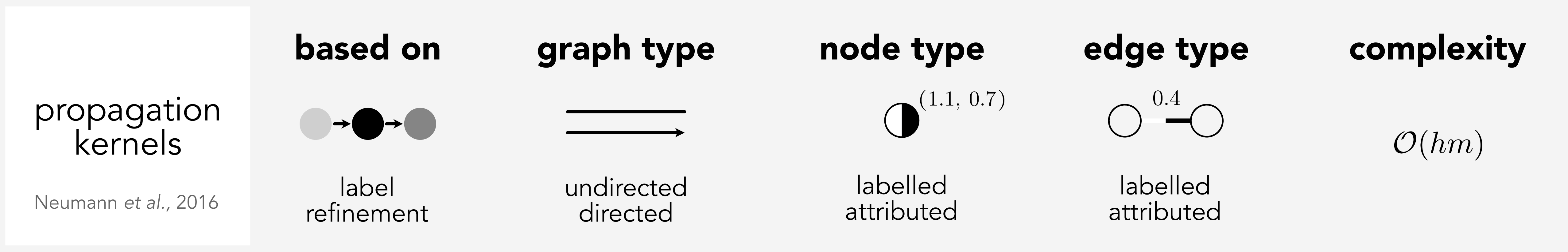}
\end{figure}

\subsubsection{Message passing graph kernels}\label{sec:Message passing graph kernels}

We conclude the section of graph kernels based on iterative label
refinement with an approach by~\citep{Nikolentzos18} that introduces the \emph{message passing} framework. Like propagation kernels, the message passing framework aims to extend existing approaches in order to handle graphs with continuous attributes. Nevertheless, this framework accomplishes this by means of a message passing step inspired by graph neural networks~(see \citet{Zhou18}
for an in-depth review of these techniques). This step, however, is defined in terms of auxiliary kernel functions, rather than parametric transformations instantiated as neural networks.
\begin{defn}[Vertex-based message passing graph kernel]
  Given a vertex-based kernel function $\kernel_{\vertex}$ and
  a neighbourhood-based kernel $\kernelNB$, the message passing graph kernel is an
  iterative scheme satisfying the recurrence formula
  \begin{equation}
    \kernel_{\vertex}^{t+1}\left(\vertex_1, \vertex_2\right) := \alpha \kernel_{\vertex}^{t}\left(\vertex_1, \vertex_2\right)
    + \beta\kernelNB\left(\neighbourhood\left(\vertex_1\right), \neighbourhood\left(\vertex_2\right)\right),
  \end{equation}
  where $t$ denotes the iteration step, $\alpha, \beta \in \real_{> 0}$ are non-negative scale factors,
  and $\neighbourhood\left(\cdot\right)$ refers to the neighbourhood of a vertex.
  \label{def:Vertex-based message passing graph kernel}
\end{defn}
This is a valid vertex kernel because of the closure properties of
kernel functions~(see Section~\ref{sec:Kernel theory} on
p.~\pageref{sec:Kernel theory} for more details). The vertex-based
kernel can be trivially extended to handle two graphs $\graph,
\graph'$ by setting
\begin{equation}
  \kernelMP\left(\graph, \graph'\right) := \kernel_{\vertices}\left(\vertices, \vertices'\right),
  \label{eq:Message passing graph kernel}
\end{equation}
where $\kernel_{\vertices}$ denotes a kernel between the vertex sets
$\vertices$ and $\vertices'$ of $\graph$ and $\graph'$,
respectively. \citet{Nikolentzos18} suggest two different families of
vertex set kernels, which make up their proposed message passing graph kernel.
%
\begin{defn}[Graph-based message passing kernel]
  Given two sets of vertices $\vertices$ and $\vertices'$ of two
  graphs $\graph$ and $\graph'$, the graph-based message passing
  kernel is defined by replacing $\kernel_{\vertices}$ in
  Eq.~\ref{eq:Message passing graph kernel} with either
  \begin{equation}
    \kernel_{\vertices}\left(\vertices, \vertices'\right) :=
    \sum_{\vertex\in\vertices} \sum_{\vertex'\in\vertices'}
    \kernel_{\vertex}\left(\vertex, \vertex'\right),
    \label{eq:Message passing kernel R-convolution}
  \end{equation}
  leading to a standard graph kernel formulation which is based on the
  \mbox{$\mathcal{R}$-convolution} framework~(see
  Section~\ref{sec:R-convolution kernels} on p.~\pageref{sec:R-convolution kernels}
  for more details),
  or, alternatively,
  \begin{equation}
    \kernel_{\vertices}\left(\vertices, \vertices'\right) := \max_{B \in
    \mathfrak{B}(\vertices, \vertices')} \sum_{\vertex, \vertex' \in B}
    \kernel_{\vertex}\left(\vertex, \vertex'\right),
    \label{eq:Message passing kernel OA}
  \end{equation}
  where $\mathfrak{B}(\vertices, \vertices')$ denotes the set of all
  bijections between $\vertices$ and $\vertices'$, leading to
  a kernel formulation based on \emph{optimal assignment}.
  \citet{Nikolentzos18} use a Dirac delta kernel for $\kernel_{\vertex}$
  in the case of graphs with discrete node labels, or a linear kernel
  for graphs with attributes.
   \label{def:Graph-based message passing graph kernel}
\end{defn}

Given a kernel function for vertex sets as shown in Definition~\ref{def:Graph-based message passing graph kernel},
Eq.~\ref{eq:Message passing graph kernel} can be easily adjusted to integrate
the recurrence formula from Definition~\ref{def:Vertex-based message passing graph kernel}.
Using a maximum of $t$ iterations, this leads to
\begin{align}
  \kernelMP\left(\graph, \graph'\right) &:= \sum_{\vertex\in\vertices}
  \sum_{\vertex'\in\vertices'} \kernel_{\vertex}^{t}\left(\vertex,
  \vertex'\right),\\
  \shortintertext{or}
  \kernelMP\left(\graph, \graph'\right) &:= \max_{B \in
  \mathfrak{B}(\vertices, \vertices')} \sum_{\vertex, \vertex' \in B}
  \kernel_{\vertex}^{t}\left(\vertex, \vertex'\right),
\end{align}
respectively.
As the subsequent discussion of optimal assignment kernels in
Section~\ref{sec:Optimal assignment kernels} on p.~\pageref{sec:Optimal
assignment kernels} will show, the kernel function $\kernel_{\vertex}$ in
Eq.~\ref{eq:Message passing kernel OA} has to satisfy certain properties---in the terminology
of \citet{Kriege16}, it has to be a \emph{strong} kernel, \ie\  one induced by
a hierarchy. To satisfy this, \citet{Nikolentzos18} suggest using a clustering scheme, such as
\mbox{$k$-means} clustering~\citep{Kanungo02}.

In terms of computational complexity, evaluating the message passing kernel based
on \mbox{$\mathcal{R}$-convolution}~(Eq.~\ref{eq:Message passing
kernel R-convolution}) has a complexity of \linebreak $\landau{|\vertices|^2}$, where $|\vertices|$
denotes the maximum cardinality of a vertex set. By contrast, an evaluation
of the optimal assignment message passing kernel~(Eq.~\ref{eq:Message passing kernel OA})
has a complexity of $\landau{\overline{d}^2}$, where $\overline{d}$ refers to the average degree of a
graph. Asymptotically, the two kernels behave the same in the case of complete graphs; in practice, however,
graph data tend to contain sparse graphs, implying $\overline{d} \ll |\vertices| - 1$.
\citet{Nikolentzos18} showed that, with a pre-calculated hierarchy, it is possible to employ a Nystr\"om method~\citep{Williams01} to compute Eq.~\ref{eq:Message passing kernel OA} with \emph{linear} time
complexity in the cardinality of the vertex set, \ie\ in $\landau{|\vertices|}$ time,
giving the message passing kernel desirable scaling properties.

The message passing kernel is linked to the Weisfeiler--Lehman subtree
kernel~(see Section~\ref{sec:Weisfeiler--Lehman kernel} on
p.~\ref{sec:Weisfeiler--Lehman kernel} for more details). In fact, as
the authors note, the message passing kernel framework \emph{includes}
the Weisfeiler--Lehman subtree kernel~\citep{Nikolentzos18}:
following the terminology of this section, the subtree kernel satisfies
the \mbox{$\mathcal{R}$-convolution} formulation from Eq.~\ref{eq:Message passing kernel R-convolution},
where the kernel between two vertices at every step $t$ of the iteration
is equal to the kernel of the previous time step followed by a Dirac
delta kernel between their new labels. However, the formulation of the
Weisfeiler--Lehman subtree kernel that we give in this survey is
computationally more efficient and to be preferred for real-world
applications.
\begin{figure}[h!]
    \centering
    \includegraphics[width=\textwidth]{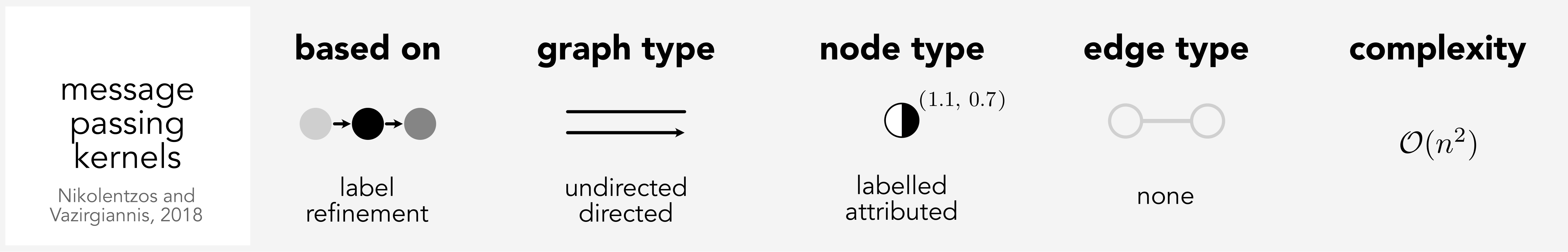}
\end{figure}

\subsection{Graph kernels based on spectral theory}\label{sec:Graph kernels
based on spectral theory}

Graph spectral theory provides a wealth of well-known results that allow
the characterisation of a graph in terms of the eigenvalues and
eigenvectors of its corresponding graph Laplacian matrix~(see
Definition~\ref{def:Graph Laplacian}). The graph Laplacian can be
intuitively understood as a discretised differential operator. Given
any function $f\colon V \to \real$, one can show
that $L := D - A$ satisfies $\left(Lf\right)(v)= \sum_{v' \in
\mathcal{N}(v)} w(v, v')(f(v) - f(v'))$, where $w(v, v')$ is the weight
of the edge between the vertices $v$ and $v'$. Moreover, one can also
show that 
\begin{equation}
\langle f, Lf \rangle = \sum_{(v, v') \in E}w(v, v') \left(f(v) - f(v')\right)^{2}
\end{equation}
holds in general. Thus, eigenvectors of $L$ corresponding to small eigenvalues can be seen as smoothly-varying functions defined on the nodes of the graph, capturing a notion of ``shape'' of the graph. These and other properties of the graph Laplacian matrix have been exploited, for example, to design kernels to quantify the similarity being nodes of a single graph~\citep{kondor2002diffusion}. The resulting kernels admit multiple theoretical interpretations, including close connections to random walks and stochastic diffusion processes~\citep[Section 3]{kondor2002diffusion}. 

Extending these approaches to instead compare graphs presents multiple challenges. Firstly, it is not immediate how to ensure that the resulting kernel will be invariant with respect to graph isomorphism. Secondly, a direct implementation would not make use of node attributes, when available. Finally, and perhaps most importantly, the graph Laplacian mostly captures ``global'' properties of the graph yet localized information about graph topology is likely to be essential for many practical applications. In the remainder of this section, we will introduce a recent graph kernel based on spectral theory that overcomes all of these drawbacks.

\subsubsection{Multiscale Laplacian graph kernel}\label{sec:Multiscale Laplacian graph kernel}

The \emph{multiscale Laplacian graph kernel} of \citet{Kondor16} solves the aforementioned limitations by combining two different contributions. Firstly, they propose the \emph{feature space Laplacian graph kernel} (FLG kernel), a novel graph kernel based on spectral theory that is able to take node attributes into account while introducing invariance to vertex permutations. However, the FLG kernel only models global aspects of the graphs being compared. In order to to capture structural information of graphs at multiple scales, \citet{Kondor16} introduce a graph kernel defined recursively in terms of a suitably-defined \emph{hierarchy} of subgraphs ``centered'' around each vertex. Typically, these subgraph hierarchies will correspond to $k$-hop neighbourhoods for increasing values of $k$. In a nutshell, the recursive construction of graph kernels in this framework builds on the FLG kernel, exploiting the fact that it admits any p.d.\ kernel to compare pairs of nodes. Thus, to obtain a graph kernel at scale $k +1$, their method uses the previous graph kernel at scale $k$ to define a p.d.\ kernel between \emph{nodes}, which compares their respective $k$-hop neighbourhoods.

We will begin by introducing the FLG kernel. The core idea behind this
approach is to define a random variable for each graph that combines
spectral information with its node attributes. Then, the problem of
comparing two graphs can be reduced to the problem of comparing the
probability densities of these random variables, for which
\citet{Kondor16} use the Bhattacharyya
kernel~\citep{jebara2003bhattacharyya}. The FLG kernel assumes the
existence of a suitable p.d.\ kernel to compare pairs of vertices,
$k_{\text{node}}$. Suppose that $\phi_{\text{node}}(\cdot)$ is the
feature map corresponding to this kernel. By definition, the 
linear combination
\begin{equation}
  \phi(G) : = \sum_{v \in V} \alpha(v) \phi_{\text{node}}(v) = \Phi_{V} \mathbf{\alpha}_{V},
\end{equation}
where $\Phi_{V}$ contains the feature map representations of all nodes
and $\mathbf{\alpha}$ is a vector of the vertex weights $\alpha(v)$,
is invariant to vertex permutations.
\citet{Kondor16} define the set $\left\{ \alpha(v) \mid
v \in V \right\}$ to be random variables distributed according to
a Gaussian Markov Random Field~\citep{koller2009probabilistic} sharing
structure with the graph. This accomplishes three objectives: (1) the
probability density of $\left\{ \alpha(v) \mid v \in V \right\}$, which
is a Normal distribution $\mathcal{N}(0, L^{-1})$, is endowed with
information about the (global) structure of the graph; (2) this density
transforms in the same way as $\left\{\phi_{\text{node}}(v) \mid v \in
V \right\}$ under vertex reorderings, achieving the desired invariance
and (3) the Bhattacharyya kernel between the densities of $\phi(G_1)
\sim \mathcal{N}(0, \Phi_{V_{1}} L_{1}^{-1} \Phi_{V_1}^{T} )$ and
$\phi(G_2) \sim \mathcal{N}(0, \Phi_{V_{2}} L_{2}^{-1} \Phi_{V_2}^{T}
)$, can be computed in closed form. This leads to the following
definition of the \emph{generalized feature space Laplacian graph
kernel}.

\begin{defn}[Generalized feature space Laplacian graph kernel]
  Let $\graph = (\vertices, \edges)$ and $\graph' = (\vertices',
  \edges')$ be two undirected graphs whose vertex sets are subsets of
  the same vertex space $\mathfrak{V}$, \ie\ $\vertices \subseteq
  \mathfrak{V}$ and $\vertices' \subseteq \mathfrak{V}$. Furthermore, let
  $\basekernel\colon\mathfrak{V}\times\mathfrak{V}\to\real$ be a base kernel
  on said vertex space. In the case that each vertex can be assigned a one-hot
  encoded feature vector, $\basekernel$ might be chosen as a linear
  kernel on these vectors, for example.
  The \emph{generalized feature space Laplacian graph kernel} induced
  by $\basekernel$ is then defined as
  \begin{equation}
    \kernelFSL^{\basekernel}\left(\graph, \graph'\right) :=
      \frac{\left|\left(\frac{1}{2}S_1^{-1} + \frac{1}{2}S_2^{-1}\right)^{-1}\right|^{\frac{1}{2}}}{\left|S_1\right|^\frac{1}{4} \left|S_2\right|^{\frac{1}{4}}},
  \end{equation}
  where $\left|\cdot\right|$ denotes the determinant of the given matrix. In the preceding
  equation, $S_1$ and $S_2$ are transformed variants of the graph Laplacian matrices that
  take the base kernel into account. More precisely, for $i \in \{1,2\}$, we have
  \begin{equation}
    S_{i} := Q_i^{\top} \laplacian_i^{-1} Q_i + \eta \ones,
  \end{equation}
  %
  where $Q_i$ contains those eigenvectors of the joint Gram matrix of the
  base kernel $\basekernel$ that correspond to the vertices of $\graph_i$,
  while $\eta$ denotes a small regularization factor, and $\ones$ is the
  identity matrix of appropriate size.
  \label{def:FSL}
\end{defn}
The generalized feature space Laplacian kernel may now be extended and
applied recursively to capture the similarity between subgraphs. We first
give the definition in terms of \emph{one} graph but will subsequently
extend it.

\begin{defn}[Multiscale Laplacian subgraph kernel]
  Let $\graph = (\vertices, \edges)$ be an undirected graph and
  $\basekernel\colon\mathfrak{V}\times\mathfrak{V}\to\real$ a base kernel on
  its nodes, with $\vertices \subseteq \mathfrak{V}$ as defined above.
  Furthermore, suppose that there is a sequence of nested neighbourhoods
  such that for each vertex $\vertex \in \vertices$,
  we have
  \begin{equation}
    \vertex\in\neighbourhood_1(\vertex)\subseteq\neighbourhood_2(\vertex)\subseteq\dots\subseteq\neighbourhood_h(\vertex)
  \end{equation}
  where $\neighbourhood_j(\vertex) \subseteq \vertices$.
  \citet{Kondor16} propose that such
  a \emph{filtration} of neighbourhoods be obtained by extending
  the usual ``hop'' neighbourhoods~(see Definition~\ref{def:Neighbourhood},
  p.~\pageref{def:Neighbourhood}) around a vertex, for example.
  Letting
  $\graph_i(\vertex)$ denote the graph that is induced by $\vertex$ and
  a certain neighbourhood, the hierarchy of \emph{multiscale Laplacian subgraph kernels}
  $\kernelMLS_1, \dots, \kernelMLS_h$ is defined recursively:
  \begin{enumerate}
    \item The first multiscale Laplacian subgraph kernel is the
    \emph{generalized feature space Laplacian} graph kernel induced by
    the base kernel $\basekernel$, \ie\
      \begin{equation}
        \kernelMLS_1(\vertex_1, \vertex_2) := \kernelFSL^{\basekernel}\left(
          \graph\left(\vertex_1\right),
          \graph\left(\vertex_2\right)
        \right),
      \end{equation}
      which is evaluated on the induced subgraphs of the first stage of the neighbourhood filtration.
      \item The higher stages of the neighbourhood filtration, by
      contrast, use the generalized feature space Laplacian kernel
      induced by lower stages, \ie\
      \begin{equation}
        \kernelMLS^{(j)}(\vertex_1, \vertex_2) := \kernelFSL^{\kernelMLS^{(j-1)}}\left(
          \graph_j\left(\vertex_1\right),
          \graph_j\left(\vertex_2\right)
        \right),
      \end{equation}
      where $j \in \{1, \dots h\}$.
  \end{enumerate}
  \label{def:MLS}
\end{defn}
Finally, to extend this definition to the comparison of two graphs
$\graph$ and $\graph'$, \citet{Kondor16} note that the kernel can
be extended under the assumption that the \emph{same} base kernel is
used. This leads to the \emph{multiscale Laplacian graph kernel}.
\begin{defn}[Multiscale Laplacian graph kernel]
  Given two graphs $\graph$ and $\graph'$ with the same vertex space
  as in Definition~\ref{def:FSL}, the \emph{multiscale Laplacian graph kernel}
  is defined as
  \begin{equation}
    \kernelML\left(\graph, \graph'\right) := \kernelFSL^{\kernelMLS^{(h)}}\left(
      \graph,
      \graph'
    \right),
  \end{equation}
  where $\kernelMLS^{(h)}$ denotes the multiscale Laplacian subgraph kernel at the
  highest level, following Definition~\ref{def:MLS}.
\end{defn}
In terms of computational complexity, it is possible to evaluate the
kernel between two graphs in time proportional to $\landau{hn^2}$, where
$h$ denotes the number of scales (recursion steps) and $n$ denotes the number of
vertices.
\begin{figure}[h!]
    \centering
    \includegraphics[width=\textwidth]{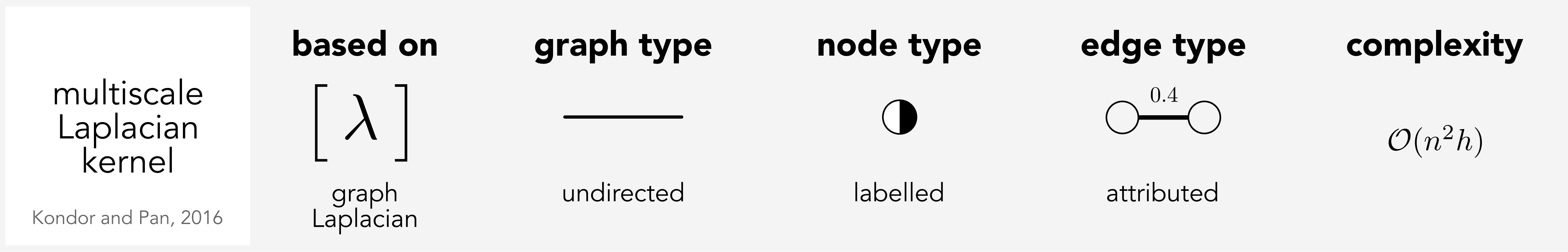}
\end{figure}

\section{Extensions of common graph kernels}\label{sec:Extensions}

Finally, some methods can be considered an extension of the
previously defined categories. We now describe two important categories
of such methods, namely approaches that extend existing methods to
continuous attributes, and methods that define graph kernels outside the popular
$\mathcal{R}$-convolution framework.

\subsection{Extending graph kernels to handle continuous attributes}\label{sec:Extending graph kernels to handle continuous attributes}

Many of the graph kernels introduced so far, including some of the most commonly used such as the Weisfeiler-Lehman framework (Section~\ref{sec:Weisfeiler--Lehman kernel}), forego the possibility to use continuous node and/or edge attributes, often due to computational considerations. Some approaches have been motivated precisely by the goal to overcome this limitation, such as the family of propagation kernels (Section~\ref{sec:Propagation kernels}) or the message passing kernels (Section~\ref{sec:Message passing graph kernels}). These new methods, however, focus on one particular type of substructure. In this section, we will instead describe three different generic approaches that allow using most graph kernels, including those limited to categorical attributes, to compare graphs that might nonetheless have continuous node and/or edge attributes.

The first method, subgraph matching kernels~(Section~\ref{sec:Subgraph
matching kernels}) counts the number of common subgraph isomorphisms
between two graphs. This kernel relies on using the direct product graph,
introduced in Section~\ref{sec:direct product graph kernel}, to identify the
maximum common subgraphs, and then weights these findings using the node
and edge labels or attributes.

The idea behind the second of these approaches, the graph invariant
kernel framework~(Section~\ref{sec:Graph invariant kernel framework}),
is to augment a simple node-only graph kernel, which can easily
accommodate continuous node attributes by defining a suitable p.d.\
kernel $k_{\text{node}}$ but misses topological information, with
a \emph{weight function}, which is computed using previously existing
graph kernels to capture higher-order graph substructures.

The third method, hash graph kernels (Section~\ref{sec:Hash graph
kernels}), instead propose to transform each input graph into an
ensemble of ``discretised'' graphs, each of them obtained by applying
a randomly sampled hash function that maps each real-valued attribute to
an integer-valued ``bin''. These graph ensembles can then be compared to
each other using existing graph kernels, and the final similarity
between any two input graphs is obtained as the average similarity
between elements of their respective ensembles.

\subsubsection{Subgraph matching kernels}\label{sec:Subgraph matching kernels}

\citet{kriege2012subgraph} provided an early approach to extend graph
kernels to graphs with continuous attributes. The foundation of the
method hinges upon counting the number of common subgraph isomorphisms
using the direct product graph. Whereas the graphlet kernel described in
Section~\ref{sec:Graphlet kernel} counts the number of matching common
subgraphs of two graphs $\graph$ and $\graph'$,
\citet{kriege2012subgraph} 
propose counting the number of matching subgraph isomorphisms, since
there can be multiple isomorphisms for a given common subgraph. After
providing a framework on how to do this for a simple graph with node and
edge labels, they provide extensions to incorporate continuous node and
label attributes. 
\begin{defn}[Common subgraph isomorphism kernel]
  Let $\graph = \left(\vertices, \edges\right)$ and $\graph'
  = \left(\vertices', \edges '\right)$ be two undirected graphs, with
  node and edge labels from the alphabets  $\Sigma_{\text{V}}$ and
  $\Sigma_{\text{E}}$ respectively. For a subset of vertices
  $\vertices_i \in \vertices$ and $\vertices_i' \in \vertices'$, and
  their corresponding induced subgraphs $\graph[\vertices_i]$ and
  $\graph'[\vertices_i']$, a common subgraph isomorphism (CSI) is the
  mapping $\varphi\colon \vertices_i \to \vertices_i'$. The
  \emph{common subgraph isomorphism kernel} can thus be defined as the
  sum over all common subgraph isomorphisms $\mathcal{I}$ in $\graph$
  and $\graph'$, weighted by some weight function $\lambda$, \ie\
  \begin{equation}
    \kernelCSI\left(\graph, \graph'\right) := \sum_{\varphi\in\mathcal{I}(\graph,\graph')} \lambda(\varphi).
  \end{equation}
\end{defn}
$\kernelCSI$ is used to then compute the subgraph matching kernel. 

\begin{defn}[Subgraph matching kernel]
  Again let $\graph = \left(\vertices, \edges\right)$ and $\graph'
  = \left(\vertices', \edges '\right)$ be two undirected graphs, with
  node and edge labels from the alphabets $\Sigma_{\text{V}}$ and
  $\Sigma_{\text{E}}$ respectively, and $\lambda$ representing all
  bijections from $\vertices_i$ to $\vertices_i'$. The subgraph matching
  kernel is thus defined as
  \begin{equation*}
    \kernelSM\left(\graph, \graph'\right) := \sum_{\varphi\in\mathfrak{B}(\graph,\graph')} \lambda(\varphi) \prod_{\vertex \in V_i}\kernelN(\vertex, \varphi(\vertex)) \prod_{e \in \vertices_i \times \vertices_i} \kernelE(e, \psi_{\varphi}(e))
  \end{equation*}
  where $\vertices_i = \domain(\varphi)$, and $\kernelN$ and $\kernelE$
  are predefined kernels on the nodes and edges. 
\end{defn}

The authors recommended a simple Dirac kernel on the node and edge
labels for $\kernelN$ and $\kernelE$ when the graphs have categorical
labels, and offered an extension to node/edge attributed graphs by
multiplying these Dirac kernel by an additional kernel evaluated on the
continuous attributes, such as the Brownian bridge kernel for node
attributes or the triangular kernel for edge attributes. Moreover, the
authors used a uniform value of $\lambda$ in their experiments.

The key step in this process is to use the result of \citet{levi1973},
who found that each maximum clique in the product graph
$\graph_{\times}$ from graphs $G$ and $G^{\prime}$ corresponds to
a maximal common subgraph of $\graph$ and $\graph^{\prime}$. Indeed,
determining all such cliques equates to finding all common subgraph
isomorphisms, and thus reduces the problem to finding all maximal
cliques in the product graph. The product graph is further extended by
creating a weighted product graph, which has the effect of assigning
a weight to the maximal cliques. 

The complexity of the subgraph matching kernel depends on the number of
nodes in the two graphs and the upper bound on the size
of subgraphs considered, $k$. This leads to a complexity of
$\landau{k(n^2)^{k+1}}$.
\begin{figure}[h!]
    \centering
    \includegraphics[width=\textwidth]{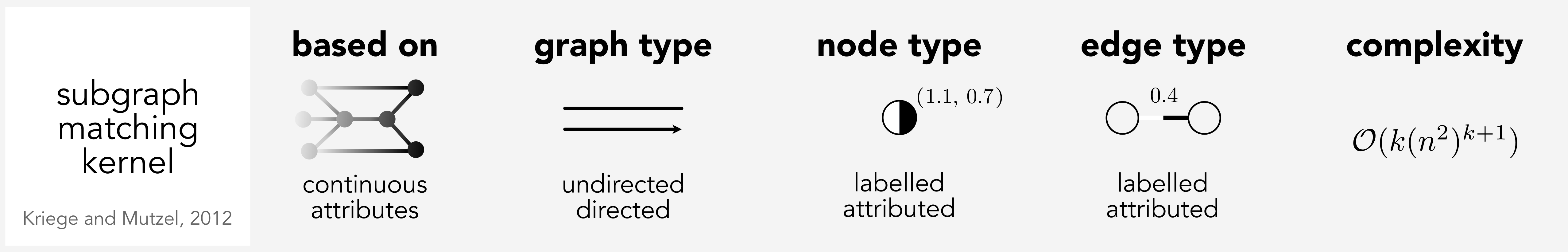}
\end{figure}
\subsubsection{Graph invariant kernel framework}\label{sec:Graph invariant kernel framework}

The framework of graph invariant kernels was developed by \citet{Orsini15}
to extend existing graph kernels to graphs with continuous attributes,
based on the calculation of certain \emph{vertex invariants}. This makes
it possible to quickly adapt certain graph kernels---most prominently
the Weisfeiler--Lehman graph kernel framework---to graphs with
continuous attributes.

The central concept of the framework by \citet{Orsini15} is the vertex
invariant. A vertex invariant is a function
\begin{equation}
  \invariant\colon\vertices\to\colours
\end{equation}
that assigns each vertex $\vertex\in\vertices$ a ``colour''
$\colour\in\colours$, also known as \emph{colour refinement}.
%
The assigned colour $\colour = \invariant(\vertex)$ needs to be
\emph{invariant} under graph isomorphism, \ie\ $\invariant\left(\vertex\right) = \invariant\left(f\left(\vertex\right)\right)$
for any isomorphism $f$. This leads to the generic formulation of the
graph invariant kernels framework.

\begin{defn}[Graph invariant kernels]
  Let $\graph = \left(\vertices, \edges\right)$ and
  $\graph' = \left(\vertices', \edges '\right)$
  be two undirected graphs, potentially with additional node attributes.
  The general \emph{graph invariant kernel} is defined as
  \begin{equation}
    \kernelGI\left(\graph, \graph'\right) := \sum_{\vertex\in\vertices} \sum_{\vertex'\in\vertices'} \fweight\left(\vertex, \vertex'\right) \kernelN\left(\vertex, \vertex'\right),
  \end{equation}
  where $\kernelN(\cdot)$ represents a suitable kernel defined on the
  nodes---and their attributes---and $\fweight(\cdot)$ is a \emph{weight
  function} that counts the number of invariants that $\graph$ and $\graph'$
  have in common.
  Following the $\mathcal{R}$-convolution framework~(see also
  Section~\ref{sec:R-convolution kernels}), it is calculated over all substructures $g \in
  \relations^{-1}(\graph)$ and $g' \in \relations^{-1}(\graph')$
  \begin{equation}
    \fweight\left(\vertex, \vertex'\right) :=
    \begin{cases}
    \sum_{(g,g')} \basekernel\left(\vertex, \vertex'\right) \frac{\match(g, g')}{|g| |g'|} & \text{if $\vertex \in g$ and $\vertex' \in g'$}\\
    0                          & \text{otherwise.}
    \end{cases}
  \end{equation}
  In the previous equation, $\basekernel(\cdot)$ refers to a kernel that
  measures the similarity of the vertex colours under a selected vertex
  invariant $\invariant$, while $\match(\cdot)$ is a function that
  determines if two substructures \emph{match}.
  Subsequently, we will describe some example choices of these two functions.
\end{defn}

We first describe suitable choices for the match function $\match(\cdot)$ because
this function admits a more intuitive explanation. In general, $\match(\cdot)$ can
be expressed as an equivalence check, \ie\
\begin{equation}
  \match(g, g') := \begin{cases}
                      1 & \text{if $g \sim g'$}\\
                      0 & \text{otherwise}.
                   \end{cases}
\end{equation}
Thus, $\match(g, g')$ checks whether the two substructures are
\emph{equivalent}. For example, if $\relations(\graph)$ decomposes
a graph with node labels into its shortest paths, $\match(g, g')$ could
be defined to check whether two shortest path have the same length \emph{and}
the same sequence of labels when following the path.
The kernel $\basekernel(\cdot)$, by contrast, can be defined by reusing
other graph kernels that are defined on substructures. For example,
suppose we are given a graph with node labels. We may then use the
Weisfeiler--Lehman framework~(see also
Section~\ref{sec:Weisfeiler--Lehman kernel}) to obtain a sequence of
neighbourhood-based labels per vertex.
Letting $\labelsequence\left(\vertex\right)$ and
$\labelsequence\left(\vertex'\right)$ refer to the Weisfeiler--Lehman
node label sequences of two vertices $\vertex \in \graph$ and $\vertex'
\in \graph'$, respectively, we can define
\begin{equation}
  \basekernel\left(\vertex, \vertex'\right) := \left|\left\{
      \labelsequence^{(i)}\left(\vertex\right),
      \labelsequence^{(i)}\left(\vertex'\right)
        \mid
      \labelsequence^{(i)}\left(\vertex\right) = \labelsequence^{(i)}\left(\vertex'\right)
  \right\}\right|,
\end{equation}
which counts the number of equal labels for the two vertices. Other kernel choices
are possible as well; please refer to \citet{Orsini15} for more information.

In terms of complexity, the graph invariant kernels framework depends on
the complexity of the node label kernel and the complexity of the match
function. In general, for an undirected graph $\graph = (\vertices,
\edges)$ with a diameter bounded by $r$ and maximum vertex degree $d$,
the complexity is of order $\landau{n^{2}\left(C_1+C_2\tau d^{4r}\right)}$, where $n$ is the number of vertices,
$C_1$ and $C_2$ are costs that depend on $\kernelN(\cdot)$ and
$\basekernel(\cdot)$, and $\tau$ is the maximum number of matching
substructures, \ie\
\begin{equation}
  \tau := \sum_{(g,g')} \match(g, g')
\end{equation}
for substructures $g \in \relations^{-1}(\graph)$ and $g' \in \relations^{-1}(\graph')$.
Typically, $C_1 = n_1$ for a node label kernel with $n_1$ features, and
$C_2 = n_2$ for an invariant kernel with  $n_2$ features. If
\emph{global} invariants, such as the degree, are used, it is possible
to simplify the equation such that $C_2 = 1$~\citep{Orsini15}.
\begin{figure}[h!]
    \centering
    \includegraphics[width=\textwidth]{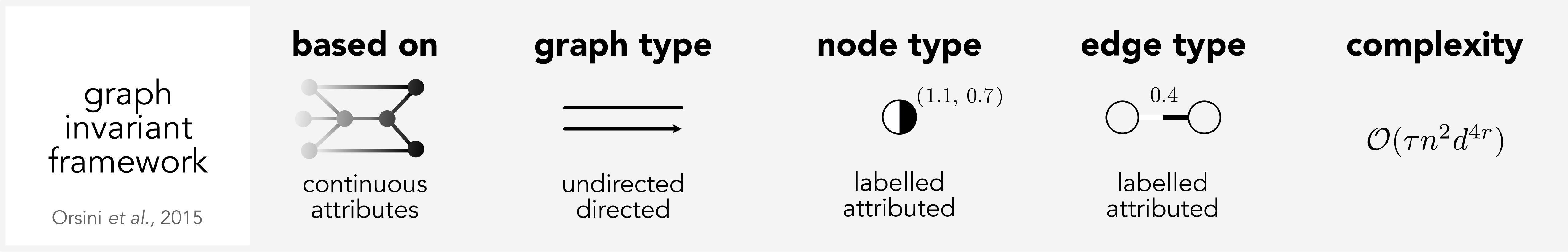}
\end{figure}
\subsubsection{Hash graph kernels}\label{sec:Hash graph kernels}

Hash functions are prevalent in several kernel frameworks---see
Section~\ref{sec:Weisfeiler--Lehman kernel} or
Section~\ref{sec:Neighbourhood hash kernel}, for example---but their
utility was mostly restricted to graphs with discrete labels.
\citet{Morris16} thus developed a framework that makes it possible to
``convert'' \emph{any} graph kernel that supports discrete attributes
to a graph kernel that supports continuous attributes.
The key concept of this framework is the use of multiple hash functions
that map continuous attributes to discrete labels, which in turn permits
the use of a discrete graph kernel.

\begin{defn}[Hash graph kernel]
  Let $\basekernel$ be a base graph kernel, such as a
  Weisfeiler--Lehman kernel~(Definition~\ref{def:Weisfeiler--Lehman
  kernel} on p.~\pageref{def:Weisfeiler--Lehman kernel}). Moreover,
  let $\hashfamily = \{\hash_1, \hash_2, \dots\}$ be a finite family of hash functions. Each element
  $\hash_i\in\hashfamily$ should be a function $\hash_i\colon\real^d \to
  \natural$, where $d$ denotes the dimensionality of the graph
  attributes. While the function $\hash_i$ is applied to individual
  components of $\graph$, such as the nodes, we will use
$\hash_i(\graph)$ to refer to the discretised graph resulting from
applying $\hash_i$ to continuous attributes of the graph.
  Given two graphs $\graph$ and $\graph'$, the \emph{hash graph
  kernel} is defined as
  \begin{equation}
    \kernelHGK\left(\graph, \graph'\right) := \frac{1}{|\hashfamily|} \sum_{i=1}^{|\hashfamily|} \basekernel\left(\hash_i\left(\graph\right), \hash_i\left(\graph'\right)\right),
  \end{equation}
  \ie\ the \emph{average} of kernel values under multiple hash
  functions, which are drawn randomly from $\hashfamily$.
\end{defn}
This formulation is advantageous because it is highly generic and
supports arbitrary kernels---moreover, explicit feature map
representations are available in case the base kernel $\basekernel$
supports them. The run time can easily be shown to only depend
on the employed base kernel, the cardinality of $\hashfamily$, and
the complexity of evaluating a single hash function~\citep{Morris16}.

Typical choices for $\hashfamily$ include, but are not limited to,
locality-sensitive hashing schemes~\citep{Datar04, Pauleve10}. As for
the base kernel function, typical choices discussed by \citet{Morris16}
include the \emph{shortest-path kernel}~(Section~\ref{sec:Shortest-path
kernel}) and the \emph{Weisfeiler--Lehman subtree
kernel}~(Section~\ref{sec:Weisfeiler--Lehman kernel}). We will
abbreviate them with \mbox{HGK-SP} and \mbox{HGK-WL}, respectively.
\begin{figure}[h!]
    \centering
    \includegraphics[width=\textwidth]{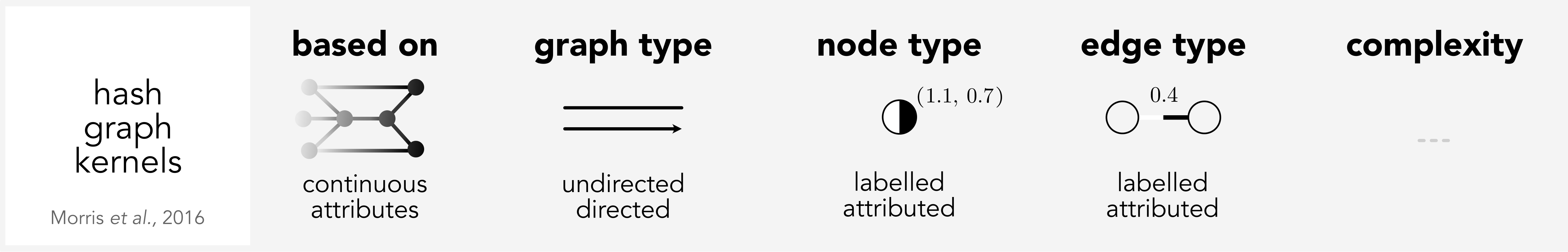}
\end{figure}

\subsection{Beyond simple instances of the $\mathcal{R}$-convolution framework}\label{sec:Beyond simple instances of the R-convolution framework}

As has been argued throughout this chapter, a large number of graph
kernels, including some of the most successful ones, belong to the
\linebreak $\mathcal{R}$-convolution framework, introduced in
Section~\ref{sec:R-convolution kernels}. The $\mathcal{R}$\-convolution
framework is extremely general. However, most
$\mathcal{R}$-convolution-based graph kernels can be seen as simple
instances of this framework that implement an ``all substructure-pairs''
kernel for a certain substructure of choice, such as walks, shortest-paths, graphlets or $k$-hop neighbourhoods. If $\mathcal{R}^{-1}(G)$ denotes the set of all such substructures in a graph $G$, these kernels could be generically expressed as
\[
\kernel(\graph, \graph') = \sum_{g \in \mathcal{R}^{-1}(\graph)}
\sum_{g' \in \mathcal{R}^{-1}(\graph')} \kernel_{\text{base}}(g, g'),
\]
where $\kernel_{\text{base}}$ is a p.d.\ kernel that quantifies the similarity between any two instances of the specific type of substructures under consideration. Moreover, for many of the graph kernels that can be expressed this way, the base kernel is defined to be a Dirac kernel, in which case the feature map of $\kernel$ can be interpreted as an unnormalised histogram that counts the number of occurrences of each possible variation of the substructure of choice. To obtain a representation of graphs that is highly expressive, often complex substructures that can present many variations, such as long walks or $k$-hop neighbourhoods, are preferred over simpler ones, such as vertices, edges or short walks. However, there is an implicit statistical trade-off incurred when making such a choice; the feature map of the kernel could be extremely high-dimensional and, most importantly, could contain a very large number of features mostly irrelevant for the task at hand. This phenomenon often manifests itself as \emph{diagonal dominance} in the resulting kernel matrix, indicating that graphs are mostly deemed similar only to themselves, leading to models prone to overfitting.

Ultimately, this limitation arises due to the underlying lack of adaptivity in the way feature maps are defined by these approaches. When large collections of annotated graph data sets are available, graph neural networks~\citep{Zhou18} are emerging as a powerful generalization of graph kernels that, borrowing many of their implicit biases, attempt to directly learn a feature map optimised end-to-end for the task of interest. Nevertheless, the graph kernels community has also explored alternative methods to alleviate these limitations by different means. In the final section of this chapter, we will describe three such approaches.

Weighted decomposition kernels (Section~\ref{sec:Weighted decomposition kernel framework}) are, in a nutshell, based on the idea of decomposing graphs into two different types of substructures simultaneously, referred to as \emph{selectors} and \emph{contexts} in their terminology. These two types serve complementary roles. Selectors are simple substructures, often chosen to be single vertices, and are compared using a Dirac kernel. In contrast, the contexts are typically chosen to be more complex substructures, such as the $k$-hop neighbourhoods surrounding the selectors, and are compared with a base kernel that accounts for partial similarities. The (partial) similarity between these contexts provides an adaptive \emph{weight} for the contribution to the kernel of the exact matching performed on their corresponding selectors, from which the method derives its name. In this way, the authors aim to define a more parsimonious feature map that retains sufficient expressivity to capture higher-order information about the graph topology.

Unlike the previous approach, the family of optimal assignment kernels
(Section~\ref{sec:Optimal assignment kernels}) represent a departure
from the $\mathcal{R}$-convolution framework. Much like the ``all
substructure-pairs'' kernel, they consider a decomposition of graphs
into a set of substructures. However, rather than computing
$\kernel(\graph, \graph')$ by comparing every substructure in graph $\graph$
to every substructure in graph $\graph'$, they compare each substructure
in $\graph$ only to one substructure in $\graph'$. This requires learning
an \emph{optimal assignment} between substructures of  $\graph$ and
$\graph'$ so as to maximise the resulting value of $\kernel(\graph,
\graph')$. By construction, this approach alleviates the diagonal dominance problem. However, as we shall see, there are important restrictions in the type of base kernels that can be used in this scheme to guarantee that the resulting kernel is positive definite.

Deep graph kernels (Section~\ref{sec:Deep graph kernels}), the last approach we will discuss, take the idea behind weighted decomposition kernels one step further. The contribution of each individual dimension in the feature maps of highly expressive graphs kernels, such as those arising from the Weisfeiler-Lehman framework, could be weighted by a free parameter learnt from the data. If suitable weights are found, this would allow prioritizing the most predictive features while down-weighting those which are irrelevant for a specific task. As will be described later, the authors propose to accomplish this with a method inspired by recent advances in natural language processing.
%
\subsubsection{Weighted decomposition kernels}\label{sec:Weighted decomposition kernel framework}

In some applications, for example molecular classification, it makes sense to assign different weights to a specific $\mathcal{R}$-decomposition in order to accentuate a given property of a molecule, for example.
\citet{Menchetti05} thus developed a \emph{weighted} variant of the
\mbox{$\mathcal{R}$-convolution} framework to handle molecular
classification better~\citep{Ceroni07}.
The central idea is to decompose a graph $\graph$ into a certain
\emph{subgraph} $s_{\graph} \subseteq \graph $ and a \emph{context} $c_{\graph} \subseteq \graph$.
The subgraph is also referred to as a \emph{selector}, while the context
subgraph typically contains $s_{\graph}$; we will also use $s$ and $c$
to denote these variables when the corresponding graph $\graph$ is
clear.
Following the notation of $\mathcal{R}$-convolution, we will denote this
by $\mathcal{R}\left(s_{\graph}, c_{\graph}, \graph\right)$.
While the subgraphs are typically only compared using a Dirac delta kernel,
the contexts are supposed to be compared in terms of their
attributes~(in the original application of \citet{Ceroni07}, edges are
assigned attributes that refer to the chemical properties of the
corresponding bonds in a molecule, for example).
To this end, \citet{Ceroni07} introduce a novel set of kernels based on
probability distributions over attributes. In the following, we assume
that we are given two graphs $\graph$ and $\graph'$ with a set of $m$
attributes $\mathcal{A} := \left\{A_1, \dots, A_m\right\}$. This
notation can be seen as a generalization of the notation presented
in Definition~\ref{def:Attributed graph} on p.~\pageref{def:Attributed
graph}, which discusses a single attribute function. Here, we assume
that more than one attribute is present, but we leave the definition of
each attribute purposefully open.

\begin{defn}[Graph probability distribution kernel]
  Let $\rho\in\real$ be a scaling parameter and $p_i(\cdot),
  p_i'(\cdot)$ be probability distributions over the individual
  attributes in $\mathcal{A}$ for $\graph$ and $\graph'$ respectively.
  Assuming that the $i$th attribute only has $n_i$ distinct values,
  a family of \emph{graph probability kernels} is obtained as
  \begin{equation}
    \basekernel_i\left(\graph, \graph'\right) :=
      \sum_{j=1}^{n_i} p_i\left(j\right)^{\rho} p_i'\left(j\right)^{\rho}.
  \end{equation}
  For \emph{continuous} attributes, \citet{Menchetti05} note that
  previous work~\citep{Jebara04} provides a theoretical framework.
  \label{defn:Graph probability distribution kernel}
\end{defn}

In the previous definition, the topology of the graph is not
used---making it possible to perform a ``soft matching'' of smaller
substructures in kernels. This results in the following generic
framework.

\begin{defn}[Weighted decomposition kernel]
  Given a decomposition of $\graph$ and $\graph'$ into subgraphs $s_{\graph}$
  and $s_{\graph'}$ with corresponding contexts $c_{\graph}$ and $c_{\graph'}$,
  the \emph{weighted decomposition kernel} is defined as
  \begin{equation}
    \kernelWD\left(\graph, \graph'\right) =
      \sum_{
        \left(s,  c \right) \in \mathcal{R}^{-1}\left(\graph\right),
      }
      \sum_{
        \left(s', c'\right) \in \mathcal{R}^{-1}\left(\graph'\right)
      }
      \delta\left(s, s'\right)
        \sum_{i=1}^{m} \basekernel_i\left(c, c'\right),
  \end{equation}
  where $\basekernel_i\left(c, c'\right)$ refers to a kernel according
  to Definition~\ref{defn:Graph probability distribution kernel}.  \end{defn}

The choice of $s$ and $c$ depends on the application. \citet{Ceroni07},
for example, propose setting $c$ to a neighbourhood of fixed radius $k$,
comprising all vertices that are \emph{reachable} with paths of length
at most $k$, while $s$ is just the source vertex~(meaning that the Dirac
delta boils down to comparing vertex labels).

The computational efficiency of this kernel depends to a large extent on the
choice of subgraph and context. Given $l$ substructures in the pre-image of the
$\mathcal{R}$-convolution relationship, the kernel between two graphs can
be computed in $\landau{l^2}$ time, which can be reduced to linear time for
sparser indices~\citep{Menchetti05}.
\begin{figure}[h!]
    \centering
    \includegraphics[width=\textwidth]{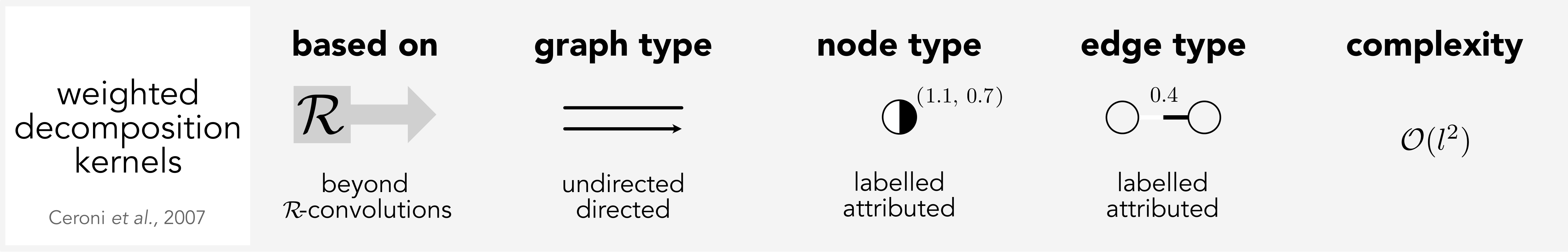}
\end{figure}

\subsubsection{Optimal assignment kernels}\label{sec:Optimal assignment kernels}

In some application domains, such as chemoinformatics, additional information for
graph nodes, is available. Molecules, for examples, might have additional information
about their structure attached. These structural information cannot always be easily
expressed in terms of continuous attributes. \citet{Froehlich05} thus developed a
kernel based on finding an optimal assignment between substructures of these graphs.
The underlying idea is reminiscent of $\mathcal{R}$-convolution
kernels~(see Section~\ref{sec:R-convolution kernels} for more
information) and requires a graph to be decomposable into a set of parts.
However, the original optimal assignment kernel is not guaranteed to be
positive semi-definite for all choices of substructure kernels~\citep{Vert08}.
\citet{Kriege16} showed that certain base kernels lead to positive semi-definite
kernels. Following their terminology, let $\mathcal{X}$ be a set~(such as all
potential vertex labels), $X, Y \subseteq \mathcal{X}$ subsets of the
same cardinality, and $\mathfrak{B}(X, Y)$ the set of all bijections
between $X$ and $Y$. This leads to the optimal assignment kernel.

\begin{defn}[Optimal assignment kernel]
  Given a base kernel $\basekernel$ between individual elements of
  $\mathcal{X}$, such as a vertex kernel, the \emph{optimal assignment
  kernel} is defined as
  \begin{equation}
    \kernelOA\left(X, Y\right) := \max_{B \in \mathfrak{B}(X, Y)} \sum_{x \in X} \basekernel\left(x,B\left(y\right)\right),
  \end{equation}
  where $\basekernel(x, y) = 0$ is used  to account for subsets of
  different cardinalities.
\end{defn}

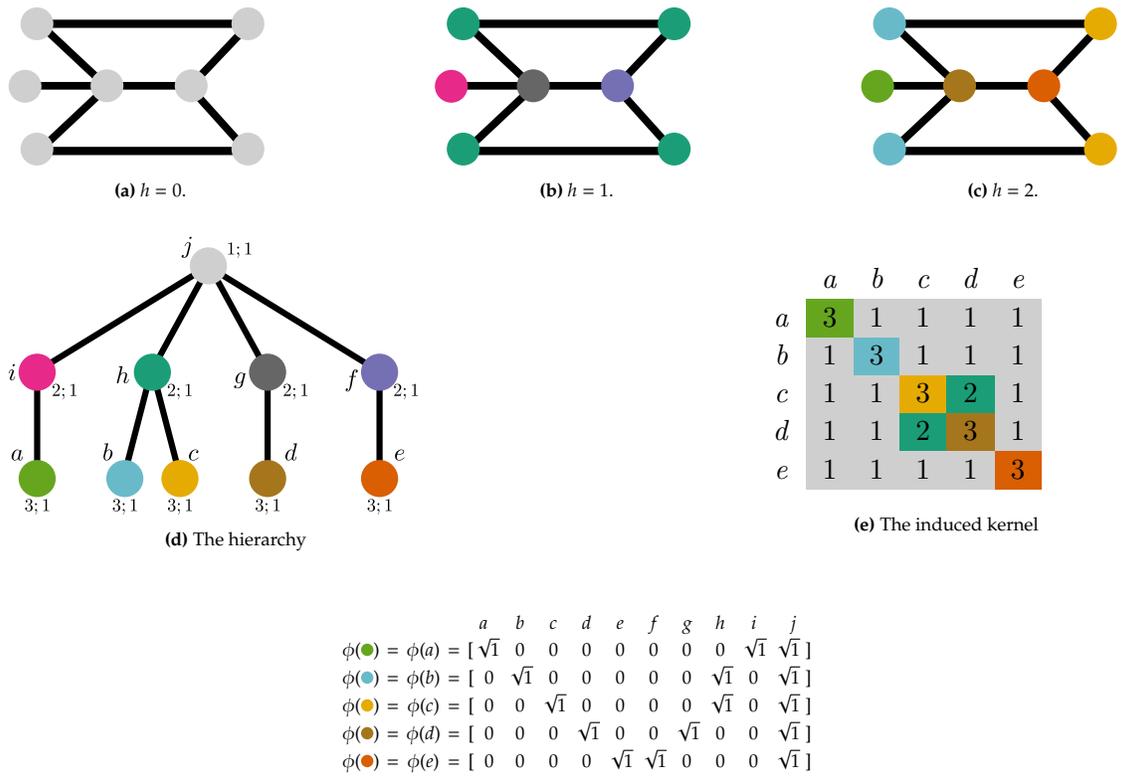
\begin{figure}[p]
    \centering
    \begin{subfigure}{0.25\textwidth}
        \includegraphics[width=0.9\linewidth]{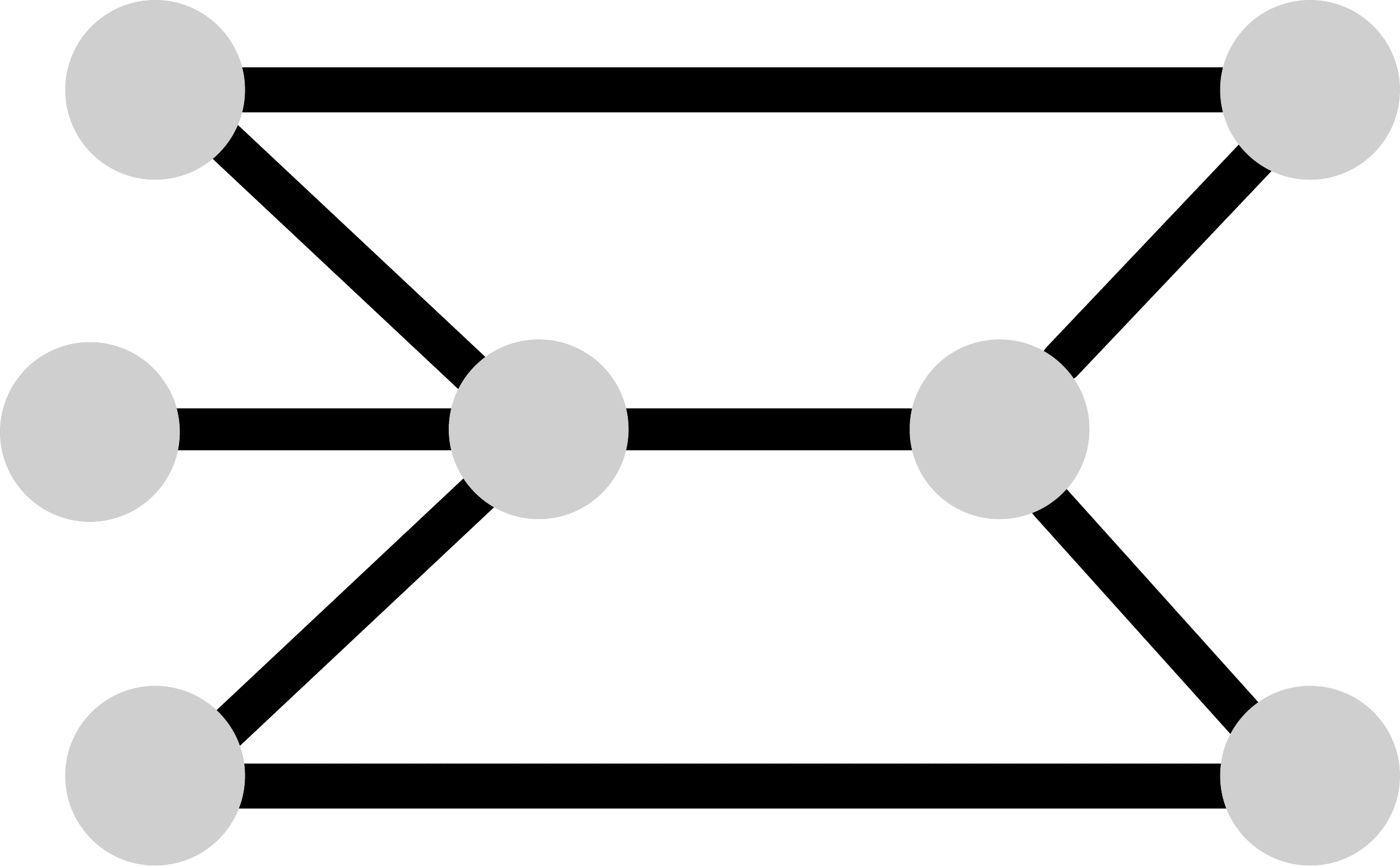}
        \caption{$h=0$.}
        \label{fig:wloa1}
    \end{subfigure}
    \hfill    
    \begin{subfigure}{0.25\textwidth}
        \includegraphics[width=0.9\linewidth]{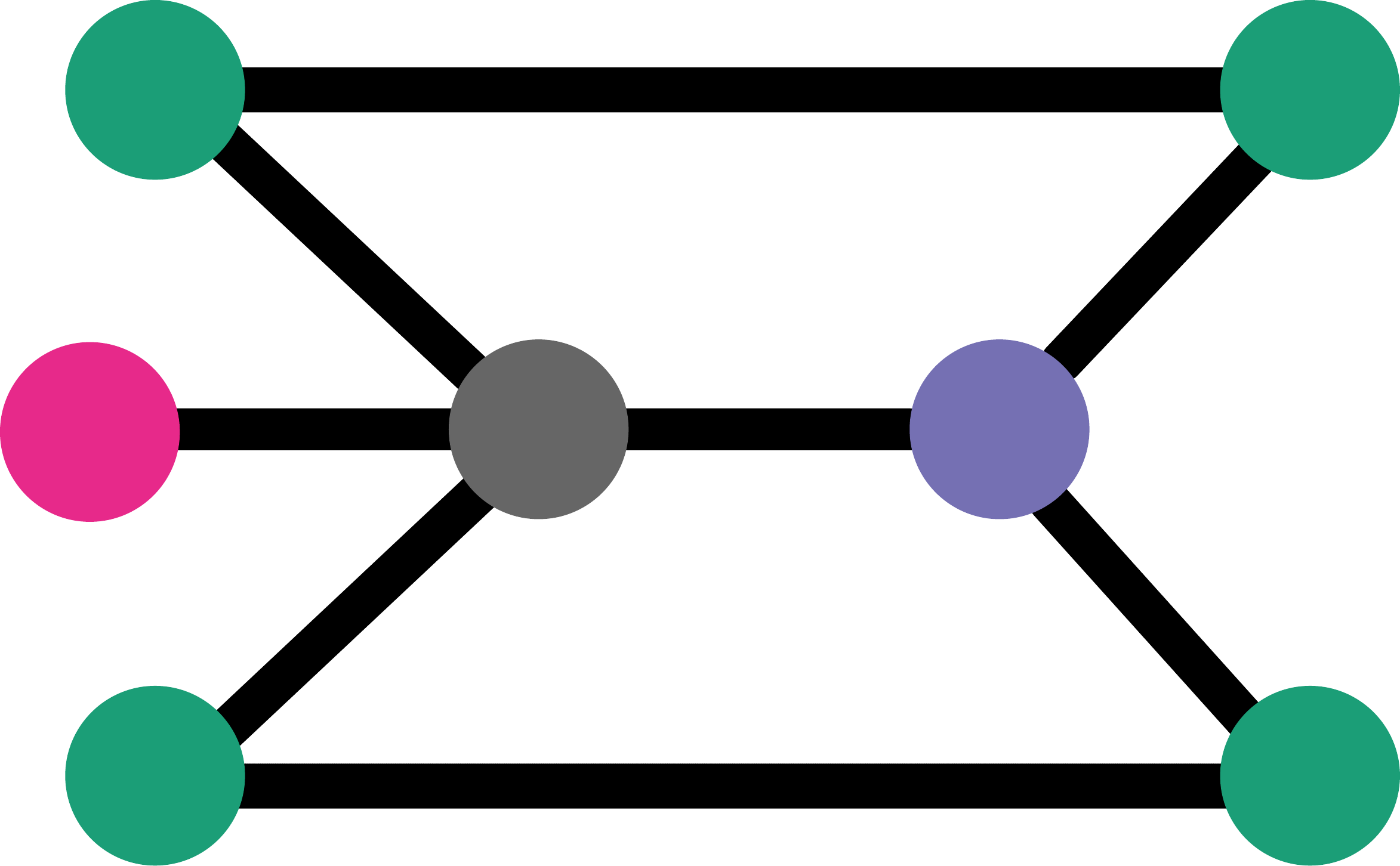}
        \caption{$h=1$.}
        \label{fig:wloa2}
    \end{subfigure}
    \hfill
    \begin{subfigure}{0.25\textwidth}
        \includegraphics[width=0.9\linewidth]{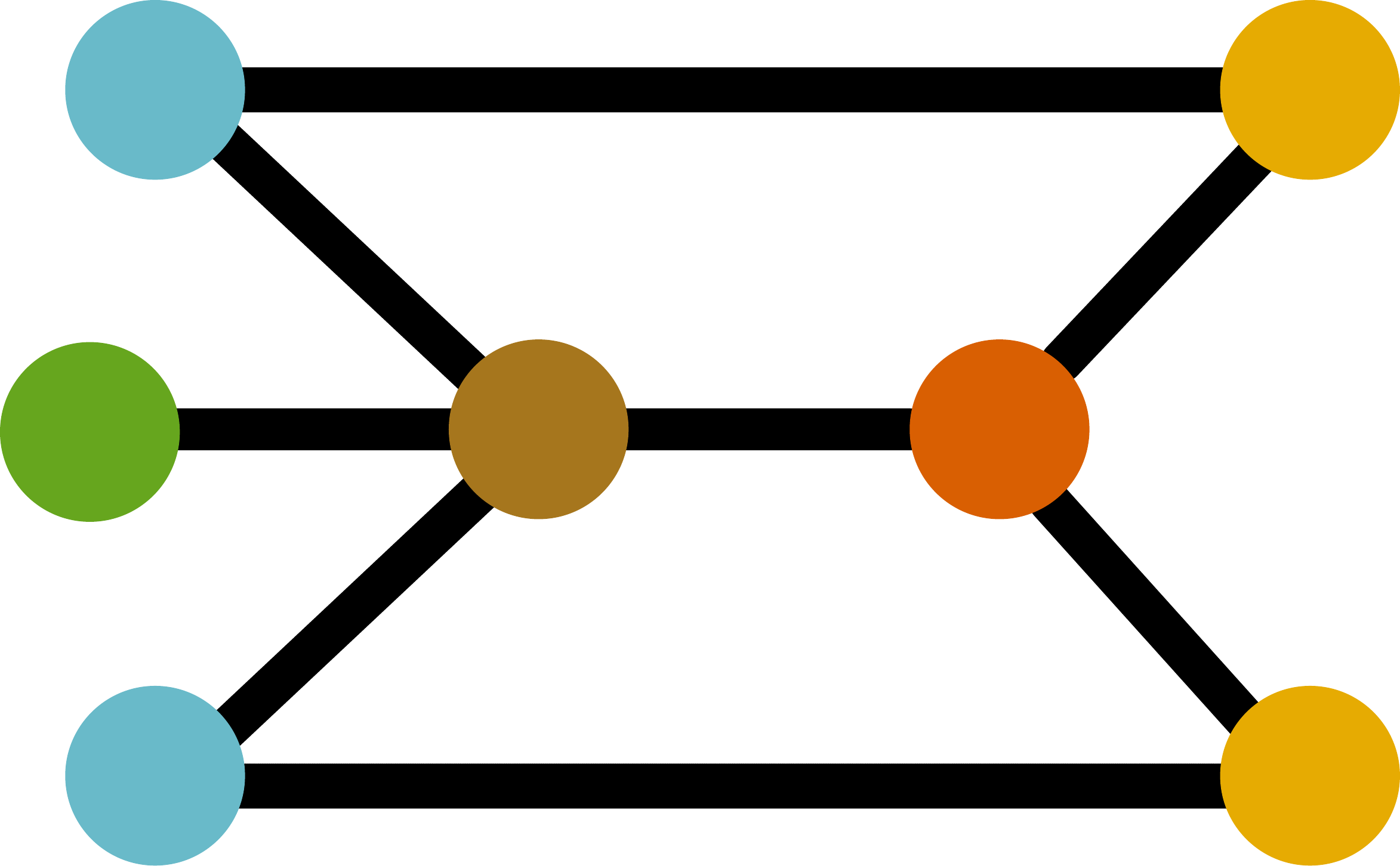}
        \caption{$h=2$.}
        \label{fig:wloa3}
    \end{subfigure}
    \par\bigskip
    \begin{subfigure}{0.4\textwidth}
        \includegraphics[width=0.9\linewidth]{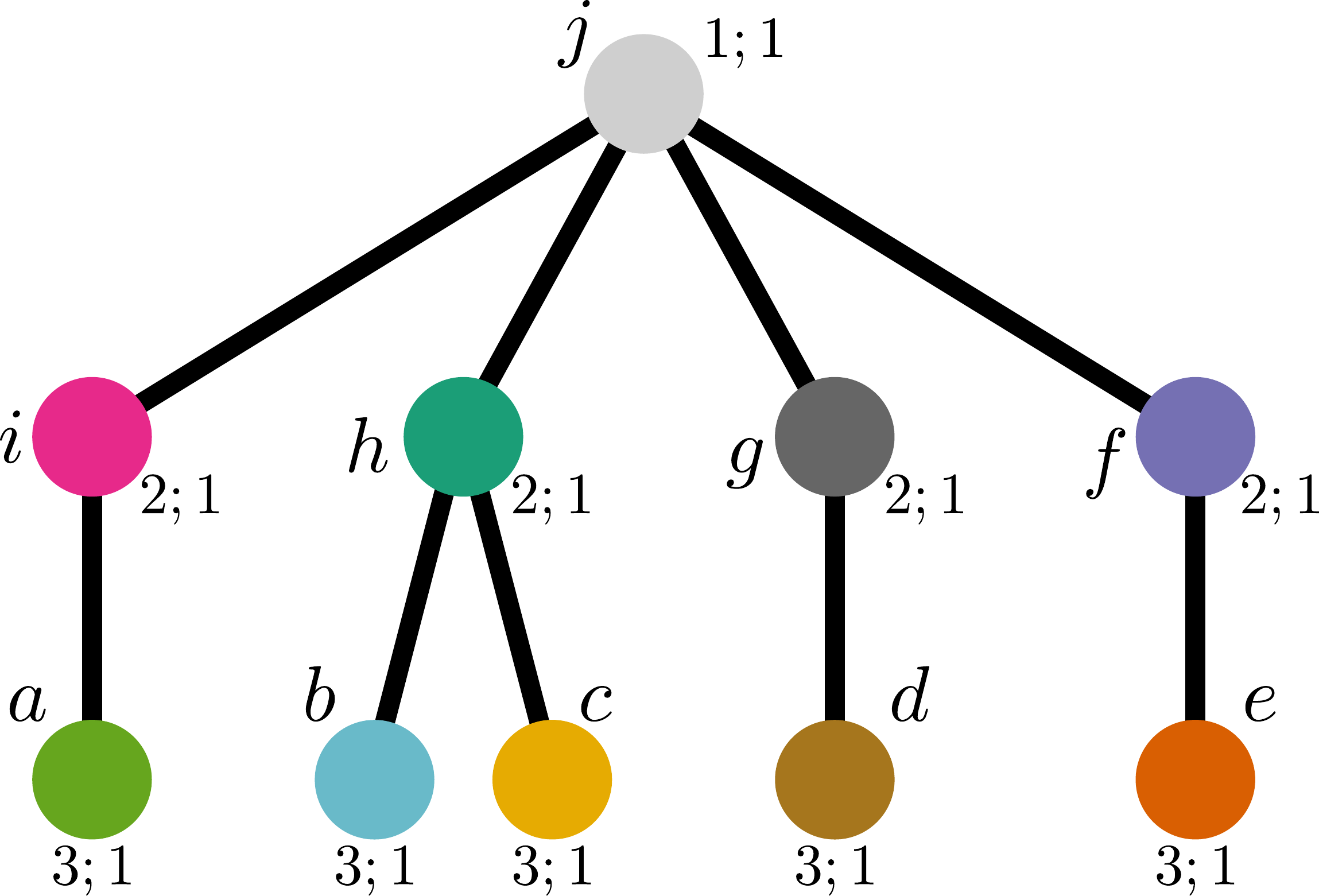}
        \caption{The hierarchy}
        \label{fig:wloa4}
    \end{subfigure}
    \hfill
    \begin{subfigure}{0.35\textwidth}
      \iffinal
        \includegraphics{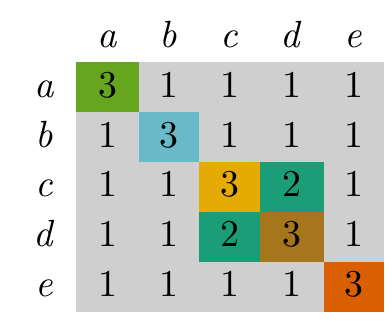}
      \else
        \begin{tikzpicture}
        \node (tbl) {
        \begin{tabularx}{.9\linewidth}{cccccc}
        \scriptsize
        & \emph{a} & \emph{b} & \emph{c} & \emph{d} & \emph{e} \\
        \emph{a} & \cellcolor[RGB]{102,166,30}3 & \cellcolor[RGB]{207,207,207}1 & \cellcolor[RGB]{207,207,207}1 & \cellcolor[RGB]{207,207,207}1 & \cellcolor[RGB]{207,207,207}1\\
        \emph{b} & \cellcolor[RGB]{207,207,207}1 & \cellcolor[RGB]{105,186,201}3 & \cellcolor[RGB]{207,207,207}1 & \cellcolor[RGB]{207,207,207}1 & \cellcolor[RGB]{207,207,207}1\\
        \emph{c} & \cellcolor[RGB]{207,207,207}1 & \cellcolor[RGB]{207,207,207}1 & \cellcolor[RGB]{230,171,2}3 & \cellcolor[RGB]{27,158,119}2 & \cellcolor[RGB]{207,207,207}1  \\
        \emph{d} & \cellcolor[RGB]{207,207,207}1 & \cellcolor[RGB]{207,207,207}1 & \cellcolor[RGB]{27,158,119}2 & \cellcolor[RGB]{166,118,29}3 & \cellcolor[RGB]{207,207,207}1\\
        \emph{e} & \cellcolor[RGB]{207,207,207}1 & \cellcolor[RGB]{207,207,207}1 & \cellcolor[RGB]{207,207,207}1 & \cellcolor[RGB]{207,207,207}1 & \cellcolor[RGB]{217,95,2}3 \\
        \end{tabularx}};
        \end{tikzpicture}
      \fi
        \caption{The induced kernel}
        \label{fig:wloa5}
        
    \end{subfigure}
    \par\bigskip
    \begin{subfigure}{0.45\textwidth}
        \small
        \begin{align*}
        \begin{smallmatrix}
        & & & & a & b & c & d & e & f & g & h & i & j \\
        \phi(\tikz {\fill[green1] (0,0) circle[radius=.5ex];}) &=&  \phi(a) &= &[ \sqrt{1} & 0 & 0 & 0 & 0 & 0 & 0 & 0 & \sqrt{1} & \sqrt{1} \;] \\
        \phi(\tikz {\fill[blue1] (0,0) circle[radius=.5ex];}) &=&  \phi(b) &= &[ \;\; 0 \; & \sqrt{1} & 0 & 0 & 0 & 0 & 0 & \sqrt{1} & 0 & \sqrt{1} \;] \\
        \phi(\tikz {\fill[yellow1] (0,0) circle[radius=.5ex];}) &=&  \phi(c) &= &[ \;\; 0 \; & 0 & \sqrt{1} & 0 & 0 & 0 & 0 & \sqrt{1} & 0 & \sqrt{1} \;] \\
        \phi(\tikz {\fill[brown1] (0,0) circle[radius=.5ex];}) &=&  \phi(d) &= &[ \;\; 0 \; & 0 & 0 & \sqrt{1} & 0 & 0 & \sqrt{1} & 0 & 0 & \sqrt{1} \;] \\
        \phi(\tikz {\fill[orange1] (0,0) circle[radius=.5ex];}) &=&  \phi(e) &= & [ \;\; 0 \; & 0 & 0 & 0 & \sqrt{1} & \sqrt{1} & 0 & 0 & 0 & \sqrt{1} \;] \\
        \end{smallmatrix}
        \end{align*}
        \caption{The representation of the leaf nodes.}
        \label{fig:wloa6}
    \end{subfigure}
    \caption{An illustration of the Weisfeiler--Lehman optimal
      assignment kernel. Assume a graph with unlabelled nodes, and its subsequent labelling
    according to the Weisfeiler--Lehman algorithm for iteration $h=0$
  (\subref{fig:wloa1}), $h=1$ (\subref{fig:wloa2}), and $h=2$
(\subref{fig:wloa3}). This relabelling process provides a hierarchy
(\subref{fig:wloa4}), where each vertex $\vertex$ is annotated with
$w(\vertex); \omega(\vertex)$, representing the weight and additive
weight of $v$ respectively. We assume an additive weight of $0$, which
provides values for the hierarchy once we assign an initial weight to
the root node $w(j)=0$. This hierarchy induces a kernel
(\subref{fig:wloa5}) on the leaf nodes, where $k(x, y) = w(c)$, if
$c$ is the lowest common ancestor of $x$ and $y$, shown by the colour in
the kernel matrix. (\subref{fig:wloa6}) shows the feature map for each
leaf node, where each element in the feature map for a node $v$ has
a value $\sqrt{\omega(m)}$ in the feature map if $m$ is on the path
between $v$ and the root node $j$. A graph $\graph$ is then represented
by the histogram of the sum of the element wise squared feature maps for
each node in (\subref{fig:wloa3}), \ie\ $H(\graph) = [1 \; 2 \; 2 \;
0 \; 1 \; 1 \; 1 \; 1 \; 4 \; 1 \; 7]$. Given a second (undepicted) graph
$\graph'$ with histogram $H(\graph')$, the histogram intersection kernel is
then just the sum of the element wise minimum of each component of the two
histograms.}
\label{fig:wloa}
\end{figure}

\citet{Kriege16} show~(by virtue of defining strong kernels, \ie\
a function $\basekernel\colon \mathcal{X} \times \mathcal{X} \to
\mathbb{R}_{\geq 0}$ such that $\basekernel(x, y) \geq
\min\{\basekernel(x, z), \basekernel(y, z)$ for all $x, y, z \in
\mathcal{X}$; notice that in a strong kernel, an object is most similar
to itself, such that $\basekernel(x, x) \geq \basekernel(x, y)$ for
objects $x$, $y$) that certain base kernels, namely the ones arising
from hierarchical partitions of the kernel domain, lead to positive
semi-definite optimal assignment kernels. This gives rise to a general
framework for kernel construction, leading to a \emph{vertex optimal
assignment kernel}, for example, which employs a Dirac delta kernel on
vertex labels. As this kernel, as well as a related \emph{edge
optimal assignment kernel}, already demonstrate better predictive
performance than regular vertex label and edge label kernels, the family
of Weisfeiler--Lehman optimal assignment kernels is particularly
interesting.

Following the notation for the Weisfeiler--Lehman iteration~(see
Section~\ref{sec:Weisfeiler--Lehman kernel}), the
\emph{Weisfeiler--Lehman optimal assignment kernel} is defined on the
vertices of a graph while using a base kernel that evaluates the
compressed subtree labels. For $h$ iterations, the kernel between two
vertices $u$ and $v$ is thus defined as
\begin{equation}
  \basekernel\left(u, v\right) := \sum_{i = 0}^{h} \kernel_{\delta}\left(\sigma_u^{(i)}, \sigma_v^{(i)}\right),
\end{equation}
where $\kernel_{\delta}$ denotes a Dirac delta kernel. We provide an
visualisation of the Weisfeiler--Lehman optimal assignment kernel in
Figure~\ref{fig:wloa}.

In addition to its highly favourable predictive
performance~\citep{Kriege16}, this kernel is advantageous because its
calculation can be done extremely efficiently using histogram
intersection. The computational complexity of evaluating the kernel for
two graphs is thus asymptotically not larger than that of the
Weisfeiler--Lehman labelling operation, leading to $\landau{hm}$, where
$m$ is the maximum number of edges.
\begin{figure}[h!]
    \centering
    \includegraphics[width=\textwidth]{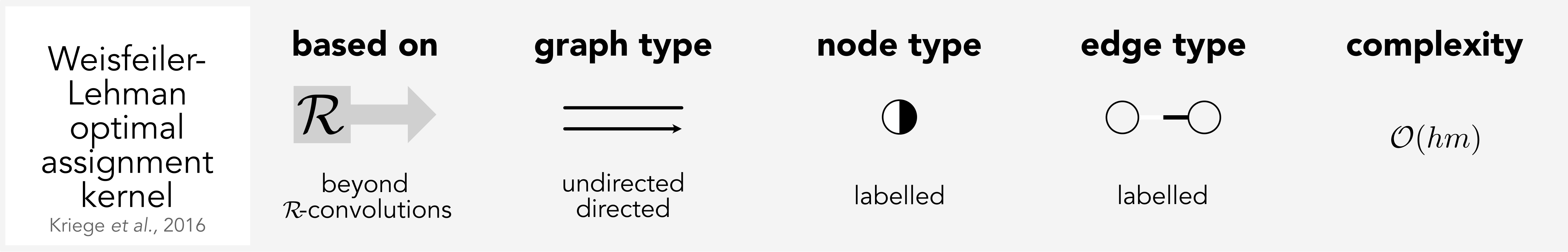}
\end{figure}
%
\subsubsection{Deep graph kernels}\label{sec:Deep graph kernels}

Inspired by new models in natural language processing that are capable of
generating word embeddings with semantic meanings~\citep{Mikolov13},
\citet{Yanardag15} developed a framework that applies the same reasoning
to substructures that arise from graph kernels. More precisely, given
a graph kernel representation in terms of an inner product of feature
vectors, \ie\
\begin{equation}
  \kernel\left(\graph, \graph'\right) = \featurevector\left(\graph\right)^{\top} \featurevector\left(\graph'\right),
\end{equation}
where $\graph$ and $\graph'$ are two graphs, they propose computing a
diagonal matrix $\mathcal{D} = \diag\left(d_1, d_2, \dots\right)$, of
an appropriate size.
The previous equation is then augmented to include weights of each
individual substructure, such that
\begin{equation}
  \kernel_{\text{deep}}\left(\graph, \graph'\right) := \featurevector\left(\graph\right)^{\top} \mathcal{D} \featurevector\left(\graph'\right),
\end{equation}
which \citet{Yanardag15} denote as a ``deep'' variant of the previous
kernel~(equivalently, the previous equation can be seen as a simple
reweighting similar to the kernels in Section~\ref{sec:Weighted decomposition kernel framework}
on p.~\pageref{sec:Weighted decomposition kernel framework}).

The intuition behind this approach is to consider substructures of
a graph kernel as ``words'' whose contexts can be calculated with
standard methods~\citep{Mikolov13}. This requires defining a proper
co-occurrence relationship between the substructures of different kernels,
and \citet{Yanardag15} define such relationships for some common kernels, namely
the shortest-path kernel~(Section~\ref{sec:Shortest-path kernel}),
the graphlet kernel~(Section~\ref{sec:Graphlet kernel}),
and the Weisfeiler--Lehman subtree kernel~(Section~\ref{sec:Weisfeiler--Lehman kernel}).
While the computational complexity of this approach is higher because of
the additional embedding calculation step, the ``deep variants'' of some
kernels are reported to achieve slightly higher
classification accuracies.
\begin{figure}[h!]
    \centering
    \includegraphics[width=\textwidth]{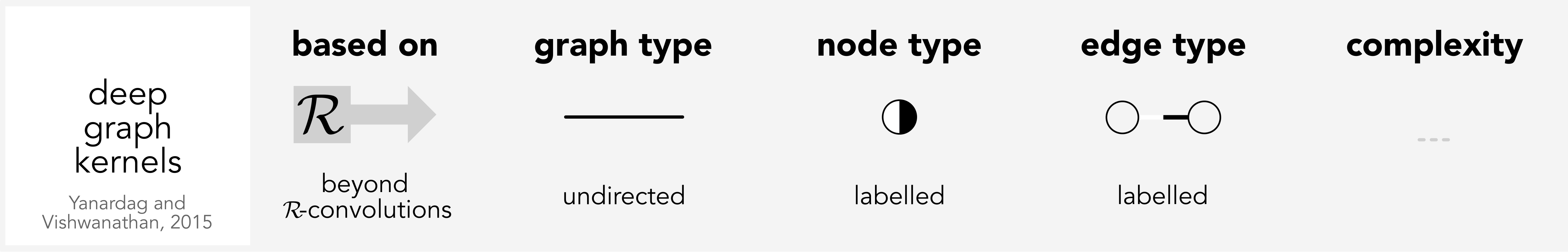}
\end{figure}

\subsubsection{Core based kernel framework}\label{sec:Core based kernel}

\citet{nikolentzos2018kcore} proposed a new framework of graph kernels
by sequentially comparing nested subgraphs formed using a classic graph
theoretic decomposition called a $k$-core decomposition. They layer
such a decomposition with a base kernel, such as the shortest path or
Weisfeiler--Lehman to generate a multi-scale view of the base kernel
evaluated alongside the decomposition, which they find improves
performance compared to the base kernels by providing a more nuanced view
of similarity across different scales of the graphs. Before defining the
kernel, we must first formally introduce a few concepts.
\begin{defn}[$k$-core]
  A \emph{$k$-core} of graph $\graph$, denoted $\graph_k = (\vertices_k,
  \edges_k)$, is the maximal subgraph of $\graph$ such that $\degree(\vertex)
  \geq k$ for all $\vertex \in \vertices_k$, where $\degree(v)$ is the degree
  of a vertex $\vertex$. The nodes in the $k$-core need not all be
  connected, and thus $\graph_k$ may contain multiple disconnected
  components.
\end{defn} 
\begin{defn}[Degeneracy]
    The \emph{degeneracy} of a graph $\graph$, $\delta^*(\graph)$, is
    determined by the largest non-empty $k$-core subgraph of $\graph$,
    \ie\ $\delta^{*}(\graph) = \max_k \colon \graph_k \neq
    \emptyset$.  
\end{defn}
\begin{defn}[$k$-core decomposition]
 The \emph{$k$-core decomposition} of $\graph$ is the nested
 sequence of $k$-cores from $k=0, \ldots, \delta^*(G) \colon \graph_0
 \subseteq \graph \subseteq \cdots \subseteq \graph_{\delta^*(\graph)}$.
 We define the set of the $k$-core subgraphs of $\graph$ as
 $\mathcal{K_{\graph}} = \{G_0, \ldots, \graph_{\delta^*(G)} \}$.
\end{defn}
Fortunately, $\mathcal{K}$ is quite efficient to compute, since one can
proceed sequentially, starting with $\graph_0$, for for each $k$ remove
any node $\vertex$ if $\degree(\vertex) < k$. 
\begin{defn}[Core based kernel]
  Given two graphs $\graph$ and $\graph'$ and their respective $k$-core
  decompositions $\mathcal{K}$ and $\mathcal{K}'$, the \emph{core based
  kernel} is defined as:
  \begin{equation}
      \kernelkcore(\graph, \graph') = \sum_{k=0}^{p} \basekernel(\graph_k, \graph_k'),
  \end{equation}
  where $\basekernel$ can be any valid base kernel, such as the shortest
  path kernel or Weisfeiler--Lehman kernel, and $p=\min(\delta^*(\graph),
  \delta^*(\graph'))$.
\end{defn}
Due to the efficiency of the $k$-core decomposition, which can be
achieved in linear time, the overall complexity for a pair of graphs
hinges upon the complexity of the chosen base kernel $\basekernel$.
\begin{figure}[h!]
    \centering
    \includegraphics[width=\textwidth]{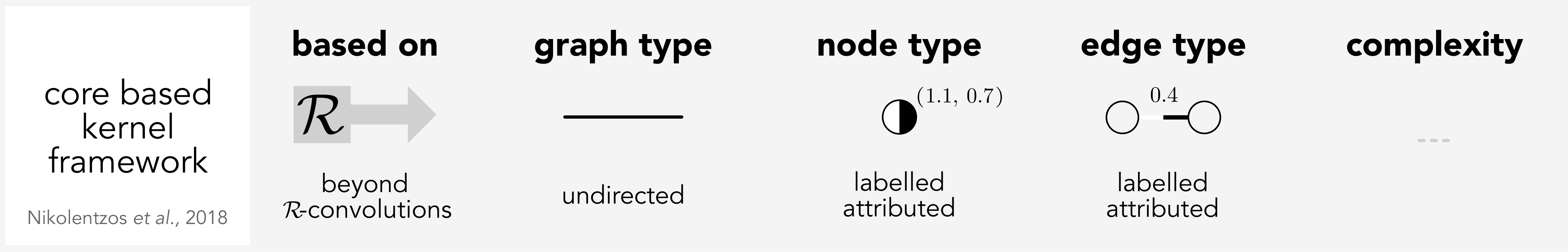}
\end{figure}

\section{Conclusion}

This chapter provided an overview of the ``zoo'' of available graph
kernels. It is clear that, due to the closure properties of kernel
functions, even more graph kernels can be constructed from the existing
ones. Certain overarching themes emerged though, one of the most common
being the idea of enumerating certain substructures---such as shortest
paths or subgraphs---and determine similarity by means of their
co-occurrence in two graphs.

Graph kernels continue to be an active research topic, with recent
publications focussing on topological attributes of
graphs~\citep{Rieck19}, or using more advanced concepts from
\emph{optimal transport theory}, such as Wasserstein
distances~\citep{Togninalli19}. We do not discuss these kernels in detail for
reasons of brevity.

Subsequently, we will discuss how to properly \emph{navigate} this
zoo, \ie\ we will discuss similarities in the practical performance of
certain graph kernels, which will make it easier to pick a suitable one
for a given application.

\chapter{Experimental evaluation of graph kernels}\label{chap:Experiments}

While the previous chapters provided a thorough overview of the rich field of
graph kernels, this chapter will focus on their practical performance. We are
mostly interested in
\begin{inparaenum}[(i)]
  \item analysing and explaining the empirical behaviour of graph kernels by
    means of numerous benchmark data sets,
  \item discussing differences and commonalities as well as other properties of
    the benchmark data sets, and
  \item ultimately providing some much-needed guidance to choose a suitable
    graph kernel in practice.
\end{inparaenum}
To this end, we provide a thorough experimental setup for assessing the
performance of individual graph kernels in a fair and comparative setup.
The insights that we gain along the way will be used to inform our predictions
and comments on future directions for
\begin{inparaenum}[(i)]
  \item potential applications,
  \item the field of graph kernels, and
  \item requirements for benchmark data sets.
\end{inparaenum}

\section{Data sets}

\begin{figure}[tbp]
  \centering
  \iffinal
  \includegraphics{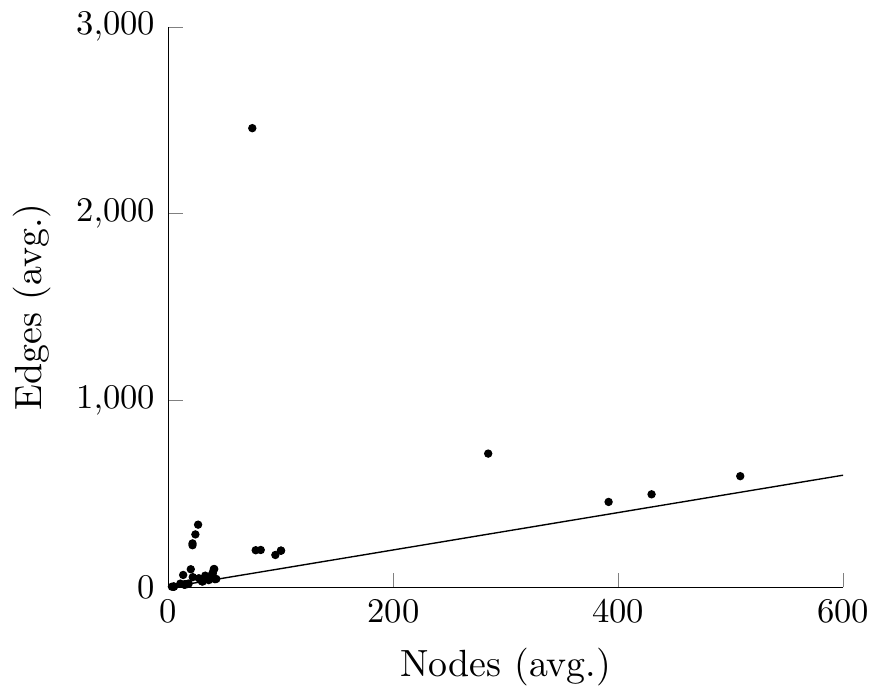}
  \else
    \begin{tikzpicture}
      \begin{axis}[%
        axis x line*     = bottom,
        axis y line*     = left,
        enlargelimits    = true,
        xlabel           = {Nodes~(avg.)},
        xtick            = {0, 200, 400, 600, 800, 1000, 1200, 1400},
        ylabel           = {Edges~(avg.)},
        xmax             = 600,
        ymin             = 0,
        ymax             = 3000,
        tick label style = {font = \small},
        enlargelimits    = false,
      ]
        \addplot[
          only marks,
          mark size = 1pt,
        ]
          table[%
            col sep = comma,
            x       = avg_nodes,
            y       = avg_edges,
          ] {Data/Summary_statistics.txt};

        \addplot[domain={0:600}] {x};
      \end{axis}
    \end{tikzpicture}
  \fi
    \caption{%
    A visualisation of the average number of nodes and the average number of
    edges for each data set. Most of the data sets are extremely \emph{sparse},
    featuring only a small number of nodes and edges on average.
  }
  \label{fig:Summary statistics}
\end{figure}
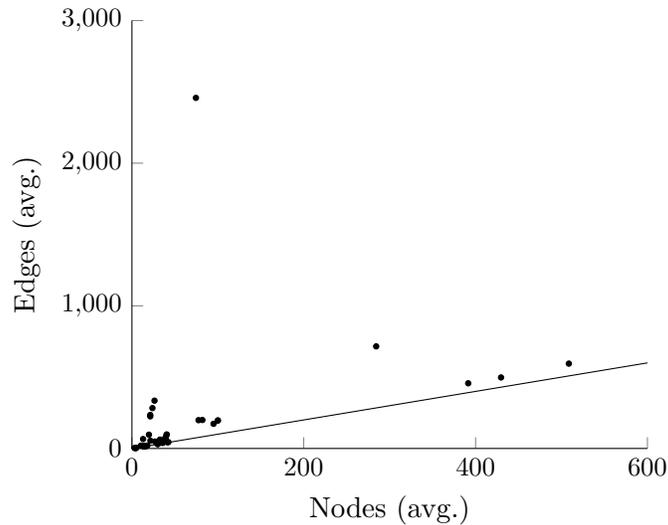

We use the data sets provided by \citet{KKMMN16}. They consist of more
than 50 different graph data sets of varying sizes and complexity.
Table~\ref{tab:Summary statistics} lists the data sets, along with some summary
statistics.
We observe that the graph data sets tend to be very sparse, \ie\ their number
of edges is roughly of the same order as their number of vertices. While this
is not a issue per se, it can have an influence on the selection of a graph
kernel. Neighbourhood-based approaches, for example, may suffer from reduced
performance---both in the computational and in the predictive sense---when
dealing with \emph{dense} graphs: as the density of a graph approaches that of
a complete graph, differences between individual node neighbourhoods start to
become indiscernible.
Figure~\ref{fig:Density distribution} depicts the distribution of density
values, whereas Figure~\ref{fig:Summary statistics} shows the average number of
nodes versus the average number of edges of each data set in order to give
a visual summary.

Moreover, we can see in the table that graphs with either node or edge
attributes are under-represented, as most data sets contain labels but
no continuous attributes.
Of the 41 data sets in this section, 27 contain node labels while 12 have
edge labels. Except for a single case, namely \texttt{COIL-DEL}, there
are no data sets that have edge labels but no node labels. \re{If
a data set did not have node labels, we generated a node label using the
degree of the node. For data sets without edge labels, we generated an
edge label as the sorted concatenation of the node labels for the nodes
incident to the edge. In the event that the data set has no node labels
or edge labels, the edge label is assigned using the node labels that
were created based on the node degree.}
The absence or presence of certain label or attributes might limit the applicability of certain graph kernels and we shall
discuss the implications of this later on when we categorise the data
sets according to which labels and attributes each has.
\paragraph{Excluded data sets}
We excluded several data sets---such as the collection of \texttt{Tox\_21} data
sets---that are present in the original repository from the subsequent analysis
because their size precludes running a sufficiently large number of graph
kernels on them. Moreover, we removed the \texttt{FIRSTMM\_DB} data set because
its small size makes it impossible to apply our training procedure that we describe
in Section~\ref{sec:Training procedure}.
\subsection{Data set categorisation}

Before discussing our experimental setup, we will first categorise the
data sets we are working with and, where applicable, explain some of their common
properties. Following the categorisation of \citet{Morris2020}, the data sets included can be grouped into five primary groups: 
\begin{inparaenum}[(i)]
  \item social networks,
  \item small molecules,
  \item bioinformatics, 
  \item computer vision, and
  \item synthetic.
\end{inparaenum}
We will now introduce each group in turn and indicate which data sets
belong in each category. 
\paragraph{Social networks} Several data sets, such as \texttt{COLLAB}, \texttt{IMDB-BINARY}, \texttt{IMDB-MULTI},
\texttt{REDDIT\-BINARY}, \texttt{REDDIT-MULTI-5K} and
\texttt{REDDIT-MULTI\-12K} represent a kind of social network. While they
often feature large graphs (with many nodes), they tend to have relatively few
edges, and therefore low density. \texttt{COLLAB} is a notable
exception to this, with a density of $0.51$. All the data sets that we
used in this category are fully unlabelled, meaning they do not have any
node or edge labels or attributes.

\paragraph{Small molecules} Another important category of the benchmark
data sets are small molecules. This group contains the following data
sets: \texttt{AIDS}, \texttt{BZR}, \texttt{BZR\_MD}, \texttt{COX2}, \texttt{COX2\_MD},
\texttt{DHFR}, \texttt{DHFR\_MD}, \texttt{ER\_MD}, \texttt{FRANKEN\-STEIN},
\texttt{MUTAG}, \texttt{Mutagenicity}, \texttt{NCI1}, \texttt{NCI109},
\texttt{PTC\_FM}, \texttt{PTC\_FR}, \texttt{PTC\_MM}, and
\texttt{PTC\_MR}. While these graphs are often small (in terms of number
of nodes) and often have a similar number of nodes and edges,
the data sets \texttt{BZR\_MD}, \texttt{COX2\_MD}, \texttt{DHFR\_MD} and
\texttt{ER\_MD} are a notable exception to that, since they are fully
connected. This presents a challenge for methods based on neighbourhood
aggregation, since each node is connected to all other nodes in the
graphs. All the data sets in the small molecules category contain some
kind of node or edge labels or attributes.

\paragraph{Bioinformatics} The third group of data sets falls under the
grouping of bioinformatics. Several data sets, such as \texttt{DD},
\texttt{ENZYMES}, \texttt{PROTEINS}, and \texttt{PROTEINS\_full} are
data sets of proteins, whereas \texttt{KKI}, \texttt{OHSU}, and
\texttt{Peking\_1} are representations of a brain. All of these data
sets have node labels (\texttt{ENZYMES} and \texttt{PROTEINS\_full} also
have node attributes), but all lack any information about their edges.

\paragraph{Computer vision} The next category falls within
the field of computer vision, where the \texttt{COIL-DEL},
\texttt{COIL-RAG}, \texttt{Letter-high}, \texttt{Letter-low},
 \texttt{Letter-med}, \texttt{MSRC\_9}, \texttt{MSRC\_21} and
\texttt{MSRC\_21C} data sets represent graphs constructed from images.
This group of data sets features a particularly large diversity among the
data sets in terms of graph size and which node and label attributes
they support, and we therefore refer the reader to
Table~\ref{tab:Summary statistics}.

\paragraph{Synthetic} Finally, there are a few datasets which were
synthetically created, namely \texttt{SYNTHETIC}, \texttt{SYNTHETICnew},
and \texttt{Synthie}. All graphs were endowed with node attributes by
design, since there are relatively few data sets containing such attributes.
%

\section{Experimental setup}
%
Having explained the data sets we use, we now detail our experimental
setup, which uses two phases:
\begin{inparaenum}[(i)]
  \item computation of kernel matrices for each graph kernel, and
  \item classifier training based on the set of kernel matrices.
\end{inparaenum}
Prior to discussing the details of these two steps, we discuss the included
kernels.

\subsection{Inclusion criteria}
%
As we have seen in Chapter~\ref{chap:Kernels}, there is a plethora of graph
kernels. To make this review feasible, we had to restrict the computations to
a subset of the existing graph kernels literature. \re{In the following
set of experiments, we tried to include a representative set of graph
kernels from the ones whose code is openly available. Specifically, we
made sure to include at least one kernel from each of the categories in
Chapter~\ref{chap:Kernels}.}
This comprises some graph kernels contained in the \texttt{graphkernels}
software package for Python~\citep{Sugiyama17}, \re{others from the \texttt{graKel} software package for Python~\citep{grakel2018}} as well as several other
kernels for which we were able to obtain an implementation. 
%
\begin{landscape}
\centering
  \setlength{\tabcolsep}{2.5pt}
  \footnotesize
  \begin{longtable}{>{\tt}lrrrS[table-format=4.2]S[table-format=4.2]rrrr}
    \caption{%
      Summary statistics of the data sets used for the subsequent
      experiments. Data set sizes vary but most of the graphs are
      comparatively small.
    }
    \label{tab:Summary statistics}\\
    \toprule
    \normalfont{Data set} & \normalfont{Graphs} & \normalfont{Classes} & \normalfont{Density} & {Nodes~(avg.)} & {Edges~(avg.)} & {N.\ attr.} &  {E.\ attr.} & {Node labels} & {Edge labels}\\
    \midrule
    \endhead
   AIDS              &       2000 &   2 &        0.19 &          15.69 &         16.20 &   4 &    &  \tableyes  & \tableyes  \\
   BZR               &        405 &   2 &        0.06 &          35.75 &         38.36 &   3 &    &  \tableyes  & \tableno   \\
   BZR\_MD           &        306 &   2 &        1.00 &          21.30 &        225.06 &     &  1 &  \tableyes  & \tableyes  \\
   COIL-DEL          &       3900 & 100 &        0.33 &          21.54 &         54.24 &   2 &    &  \tableno   & \tableyes  \\
   COIL-RAG          &       3900 & 100 &        0.92 &           3.01 &          3.02 &  64 &  1 &  \tableno   & \tableno   \\
   COLLAB            &       5000 &   3 &        0.51 &          74.49 &       2457.78 &     &    &  \tableno   & \tableno   \\
   COX2              &        467 &   2 &        0.05 &          41.22 &         43.45 &   3 &    &  \tableyes  & \tableno   \\
   COX2\_MD          &        303 &   2 &        1.00 &          26.28 &        335.12 &     &  1 &  \tableyes  & \tableyes  \\
   DD                &       1178 &   2 &        0.03 &         284.32 &        715.66 &     &    &  \tableyes  & \tableno   \\
   DHFR              &        756 &   2 &        0.05 &          42.43 &         44.54 &   3 &    &  \tableyes  & \tableno   \\
   DHFR\_MD          &        393 &   2 &        1.00 &          23.87 &        283.02 &     &  1 &  \tableyes  & \tableyes  \\
   ENZYMES           &        600 &   6 &        0.16 &          32.63 &         62.14 &  18 &    &  \tableyes  & \tableno   \\
   ER\_MD            &        446 &   2 &        1.00 &          21.33 &        234.85 &     &  1 &  \tableyes  & \tableyes  \\
   FRANKENSTEIN      &       4337 &   2 &        0.17 &          16.90 &         17.88 & 780 &    &  \tableno   & \tableno   \\
   IMDB-BINARY       &       1000 &   2 &        0.52 &          19.77 &         96.53 &     &    &  \tableno   & \tableno   \\
   IMDB-MULTI        &       1500 &   3 &        0.77 &          13.00 &         65.94 &     &    &  \tableno   & \tableno   \\
   KKI               &         83 &   2 &        0.18 &          26.96 &         48.42 &     &    &  \tableyes  & \tableno   \\
   Letter-high       &       2250 &  15 &        0.58 &           4.67 &          4.50 &   2 &    &  \tableno   & \tableno   \\
   Letter-low        &       2250 &  15 &        0.42 &           4.68 &          3.13 &   2 &    &  \tableno   & \tableno   \\
   Letter-med        &       2250 &  15 &        0.42 &           4.67 &          3.21 &   2 &    &  \tableno   & \tableno   \\
   MSRC\_21          &        563 &  20 &        0.07 &          77.52 &        198.32 &     &    &  \tableyes  & \tableno   \\
   MSRC\_21C         &        209 &  17 &        0.12 &          40.28 &         96.60 &     &    &  \tableyes  & \tableno   \\
   MSRC\_9           &        221 &   8 &        0.12 &          40.58 &         97.94 &     &    &  \tableyes  & \tableno   \\
   MUTAG             &        188 &   2 &        0.14 &          17.93 &         19.79 &     &    &  \tableyes  & \tableyes  \\
   Mutagenicity      &       4337 &   2 &        0.09 &          30.32 &         30.77 &     &    &  \tableyes  & \tableyes  \\
   NCI1              &       4110 &   2 &        0.09 &          29.87 &         32.30 &     &    &  \tableyes  & \tableno   \\
   NCI109            &       4127 &   2 &        0.09 &          29.68 &         32.13 &     &    &  \tableyes  & \tableno   \\
   OHSU              &         79 &   2 &        0.08 &          82.01 &        199.66 &     &    &  \tableyes  & \tableno   \\
   PROTEINS          &       1113 &   2 &        0.21 &          39.06 &         72.82 &     &    &  \tableyes  & \tableno   \\
   PROTEINS\_full    &       1113 &   2 &        0.21 &          39.06 &         72.82 &  29 &    &  \tableyes  & \tableno   \\
   PTC\_FM           &        349 &   2 &        0.22 &          14.11 &         14.48 &     &    &  \tableyes  & \tableyes  \\
   PTC\_FR           &        351 &   2 &        0.21 &          14.56 &         15.00 &     &    &  \tableyes  & \tableyes  \\
   PTC\_MM           &        336 &   2 &        0.22 &          13.97 &         14.32 &     &    &  \tableyes  & \tableyes  \\
   PTC\_MR           &        344 &   2 &        0.21 &          14.29 &         14.69 &     &    &  \tableyes  & \tableyes  \\
   Peking\_1         &         85 &   2 &        0.13 &          39.31 &         77.35 &     &    &  \tableyes  & \tableno   \\
   REDDIT-BINARY     &       2000 &   2 &        0.02 &         429.63 &        497.75 &     &    &  \tableno   & \tableno   \\
   REDDIT-MULTI-12K  &      11929 &  11 &        0.02 &         391.41 &        456.89 &     &    &  \tableno   & \tableno   \\
   REDDIT-MULTI-5K   &       4999 &   5 &        0.01 &         508.52 &        594.87 &     &    &  \tableno   & \tableno   \\
   SYNTHETIC         &        300 &   2 &        0.04 &         100.00 &        196.00 &   1 &    &  \tableyes  & \tableno   \\
   SYNTHETICnew      &        300 &   2 &        0.04 &         100.00 &        196.25 &   1 &    &  \tableno   & \tableno   \\
   Synthie           &        400 &   4 &        0.04 &          95.00 &        172.93 &  15 &    &  \tableno   & \tableno   \\
    \bottomrule
   \end{longtable}
  \end{landscape}

\begin{figure}[tbp]
  \centering
  \iffinal
    \includegraphics{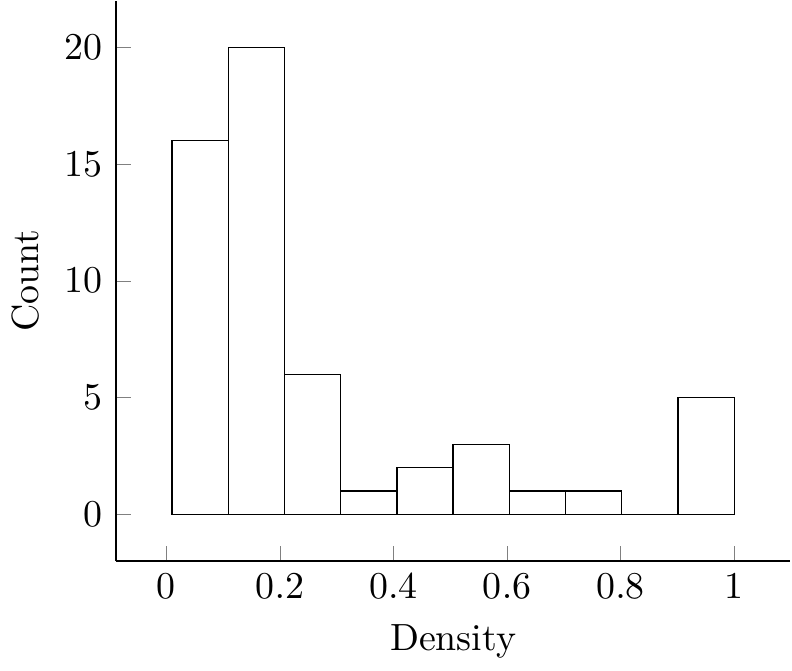}
  \else
    \begin{tikzpicture}
      \begin{axis}[
        axis x line*  = bottom,
        axis y line*  = left,
        enlargelimits = true,
        xlabel        = {Density},
        ylabel        = {Count},
      ]
        \addplot[hist = {%
            bins = 10,
            data = x,
          }
        ] file {Data/Density_distribution.txt};
      \end{axis}
    \end{tikzpicture}
  \fi
  \caption{%
    A histogram of the density values of the graphs. Few dense or complete graphs
    can be found among the benchmark data sets.
  }
  \label{fig:Density distribution}
\end{figure}
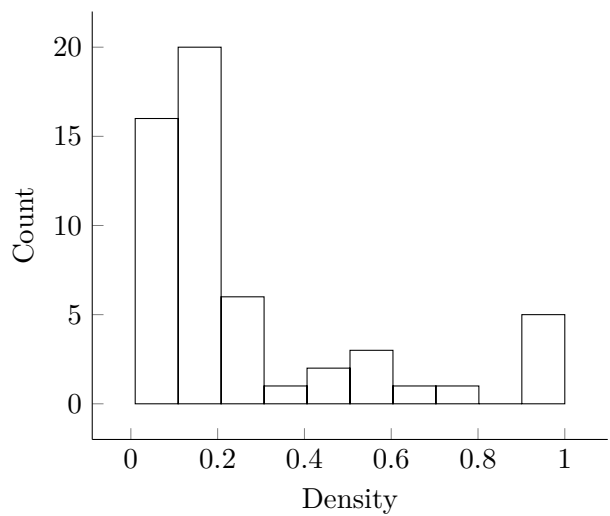

\begin{table}[tb]
  \centering
  \setlength{\tabcolsep}{3 pt}
  \footnotesize 
  \renewcommand\tabularxcolumn[1]{m{#1}}
  \begin{tabularx}{\textwidth}{XlXcccc}
    \toprule
    Kernel & Reference & Parameters & Node labels & Node attr. & Edge labels
                       & Edge attr. \\
    \midrule
    GraphHopper  & p.\ \pageref{sec:GraphHopper kernel} & $\gamma
    = \frac{1}{d}$ &  & \tableyes$^{\ast}$\\
    \midrule
    Graphlets & p.\ \pageref{sec:Graphlet kernel} & $k \in \{3, 4, 5\}$\\
    \midrule
    HGK-SP & p.\ \pageref{sec:Hash graph kernels} & $\emptyset$
                                                  & \tableyes
                                                  & \tableyes\\
    HGK-WL & p.\ \pageref{sec:Hash graph kernels} & $h \in \{0, 1, 2, 3,
    4, 5, 6, 7\}$ & \tableyes & \tableyes \\
    \midrule
    Histogram~($\vertices$) & p.\ \pageref{Node histogram kernel}
                  & $\emptyset$ & \tableyes \\
    Histogram~($\edges$)  & p.\ \pageref{Edge histogram kernel}
                  & $\emptyset$ & & & \tableyes \\
    \midrule
    %
    Message passing & p.~\pageref{sec:Message passing graph kernels}
                    & $T \in \{1, 2, 3, 4\}$, \newline $\alpha = 0.8$,
    $\beta = 0.2$ & \tableyes \\
    \midrule
    %
    Multiscale Laplacian & p.\ \pageref{sec:Multiscale Laplacian graph
  kernel} & $\eta, \gamma \in \{0.01, 0.1\}$, \newline $r, l \in \{2,
3\}$ & \tableyes \\
    \midrule
Shortest-path & p.\ \pageref{sec:Shortest-path kernel}
                     & $\emptyset$ & \tableyes
                     \\
    \midrule
Subgraph matching & p.\ \pageref{sec:Subgraph matching kernels} & $k
    \in \{3, 4, 5\}$ \newline BB: $c=3$ \newline TP:
  $c \in \{0.1, 0.5, 1.0\}$ & \tableyes & \tableyes & \tableyes
                            & \tableyes\\
    \midrule
  Random walk  & p.\ \pageref{sec:Fast computation of walk-based
kernels} & \tableyes
    \\
    \midrule
WL subtree & p.\ \pageref{sec:Weisfeiler--Lehman
  kernel} & $h \in \{0, 1, 2, 3, 4, 5, 6, 7\}$
           & \tableyes \\
  WL-OA & p.\ \pageref{sec:Optimal assignment kernels} & $h \in \{0, 1,
  2, 3, 4, 5, 6, 7\}$ & \tableyes \\
    \bottomrule
  \end{tabularx}
  %
  \caption{%
    Selected parameters for each of the graph kernels. TP refers to the
    triangular kernel, and BB is the Brownian bridge kernel. Please refer to
    the indicated page for more details about the parameters. $^\ast$
    indicates that node labels were used if there were no node
    attributes. 
  }
  \label{tab:Parameters}
\end{table}

Specifically, we
included the following kernels:
\begin{compactenum}[(i)]
  \item the GraphHopper kernel~\citep[GH]{Feragen13},
  \item the graphlets kernel~\citep[GL]{Shervashidze09a},
  \item two histogram kernels~(based on vertex~(V) and edge labels~(E), respectively),
  \item two instances of the hash graph kernels framework~\citep[HGK-SP, HGK-WL]{Morris16},
  \item the message passing kernel~\citep[MP]{Nikolentzos18},
  \item the multiscale Laplacian graph kernel~\citep[MLG]{Kondor16},
  \item the random walk kernel~\citep[RW]{Vishwanathan06},
  \item the shortest-path kernel~\citep[SP]{Borgwardt05},
  \item the subgraph matching kernel~\citep[CSM]{kriege2012subgraph},
  \item the Weisfeiler--Lehman subtree kernel~\citep[WL]{Shervashidze11}, and
  \item the Weisfeiler--Lehman optimal assignment kernel~\citep[WL-OA]{Kriege16}.
\end{compactenum}

\subsection{Kernel matrix computation}
%
For each of the included graph kernels, we generate a set of full kernel
matrices, \ie\ kernel matrices between all pairs of graphs of the input data
set. We account for the parameters of a graph kernel by generating a new matrix
for all possible combinations of parameter values. While seemingly wasteful,
this ensures that we are able to perform a proper hyperparameter optimisation
of each kernel and select the \emph{best}---in terms of predictive
performance---parameter set for each algorithm.
As a brief example, consider the Weisfeiler--Lehman subtree kernel. Its single
parameter is $h$, the subtree depth. In this case, we calculate a collection of
kernel matrices that are indexed by the respective value of $h$.
Table~\ref{tab:Parameters} lists the parameters used to create the set
of kernel matrices and which information from the graph is processed by
each kernel. For some of the graph kernels, parameter selection
is dictated by computational efficiency~(graphlet enumeration does not
scale well to higher-order graphlets, for example). For other kernels,
we used parameter ranges suggested by the authors. \re{If the parameter
ranges suggested by the authors were too expansive to successfully run
in 120 hours, we resorted to a subset of the original parameters in order to
obtain results. Additionally, some kernels allowed for the specification
of additional kernels on node and/or edge attributes. In the GraphHopper
kernel, we used a gaussian kernel on the node attributes, 
with $\gamma=\frac{1}{d}$, where $d$ is the dimension of the node
attributes (or one-hot representations of the node labels, if there were
no node attributes), as was suggested by the authors.  In the subgraph matching kernel, we used the Brownian
bridge kernel for node attributes (when present) and the triangular
kernel for edge attributes (when present), with the parameters specified
in Table~\ref{tab:Parameters}. While some kernels can theoretically
incorporate more graph information, we found that many of the available
implementations did not allow for it out of the box. Our results
accordingly largely reflect what the implementations currently natively support.}

\subsection{Training procedure}\label{sec:Training procedure}
%
Given a set of kernel matrices that belong to a certain graph kernel, we
use a nested cross-validation procedure, as this ensures that we obtain
a suitable assessment of the generalisation performance of a kernel
without risking to suffer from overfitting.
Our outer cross-validation loop employs a stratified \mbox{$10$-fold}
cross-validation~(randomly shuffled), while the inner loop uses
a \mbox{$5$-fold} cross-validated grid search, which is used to
determine the \emph{best} parameter set of a kernel. In addition, this
procedure is repeated $10$ times so that we can report an average
performance value and its standard deviation.
We use a standard support vector machine~(SVM) classifier with
a precomputed kernel matrix. The setup and the training times
are thus comparable---by contrast, it is possible to use different
implementations in different programming languages.
The SVM uses $C \in \{10^{-3}, 10^{-2}, \dots, 10^2, 10^3\}$.
In addition, we consider \emph{normalising} each kernel matrix to be a
trainable hyperparameter. Hence, we make it possible to replace each
kernel matrix $K
= \left(k_{ij}\right)_{i, j \in \{1, \dots, n\}}$ by $K'
= \left(k'_{ij}\right)_{i, j \in \{1, \dots, n\}}$, where
\begin{equation}
  k'_{ij} = \frac{k_{ij}}{\sqrt{k_{ii} k_{jj}}},
\end{equation}
which is a non-linear normalisation. Intuitively, this can can be seen
as the ``kernel variant'' of restricting \mbox{$d$-dimensional} feature
vectors to lie on a unit hypersphere; in this case, however, the hypersphere
is ``measured'' via the kernel function, instead of the usual metric of
a space.
Such a normalisation can have a positive impact on SVM
classifiers~\citep{Graf01}. Our experiments indicate, however, that
normalisation is rarely required to obtain good performance values.

\subsection{Training environment}%
%
Training is performed on a multi-core cluster system. Each kernel is
allocated \SI{120}{\hour} of multi-core processing time for its graph
kernel matrix computation, followed by an additional \SI{120}{\hour}
to perform hyperparameter search and model fitting. Moreover, we use
a maximum of $500$ iterations to fit the SVM classifier. Convergence is
usually obtained much more rapidly, but the restriction ensures that we
are able to train all graph kernels.

Our training environment necessitated the cross-validation setup described
above; with a nested \mbox{$10$-fold} cross-validation in which both the
inner cross-validation loop and the outer loop use $10$ folds, some
graph kernels would be penalised because their hyperparameter search
procedure would not converge. In the interest of fairness, we selected
parameters that allowed the majority of graph kernels to be trained,
despite some of them---such as the multiscale Laplacian graph
kernel---requiring very dense parameter grids. We thus opted for
a \mbox{$5$-fold} cross-validated grid search, as described above.
Moreover, every kernel is allocated a computational budget of
\SI{128}{\giga\byte} of RAM, in order to specify a realistic training
environment. Any runs that fail to satisfy these requirements have been marked
with ``OOM''~(short for ``out-of-memory''). \re{In some cases, the kernel matrices
  did not finish computing in 120h, and thus have been marked with
``OOT''~(short for ``out of time'').}
Graph kernels whose implementations preclude handling a certain data
set have been marked with ``NA'' to indicate that results are ``not
available''. The code used for our experiments can be found at 
\url{https://github.com/BorgwardtLab/graphkernels-review/}.

\begin{table}[tbp]
  \centering
  \footnotesize
  \begin{tabular}{>{\ttfamily}lrrr}
    \toprule
    \normalfont{Data set} & \normalfont{Graphs} & \normalfont{Classes} & \normalfont{Class balance}\\
    \midrule
    AIDS                  &       2000 &   2 &        $4 : 1$\\
    BZR                   &        405 &   2 &        $\approx 4 : 1$\\
    BZR\_MD$\ast$         &        306 &   2 &        $\approx 1 : 1$\\
    COIL-DEL$\ast$        &       3900 & 100 &        $1:1$\\
    COIL-RAG$\ast$        &       3900 & 100 &        $1:1$\\
    COLLAB                &       5000 &   3 &        $2600 : 1625: 775$\\
    COX2                  &        467 &   2 &        $\approx 4:1$\\
    COX2\_MD$\ast$        &        303 &   2 &        $\approx 1:1$\\
    DD                    &       1178 &   2 &        $\approx 2:1$\\
    DHFR                  &        756 &   2 &        $\approx 2:1$\\
    DHFR\_MD              &        393 &   2 &        $\approx 2:1$\\
    ENZYMES$\ast$         &        600 &   6 &        $1:1$\\
    ER\_MD                &        446 &   2 &        $\approx 2:1$\\
    FRANKENSTEIN$\ast$    &       4337 &   2 &        $\approx 1:1$\\
    IMDB-BINARY$\ast$     &       1000 &   2 &        $1:1$\\
    IMDB-MULTI$\ast$      &       1500 &   3 &        $1:1$\\
    KKI$\ast$             &         83 &   2 &        $\approx 1:1$\\
    Letter-high$\ast$     &       2250 &  15 &        $1:1$\\
    Letter-low$\ast$      &       2250 &  15 &        $1:1$\\
    Letter-med$\ast$      &       2250 &  15 &        $1:1$\\
    MSRC\_21$\ast$        &        563 &  20 &        $\approx 1:1$\\
    MSRC\_21C             &        209 &  17 &        $\approx 10:1$\\
    MSRC\_9$\ast$         &        221 &   8 &        $\approx 1:1$\\
    MUTAG                 &        188 &   2 &        $\approx 2:1$\\
    Mutagenicity$\ast$    &       4337 &   2 &        $\approx 1:1$\\
    NCI1$\ast$            &       4110 &   2 &        $\approx 1:1$\\
    NCI109$\ast$          &       4127 &   2 &        $\approx 1:1$\\
    OHSU$\ast$            &         79 &   2 &        $\approx 1:1$\\
    PROTEINS              &       1113 &   2 &        $\approx 2:1$\\
    PROTEINS\_full        &       1113 &   2 &        $\approx 2:1$\\
    PTC\_FM               &        349 &   2 &        $\approx 2:1$\\
    PTC\_FR               &        351 &   2 &        $\approx 2:1$\\
    PTC\_MM               &        336 &   2 &        $\approx 2:1$\\
    PTC\_MR$\ast$         &        344 &   2 &        $\approx 1:1$\\
    Peking\_1$\ast$       &         85 &   2 &        $\approx 1:1$\\
    REDDIT-BINARY$\ast$   &       2000 &   2 &        $1:1$\\
    REDDIT-MULTI-12K      &      11929 &  11 &        $\approx 3:1$\\
    REDDIT-MULTI-5K$\ast$ &       4999 &   5 &        $1:1$\\
    SYNTHETIC$\ast$       &        300 &   2 &        $1:1$\\
    SYNTHETICnew$\ast$    &        300 &   2 &        $1:1$\\
    Synthie$\ast$         &        400 &   4 &        $\approx 1:1$\\
    \bottomrule
  \end{tabular}
  \caption{%
    Class ratio of the data sets that we used in the subsequent
    experiments. For binary classification problems, we provide
    an~(approximate) class ratio, whereas for multi-class problems we
    either list label counts individually or give the imbalance with
    respect to the largest class. A value of $1:1$ means that the data
    set is~(almost) perfectly balanced. Such data sets are printed with an
    additional asterisk~($\ast$) after their name.
  }
  \label{tab:Class imbalance}
\end{table}

\subsection{Evaluation procedure}%
%
As the listing of class ratios in Table~\ref{tab:Class imbalance} shows,
most of the data sets are balanced and consist of two classes. We thus use
\emph{accuracy} as our main evaluation measure.
For multi-class data sets, an evaluation in terms of the area under the
Receiver Operating Characteristic curve~(AUROC). For two classes~(which
are by custom termed ``positive'' and ``negative''), this measure
defines the probability that a randomly-selected positive example is
assigned a \emph{higher} score than a randomly-selected negative
example. AUROC can therefore be understood as a measure of how well
a classifier is able to distinguish between two classes. Being
scale-invariant, AUROC is suitable for comparing classifiers
\emph{across} data sets. In order to extend this binary measure to
a multi-class situation, we calculate AUROC values for all
one-versus-rest classification scenarios and calculate their mean. This
is also know as ``macro averaging''~\citep{Yang99}.
We provide macro-averaged AUROC values in an additional table and for
all analyses that compare kernels \emph{across} data sets.

\section{Classification performance}

This section details the performance of numerous graph kernels on
all data sets.
Table~\ref{tab:Classification accuracy baseline} provides the accuracies
and standard deviations of what we
consider to be the baseline graph kernels, namely the node and edge
label histograms. Table~\ref{tab:Empirical performance} summarises the
additional, more sophisticated graph kernel results in terms of
mean accuracy and standard deviations, calculated over the $10$ iterations of
our training procedure.
In addition, Table~\ref{tab:Classification auroc baseline} and
Table~\ref{tab:Empirical performance AUROC} provide a similar summary of
the performance in terms of AUROC.
Given the density of the tables, we will subsequently discuss the results in
more detail and under different aspects, such as a breaking down values by data
set type.

\paragraph{Overall assessment}
Before delving into various aspects of the performance of these kernels, we
want to briefly comment on some global patterns that are observable in the
table. First, in Table~\ref{tab:Empirical performance}, it is striking to
observe that the best-performing algorithms on every data set tend to employ
the Weisfeiler--Lehman relabelling procedure, or a variant thereof:
out of 41 data sets, \mbox{MP}~(the message passing kernel, which is based on the
Weisfeiler--Lehman relabelling procedure) exhibits the best performance
in 10 data sets. In addition, \mbox{WL-OA}~(the optimal assignment variant of the
Weisfeiler--Lehman subtree scheme) outperforms all other kernels on 6 data sets,
while \mbox{HGK-WL}~(the hash graph kernel variant of the
Weisfeiler--Lehman subtree kernel) does so on 4 data sets.
Last, the Weisfeiler--Lehman subtree kernel leads on a single data set.
Consequently, on 21 out of 41 data sets, kernels based on the principles
of the Weisfeiler--Lehman relabelling procedure outperform all other
kernels.
On the outset, it thus seems that higher performance can be achieved by looking
at decompositions of the graph---for the Weisfeiler--Lehman procedure, these
decompositions are provided by increasing neighbourhoods. In the following, we
will look more closely at this aspect.

\begin{figure}[tbp]
  \centering
  \iffinal
    \includegraphics{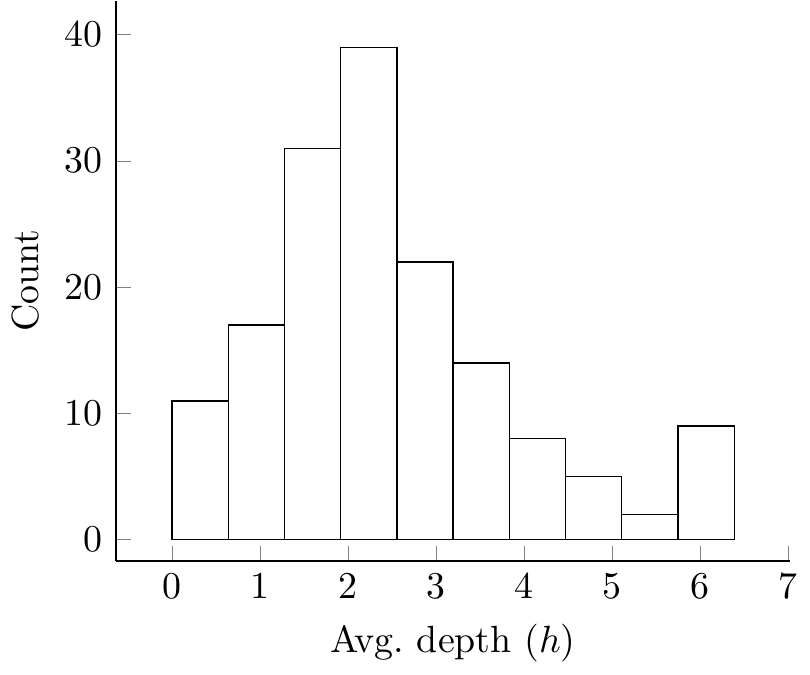}
  \else
  \begin{tikzpicture}
    \begin{axis}[
      axis x line*  = bottom,
      axis y line*  = left,
      xlabel        = {Avg.\ depth~($h$)},
      ylabel        = {Count},
      xtick         = {0, 1, 2, 3, 4, 5, 6, 7},
      enlargelimits = true,
      xmin          = 0,
    ]
      \addplot[hist = {%
          bins = 10,
          data = x,
        }
      ] table[col sep=comma, x=mean_depth] {Data/Accuracy_vs_depth.csv};

    \end{axis}
  \end{tikzpicture}
  \fi
  \caption{%
    A histogram of the average depth, \ie\ the average number of
    iterations $h$ over all folds of the training, of graph kernels that
    are based on a variant of the Weisfeiler--Lehman framework.
  }
  \label{fig:Accuracy vs. depth}
\end{figure}
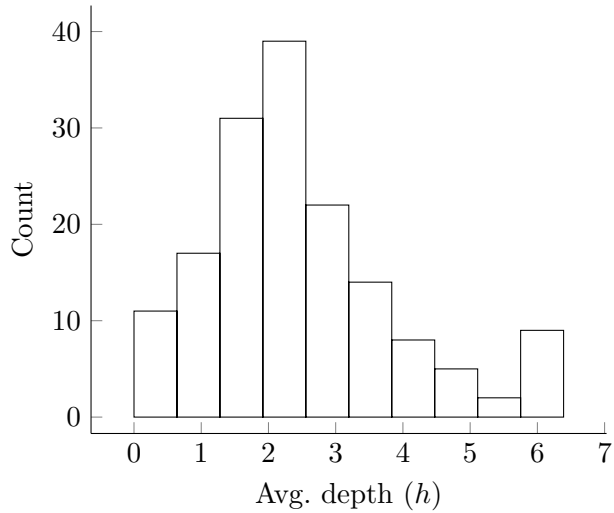

\paragraph{The impact of $h$}
Given that algorithms based on Weisfeiler--Lehman relabelling are among the top
performers on the benchmark data sets, we analyse their performance in more
detail. Our main question is to what extent a sufficient propagation depth into
a graph is required in order to achieve good performance.
To this end, Figure~\ref{fig:Accuracy vs. depth} depicts the mean
``depth''---\ie\ the mean number of iterations $h$---over the training process
of these graph kernels.
A value of $h = 0$ would be tantamount to \emph{not} using any neighbourhood
information at all; higher values indicate that information from more distant
neighbourhoods is  propagated.
Overall, we observe that all graph kernels that are capable of propagating
information across neighbourhoods---which can also be seen as diffusing
information of the graph over different scales---tend to do so to in order
to reach their final accuracy values.
This demonstrates that graph kernels need to be capable of analysing the graph
at multiple scales.

\begin{table}
  \footnotesize
  \centering
  \setlength{\tabcolsep}{1.5pt}
  \sisetup{
    detect-weight           = true,
    detect-inline-weight    = math,
    table-format            = 2.2(3),
    separate-uncertainty    = true,
    table-align-uncertainty = true,
    tight-spacing           = true,
  }
  \newcommand{\w}{\color{red}}
  \newcommand{\f}{\color{gray}}
  \begin{tabular}{>{\ttfamily}lSS}
    \toprule
    \normalfont{Data set} & {H($\vertices$)} & {H($\edges$)} \\
    \midrule
    AIDS              & \w 99.70 \pm 0.00  & 99.28 \pm 0.07\\
    BZR               & 65.18 \pm 0.93  & 77.36 \pm 1.11\\
    BZR\_MD           & 70.44 \pm 1.25  & 65.49 \pm 1.76\\
    COIL-DEL          & 15.03 \pm 0.27  & 7.81 \pm 0.28\\
    COIL-RAG          & 7.79 \pm 0.07   & \f {NA} \\
    COLLAB            & 31.32 \pm 2.53  & 79.52 \pm 0.44\\
    COX2              & 59.71 \pm 0.92  & 73.89 \pm 1.02\\
    COX2\_MD          & 62.80 \pm 0.93  & 60.03 \pm 1.16\\
    DD                & 68.68 \pm 3.04  & 78.52 \pm 0.34\\
    DHFR              & 60.25 \pm 0.36  & 67.55 \pm 0.86\\
    DHFR\_MD          & 67.92 \pm 0.08  & 62.24 \pm 1.78\\
    ENZYMES           & 25.20 \pm 0.96  & 27.33 \pm 0.80\\
    ER\_MD            & 66.88 \pm 1.00 & 69.79 \pm 1.35\\
    FRANKENSTEIN      & 64.96 \pm 0.34  & 60.33 \pm 1.90\\
    IMDB-BINARY       & 50.58 \pm 0.20  & 73.46 \pm 0.60\\
    IMDB-MULTI        & 34.90 \pm 0.31  & 50.38 \pm 0.55\\
    KKI               & 52.40 \pm 2.66  & 48.97 \pm 2.32\\
    Letter-high       & 36.69 \pm 0.52  & \f {NA}\\
    Letter-low        & 47.93 \pm 0.33  & \f {NA}\\
    Letter-med        & 43.93 \pm 0.55  & \f {NA}\\
    MSRC\_21          & 89.45 \pm 0.41  & 90.04 \pm 0.84\\
    MSRC\_21C         & 83.36 \pm 1.39  & 84.68 \pm 0.77\\
    MSRC\_9           & 89.36 \pm 0.84  & \w 93.17 \pm 1.09\\
    MUTAG             & 85.98 \pm 0.40  & 85.14 \pm 1.03\\
    Mutagenicity      & 67.01 \pm 0.83  & 49.13 \pm 1.75\\
    NCI1              & 64.66 \pm 0.53  & 51.71 \pm 1.45\\
    NCI109            & 63.24 \pm 0.54  & 51.45 \pm 2.06\\
    OHSU              & 51.86 \pm 3.05  & 54.23 \pm 3.59\\
    PROTEINS          & 70.13 \pm 1.61  & 69.32 \pm 0.92\\
    PROTEINS\_full    & 70.13 \pm 1.61 & 69.32 \pm 0.92\\
    PTC\_FM           & 58.12 \pm 0.85 & 57.57 \pm 1.40\\
    PTC\_FR           & \w 67.84 \pm 0.16 & 40.26 \pm 2.03\\
    PTC\_MM           & \w 66.55 \pm 0.61 & 41.36 \pm 1.13\\
    PTC\_MR           & 58.46 \pm 0.27 & 53.03 \pm 2.21\\
    Peking\_1         & 57.75 \pm 3.03 & 55.12 \pm 4.12\\
    REDDIT-BINARY     & 50.03 \pm 2.24  & 78.94 \pm 0.60\\
    REDDIT-MULTI-12K  & 7.34 \pm 1.59   & 31.06 \pm 0.31\\
    REDDIT-MULTI-5K   & 17.86 \pm 2.13  & 45.72 \pm 0.22\\
    SYNTHETIC         & 50.00 \pm 0.00  & 50.00 \pm 0.00\\
    SYNTHETICnew      & 62.30 \pm 0.55  & 71.20 \pm 1.69\\
    Synthie           & 48.78 \pm 1.33  & 47.17 \pm 1.81\\
    \bottomrule
  \end{tabular}
  \caption{Empirical performance of the baseline graph kernels, the
  vertex histogram (H($\vertices$)) and edge histogram (H($\edges$)) on
the benchmark data sets. The mean accuracy over 10 iterations of
a nested cross-validation procedure is shown along with its
corresponding standard deviation. The highest mean accuracy of each
data set across the baseline kernels and the normal kernels is shown in
\textcolor{red}{red}.  OOM means ``out-of-memory", OOT means
``out-of-time", and NA indicates that the given implementation could not
handle a certain data set.}
  \label{tab:Classification accuracy baseline}
\end{table}


\begin{landscape}
  \setlength{\tabcolsep}{1.5pt}
  \sisetup{
    detect-weight           = true,
    detect-inline-weight    = math,
    table-format            = 2.2(3),
    separate-uncertainty    = true,
    table-align-uncertainty = true,
    tight-spacing           = true,
  }
  \footnotesize
  \newcommand{\w}{\color{red}}
  \newcommand{\f}{\color{gray}}
  \begin{longtable}{>{\tt}lSSSSSSSSSSS}
  \caption{%
      Empirical performance of graph kernels on the benchmark data sets.
      The mean accuracy over $10$ iterations of a nested cross-validation
      procedure is shown along with its corresponding standard deviation.
      The highest mean accuracy of each
data set across the baseline kernels and the normal kernels
\textcolor{red}{red}. OOM means ``out-of-memory", OOT means
``out-of-time", and NA indicates that the given implementation could not
handle a certain data set.}
    \label{tab:Empirical performance}\\
    \toprule
    \normalfont{Data set} & {CSM} & {GH}              & {Graphlet} & {HGK-SP} & {HGK-WL} & {MLG} & {MP}             & {SP}                & {RW}           & {WL}              & {WL-OA}\\
    \midrule
    \endhead
    AIDS              & 99.47 \pm 0.05  & 99.13 \pm 0.07  & 98.05 \pm 0.14  & 99.38 \pm 0.02  & 98.76 \pm 0.09  & 98.38 \pm 0.14  & 99.46 \pm 0.06  & 99.64 \pm 0.02  & \w 99.70 \pm 0.00   & 99.52 \pm 0.10 & 99.69 \pm 0.02 \\
    BZR               & 84.54 \pm 0.65  & 79.32 \pm 0.48  & 79.20 \pm 0.65  & 81.99 \pm 0.30  & 81.42 \pm 0.60  & 88.04 \pm 0.70  & \w 88.08 \pm 0.93  & 81.58 \pm 0.75  & 76.63 \pm 1.76   & 87.16 \pm 0.97 & 87.43 \pm 0.81 \\
    BZR\_MD           & \w 77.63 \pm 1.29  & 51.99 \pm 0.25  & 52.71 \pm 0.24  & 60.08 \pm 0.88  & 52.64 \pm 1.20  & 51.46 \pm 0.61  & 65.11 \pm 1.59  & 68.90 \pm 1.61  & 68.98 \pm 1.28   & 67.45 \pm 1.20 & 68.19 \pm 1.09 \\
    COIL-DEL          & \f {OOT}        & \f {OOM}        & \f {NA} & 66.19 \pm 0.20  & 61.39 \pm 0.29  & 3.90 \pm 0.19   & \w 80.94 \pm 0.49 & 17.32 \pm 0.23  & \f {OOT}             & 16.68 \pm 0.29 & 16.56 \pm 0.31 \\
    COIL-RAG          & \f {NA}  & \f {OOM}        & 4.91 \pm 0.10 & \f 91.65 \pm 0.26  & \w 91.11 \pm 0.47  & \f {OOM} & 83.50 \pm 0.39  & \f {NA}   & \f {NA}   & 7.83 \pm 0.08 & 7.90 \pm 0.10 \\
    COLLAB            & \f {OOT}        & \f {OOM}        & \f {NA} & 77.15 \pm 0.15  & \f {OOM}             & \f {OOM} & \w 80.93 \pm 0.28  & 79.92 \pm 0.37  & \f {OOM} & 68.25 \pm 1.50  & 80.18 \pm 0.25 \\ 
    COX2              & 79.78 \pm 1.04  & 78.16 \pm 0.00  & 49.90 \pm 0.00 & 78.16 \pm 0.00  & 78.16 \pm 0.00  & 76.76 \pm 0.87  & 80.76 \pm 0.78 & 78.03 \pm 1.10  & 63.21 \pm 5.27  & 79.67 \pm 1.32  & \w 81.08 \pm 0.89 \\
    COX2\_MD          & \f {OOT}        & 60.99 \pm 1.06  & 51.15 \pm 0.00 & 59.92 \pm 0.66  & 57.15 \pm 1.20  & 51.15 \pm 0.00  & 61.81 \pm 1.51 & \w 64.95 \pm 1.09  & 63.14 \pm 1.57  & 60.07 \pm 2.22  & 62.37 \pm 2.11 \\
    DD                & \f {OOM}        & \f {OOM}        & \f {NA} & \f {OOM}             & 75.35 \pm 0.94  & 75.68 \pm 0.84  & 79.02 \pm 0.25 & \w 80.22 \pm 0.51  & \f {OOM}             & 77.73 \pm 1.97  & 77.78 \pm 1.22 \\
    DHFR              & 77.99 \pm 0.96  & 68.70 \pm 0.86  & 60.98 \pm 0.00 & 72.48 \pm 0.65  & 75.35 \pm 0.66  & \w 83.22 \pm 0.94  & 80.47 \pm 0.92 & 79.35 \pm 1.55  & \f {OOT}             & 81.72 \pm 0.80  & 82.40 \pm 0.97 \\
    DHFR\_MD          & \f {OOT}        & \w 67.95 \pm 0.00  & \w 67.95 \pm 0.00 & \w 67.95 \pm 0.00  & 66.08 \pm 1.02  & \w 67.95 \pm 0.00  & 66.07 \pm 0.88 & 63.76 \pm 2.15  & 64.35 \pm 1.43  & 62.56 \pm 1.51  & 64.10 \pm 1.70 \\
    ENZYMES           & \w 66.38 \pm 1.14  & 46.53 \pm 1.26  & \f {NA} & 63.07 \pm 0.69  & 54.53 \pm 1.34  & 51.17 \pm 1.59  & 60.02 \pm 0.75 & 41.27 \pm 1.18  & 26.10 \pm 1.15  & 54.27 \pm 0.94  & 58.88 \pm 0.85\\ 
    ER\_MD            & \f {OOT}        & 59.42 \pm 0.00  & 59.42 \pm 0.00 & 59.42 \pm 0.00  & 66.72 \pm 1.28  & 60.72 \pm 0.69  & \w 71.62 \pm 1.20 & 70.55 \pm 0.86  & \f {NA}             & 70.35 \pm 1.01  & 70.96 \pm 0.75 \\
    FRANKENSTEIN      & \f {OOT}        & \f {OOM}        & 56.28 \pm 0.19 & 56.05 \pm 0.10  & 63.69 \pm 0.51  & \w 72.36 \pm 0.22  & 57.01 \pm 0.66 & 58.85 \pm 1.96  & \f {OOT}             & 71.81 \pm 0.31 & 72.02 \pm 0.30 \\
    IMDB-BINARY       & \f {OOT}        & \f {OOM}        & \f {NA} & 73.34 \pm 0.47  & 72.75 \pm 1.02  & 52.56 \pm 0.42  & 72.28 \pm 0.75 & 72.23 \pm 0.78  & \w 74.20 \pm 0.76  & 71.15 \pm 0.47  & 74.01 \pm 0.66 \\
    IMDB-MULTI        & \f {OOT}        & \f {OOM}        & \f {NA} & \w 51.58 \pm 0.42  & 50.73 \pm 0.63  & 34.27 \pm 0.33  & 49.55 \pm 0.68 & 51.31 \pm 0.28  & 50.13 \pm 0.50  & 50.25 \pm 0.72  & 49.95 \pm 0.46 \\
    KKI               & 49.56 \pm 3.50  & \w 55.22 \pm 0.75  & \f {NA} & 53.07 \pm 2.52  & 52.64 \pm 1.67  & 53.59 \pm 3.44  & 46.53 \pm 5.95 & 47.21 \pm 3.05  & 48.26 \pm 4.51  & 53.90 \pm 2.42  & 54.05 \pm 2.27 \\
    Letter-high       & \f {NA}  & \f {OOM}        & 33.64 \pm 0.91 & 87.96 \pm 0.31  & \w 90.54 \pm 0.32  & 16.08 \pm 1.18  & 86.07 \pm 0.50 & \f {NA}  & \f {NA}  & 37.56 \pm 0.26  & 37.41 \pm 0.39 \\
    Letter-low        & \f {NA }& \f {OOM}        & 45.94 \pm 0.54 & 99.29 \pm 0.04  & \w 99.66 \pm 0.03  & 22.01 \pm 0.35  & 99.52 \pm 0.12 & \f {NA} & \f {NA}  & 49.94 \pm 0.31  & 49.89 \pm 0.22 \\
    Letter-med        & \f {NA} & \f {OOM}        & 41.31 \pm 0.38 & 92.25 \pm 0.19  & 93.82 \pm 0.17  & 19.03 \pm 1.16  & \w 94.56 \pm 0.27 & \f {NA}  & \f {NA} & 45.47 \pm 0.28  & 45.84 \pm 0.39 \\
    MSRC\_21          & \f {OOT}        & 86.66 \pm 0.57  & \f {NA} & 84.36 \pm 0.41  & 78.18 \pm 0.54  & 67.67 \pm 0.67  & \w 90.47 \pm 0.75 & 90.04 \pm 0.74  & \f {OOT}             & 89.19 \pm 0.63  & 89.42 \pm 0.55 \\
    MSRC\_21C         & 83.24 \pm 0.80  & 81.35 \pm 1.32  & \f {NA} & 81.21 \pm 0.76  & 75.18 \pm 0.99  & 60.56 \pm 2.13  & 84.55 \pm 0.45 & 84.49 \pm 0.90  & 82.13 \pm 1.33  & 82.96 \pm 1.45  & \w 85.70 \pm 1.56 \\
    MSRC\_9           & 92.50 \pm 1.30  & 89.85 \pm 0.65  & \f {NA} & 89.26 \pm 0.80  & 88.46 \pm 0.57  & 79.74 \pm 2.51  & 90.21 \pm 0.67 & 92.59 \pm 0.67  & 91.18 \pm 1.18  & 89.87 \pm 1.07  & 91.16 \pm 1.07 \\
    MUTAG             & \w 87.29 \pm 1.25  & 81.07 \pm 0.45  & 67.38 \pm 0.45 & 80.90 \pm 0.48  & 75.51 \pm 1.34  & 78.53 \pm 2.25  & 86.98 \pm 1.09 & 85.06 \pm 1.28  & 85.62 \pm 1.84  & 85.75 \pm 1.96  & 86.10 \pm 1.95 \\
    Mutagenicity      & \f {OOT}        & \f {OOM}        & 55.31 \pm 0.03 & 71.83 \pm 0.15  & 80.12 \pm 0.39  & \f {OOM} & 79.03 \pm 0.32 & 50.75 \pm 2.50  & \f {OOT}             & 82.03 \pm 0.44  & \w 83.24 \pm 0.65 \\
    NCI1              & \f {OOT}        & \f {OOM}        & 50.98 \pm 0.47 & 69.55 \pm 0.16  & 81.26 \pm 0.21  & 78.17 \pm 0.33  & 78.05 \pm 0.76 & 55.76 \pm 1.68  & \f {OOT}             & 85.60 \pm 0.36 & \w 85.95 \pm 0.23 \\
    NCI109            & \f {OOT}        & \f {OOM}        & 50.55 \pm 0.07 & 69.56 \pm 0.19  & 80.69 \pm 0.19  & 78.01 \pm 0.58  & 76.75 \pm 0.39 & 55.78 \pm 2.01  & \f {OOT}             & 85.76 \pm 0.22 & \w 86.17 \pm 0.19 \\
    OHSU              & 52.66 \pm 3.37  & \w 55.79 \pm 0.00  & \f {NA} & 50.74 \pm 1.73  & 55.40 \pm 0.64  & 54.80 \pm 3.18  & 54.63 \pm 2.22 & 52.16 \pm 2.81  & \f {OOT}             & 52.36 \pm 3.08  & 52.36 \pm 3.08 \\
    PROTEINS          & \f {OOT}        & 75.08 \pm 0.29  & \f {NA} & 74.53 \pm 0.35  & 73.65 \pm 0.72  & 75.55 \pm 0.71  & 73.30 \pm 0.64 & \w 75.72 \pm 0.42  & \f {OOM}             & 73.06 \pm 0.47  & 73.50 \pm 0.87 \\
    PROTEINS\_full    & 69.35 \pm 0.25  & \f {OOM}        & \f {NA} & \w 75.92 \pm 0.49  & 75.57 \pm 0.48  & 75.55 \pm 0.71  & 73.86 \pm 0.83 & 75.72 \pm 0.42  & \f {OOM}             & 73.06 \pm 0.47  & 73.50 \pm 0.87 \\
    \pagebreak
    PTC\_FM           & 59.75 \pm 1.90  & 63.63 \pm 0.62  & 58.59 \pm 0.50 & 63.78 \pm 0.33  & \w 64.38 \pm 0.71  & 59.87 \pm 1.64  & 63.09 \pm 1.97 & 62.44 \pm 0.84  & 62.35 \pm 1.00  & 61.77 \pm 1.92  & 61.98 \pm 2.22 \\
    PTC\_FR           & 59.97 \pm 2.57  & 66.53 \pm 0.31  & 65.53 \pm 0.00 & 67.50 \pm 0.25  & 67.81 \pm 0.24  & 66.25 \pm 0.72  & 64.76 \pm 1.70 & 64.56 \pm 1.57  & 39.10 \pm 2.76  & 65.68 \pm 1.59  & 65.17 \pm 1.41 \\
    PTC\_MM           & 62.34 \pm 0.90  & 64.23 \pm 0.60  & 60.02 \pm 0.40 & 64.71 \pm 0.78  & 65.69 \pm 0.97  & 60.71 \pm 0.80  & 65.19 \pm 1.68 & 64.23 \pm 0.64  & 56.22 \pm 1.53  & 64.36 \pm 2.36  & 64.18 \pm 1.18 \\
    PTC\_MR           & 59.40 \pm 1.88  & 57.76 \pm 1.44  & 54.45 \pm 0.37 & 57.48 \pm 1.14  & 59.87 \pm 1.00  & 60.37 \pm 1.57  & \w 61.24 \pm 2.21 & 57.26 \pm 2.43  & 53.25 \pm 1.63  & 59.97 \pm 1.95  & 60.38 \pm 1.40 \\
    Peking\_1         & 55.86 \pm 3.49  & \f {OOM}        & \f {NA} & \w 58.74 \pm 2.11  & 56.13 \pm 2.39  & 53.82 \pm 2.79  & 57.70 \pm 4.04 & 54.79 \pm 2.42  & \f {OOT}             & 56.90 \pm 2.02  & 56.22 \pm 2.52 \\
    REDDIT-BINARY     & \f {OOM}        & \f {OOM}        & \f {NA} & \f {OOM}             & \f {OOM}             & \f {OOM} & \w 89.24 \pm 0.48 & 89.20 \pm 0.37  & \f {OOM}             & 77.95 \pm 0.60  & 87.60 \pm 0.33 \\
    REDDIT-MULTI-12K  & \f {OOM}        & \f {OOM}        & \f {NA} & \f {OOM}             & \f {OOM}             & \f {OOM} & \w 42.09 \pm 0.31 & 31.06 \pm 0.63  & \f {OOM}             & 26.88 \pm 0.62  & \f {OOM} \\
    REDDIT-MULTI-5K   & \f {OOM}        & \f {OOM}        & \f {NA} & \f {OOM}             & \f {OOM}             & \f {OOM} &50.85 \pm 0.47 & 45.54 \pm 0.75  & \f {OOM}             & \w 51.63 \pm 0.37  & \f {OOM} \\
    SYNTHETIC         & 66.27 \pm 1.71  & 58.83 \pm 0.98  & 50.00 \pm 0.00 & 59.93 \pm 1.24  & \w 74.07 \pm 1.13  & 50.00 \pm 0.00  & 50.00 \pm 0.00 & 50.00 \pm 0.00  & \f {OOT}             & 50.00 \pm 0.00 & 50.00 \pm 0.00 \\
    SYNTHETICnew      & 94.47 \pm 0.50  & 58.73 \pm 0.93  & \f {NA} & 63.73 \pm 1.59  & 71.60 \pm 1.94  & 82.97 \pm 2.06  & 62.00 \pm 2.38 & 82.97 \pm 1.07  & \f {OOT} & 97.87 \pm 0.61  & \w 98.10 \pm 0.39 \\
    Synthie           & 51.05 \pm 1.70  & 68.07 \pm 0.79  & \f {NA} & 82.38 \pm 0.63  & 50.04 \pm 0.69  & 50.26 \pm 2.83  & \w 96.57 \pm 0.50 & 47.38 \pm 2.23  & \f {OOT}             & 48.53 \pm 2.01  & 48.88 \pm 2.40 \\
    \bottomrule
      \end{longtable}
\end{landscape}

\begin{table}
  \footnotesize
  \centering
  \setlength{\tabcolsep}{1.5pt}
  \sisetup{
    detect-weight           = true,
    detect-inline-weight    = math,
    table-format            = 2.2(3),
    separate-uncertainty    = true,
    table-align-uncertainty = true,
    tight-spacing           = true,
  }
  \newcommand{\w}{\color{red}}
  \newcommand{\f}{\color{gray}}
  \begin{tabular}{>{\ttfamily}lSS}
    \toprule
    \normalfont{Dataset} & {H($\vertices$)} & {H($\edges$)} \\
    \midrule
    AIDS              & 99.62 \pm 0.02  & 99.64 \pm 0.02\\
    BZR               & 72.85 \pm 0.60  & 57.29 \pm 2.23\\
    BZR\_MD           & 75.86 \pm 0.94  & 67.83 \pm 1.36\\
    COIL-DEL          & 88.82 \pm 0.05  & 82.94 \pm 0.08\\
    COIL-RAG          & 76.40 \pm 0.12  & \f {NA}\\
    COLLAB            & 50.29 \pm 2.37  & 90.78 \pm 0.17\\
    COX2              & 64.65 \pm 0.71  & 69.86 \pm 2.16\\
    COX2\_MD          & 67.54 \pm 0.66  & 57.94 \pm 1.68\\
    DD                & 76.40 \pm 4.99  & 84.87 \pm 0.25\\
    DHFR              & 42.49 \pm 2.50  & 72.41 \pm 0.70\\
    DHFR\_MD          & 47.11 \pm 2.92  & 52.50 \pm 2.63\\
    ENZYMES           & 59.47 \pm 0.32  & 62.36 \pm 0.67\\
    ER\_MD            & 75.17 \pm 0.53  & \w 79.10 \pm 0.73\\
    FRANKENSTEIN      & 71.23 \pm 0.11  & 38.01 \pm 3.41\\
    IMDB-BINARY       & 49.62 \pm 0.41  & 81.25 \pm 0.57\\
    IMDB-MULTI        & 50.66 \pm 0.84  & \w 67.26 \pm 0.51\\
    KKI               & 47.48 \pm 5.02  & 50.41 \pm 4.20\\
    Letter-high       & 87.13 \pm 0.03  & \f {NA}\\
    Letter-low        & 93.55 \pm 0.05  & \f {NA}\\
    Letter-med        & 91.55 \pm 0.04  & \f {NA}\\
    MSRC\_21          & 99.22 \pm 0.05  & 99.14 \pm 0.07\\
    MSRC\_21C         & 95.22 \pm 1.27  & 94.63 \pm 0.58\\
    MSRC\_9           & 98.58 \pm 0.18  & 99.05 \pm 0.14\\
    MUTAG             & 89.80 \pm 1.05  & 89.20 \pm 0.79\\
    Mutagenicity      & 73.70 \pm 0.37  & 50.80 \pm 2.40\\
    NCI1              & 70.38 \pm 0.25  & 50.78 \pm 4.44\\
    NCI109            & 69.19 \pm 0.68  & 49.40 \pm 3.44\\
    OHSU              & 48.28 \pm 5.97  & 41.45 \pm 3.81\\
    PROTEINS          & 77.68 \pm 0.37  & 75.46 \pm 0.29\\
    PROTEINS\_full     & 77.68 \pm 0.37 & 75.46 \pm 0.29\\
    PTC\_FM            & 50.02 \pm 4.26  & 44.69 \pm 2.42\\
    PTC\_FR            & 53.48 \pm 1.81  & 55.91 \pm 2.99\\
    PTC\_MM            & 63.53 \pm 1.71  & 47.29 \pm 2.98\\
    PTC\_MR            & 50.46 \pm 2.95  & 45.86 \pm 2.93\\
    Peking\_1          & 40.85 \pm 5.02  & 46.89 \pm 7.56\\
    REDDIT-BINARY     & 49.12 \pm 6.96  & 84.81 \pm 0.31\\
    REDDIT-MULTI-12K  & 44.52 \pm 1.75  & 77.65 \pm 0.08\\
    REDDIT-MULTI-5K   & 44.78 \pm 4.14  & 78.01 \pm 0.14\\
    SYNTHETIC         & 50.00 \pm 0.00  & 50.00 \pm 0.00\\
    SYNTHETICnew      & 62.78 \pm 0.52  & 76.18 \pm 0.64\\
    Synthie           & 83.20 \pm 0.82  & 82.97 \pm 0.76\\
    \bottomrule
  \end{tabular}
  \caption{Empirical performance of the baseline graph kernels on the benchmark data sets.  The
      mean area under the Receiver Operating Characteristic curve~(AUROC) over
      $10$ iterations of a nested cross-validation procedure is shown along
      with its corresponding standard deviation.
      The highest AUROC value for each data set (considering the baseline
      kernels and normal kernels) is shown in \textcolor{red}{red}. OOM
      means ``out-of-memory", OOT means ``out-of-time", and NA indicates that the given implementation could not
handle a certain data set.}
  \label{tab:Classification auroc baseline}
\end{table}


\begin{landscape}
  \setlength{\tabcolsep}{1.50pt}
  \sisetup{
    detect-weight           = true,
    detect-inline-weight    = math,
    table-format            = 2.2(3),
    separate-uncertainty    = true,
    table-align-uncertainty = true,
    tight-spacing           = true,
  }
  \footnotesize
  \newcommand{\w}{\color{red}}
  \newcommand{\f}{\color{gray}}
  \begin{longtable}{>{\tt}lSSSSSSSSSSS}
  \caption{%
      Empirical performance of graph kernels on the benchmark data sets.  The
      mean area under the Receiver Operating Characteristic curve~(AUROC) over
      $10$ iterations of a nested cross-validation procedure is shown along
      with its corresponding standard deviation.
      The highest AUROC value for each data set (considering the baseline
      kernels and normal kernels) is shown in \textcolor{red}{red} OOM
      means ``out-of-memory", OOT means ``out-of-time", and NA indicates that the given implementation could not
handle a certain data set.
    }
    \label{tab:Empirical performance AUROC}\\
    \toprule
    \normalfont{Data set} & {CSM}       & {GH}              & {Graphlet}
                          & {HGK-SP}           & {HGK-WL}
                          & {MLG}              & {MP}
                          & {SP}                & {RW} & {WL}
                          & {WL-OA} \\
    \midrule
    \endhead
      AIDS             & 99.64 \pm 0.04  & 99.57 \pm 0.02  & 99.56 \pm 0.06  & 99.61 \pm 0.02   & 99.25 \pm 0.05   & 98.57 \pm 0.10 & 99.64 \pm 0.02   & 99.64 \pm 0.02  & \w 99.67 \pm 0.01  & 99.61 \pm 0.04  & 99.63 \pm 0.03\\
      BZR               & 85.17 \pm 1.14  & 73.06 \pm 1.20  & 49.81 \pm 3.39 & 79.45 \pm 0.80   & 83.77 \pm 0.88   & \w 89.49 \pm 1.14  & 89.03 \pm 1.07   & 76.05 \pm 3.13  & 60.53 \pm 1.43  & 87.16 \pm 0.89  & 87.77 \pm 1.36\\
      BZR\_MD            & \w 85.53 \pm 1.05  & 48.40 \pm 7.05  & 35.35 \pm 3.32 & 68.96 \pm 2.09   & 48.14 \pm 8.64   & 35.11 \pm 2.89  & 67.41 \pm 1.41   & 75.37 \pm 1.69  & 75.70 \pm 1.05  & 70.24 \pm 1.00  & 73.13 \pm 1.45\\
      COIL-DEL          & \f {OOT}             & \f {OOM} & \f {NA} & 98.93 \pm 0.02   & 98.13 \pm 0.03   & 53.91 \pm 5.43  & \w 99.43 \pm 0.02   & 89.34 \pm 0.07  & \f {OOT}             & 88.34 \pm 0.14  & 87.85 \pm 0.12\\
      COIL-RAG          & \f {NA}  & \f {OOM}             & 73.16 \pm 0.15 & \w 99.85 \pm 0.01   & 99.83 \pm 0.01   & \f {OOM} & 99.36 \pm 0.02   & \f {NA}  & \f {NA} & 76.57 \pm 0.14  & 76.73 \pm 0.14\\
      COLLAB            & \f {OOT}             & \f {OOM} & \f {NA} & 91.17 \pm 0.21   & \f {OOM}              & \f {OOM}             & \w 92.97 \pm 0.20   & 91.87 \pm 0.20  & \f {OOM}             & 81.48 \pm 1.66  & 92.70 \pm 0.09\\
      COX2              & 81.63 \pm 1.00  & 69.65 \pm 0.79  & 44.98 \pm 0.43 & 73.47 \pm 1.00   & 73.19 \pm 1.32   & 74.32 \pm 0.93  & \w 81.82 \pm 1.50   & 75.70 \pm 4.88  & 55.10 \pm 3.68  & 79.87 \pm 1.01  & 79.96 \pm 0.51\\
      COX2\_MD           & \f {OOT}             & 59.61 \pm 3.86 & 48.76 \pm 2.05 & 64.54 \pm 0.65   & 53.29 \pm 2.53   & 48.54 \pm 1.17   & 63.78 \pm 1.32    & \w 69.12 \pm 0.78  & 68.57 \pm 1.64  & 63.37 \pm 2.87  & 65.38 \pm 2.59\\
      DD                & \f {OOM}             & \f {OOM} & \f {NA} & \f {OOM} & 84.51 \pm 0.48   & 82.16 \pm 0.39  & 85.51 \pm 0.44 & \w 87.44 \pm 0.38  & \f {OOM} & 83.93 \pm 1.90  & 84.75 \pm 1.41\\
      DHFR              & 85.40 \pm 0.80  & 76.21 \pm 0.46  & 45.44 \pm 1.47 & 77.95 \pm 0.48   & 80.39 \pm 0.66   & 88.66 \pm 0.71 & 88.41 \pm 0.58   & 85.79 \pm 0.94  & \f {OOT} & 89.49 \pm 0.65  & \w 90.17 \pm 0.74\\
      DHFR\_MD           & \f {OOT}             & 54.50 \pm 2.69 & 47.62 \pm 4.04 & \w 62.65 \pm 4.39   & 56.66 \pm 1.79   & 54.69 \pm 1.85  & 60.05 \pm 2.21   & 54.37 \pm 2.82  & 57.92 \pm 2.96  & 55.72 \pm 2.46  & 61.84 \pm 1.44\\
      ENZYMES           & \w 91.73 \pm 0.47  & 78.97 \pm 0.41  & \f {NA} & 87.97 \pm 0.36   & 84.00 \pm 0.39   & 79.79 \pm 0.82  & 85.37 \pm 0.57   & 74.67 \pm 0.53   & 62.56 \pm 0.93  & 80.45 \pm 0.46  & 84.86 \pm 0.48\\
      ER\_MD             & \f {OOT}             & 72.86 \pm 0.28 & 70.39 \pm 0.35 & 72.58 \pm 0.34   & 71.13 \pm 1.10   & 61.42 \pm 1.16  & 73.28 \pm 1.17   & 76.18 \pm 0.91  & \f {NA}             & 73.28 \pm 1.16  & 75.84 \pm 0.98\\
      FRANKENSTEIN      & \f {OOT}             & \f {OOM} & 60.09 \pm 2.41 & 65.14 \pm 1.21   & \w 77.25 \pm 0.26   & 77.22 \pm 0.19  & 59.83 \pm 0.61   & 52.10 \pm 4.09  & \f {OOI}             & 76.58 \pm 0.35  & 77.09 \pm 0.50\\
      IMDB-BINARY       & \f {OOT}             & \f {OOM} & \f {NA} & 80.88 \pm 0.25   & 80.08 \pm 0.54   & 44.32 \pm 2.69   & 77.07 \pm 1.29    & 80.21 \pm 0.61  & \w 81.87 \pm 0.46  & 78.74 \pm 0.99  & 80.08 \pm 0.89\\
      IMDB-MULTI        & \f {OOT}             & \f {OOM} & \f {NA} & 66.92 \pm 0.29   & 66.24 \pm 0.46   & 44.93 \pm 0.72  & 65.41 \pm 0.46   & 66.95 \pm 0.38  & 62.37 \pm 0.40  & 65.87 \pm 0.70  & 65.78 \pm 0.50\\
      KKI               & 48.72 \pm 6.41  & \w 59.90 \pm 4.85  & \f {NA} & 52.49 \pm 4.29   & 42.33 \pm 4.20   & 48.41 \pm 6.48  & 53.22 \pm 7.32   & 56.60 \pm 4.98  & 54.61 \pm 5.11  & 51.20 \pm 6.00  & 49.31 \pm 6.26\\
      Letter-high       & \f {NA}  & \f {OOM}             & 84.96 \pm 0.71 & 99.39 \pm 0.02   & \w 99.49 \pm 0.02   & 73.36 \pm 0.50 & 99.28 \pm 0.05    & \f {NA}  & \f {NA}  & 86.78 \pm 0.08  & 86.86 \pm 0.13\\
      Letter-low        & \f {NA}  & \f {OOM}             & 93.21 \pm 0.01 & \w 100.00 \pm 0.00  & \w 100.00 \pm 0.00  & 78.12 \pm 0.42  & \w 100.00 \pm 0.00  & \f {NA}  & \f {NA}  & 93.63 \pm 0.05  & 93.61 \pm 0.05\\
      Letter-med        & \f {NA}  & \f {OOM}             & 90.91 \pm 0.04 & 99.72 \pm 0.02   & 99.78 \pm 0.01   & 74.46 \pm 1.20 & \w 99.85 \pm 0.01   & \f {NA}  & \f {NA} & 91.77 \pm 0.06  & 91.72 \pm 0.11\\
      MSRC\_21           & \f {OOT}             & 99.38 \pm 0.05  & \f {NA} & 99.16 \pm 0.05   & 98.75 \pm 0.06   & 95.75 \pm 0.10 & 99.45 \pm 0.05   & 99.31 \pm 0.07  & \f {OOT} & 99.20 \pm 0.06  & \w 99.47 \pm 0.04\\
      MSRC\_21C          & 94.04 \pm 0.69  & 94.96 \pm 1.30  & \f {NA} & 95.40 \pm 1.03   & 94.63 \pm 0.99   & 90.63 \pm 1.03  & 94.55 \pm 1.38   & 95.06 \pm 0.42  & 94.37 \pm 0.65 & 95.19 \pm 1.43  & \w 95.55 \pm 1.23\\
      MSRC\_9            & 99.03 \pm 0.27  & \w 99.13 \pm 0.08  & \f {NA} & 98.64 \pm 0.29   & 98.51 \pm 0.22   & 98.21 \pm 0.39 & 98.84 \pm 0.09   & \w 99.13 \pm 0.13  & 98.81 \pm 0.10  & 98.71 \pm 0.11  & 98.93 \pm 0.16\\
      MUTAG             & \w 93.64 \pm 0.95  & 83.62 \pm 0.85  & 85.72 \pm 0.81 & 87.87 \pm 0.79   & 88.11 \pm 0.52   & 86.43 \pm 1.29  & 92.65 \pm 1.26   & 91.16 \pm 0.93  & 91.62 \pm 1.09  & 91.54 \pm 1.46  & 90.03 \pm 2.56\\
      Mutagenicity      & \f {OOT}             & \f {OOM} & 46.84 \pm 2.78 & 79.93 \pm 0.05   & 87.34 \pm 0.33   & \f {OOM} & 86.12 \pm 0.29   & 56.35 \pm 2.93 & \f {OOT} & 88.93 \pm 0.34  & \w 90.12 \pm 0.46\\
      NCI1              & \f {OOT}             & \f {OOM} & 49.98 \pm 4.29 & 76.31 \pm 0.08   & 87.19 \pm 0.13   & 84.41 \pm 0.35  & 85.27 \pm 0.66   & 58.33 \pm 3.49  & \f {OOT}             & 91.02 \pm 0.22  & \w 92.14 \pm 0.17\\
      NCI109            & \f {OOT}             & \f {OOM} & 51.74 \pm 0.26 & 76.65 \pm 0.06   & 86.72 \pm 0.14   & 83.98 \pm 0.38  & 83.96 \pm 0.39   & 57.50 \pm 4.18  & \f {OOT}             & 91.19 \pm 0.16  & \w 91.76 \pm 0.07\\
      OHSU              & 41.51 \pm 8.60  & \w 58.75 \pm 4.30  & \f {NA} & 47.59 \pm 5.34   & 47.70 \pm 7.30   & 45.34 \pm 6.36  & 47.78 \pm 7.08   & 45.62 \pm 2.65  & \f {OOT}             & 44.25 \pm 4.75  & 42.41 \pm 4.92\\
      PROTEINS          & \f {OOT}             & 80.82 \pm 0.07  & \f {NA} & 79.65 \pm 0.16   & 80.78 \pm 0.37   & 80.67 \pm 0.47 & 78.33 \pm 0.54   & \w 82.20 \pm 0.25  & \f {OOM}             & 79.37 \pm 0.77  & 78.96 \pm 0.93\\
      PROTEINS\_full     & 70.67 \pm 0.45  & \f {OOM}             & \f {NA} & 81.18 \pm 0.15   & 81.76 \pm 0.24   & 80.67 \pm 0.47 & 78.99 \pm 0.59   & \w 82.20 \pm 0.25  & \f {OOM}             & 79.38 \pm 0.77  & 78.97 \pm 0.93\\
      \pagebreak
      PTC\_FM            & 51.98 \pm 3.51  & 61.00 \pm 0.75  & 53.14 \pm 2.44 & 61.75 \pm 1.21   & 64.06 \pm 1.68   & 54.77 \pm 4.33  & \w 66.27 \pm 1.94   & 57.40 \pm 4.02  & 52.70 \pm 3.10  & 62.41 \pm 2.47  & 64.44 \pm 2.46\\
      PTC\_FR            & 59.43 \pm 3.24  & 62.00 \pm 1.39  & 51.12 \pm 2.18 & 63.04 \pm 0.92   & 64.04 \pm 2.03   & 50.60 \pm 2.90 & 65.28 \pm 1.64   & \w 66.00 \pm 1.75  & 49.06 \pm 1.81  & 63.05 \pm 2.99 & 63.65 \pm 1.62\\
      PTC\_MM            & 53.62 \pm 3.23  & 65.17 \pm 1.39  & 47.43 \pm 2.79 & 63.71 \pm 0.97   & \w 68.12 \pm 1.30   & 49.60 \pm 4.29  & 66.16 \pm 1.32   & 63.38 \pm 3.64  & 61.47 \pm 2.12  & 63.56 \pm 2.24  & 61.79 \pm 2.34\\
      PTC\_MR            & 62.14 \pm 2.36  & 61.32 \pm 1.35  & 48.44 \pm 3.79 & 61.54 \pm 1.04   & 63.29 \pm 1.80   & 63.40 \pm 3.31  & \w 64.90 \pm 1.26   & 61.68 \pm 1.97  & 53.17 \pm 3.13  & 63.07 \pm 1.27  & 64.33 \pm 2.06\\
      Peking\_1          & 45.34 \pm 5.56  & \f {OOM}             & \f {NA} & \w 54.80 \pm 7.22   & 49.78 \pm 7.76   & 46.47 \pm 5.09  & 52.99 \pm 6.60   & 42.86 \pm 6.70  & \f {OOT}             & 49.86 \pm 7.52  & 49.99 \pm 6.37\\
      REDDIT-BINARY     & \f {OOM}             & \f {OOM} & \f {NA} & \f {OOM} & \f {OOM}              & \f {OOM}             & \w 95.54 \pm 0.26 & 95.22 \pm 0.15  & \f {OOM} & 83.51 \pm 2.72  & 94.62 \pm 0.13\\
      REDDIT-MULTI-12K  & \f {OOM}             & \f {OOM} & \f {NA} & \f {OOM} & \f {OOM}              & \f {OOM} & \w 85.66 \pm 0.09 & 80.78 \pm 0.30  & \f {OOM} & 75.36 \pm 0.32  & \f {OOM}\\
      REDDIT-MULTI-5K   & \f {OOM}             & \f {OOM} & \f {NA} & \f {OOM} & \f {OOM}              & \f {OOM}             & \w \w 81.77 \pm 0.16 & 79.80 \pm 0.22  & \f {OOM} & 81.59 \pm 0.13  & \f {OOM}\\
      SYNTHETIC         & \w 71.96 \pm 1.30  & 39.20 \pm 1.81  & 50.00 \pm 0.00 & 39.27 \pm 3.08    & 27.86 \pm 5.01   & 50.00 \pm 0.00  & 50.00 \pm 0.00   & 50.00 \pm 0.00  & \f {OOT}             & 50.00 \pm 0.00  & 50.00 \pm 0.00\\
      SYNTHETICnew      & 98.69 \pm 0.30  & 38.00 \pm 1.86  & \f {NA} & 33.72 \pm 2.60   & 29.92 \pm 6.00   & 90.91 \pm 1.26  & 68.43 \pm 2.23   & 91.19 \pm 1.09  & \f {OOT} & \w 99.90 \pm 0.11  & 99.87 \pm 0.13\\
      Synthie           & 82.64 \pm 0.80  & 91.43 \pm 0.21  & \f {NA} & 97.08 \pm 0.08   & 86.60 \pm 0.36   & 83.11
      \pm 0.99  & \w 99.74 \pm 0.10   & 82.95 \pm 1.07  & \f {OOT}             & 82.56 \pm 0.80  & 82.93 \pm 0.96\\
  \bottomrule
      \end{longtable}
\end{landscape}

\begin{table}[tbp]
  \centering
  \setlength{\tabcolsep}{2.5pt}
  \sisetup{
    detect-weight           = true,
    detect-inline-weight    = math,
    table-format            = 2.2(3),
    separate-uncertainty    = true,
    table-align-uncertainty = true,
    tight-spacing           = false,
  }
  \begin{tabular}{lS}
    \toprule
    Kernel & Rank\\
    \midrule
    Message passing$^\ast$                         & 4.27 \pm 2.49\\
    Weisfeiler--Lehman~(optimal assignment)$^\ast$ & 4.51 \pm 2.34\\
    Weisfeiler--Lehman$^\ast$                      & 5.04 \pm 2.05\\
    HGK-WL$^\ast$                                  & 5.85 \pm 3.26\\
    HGK-SP                                         & 5.85 \pm 3.02\\
    Shortest--path                                 & 5.89 \pm 3.02\\
    Histogram~($\vertices$)                        & 6.74 \pm 2.87\\
    Multiscale Laplacian                           & 7.57 \pm 3.30\\
    Histogram~($\edges$)                           & 7.74 \pm 3.20\\
    Subgraph Matching                              & 8.48 \pm 3.69\\
    GraphHopper                                    & 8.67 \pm 3.18\\
    Random Walk                                    & 9.88 \pm 3.20\\
    Graphlet                                       & 10.49 \pm 2.30\\

    \bottomrule
  \end{tabular}
  \caption{%
    Mean rank~(and standard deviation) for each graph kernel. The
    ranking has been obtained by simulating accuracy distributions.
    Methods based on the Weisfeiler--Lehman relabelling scheme or one of
    its variants are printed with an additional asterisk~($\ast$) after
    their name.
  }
  \label{tab:Ranks}
\end{table}

\paragraph{Ranking}

As an additional simple summary of Table~\ref{tab:Empirical performance}, we
 also calculate the \emph{ranks} of individual algorithms.
To make proper use of the standard deviation, we consider each kernel's
accuracy values to be normally distributed, which is a standard assumption
when employing a cross-validation setup. Using the corresponding mean
accuracy and standard deviation as the respective mean and standard deviation
of such a normal distribution, we simulate $10,000$ independent ``draws'' of
accuracy values from these distributions. Ultimately, this will permit us
to calculate a mean rank and standard deviation for each kernel.
The results of this Monte Carlo simulation are shown in Table~\ref{tab:Ranks}.
In some sense, the table confirms our observation from above: approaches based
on the Weisfeiler--Lehman relabelling scheme tend to outperform other kernels.
Somewhat surprisingly, the vertex histogram kernel is not among the
worst-performing kernels---we would expect that this kernel, which after
all is not using \emph{any} connectivity information of a graph, cannot
reach a competitive performance. As this kernel, and edge histogram
kernels, are often used as a baseline kernel, we consider it necessary
to perform a more detailed analysis here.
Hence, after giving a breakdown by graph data set type for each kernel
in Section~\ref{sec:Breakdown}, we will assess the performance of
histogram kernels in a more detailed manner in
Section~\ref{sec:Histogram kernels}.

\subsection{Breakdown of results per data set type}\label{sec:Breakdown}

As a precursor to deciding which kernel to apply for a \emph{new} data
set, we first use Table~\ref{tab:Summary statistics} to perform a coarse
type assessment. Looking at the presence and absence of either node or edge labels or attributes, we
create the following classes of data sets:
\begin{compactenum}[(i)]
  \item \emph{fully unlabelled}: data sets that do not have \emph{any}
    attributes or labels
  \item \emph{node labels}: data sets that \emph{only} have node labels
  \item \emph{fully labelled}: data sets that have node labels and edge
    labels but no other information
  \item \emph{only node attributes}: data sets that contain node
    attributes and nothing else; for reasons of simplicity, we also
    include \texttt{COIL-DEL} in this category because said data set
    would otherwise constitute a category of its own.
  \item \emph{full node information}: data sets that contain node labels
    and node attributes, but nothing else.
  \item \emph{everything but node attributes}: data sets that
    \emph{lack} node attributes, but feature edge attributes as well as
    node labels and edge labels.
\end{compactenum}
This list is exhaustive concerning the benchmark data sets. Technically,
other categories are possible---such as data sets that only contain edge
labels and nothing else---but there are no examples of such data sets in
the repository at present.
We provide a ranking based on the \emph{average} performance of a graph kernel on
some data set---this will promote graph kernels whose performance on the
data set is \emph{consistent}, \ie\ that are capable of classifying
graphs from that data set to a similar extent.
Table~\ref{tab:Breakdown mean accuracy ranking} depicts the resulting
table. The dominance of Weisfeiler--Lehman approaches is now even more
evident---these types of graph kernels perform consistently well for
several types of data sets. Class~vi~(\emph{everything but node attributes})
of data sets is special and surprising in the sense that vertex histogram kernels
exhibit suitable overall performance here. In total, the table seems to suggest that, at
least for data sets of types~i, ii, and iii, the Weisfeiler--Lehman
subtree kernel or its optimal assignment variant are most suitable;
alternatively, the message passing~(MP) kernel, with a somewhat higher
complexity, can be used to achieve a good performance.

\begin{table}[tbp]
  \centering
  \begin{tabular}{llll}
    \toprule
    Type & 1\st   & 2\nd   & 3\rd\\
    \midrule
    i    & SP     & MP     & WL\\
    ii   & WL-OA  & WL     & HGK-WL\\
    iii  & WL-OA  & MP     & WL\\
    iv   & MP     & HGK-SP & HGK-WL\\
    v    & WL-OA  & MP     & CSM\\
    vi   & SP     & VH  & WL-OA\\
    \bottomrule
  \end{tabular}
  \caption{%
    A ranking of graph kernels based on their mean accuracy achieved on
    a specific type of data set.
  }
  \label{tab:Breakdown mean accuracy ranking}
\end{table}

\begin{figure}[tbp]


  \centering
  \tikzstyle{every node}=[font=\tiny]

  \pgfplotsset{every axis/.style = {%
      tick align = outside,
      axis x line*   = bottom,
      axis y line*   = left,
      width          = 0.50\linewidth,
      ylabel         = {AUROC~(in \si{\percent})},
      xmin           = 0,
      mark size      = 1pt,
      xtick          = {1, 2, 3, 4, 5, 6, 7, 8, 9, 10, 11, 12, 13},
      xticklabels    = {
        MP, WL-OA, WL, HGK-WL, HGK-SP, SP,
        H~($\vertices$), MLG, H~($\edges$),
        CSM, GH, RW, GL
      },
      ytick          = {0, 10, 20, 30, 40, 50, 60, 70, 80, 90, 100},
      boxplot/draw direction = y,
      xticklabel style       = {
        rotate = 90,
      },
      baseline,
    }
  }

  \subcaptionbox{\emph{Fully unlabelled}}{%
  \iffinal
  \includegraphics[width=0.4\textwidth]{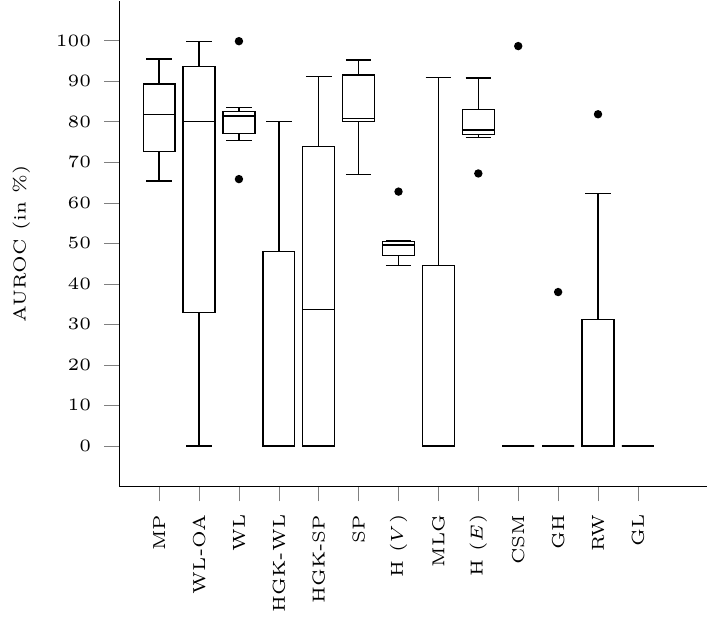}
  \else
    \begin{tikzpicture}
      \begin{axis}
        \addplot[boxplot] table[col sep = comma, x expr =  1, y = accuracy] {Data/aurocs_per_class/i_MP.csv};
        \addplot[boxplot] table[col sep = comma, x expr =  2, y = accuracy] {Data/aurocs_per_class/i_WLOA.csv};
        \addplot[boxplot] table[col sep = comma, x expr =  3, y = accuracy] {Data/aurocs_per_class/i_WL.csv};
        \addplot[boxplot] table[col sep = comma, x expr =  4, y = accuracy] {Data/aurocs_per_class/i_HGKWL_seed0.csv};
        \addplot[boxplot] table[col sep = comma, x expr =  5, y = accuracy] {Data/aurocs_per_class/i_HGKSP_seed0.csv};
        \addplot[boxplot] table[col sep = comma, x expr =  6, y = accuracy] {Data/aurocs_per_class/i_SP_gkl.csv};
        \addplot[boxplot] table[col sep = comma, x expr =  7, y = accuracy] {Data/aurocs_per_class/i_VH.csv};
        \addplot[boxplot] table[col sep = comma, x expr =  8, y = accuracy] {Data/aurocs_per_class/i_MLG.csv};
        \addplot[boxplot] table[col sep = comma, x expr =  9, y = accuracy] {Data/aurocs_per_class/i_EH_gkl.csv};
        \addplot[boxplot] table[col sep = comma, x expr = 10, y = accuracy] {Data/aurocs_per_class/i_CSM_gkl.csv};
        \addplot[boxplot] table[col sep = comma, x expr = 11, y = accuracy] {Data/aurocs_per_class/i_GH.csv};
        \addplot[boxplot] table[col sep = comma, x expr = 12, y = accuracy] {Data/aurocs_per_class/i_RW_gkl.csv};
        \addplot[boxplot] table[col sep = comma, x expr = 13, y = accuracy] {Data/aurocs_per_class/i_GL.csv};
      \end{axis}
    \end{tikzpicture}
    \fi
  }
  \subcaptionbox{\emph{Node labels}}{%
    \iffinal
    \includegraphics[width=0.4\textwidth]{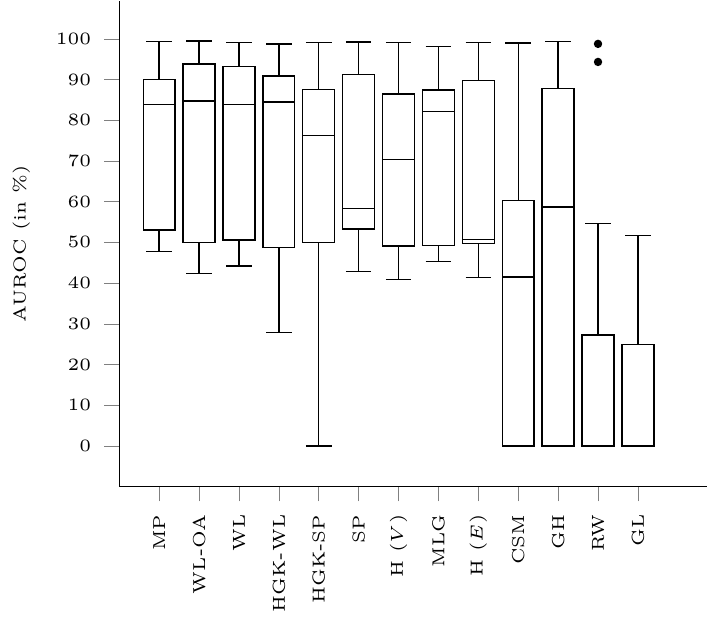}
    \else
    \begin{tikzpicture}
      \begin{axis}
        \addplot[boxplot] table[col sep = comma, x expr =  1, y = accuracy] {Data/aurocs_per_class/ii_MP.csv};
        \addplot[boxplot] table[col sep = comma, x expr =  2, y = accuracy] {Data/aurocs_per_class/ii_WLOA.csv};
        \addplot[boxplot] table[col sep = comma, x expr =  3, y = accuracy] {Data/aurocs_per_class/ii_WL.csv};
        \addplot[boxplot] table[col sep = comma, x expr =  4, y = accuracy] {Data/aurocs_per_class/ii_HGKWL_seed0.csv};
        \addplot[boxplot] table[col sep = comma, x expr =  5, y = accuracy] {Data/aurocs_per_class/ii_HGKSP_seed0.csv};
        \addplot[boxplot] table[col sep = comma, x expr =  6, y = accuracy] {Data/aurocs_per_class/ii_SP_gkl.csv};
        \addplot[boxplot] table[col sep = comma, x expr =  7, y = accuracy] {Data/aurocs_per_class/ii_VH.csv};
        \addplot[boxplot] table[col sep = comma, x expr =  8, y = accuracy] {Data/aurocs_per_class/ii_MLG.csv};
        \addplot[boxplot] table[col sep = comma, x expr =  9, y = accuracy] {Data/aurocs_per_class/ii_EH_gkl.csv};
        \addplot[boxplot] table[col sep = comma, x expr = 10, y = accuracy] {Data/aurocs_per_class/ii_CSM_gkl.csv};
        \addplot[boxplot] table[col sep = comma, x expr = 11, y = accuracy] {Data/aurocs_per_class/ii_GH.csv};
        \addplot[boxplot] table[col sep = comma, x expr = 12, y = accuracy] {Data/aurocs_per_class/ii_RW_gkl.csv};
        \addplot[boxplot] table[col sep = comma, x expr = 13, y = accuracy] {Data/aurocs_per_class/ii_GL.csv};
      \end{axis}
    \end{tikzpicture}
    \fi
  }
  \subcaptionbox{\emph{Fully labelled}}{%
    \iffinal
    \includegraphics[width=0.4\textwidth]{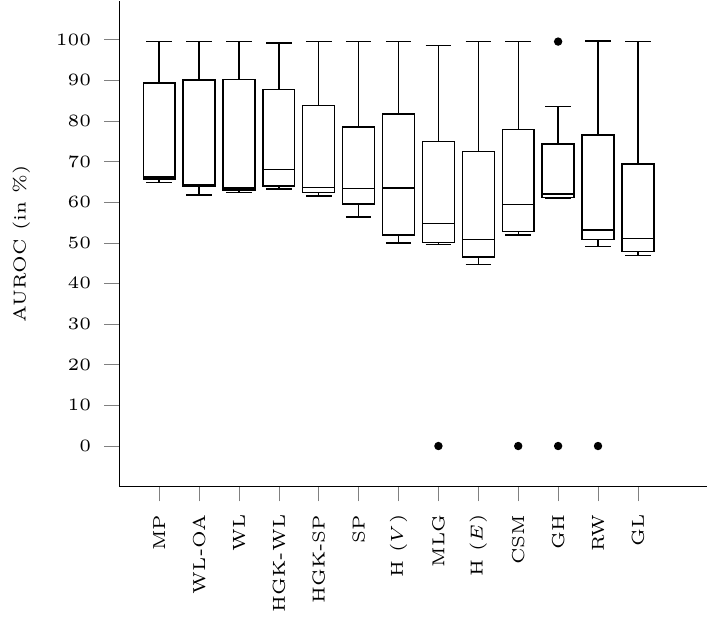}
    \else
    \begin{tikzpicture}
      \begin{axis}
        \addplot[boxplot] table[col sep = comma, x expr =  1, y = accuracy] {Data/aurocs_per_class/iii_MP.csv};
        \addplot[boxplot] table[col sep = comma, x expr =  2, y = accuracy] {Data/aurocs_per_class/iii_WLOA.csv};
        \addplot[boxplot] table[col sep = comma, x expr =  3, y = accuracy] {Data/aurocs_per_class/iii_WL.csv};
        \addplot[boxplot] table[col sep = comma, x expr =  4, y = accuracy] {Data/aurocs_per_class/iii_HGKWL_seed0.csv};
        \addplot[boxplot] table[col sep = comma, x expr =  5, y = accuracy] {Data/aurocs_per_class/iii_HGKSP_seed0.csv};
        \addplot[boxplot] table[col sep = comma, x expr =  6, y = accuracy] {Data/aurocs_per_class/iii_SP_gkl.csv};
        \addplot[boxplot] table[col sep = comma, x expr =  7, y = accuracy] {Data/aurocs_per_class/iii_VH.csv};
        \addplot[boxplot] table[col sep = comma, x expr =  8, y = accuracy] {Data/aurocs_per_class/iii_MLG.csv};
        \addplot[boxplot] table[col sep = comma, x expr =  9, y = accuracy] {Data/aurocs_per_class/iii_EH_gkl.csv};
        \addplot[boxplot] table[col sep = comma, x expr = 10, y = accuracy] {Data/aurocs_per_class/iii_CSM_gkl.csv};
        \addplot[boxplot] table[col sep = comma, x expr = 11, y = accuracy] {Data/aurocs_per_class/iii_GH.csv};
        \addplot[boxplot] table[col sep = comma, x expr = 12, y = accuracy] {Data/aurocs_per_class/iii_RW_gkl.csv};
        \addplot[boxplot] table[col sep = comma, x expr = 13, y = accuracy] {Data/aurocs_per_class/iii_GL.csv};
      \end{axis}
    \end{tikzpicture}
    \fi
  }
  \subcaptionbox{\emph{Only node attributes}}{%
    \iffinal
    \includegraphics[width=0.4\textwidth]{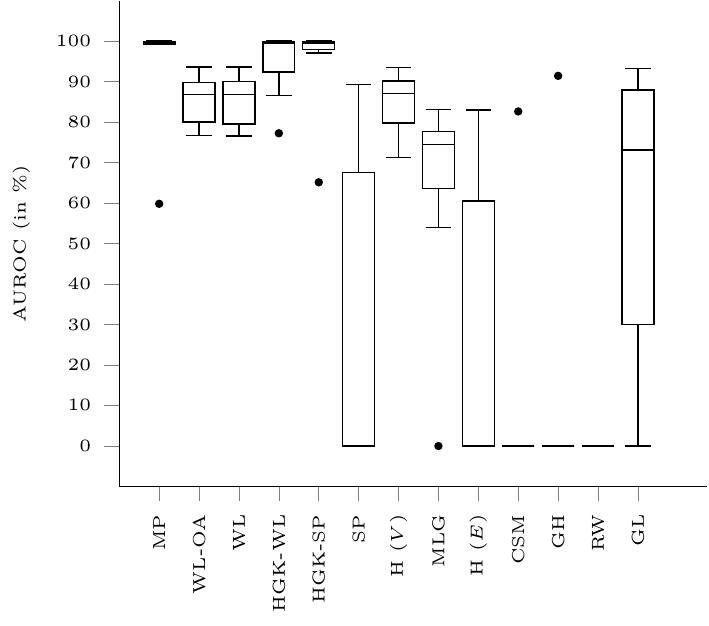}
    \else
    \begin{tikzpicture}
      \begin{axis}
        \addplot[boxplot] table[col sep = comma, x expr =  1, y = accuracy] {Data/aurocs_per_class/iv_MP.csv};
        \addplot[boxplot] table[col sep = comma, x expr =  2, y = accuracy] {Data/aurocs_per_class/iv_WLOA.csv};
        \addplot[boxplot] table[col sep = comma, x expr =  3, y = accuracy] {Data/aurocs_per_class/iv_WL.csv};
        \addplot[boxplot] table[col sep = comma, x expr =  4, y = accuracy] {Data/aurocs_per_class/iv_HGKWL_seed0.csv};
        \addplot[boxplot] table[col sep = comma, x expr =  5, y = accuracy] {Data/aurocs_per_class/iv_HGKSP_seed0.csv};
        \addplot[boxplot] table[col sep = comma, x expr =  6, y = accuracy] {Data/aurocs_per_class/iv_SP_gkl.csv};
        \addplot[boxplot] table[col sep = comma, x expr =  7, y = accuracy] {Data/aurocs_per_class/iv_VH.csv};
        \addplot[boxplot] table[col sep = comma, x expr =  8, y = accuracy] {Data/aurocs_per_class/iv_MLG.csv};
        \addplot[boxplot] table[col sep = comma, x expr =  9, y = accuracy] {Data/aurocs_per_class/iv_EH_gkl.csv};
        \addplot[boxplot] table[col sep = comma, x expr = 10, y = accuracy] {Data/aurocs_per_class/iv_CSM_gkl.csv};
        \addplot[boxplot] table[col sep = comma, x expr = 11, y = accuracy] {Data/aurocs_per_class/iv_GH.csv};
        \addplot[boxplot] table[col sep = comma, x expr = 12, y = accuracy] {Data/aurocs_per_class/iv_RW_gkl.csv};
        \addplot[boxplot] table[col sep = comma, x expr = 13, y = accuracy] {Data/aurocs_per_class/iv_GL.csv};
      \end{axis}
    \end{tikzpicture}
    \fi
  }
  \subcaptionbox{\emph{Full node information}}{%
    \iffinal
    \includegraphics[width=0.4\textwidth]{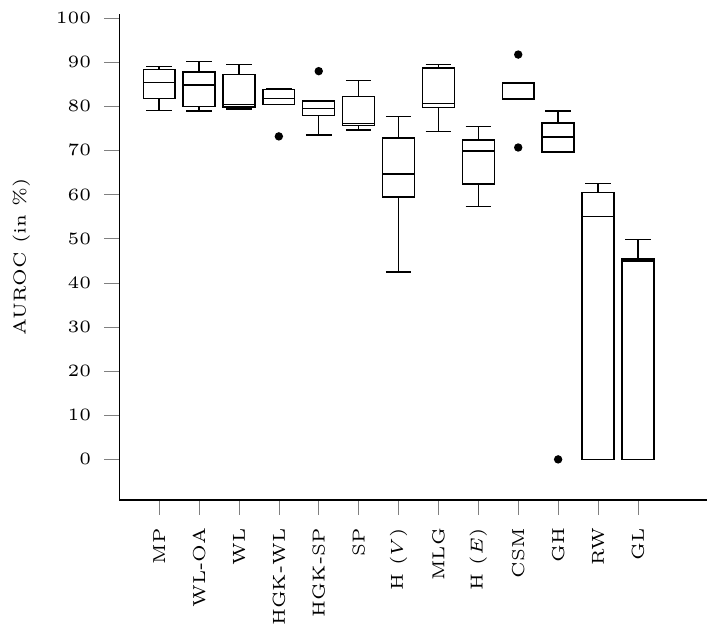}
    \else
    \begin{tikzpicture}
      \begin{axis}
        \addplot[boxplot] table[col sep = comma, x expr =  1, y = accuracy] {Data/aurocs_per_class/v_MP.csv};
        \addplot[boxplot] table[col sep = comma, x expr =  2, y = accuracy] {Data/aurocs_per_class/v_WLOA.csv};
        \addplot[boxplot] table[col sep = comma, x expr =  3, y = accuracy] {Data/aurocs_per_class/v_WL.csv};
        \addplot[boxplot] table[col sep = comma, x expr =  4, y = accuracy] {Data/aurocs_per_class/v_HGKWL_seed0.csv};
        \addplot[boxplot] table[col sep = comma, x expr =  5, y = accuracy] {Data/aurocs_per_class/v_HGKSP_seed0.csv};
        \addplot[boxplot] table[col sep = comma, x expr =  6, y = accuracy] {Data/aurocs_per_class/v_SP_gkl.csv};
        \addplot[boxplot] table[col sep = comma, x expr =  7, y = accuracy] {Data/aurocs_per_class/v_VH.csv};
        \addplot[boxplot] table[col sep = comma, x expr =  8, y = accuracy] {Data/aurocs_per_class/v_MLG.csv};
        \addplot[boxplot] table[col sep = comma, x expr =  9, y = accuracy] {Data/aurocs_per_class/v_EH_gkl.csv};
        \addplot[boxplot] table[col sep = comma, x expr = 10, y = accuracy] {Data/aurocs_per_class/v_CSM_gkl.csv};
        \addplot[boxplot] table[col sep = comma, x expr = 11, y = accuracy] {Data/aurocs_per_class/v_GH.csv};
        \addplot[boxplot] table[col sep = comma, x expr = 12, y = accuracy] {Data/aurocs_per_class/v_RW_gkl.csv};
        \addplot[boxplot] table[col sep = comma, x expr = 13, y = accuracy] {Data/aurocs_per_class/v_GL.csv};
      \end{axis}
    \end{tikzpicture}
    \fi
  }
  \subcaptionbox{\emph{Everything but node attributes}}{%
    \iffinal
    \includegraphics[width=0.4\textwidth]{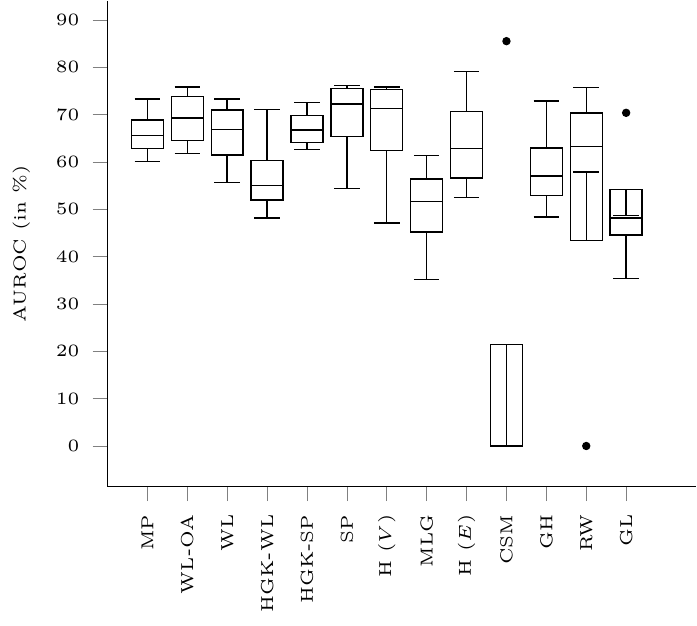}
    \else
    \begin{tikzpicture}
      \begin{axis}
        \addplot[boxplot] table[col sep = comma, x expr =  1, y = accuracy] {Data/aurocs_per_class/vi_MP.csv};
        \addplot[boxplot] table[col sep = comma, x expr =  2, y = accuracy] {Data/aurocs_per_class/vi_WLOA.csv};
        \addplot[boxplot] table[col sep = comma, x expr =  3, y = accuracy] {Data/aurocs_per_class/vi_WL.csv};
        \addplot[boxplot] table[col sep = comma, x expr =  4, y = accuracy] {Data/aurocs_per_class/vi_HGKWL_seed0.csv};
        \addplot[boxplot] table[col sep = comma, x expr =  5, y = accuracy] {Data/aurocs_per_class/vi_HGKSP_seed0.csv};
        \addplot[boxplot] table[col sep = comma, x expr =  6, y = accuracy] {Data/aurocs_per_class/vi_SP_gkl.csv};
        \addplot[boxplot] table[col sep = comma, x expr =  7, y = accuracy] {Data/aurocs_per_class/vi_VH.csv};
        \addplot[boxplot] table[col sep = comma, x expr =  8, y = accuracy] {Data/aurocs_per_class/vi_MLG.csv};
        \addplot[boxplot] table[col sep = comma, x expr =  9, y = accuracy] {Data/aurocs_per_class/vi_EH_gkl.csv};
        \addplot[boxplot] table[col sep = comma, x expr = 10, y = accuracy] {Data/aurocs_per_class/vi_CSM_gkl.csv};
        \addplot[boxplot] table[col sep = comma, x expr = 11, y = accuracy] {Data/aurocs_per_class/vi_GH.csv};
        \addplot[boxplot] table[col sep = comma, x expr = 12, y = accuracy] {Data/aurocs_per_class/vi_RW_gkl.csv};
        \addplot[boxplot] table[col sep = comma, x expr = 13, y = accuracy] {Data/aurocs_per_class/vi_GL.csv};
      \end{axis}
    \end{tikzpicture}
    \fi
  }
  \caption{%
    Accuracy values achieved by each kernel on data sets of a specific class.
    The \mbox{$x$-axis} is ordered according to the average accuracy on the
    complete data set.
  }
  \label{fig:Accuracy breakdown boxplots}
\end{figure}
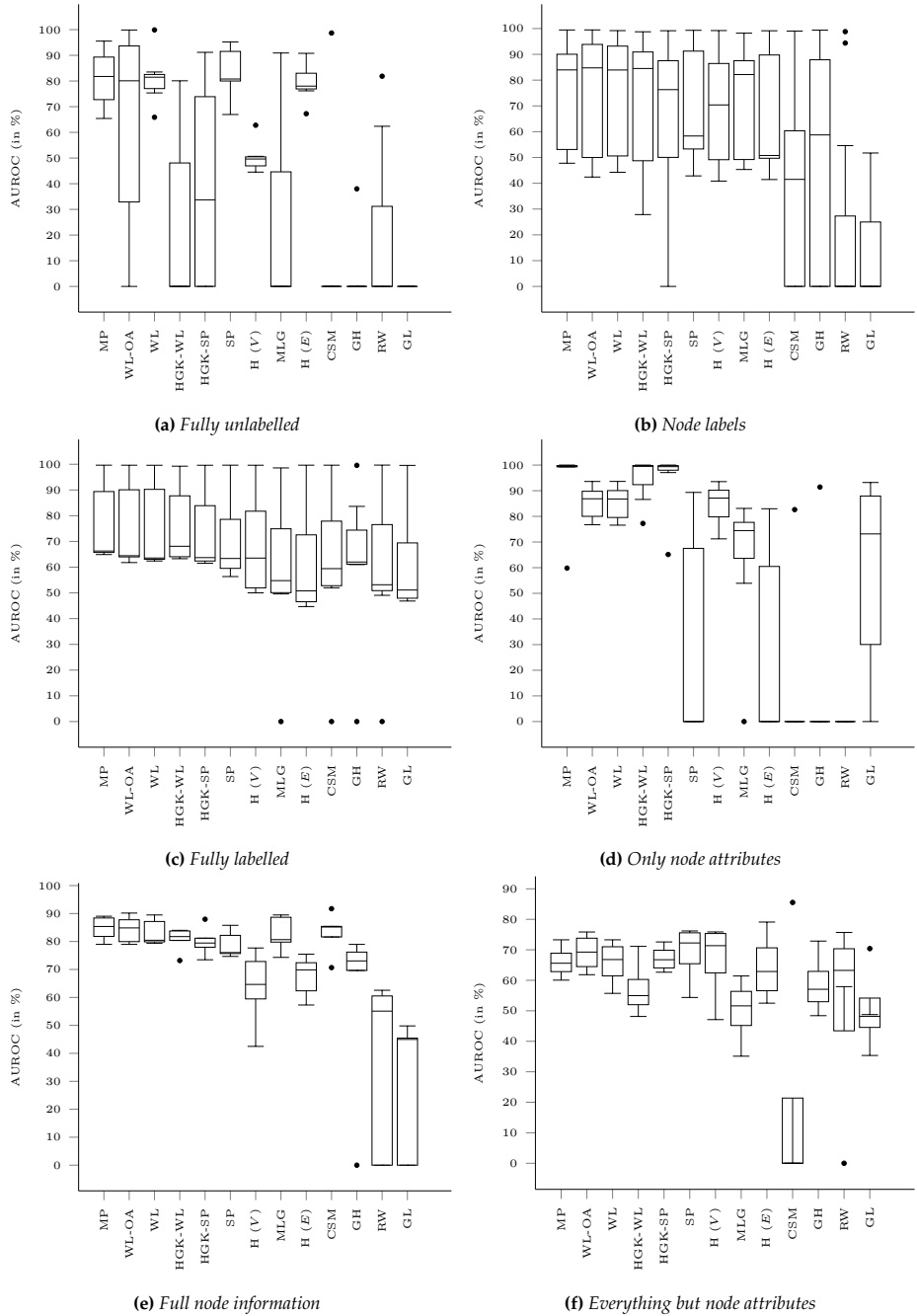

Finally, Figure~\ref{fig:Accuracy breakdown boxplots} provides a visual depiction of the
accuracies of each kernel, broken down by the classes defined above. To provide
a consistent visual design, the graph kernels have been sorted by their ranking
according to Table~\ref{tab:Ranks}.
Boxplots are generated using the AUROC of a specific graph kernel on a data set
in order to ensure comparability.
If all graph kernels would deal with all types of data sets in a consistent
manner, the \emph{mean} of every boxplot should reflect this trend for each
class, and we would be able to observe similar boxplots across all classes.
As the figure demonstrates, this is not the case---the performance of graph
kernels varies depending on the type of data set.
There are several other interesting patterns that emerge from the plot,
though:
\begin{compactenum}
  \item The performance of all graph kernels for the \emph{fully labelled}
    graphs is much more consistent than on other types of data sets.
  \item For graphs with \emph{full node information}, there is a clear
    division between the graph kernels in terms of their classification
    accuracy.
  \item A similar observation holds for the \emph{node-labelled} graphs,
    even though this type of data set exhibits large variances in AUROC.
  \item In general, the \emph{overall} ranking---which was calculated in
    terms of mean accuracy---is not evident in the mean AUROC
    distributions over different data sets.
\end{compactenum}

We will return to this discussion later on when we give suggestions for
\emph{choosing} a suitable graph kernel. Prior to that, we first discuss
the performance of baseline kernels, as well as the data set difficulty
in general.

\subsection{Performance of histogram kernels}\label{sec:Histogram kernels}

The performance of pure label histogram kernels, either based on vertex
labels~($\vertices$) or on edge labels~($\edges$) is noteworthy because they
outperform \emph{all} remaining kernels on 4 data sets out of the 41~(out of
these, the vertex histogram kernel performs best for 3 of them).
The performance seems to suggest that for some of the benchmark data
sets, the underlying structure of the graph is \emph{irrelevant}, at
least insofar as it is not required to obtain good predictive
performance. This raises the issue of assessing the overall difficulty
of the benchmark data sets; we will discuss it further in
Section~\ref{sec:Difficulty}, whereas this section will focus on
a comparison of histogram kernels with other graph kernels.

\subsubsection{Distribution analysis}

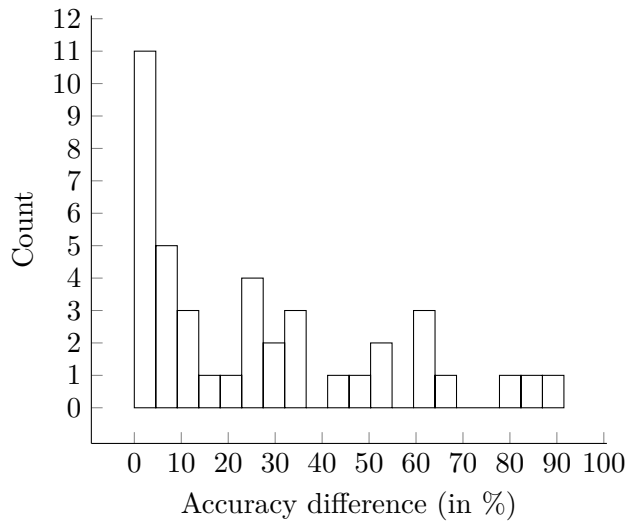
\begin{figure}[tb]
  \centering
    \iffinal
      \includegraphics{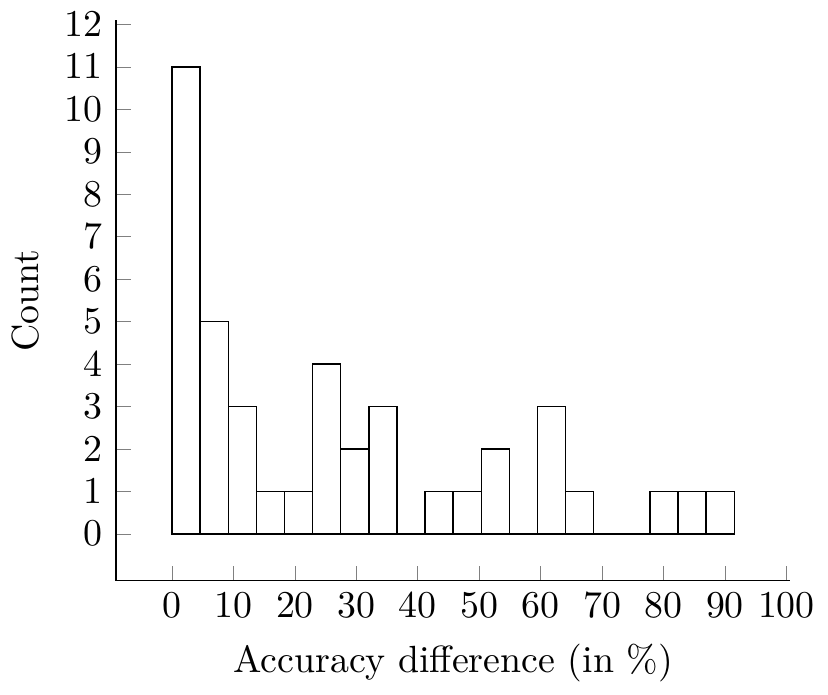}
    \else
  \begin{tikzpicture}
    \begin{axis}[
      axis x line*  = bottom,
      axis y line*  = left,
      enlargelimits = true,
      xlabel        = {Accuracy difference~(in \si{\percent})},
      xtick         = {0, 10, 20, 30, 40, 50, 60, 70, 80, 90, 100},
      ytick         = {0, 1, 2, 3, 4, 5, 6, 7, 8, 9, 10, 11, 12, 13, 14, 15},
      ylabel        = {Count},
    ]
      \addplot[hist = {%
          bins = 20,
          data = x,
        }
      ] file {Data/Vertex_histogram_kernel_accuracy_differences.txt};
    \end{axis}
  \end{tikzpicture}
  \fi
  \caption{%
    A distribution of the relative differences in mean accuracy of the
    vertex histogram kernel and the corresponding best-performing graph
    kernel.
  }
  \label{fig:Vertex histogram kernel accuracy distribution}
\end{figure}

As a first analysis, we look at the vertex histogram kernels.
Figure~\ref{fig:Vertex histogram kernel accuracy distribution} depicts
the \emph{relative} accuracy differences of the vertex histogram kernel,
with respect to the best-performing graph kernel for each data set. To
this end, letting $x_b$ refer to the accuracy of the best-performing graph
kernel on a data set, and $x_v$ refer to the accuracy of the vertex
histogram graph kernel, we calculate
\begin{equation}
  \Delta = \frac{100 \left(x_b - x_v\right)}{x_b}
\end{equation}
and plot it in a histogram.
We observe that for 16 out of the 41 of the data sets, the relative
differences are below 10\%.  This suggests that a more complex method
that takes edges, nodes, and their connectivity---as  well as their
labels or attributes---into account does not necessarily yield
a classification accuracy that is higher by a large margin.
Moreover, this indicates that some data sets do \emph{not} require the
usage of a complex graph kernel, because they do not contain graphs
whose structure needs to be exploited in order to obtain good
classification results.

\subsubsection{Visual analysis}

\begin{figure}[p]
  \centering
    \iffinal
    \includegraphics[width=0.6\textwidth]{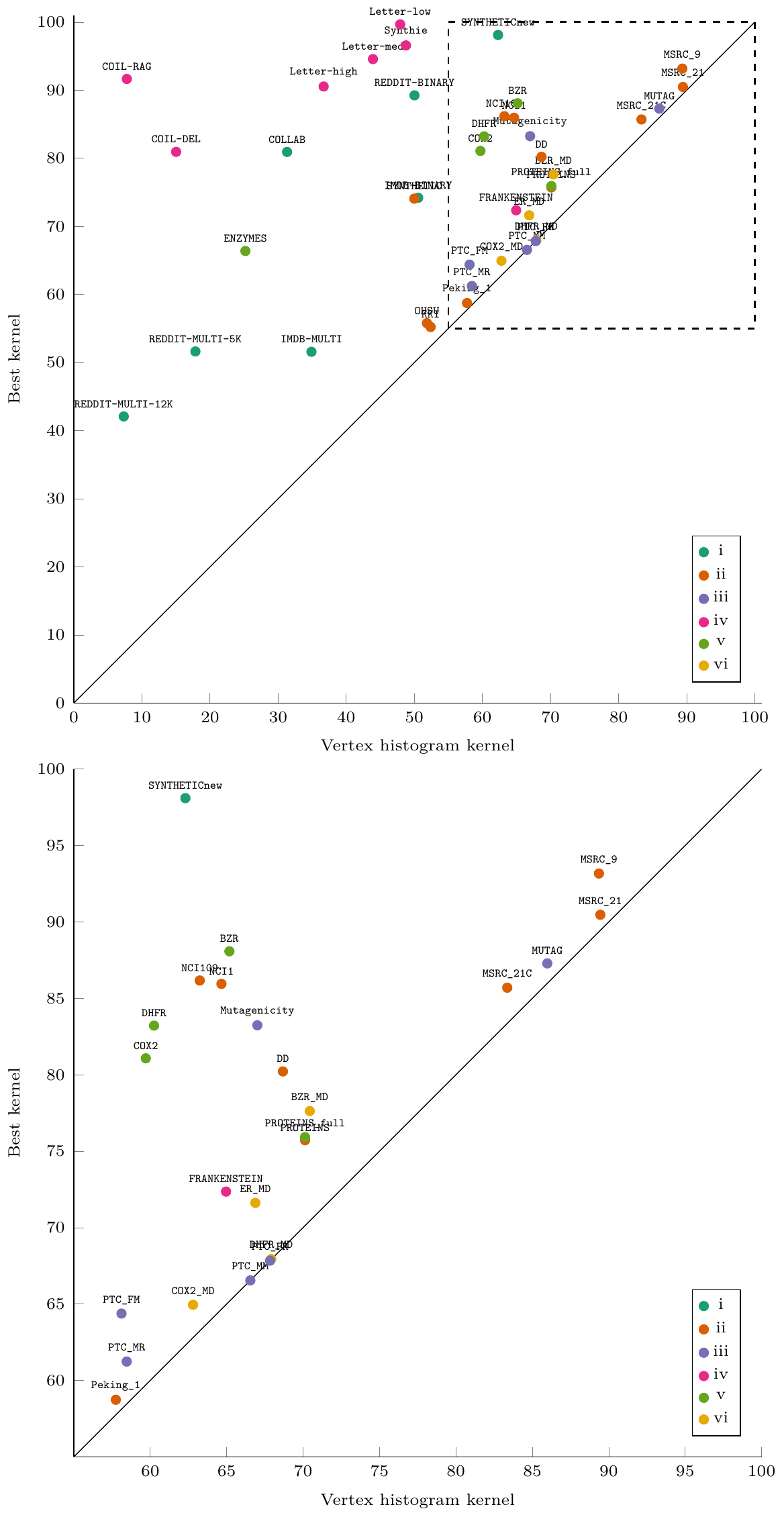}
    \else
  \tikzstyle{every node}=[font=\scriptsize]

  \begin{tikzpicture}

    \pgfplotsset{
      every node near coord/.append style = {%
        font = \tiny\ttfamily,
      },
    }

    \begin{groupplot}[
      group style= {%
        group size = 1 by 2,
      },
      axis x line*  = bottom,
      axis y line*  = left,
      enlargelimits = false,
      xlabel        = {Vertex histogram kernel},
      ylabel        = {Best kernel},
      xmin          =   0.0,
      xmax          = 101.0,
      ymin          =   0.0,
      ymax          = 101.0,
      %
      unit vector ratio* = 1 1 1,
      width         = 0.92\linewidth,
      %
      scatter/classes = {%
        1={Dark2-A},
        2={Dark2-B},
        3={Dark2-C},
        4={Dark2-D},
        5={Dark2-E},
        6={Dark2-F},
        7={Dark2-G}
      },
      legend pos = south east,
    ]
      \nextgroupplot

      \addlegendimage{Dark2-A, only marks, mark=*}
      \addlegendimage{Dark2-B, only marks, mark=*}
      \addlegendimage{Dark2-C, only marks, mark=*}
      \addlegendimage{Dark2-D, only marks, mark=*}
      \addlegendimage{Dark2-E, only marks, mark=*}
      \addlegendimage{Dark2-F, only marks, mark=*}

      \legend{i, ii, iii, iv, v, vi}

      \addplot[domain={0:100}] {x};

      \addplot[%
        nodes near coords,
        only marks,
        mark       = none,
        point meta = explicit symbolic
      ] table[%
        x index    = 5,
        y index    = 2,
        meta index = 6
      ] {Data/Vertex_histogram_kernel_accuracy_differences.txt};

      \addplot[%
         scatter, only marks,
         scatter src = explicit symbolic,
       ] table[%
         x index    = 5,
         y index    = 2,
         meta index = 7
       ] {Data/Vertex_histogram_kernel_accuracy_differences.txt};

      \draw [black, dashed, thick] (55,55) rectangle (100,100);

      \nextgroupplot[%
        xmin = 55, xmax = 100, ymin = 55, ymax = 100,
        ytick = {60, 65, 70, 75, 80, 85, 90, 95, 100},
        xtick = {60, 65, 70, 75, 80, 85, 90, 95, 100},
      ]

      \addlegendimage{Dark2-A, only marks, mark=*}
      \addlegendimage{Dark2-B, only marks, mark=*}
      \addlegendimage{Dark2-C, only marks, mark=*}
      \addlegendimage{Dark2-D, only marks, mark=*}
      \addlegendimage{Dark2-E, only marks, mark=*}
      \addlegendimage{Dark2-F, only marks, mark=*}

      \legend{i, ii, iii, iv, v, vi}

      \addplot[%
         nodes near coords,
         only marks,
         mark       = none,
         point meta = explicit symbolic
       ] table[%
         x index    = 5,
         y index    = 2,
         meta index = 6
       ] {Data/Vertex_histogram_kernel_accuracy_differences.txt};

      \addplot[%
         scatter, only marks,
         scatter src = explicit symbolic,
       ] table[%
         x index    = 5,
         y index    = 2,
         meta index = 7
       ] {Data/Vertex_histogram_kernel_accuracy_differences.txt};

       \addplot[domain={0:100}] {x};

    \end{groupplot}
  \end{tikzpicture}
  \fi
  \caption{%
    The performance of the vertex histogram kernel plotted against the
    respective \emph{best} kernel on every benchmark data set. We also
    provide a second plot that ``zooms'' into the marked region to
    decrease the clutter.
    Points have been coloured according to the data set type introduced
    in Section~\protect\ref{sec:Breakdown}.
    %
  }
  \label{fig:Vertex histogram kernel scatterplot}
\end{figure}
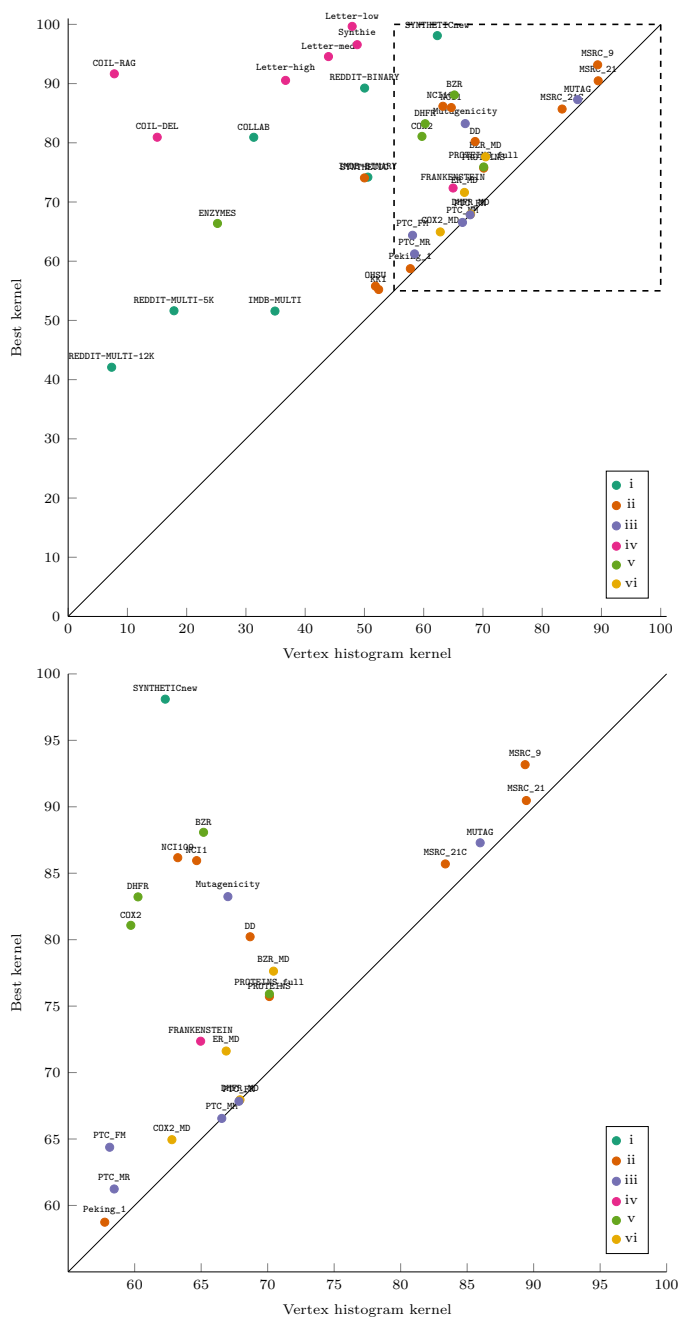

To analyse this situation from a more detailed perspective, we need
a more detailed visualisation that is capable of showing individual data
sets.
Hence, Figure~\ref{fig:Vertex histogram kernel scatterplot} depicts
these data from another perspective: it shows a scatterplot with the
vertex histogram kernel performance on the $x$-axis and the performance
of the respective \emph{best} graph kernel on the $y$-axis.
Furthermore, data sets have been colour-coded according to the types
introduced in Section~\ref{sec:Breakdown}.
In this plot, the distance to the diagonal can thus be seen as
indicating to what extent a data set is ``graphical,'' meaning that
there is a gap in classification performance between simple
histogram-based methods and more complex graph kernels.
Since accuracy was used to measure performance, the relative distance
between data sets has no meaning in this plot and should not be taken to
indicate similarities between data sets.
A second variant of this plot is shown in Figure~\ref{fig:Vertex
histogram kernel scatterplot AUROC}; since it uses AUROC as the main
comparison measure, relative differences can be compared more easily.
Even though the placement of several data sets is slightly different in
comparison to Figure~\ref{fig:Vertex histogram kernel scatterplot}, the
same observations as for Figure~\ref{fig:Vertex histogram kernel
scatterplot} apply.

\begin{figure}[p]
  \centering
    \iffinal
    \includegraphics[width=0.6\textwidth]{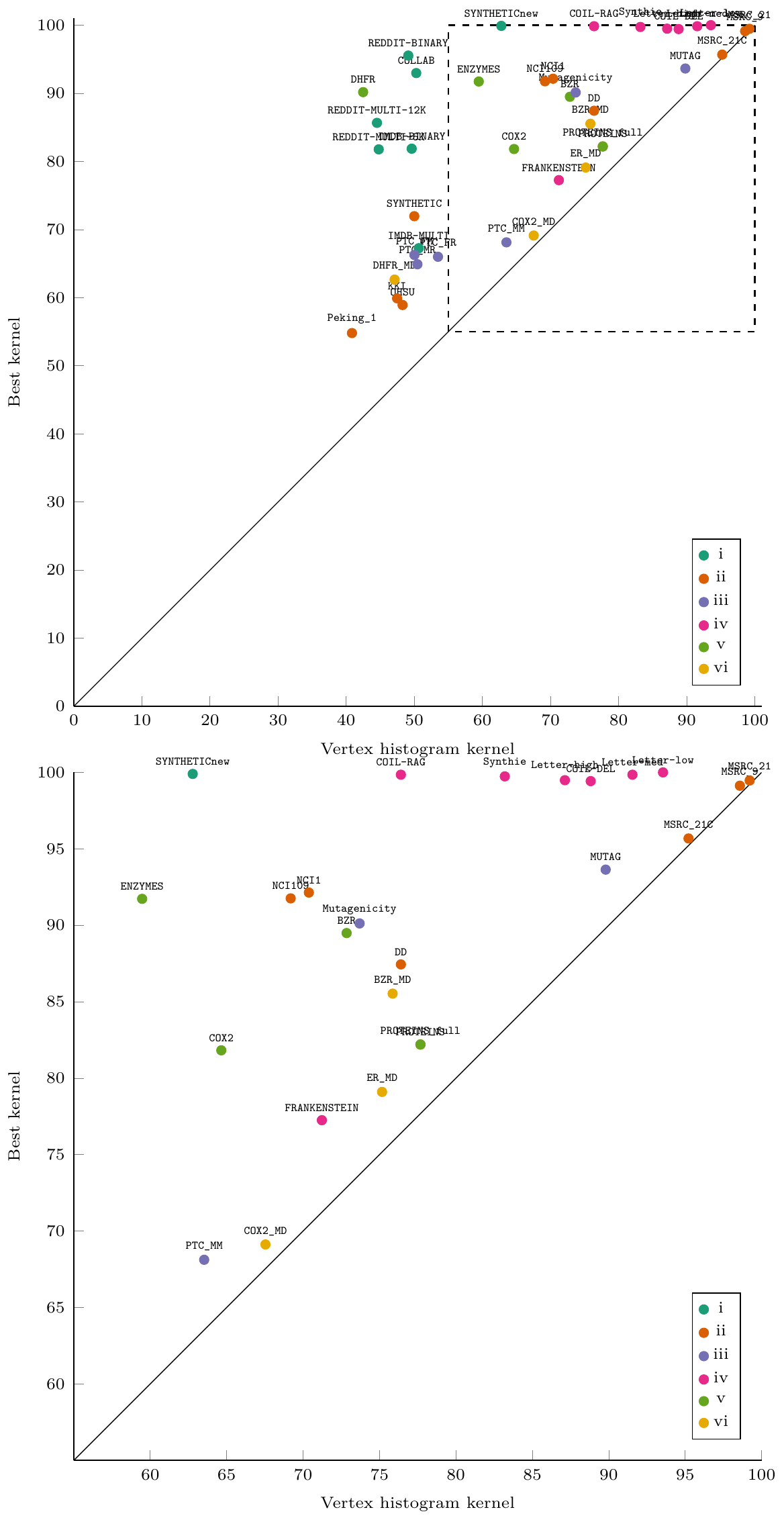}
    \else
  \tikzstyle{every node}=[font=\scriptsize]

  \begin{tikzpicture}

    \pgfplotsset{
      every node near coord/.append style = {%
        font = \tiny\ttfamily,
      },
    }

    \begin{groupplot}[
      group style= {%
        group size = 1 by 2,
      },
      axis x line*  = bottom,
      axis y line*  = left,
      enlargelimits = false,
      xlabel        = {Vertex histogram kernel},
      ylabel        = {Best kernel},
      xmin          =   0.0,
      xmax          = 101.0,
      ymin          =   0.0,
      ymax          = 101.0,
      %
      unit vector ratio* = 1 1 1,
      width         = 0.92\linewidth,
      %
      scatter/classes = {%
        1={Dark2-A},
        2={Dark2-B},
        3={Dark2-C},
        4={Dark2-D},
        5={Dark2-E},
        6={Dark2-F},
        7={Dark2-G}
      },
      legend pos = south east,
    ]
      \nextgroupplot

      \addlegendimage{Dark2-A, only marks, mark=*}
      \addlegendimage{Dark2-B, only marks, mark=*}
      \addlegendimage{Dark2-C, only marks, mark=*}
      \addlegendimage{Dark2-D, only marks, mark=*}
      \addlegendimage{Dark2-E, only marks, mark=*}
      \addlegendimage{Dark2-F, only marks, mark=*}

      \legend{i, ii, iii, iv, v, vi}

      \addplot[domain={0:100}] {x};

      \addplot[%
        nodes near coords,
        only marks,
        mark       = none,
        point meta = explicit symbolic
      ] table[%
        x index    = 5,
        y index    = 2,
        meta index = 6
      ] {Data/Vertex_histogram_kernel_auroc_differences.txt};

      \addplot[%
         scatter, only marks,
         scatter src = explicit symbolic,
       ] table[%
         x index    = 5,
         y index    = 2,
         meta index = 7
       ] {Data/Vertex_histogram_kernel_auroc_differences.txt};

      \draw [black, dashed, thick] (55,55) rectangle (100,100);

      \nextgroupplot[%
        xmin = 55, xmax = 100, ymin = 55, ymax = 100,
        ytick = {60, 65, 70, 75, 80, 85, 90, 95, 100},
        xtick = {60, 65, 70, 75, 80, 85, 90, 95, 100},
      ]

      \addlegendimage{Dark2-A, only marks, mark=*}
      \addlegendimage{Dark2-B, only marks, mark=*}
      \addlegendimage{Dark2-C, only marks, mark=*}
      \addlegendimage{Dark2-D, only marks, mark=*}
      \addlegendimage{Dark2-E, only marks, mark=*}
      \addlegendimage{Dark2-F, only marks, mark=*}

      \legend{i, ii, iii, iv, v, vi}

      \addplot[%
         nodes near coords,
         only marks,
         mark       = none,
         point meta = explicit symbolic
       ] table[%
         x index    = 5,
         y index    = 2,
         meta index = 6
       ] {Data/Vertex_histogram_kernel_auroc_differences.txt};

      \addplot[%
         scatter, only marks,
         scatter src = explicit symbolic,
       ] table[%
         x index    = 5,
         y index    = 2,
         meta index = 7
       ] {Data/Vertex_histogram_kernel_auroc_differences.txt};

       \addplot[domain={0:100}] {x};

    \end{groupplot}
  \end{tikzpicture}
  \fi
  \caption{%
    Similar to Figure~\protect\ref{fig:Vertex histogram kernel scatterplot},
    we depict the performance of the vertex histogram kernel compared to
    the respective \emph{best} kernel on every benchmark data set.
    Here, classification performance is measured in terms of AUROC.
  }
  \label{fig:Vertex histogram kernel scatterplot AUROC}
\end{figure}
The scatterplot visualisation gives rise to several interesting
observations. First of all, we see that several data sets are situated
well above the diagonal. This includes data sets of class~i and class~iv
that miss node labels---in which case the histogram kernel boils down
to comparing node degrees---but also more ``rich''~(in terms of
features) data sets such as \texttt{ENYZYMES} or the
\texttt{Letter-$\ast$} data sets.
The other classes are closer to the diagonal, though. A ``zoomed-in''
version of the plot shows a portion of them in greater detail. In
general, the vertex histogram kernel provides a useful baseline for
them, and provides a surprisingly competitive performance for several
data sets. 
%
This makes it clear that the performance of the vertex histogram kernel
is caused by selecting data sets that do not feature any ``graphical''
structure; we will get back to this point in a subsequent section, when
we discuss the suitability of some benchmark data sets.

\subsubsection{Histogram kernels compared to other kernels}

As a last item in our analysis of histogram kernels, we observe to what
extent it is possible to distinguish histogram-based kernels from other
graph kernels.
%
\begin{figure}[tbp]
    \iffinal
    \includegraphics[width=0.95\textwidth]{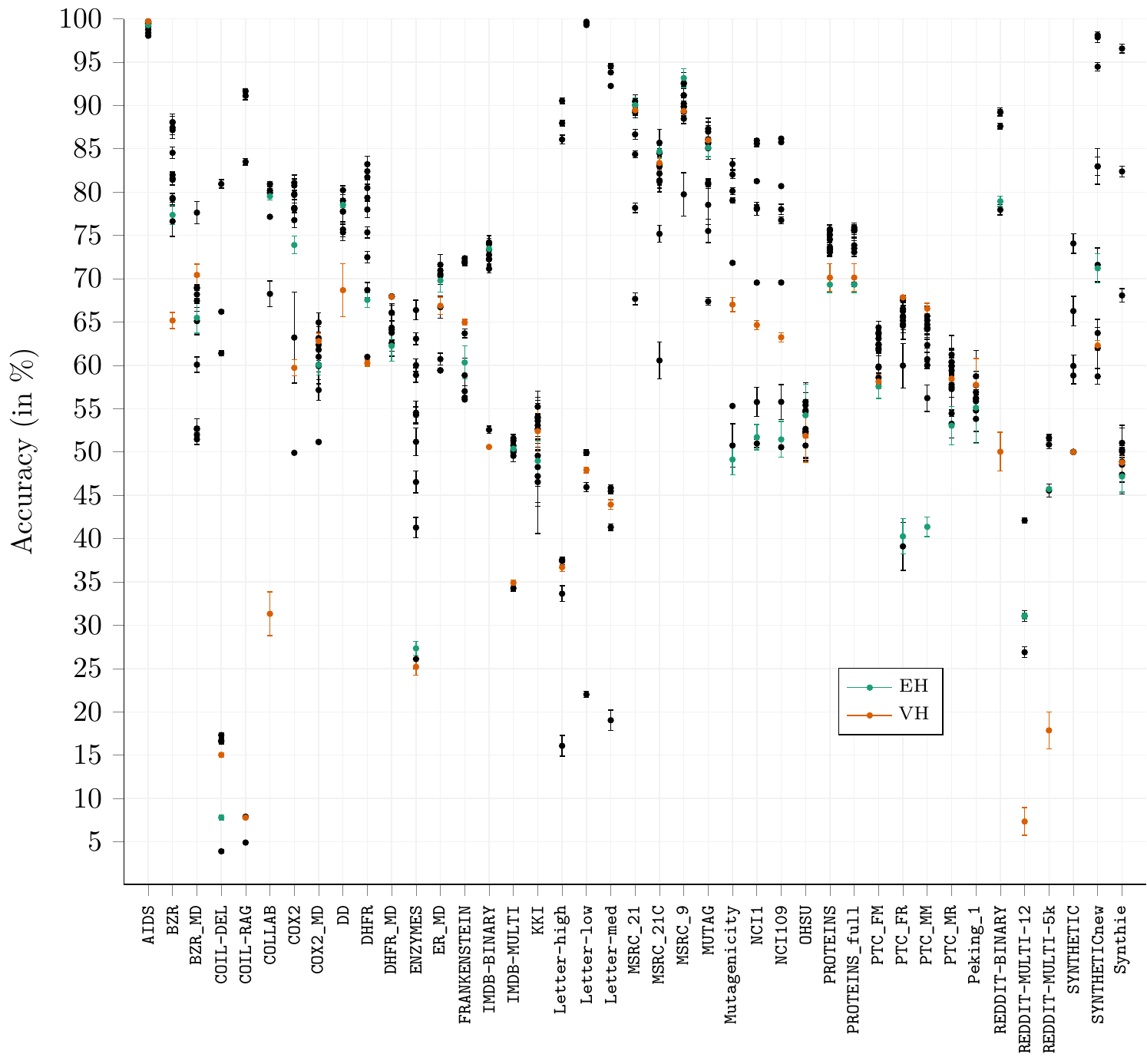}
    \else
  \centering
  \begin{tikzpicture}
    \pgfplotsset{%
      major grid style = {%
        gray!10,
      },
      error bars/.cd,
        y dir = both,
        y explicit,
        %
        %
    }
    \begin{axis}[%
      axis x line*                    = bottom,
      axis y line*                    = left,
      xmajorgrids                     = true,
      ymajorgrids                     = true,
      enlarge x limits                = false,
      enlarge y limits                = false,
      width                           = \textwidth,
      xtick = {%
        0,
        1,
        2,
        3,
        4,
        5,
        6,
        7,
        8,
        9,
        10,
        11,
        12,
        13,
        14,
        15,
        16,
        17,
        18,
        19,
        20,
        21,
        22,
        23,
        24,
        25,
        26,
        27,
        28,
        29,
        30,
        31,
        32,
        33,
        34,
        35,
        36,
        37,
        38,
        39,
        40
      },
      xticklabels = {%
          AIDS,
          BZR,
          BZR\_MD,
          COIL-DEL,
          COIL-RAG,
          COLLAB,
          COX2,
          COX2\_MD,
          DD,
          DHFR,
          DHFR\_MD,
          ENZYMES,
          ER\_MD,
          FRANKENSTEIN,
          IMDB-BINARY,
          IMDB-MULTI,
          KKI,
          Letter-high,
          Letter-low,
          Letter-med,
          MSRC\_21,
          MSRC\_21C,
          MSRC\_9,
          MUTAG,
          Mutagenicity,
          NCI1,
          NCI109,
          OHSU,
          PROTEINS,
          PROTEINS\_full,
          PTC\_FM,
          PTC\_FR,
          PTC\_MM,
          PTC\_MR,
          Peking\_1,
          REDDIT-BINARY,
          REDDIT-MULTI-12,
          REDDIT-MULTI-5k,
          SYNTHETIC,
          SYNTHETICnew,
          Synthie
      },
      tick align = outside,
      xmin  = -1,
      xmax  = 41,
      ymin  = 0.1,
      ymax  = 100,
      ytick = {%
          0,
          5,
         10,
         15,
         20,
         25,
         30,
         35,
         40,
         45,
         50,
         55,
         60,
         65,
         70,
         75,
         80,
         85,
         90,
         95,
        100
      },
      ylabel = {Accuracy~(in \si{\percent})},
      xticklabel style = {
        font   = \scriptsize\ttfamily,
        rotate = 90,
      },
      width = \textwidth,
      legend style = {%
        at     = {(0.75, 0.25)},
        anchor = north,
        font   = \scriptsize,
      },
      %
      legend image post style = {
        mark      = *,
        mark size = 1.0pt,
      },
    ]
      \addlegendimage{Dark2-A}
      \addlegendimage{Dark2-B}

      \legend{EH, VH}

      \addplot[mark=*, mark size=1.0pt, only marks, black] table[y error index=2] {Data/accuracies/GH_accuracies.csv};
      \addplot[mark=*, mark size=1.0pt, only marks, black] table[y error index=2] {Data/accuracies/GL_accuracies.csv};
      \addplot[mark=*, mark size=1.0pt, only marks, black] table[y error index=2] {Data/accuracies/HGKSP_seed0_accuracies.csv};
      \addplot[mark=*, mark size=1.0pt, only marks, black] table[y error index=2] {Data/accuracies/HGKWL_seed0_accuracies.csv};
      \addplot[mark=*, mark size=1.0pt, only marks, black] table[y error index=2] {Data/accuracies/MLG_accuracies.csv};
      \addplot[mark=*, mark size=1.0pt, only marks, black] table[y error index=2] {Data/accuracies/MP_accuracies.csv};
      \addplot[mark=*, mark size=1.0pt, only marks, black] table[y error index=2] {Data/accuracies/SP_gkl_accuracies.csv};
      \addplot[mark=*, mark size=1.0pt, only marks, black] table[y error index=2] {Data/accuracies/WL_accuracies.csv};
      \addplot[mark=*, mark size=1.0pt, only marks, black] table[y error index=2] {Data/accuracies/WLOA_accuracies.csv};
      \addplot[mark=*, mark size=1.0pt, only marks, black] table[y error index=2] {Data/accuracies/RW_gkl_accuracies.csv};
      \addplot[mark=*, mark size=1.0pt, only marks, black] table[y error index=2] {Data/accuracies/CSM_gkl_accuracies.csv};
      \addplot[mark=*, mark size=1.0pt, only marks, Dark2-A] table[y error index=2] {Data/accuracies/EH_gkl_accuracies.csv};
      \addplot[mark=*, mark size=1.0pt, only marks, Dark2-B] table[y error index=2] {Data/accuracies/VH_accuracies.csv};

    \end{axis}
  \end{tikzpicture}
  \fi
  \caption{%
    A visualisation of \emph{all} accuracies~(including the standard
    deviation) on all benchmark data sets. Each individual kernel is
    represented by a ``dot'' that includes error bars.
    The histogram kernels are highlighted in this visualisation; they
    can be clearly distinguished from the remaining data sets through
    their performance.
  }
  \label{fig:Histogram kernels vs. rest}
\end{figure}
%
This is demonstrated in Figure~\ref{fig:Histogram kernels vs. rest},
which provides a dense visualisation of the accuracies~(including
standard deviations) of all data sets, while highlighting the histogram
kernels. For most of the data sets, there is a clear gap between the
performance of histogram-based kernels and other kernels, with the other
kernels typically outperforming the histogram-based ones. 

To summarise our analysis so far: we have seen that the performance of
graph kernels hinges largely on the type of data set. For some of the
data sets, simple histogram-based kernels, which are incapable of
exploiting any structures of a graph beyond vertex/edge  labels, are
sufficient to obtain suitable---and in some cases even
competitive---performance values. Considering that these histogram-based
graph kernels represent a \emph{baseline} and not a regular choice of
algorithm to be used in practice, the preceding discussion raises the
question of the \emph{difficulty} of the benchmark data sets. We will
discuss this in the next section.

\section{Analysing the difficulty of data sets}\label{sec:Difficulty}

To further examine the behaviour of \emph{all} kernels---not only the
histogram-based ones---we now turn to analysing the \emph{difficulty} of
available graph benchmark data sets.
We will present multiple ways of depicting the difficulty, starting with
a high-level discussion of accuracy distributions, which is followed by
increasingly detailed discussions on optimal classification accuracy
estimates.

\subsection{Accuracies and standard deviations}

We begin our analysis with a visualisation in the style of
Figure~\ref{fig:Histogram kernels vs. rest}. However, instead of
highlighting histogram-based kernels, we show the accuracies and
standard deviations of all kernels without labels. This will make it
easier to see to what extent there are different groups of kernels for
specific data sets. Moreover, including the standard deviation also
provides us with information about the performance across different
\re{iterations of the cross-validation training procedure}.

\begin{figure}[p]
  \centering
    \iffinal
    \includegraphics[width=0.95\textwidth]{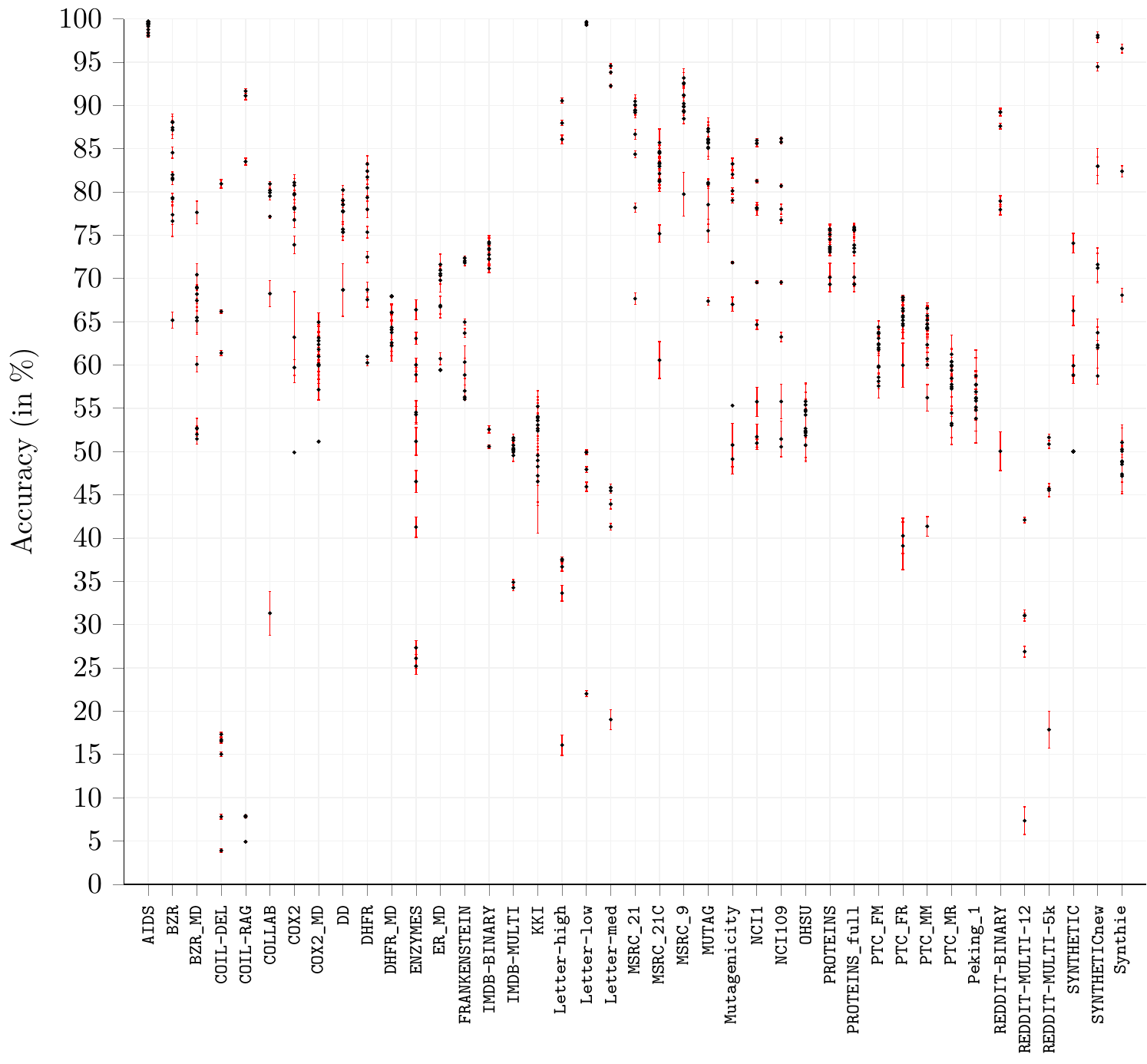}
    \else
  \begin{tikzpicture}
    \pgfplotsset{%
      major grid style = {%
        gray!10,
      }
    }
    \begin{axis}[
      axis x line*                    = bottom,
      axis y line*                    = left,
      xmajorgrids                     = true,
      ymajorgrids                     = true,
      enlarge x limits                = false,
      enlarge y limits                = false,
      width                           = \textwidth,
      xtick = {%
        1, 2, 3, 4, 5, 6, 7, 8, 9, 10, 11, 12, 13, 14, 15, 16, 17, 18, 19,
        20, 21, 22, 23, 24, 25, 26, 27, 28, 29, 30, 31, 32, 33, 34, 35, 36,
        37, 38, 39, 40, 41
      },
      tick align = outside,
      xmin  = 0,
      xmax  = 42,
      ymin  = 0,
      ymax  = 100,
      ytick = {%
          0, 5, 10, 15, 20, 25, 30, 35, 40, 45, 50, 55, 60, 65, 70, 75,
         80, 85, 90, 95, 100
      },
      ylabel = {Accuracy~(in \si{\percent})},
      xticklabels = {%
        AIDS,
        BZR,
        BZR\_MD,
        COIL-DEL,
        COIL-RAG,
        COLLAB,
        COX2,
        COX2\_MD,
        DD,
        DHFR,
        DHFR\_MD,
        ENZYMES,
        ER\_MD,
        FRANKENSTEIN,
        IMDB-BINARY,
        IMDB-MULTI,
        KKI,
        Letter-high,
        Letter-low,
        Letter-med,
        MSRC\_21,
        MSRC\_21C,
        MSRC\_9,
        MUTAG,
        Mutagenicity,
        NCI1,
        NCI109,
        OHSU,
        PROTEINS,
        PROTEINS\_full,
        PTC\_FM,
        PTC\_FR,
        PTC\_MM,
        PTC\_MR,
        Peking\_1,
        REDDIT-BINARY,
        REDDIT-MULTI-12,
        REDDIT-MULTI-5k,
        SYNTHETIC,
        SYNTHETICnew,
        Synthie
      },
      xticklabel style = {
        font   = \scriptsize\ttfamily,
        rotate = 90,
      },
    ]

      \pgfplotsset{%
        error bars/.cd,
          y dir = both,
          y explicit,
          error bar style = {%
            red
          },
      }

      \addplot[mark=*, only marks, mark size = 0.5pt, x filter/.expression = { 1}] table[y index=0, y error index=1] {Data/sdev/AIDS.txt};
      \addplot[mark=*, only marks, mark size = 0.5pt, x filter/.expression = { 2}] table[y index=0, y error index=1] {Data/sdev/BZR.txt};
      \addplot[mark=*, only marks, mark size = 0.5pt, x filter/.expression = { 3}] table[y index=0, y error index=1] {Data/sdev/BZR_MD.txt};
      \addplot[mark=*, only marks, mark size = 0.5pt, x filter/.expression = { 4}] table[y index=0, y error index=1] {Data/sdev/COIL-DEL.txt};
      \addplot[mark=*, only marks, mark size = 0.5pt, x filter/.expression = { 5}] table[y index=0, y error index=1] {Data/sdev/COIL-RAG.txt};
      \addplot[mark=*, only marks, mark size = 0.5pt, x filter/.expression = { 6}] table[y index=0, y error index=1] {Data/sdev/COLLAB.txt};
      \addplot[mark=*, only marks, mark size = 0.5pt, x filter/.expression = { 7}] table[y index=0, y error index=1] {Data/sdev/COX2.txt};
      \addplot[mark=*, only marks, mark size = 0.5pt, x filter/.expression = { 8}] table[y index=0, y error index=1] {Data/sdev/COX2_MD.txt};
      \addplot[mark=*, only marks, mark size = 0.5pt, x filter/.expression = { 9}] table[y index=0, y error index=1] {Data/sdev/DD.txt};
      \addplot[mark=*, only marks, mark size = 0.5pt, x filter/.expression = { 9}] table[y index=0, y error index=1] {Data/sdev/DD.txt};
      \addplot[mark=*, only marks, mark size = 0.5pt, x filter/.expression = {10}] table[y index=0, y error index=1] {Data/sdev/DHFR.txt};
      \addplot[mark=*, only marks, mark size = 0.5pt, x filter/.expression = {11}] table[y index=0, y error index=1] {Data/sdev/DHFR_MD.txt};
      \addplot[mark=*, only marks, mark size = 0.5pt, x filter/.expression = {12}] table[y index=0, y error index=1] {Data/sdev/ENZYMES.txt};
      \addplot[mark=*, only marks, mark size = 0.5pt, x filter/.expression = {13}] table[y index=0, y error index=1] {Data/sdev/ER_MD.txt};
      \addplot[mark=*, only marks, mark size = 0.5pt, x filter/.expression = {14}] table[y index=0, y error index=1] {Data/sdev/FRANKENSTEIN.txt};
      \addplot[mark=*, only marks, mark size = 0.5pt, x filter/.expression = {15}] table[y index=0, y error index=1] {Data/sdev/IMDB-BINARY.txt};
      \addplot[mark=*, only marks, mark size = 0.5pt, x filter/.expression = {16}] table[y index=0, y error index=1] {Data/sdev/IMDB-MULTI.txt};
      \addplot[mark=*, only marks, mark size = 0.5pt, x filter/.expression = {17}] table[y index=0, y error index=1] {Data/sdev/KKI.txt};
      \addplot[mark=*, only marks, mark size = 0.5pt, x filter/.expression = {18}] table[y index=0, y error index=1] {Data/sdev/Letter-high.txt};
      \addplot[mark=*, only marks, mark size = 0.5pt, x filter/.expression = {19}] table[y index=0, y error index=1] {Data/sdev/Letter-low.txt};
      \addplot[mark=*, only marks, mark size = 0.5pt, x filter/.expression = {20}] table[y index=0, y error index=1] {Data/sdev/Letter-med.txt};
      \addplot[mark=*, only marks, mark size = 0.5pt, x filter/.expression = {21}] table[y index=0, y error index=1] {Data/sdev/MSRC_21.txt};
      \addplot[mark=*, only marks, mark size = 0.5pt, x filter/.expression = {22}] table[y index=0, y error index=1] {Data/sdev/MSRC_21C.txt};
      \addplot[mark=*, only marks, mark size = 0.5pt, x filter/.expression = {23}] table[y index=0, y error index=1] {Data/sdev/MSRC_9.txt};
      \addplot[mark=*, only marks, mark size = 0.5pt, x filter/.expression = {24}] table[y index=0, y error index=1] {Data/sdev/MUTAG.txt};
      \addplot[mark=*, only marks, mark size = 0.5pt, x filter/.expression = {25}] table[y index=0, y error index=1] {Data/sdev/Mutagenicity.txt};
      \addplot[mark=*, only marks, mark size = 0.5pt, x filter/.expression = {26}] table[y index=0, y error index=1] {Data/sdev/NCI1.txt};
      \addplot[mark=*, only marks, mark size = 0.5pt, x filter/.expression = {27}] table[y index=0, y error index=1] {Data/sdev/NCI109.txt};
      \addplot[mark=*, only marks, mark size = 0.5pt, x filter/.expression = {28}] table[y index=0, y error index=1] {Data/sdev/OHSU.txt};
      \addplot[mark=*, only marks, mark size = 0.5pt, x filter/.expression = {29}] table[y index=0, y error index=1] {Data/sdev/PROTEINS.txt};
      \addplot[mark=*, only marks, mark size = 0.5pt, x filter/.expression = {30}] table[y index=0, y error index=1] {Data/sdev/PROTEINS_full.txt};
      \addplot[mark=*, only marks, mark size = 0.5pt, x filter/.expression = {31}] table[y index=0, y error index=1] {Data/sdev/PTC_FM.txt};
      \addplot[mark=*, only marks, mark size = 0.5pt, x filter/.expression = {32}] table[y index=0, y error index=1] {Data/sdev/PTC_FR.txt};
      \addplot[mark=*, only marks, mark size = 0.5pt, x filter/.expression = {33}] table[y index=0, y error index=1] {Data/sdev/PTC_MM.txt};
      \addplot[mark=*, only marks, mark size = 0.5pt, x filter/.expression = {34}] table[y index=0, y error index=1] {Data/sdev/PTC_MR.txt};
      \addplot[mark=*, only marks, mark size = 0.5pt, x filter/.expression = {35}] table[y index=0, y error index=1] {Data/sdev/Peking_1.txt};
      \addplot[mark=*, only marks, mark size = 0.5pt, x filter/.expression = {36}] table[y index=0, y error index=1] {Data/sdev/REDDIT-BINARY.txt};
      \addplot[mark=*, only marks, mark size = 0.5pt, x filter/.expression = {37}] table[y index=0, y error index=1] {Data/sdev/REDDIT-MULTI-12K.txt};
      \addplot[mark=*, only marks, mark size = 0.5pt, x filter/.expression = {38}] table[y index=0, y error index=1] {Data/sdev/REDDIT-MULTI-5K.txt};
      \addplot[mark=*, only marks, mark size = 0.5pt, x filter/.expression = {39}] table[y index=0, y error index=1] {Data/sdev/SYNTHETIC.txt};
      \addplot[mark=*, only marks, mark size = 0.5pt, x filter/.expression = {40}] table[y index=0, y error index=1] {Data/sdev/SYNTHETICnew.txt};
      \addplot[mark=*, only marks, mark size = 0.5pt, x filter/.expression = {41}] table[y index=0, y error index=1] {Data/sdev/Synthie.txt};
    \end{axis}
  \end{tikzpicture}
  \fi
  \caption{%
    \re{Mean} accuracy values along with their standard deviations~(plotted as
    error bars) \re{over the different iterations of cross-validation} for all graphs, separated by data set.
    A gap indicates that performance values do not overlap for different
    repetitions of the training.
  }
  \label{fig:Accuracies and sdev}
\end{figure}
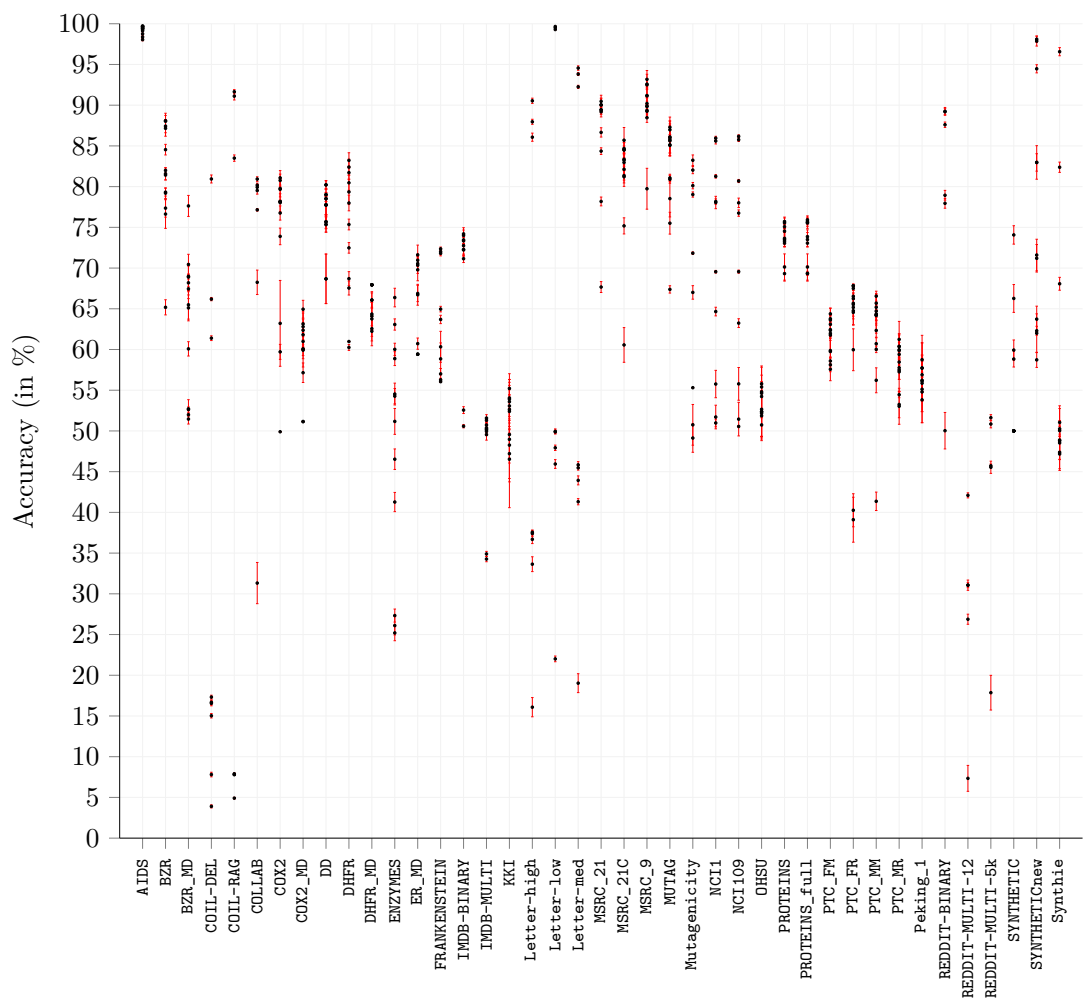

Figure~\ref{fig:Accuracies and sdev} depicts the resulting
visualisation. Each point corresponds to the mean accuracy
of a specific graph kernel, whereas each bar indicates its
standard deviation across the different \re{repetitions} of the cross-validation process. Intuitively, this bar can also be seen
as an \emph{uncertainty} of the ``true'' performance on an
unseen part of a specific data set.
This data-rich visualisation depicts certain idiosyncrasies of the data
sets in an intuitive manner: for each data set, the presence of a \emph{single}
region, \ie\ a region of---either pairwise or mutually---overlapping error bars,
indicates that the performance of a specific graph kernel varies too much between
folds and thus cannot be easily distinguished from the remaining kernels.
By contrast, a \emph{gap} indicates that there is a subset of kernels
that exhibits a markedly different performance over all \re{iterations}.

\paragraph{Visual analysis of overlaps}
%
Out of all the data sets described here, \texttt{REDDIT-MULTI-12K} is the only one
for which we observe \emph{no} overlaps at all---different graph kernels are
thus perfectly separable from each other. Similarly, for the data sets
\begin{inparaenum}[(i)]
  \item \texttt{REDDIT-BINARY}, and
  \item \texttt{COIL-DEL},
\end{inparaenum}
only a single overlap occurs. For all of these data sets, different
graph kernels can be easily separated from each other in terms of their
performance. By contrast, \texttt{Peking\_1} exhibits the largest number
of pairwise overlaps; here, almost \emph{all} standard deviation
intervals exhibit a mutual overlap.
We will subsequently quantify these observations.

\begin{figure}
  \centering
    \iffinal
      \includegraphics{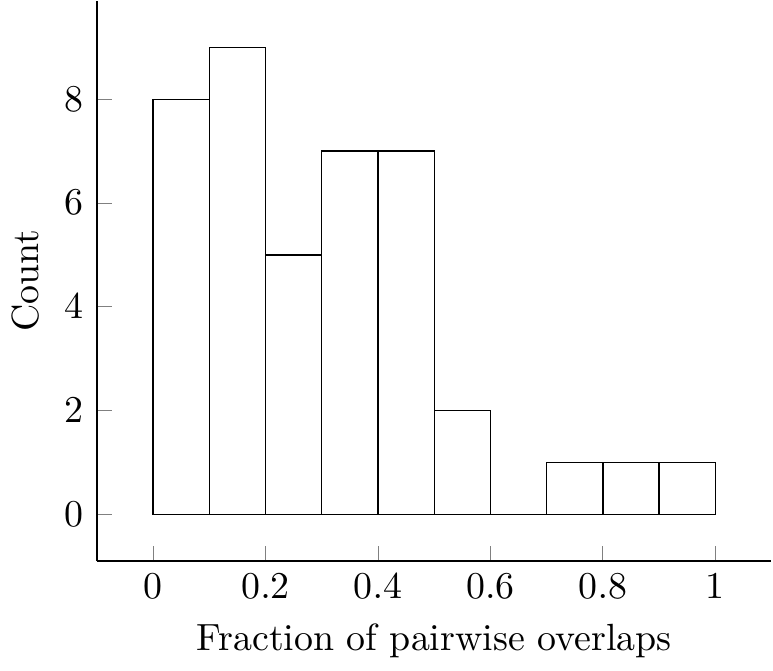}
    \else
  \begin{tikzpicture}
    \begin{axis}[%
      axis x line*  = bottom,
      axis y line*  = left,
      enlargelimits = true,
      ylabel        = {Count},
      xlabel        = {Fraction of pairwise overlaps},
      ymin          = 0,
      xmin          = 0,
      xmax          = 1,
    ]
      \addplot[
        hist = {%
          bins = 10,
          data = {y / x},
        },
      ] table[
        col sep = comma,
        y       = n_overlaps,
        x       = n_pairs,
      ] {Data/Overlaps.csv};
    \end{axis}
  \end{tikzpicture}
  \fi
  \caption{%
    A histogram of the fraction of pairwise overlaps between the
    mean accuracies and standard deviations shown in Figure~\protect\ref{fig:Accuracies and sdev}.
    A small fraction of pairwise overlaps is desirable because it
    simplifies comparing the performance of different graph kernels.
  }
  \label{fig:Overlaps}
\end{figure}
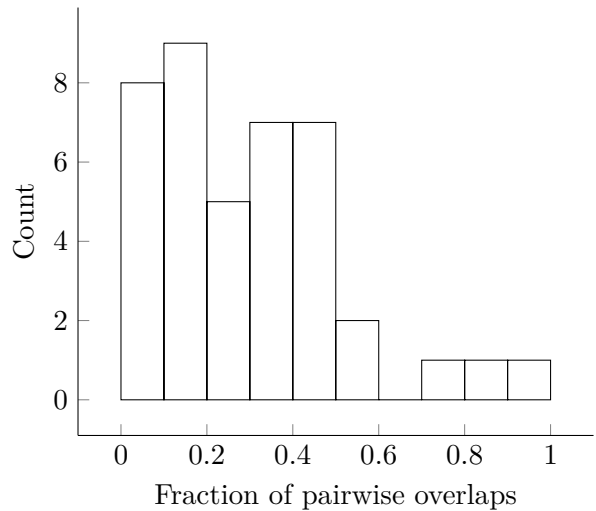

\paragraph{Histogram analysis of overlaps}
%
As a histogram of all pairwise overlaps in Figure~\ref{fig:Overlaps}
shows, most of the data sets exhibit a fraction of \SIrange{0}{50}{\percent} of
pairwise overlaps. This is not necessarily problematic, but as we will
see in Section~\ref{sec:Critical difference analysis} on
p.~\pageref{sec:Critical difference analysis}, it will slightly decrease
statistical power if the \emph{full} data set is selected to
make claims on the statistical superiority of specific graph kernels.

\paragraph{Discussion}
In general, care must be taken when considering these analyses; a low
number of overlaps can have multiple causes, among them being
\begin{inparaenum}[(i)]
  \item the data set could be too easy, making it impossible to
    distinguish between different graph kernels in terms of their
    predictive performance, or
  \item the data set could be too small, making predictive performance
    highly vary depending on the training procedure and the fold
    assignment, or
  \item the data set could be too hard to classify in general, leading
    to a high variance of predictive values.
\end{inparaenum}
Hence, we will subsequently discuss the difficulty and general
suitability of the benchmark data sets under different perspectives,
before giving our recommendations.

\begin{figure}[p]
  \centering
    \iffinal
    \includegraphics[width=0.95\textwidth]{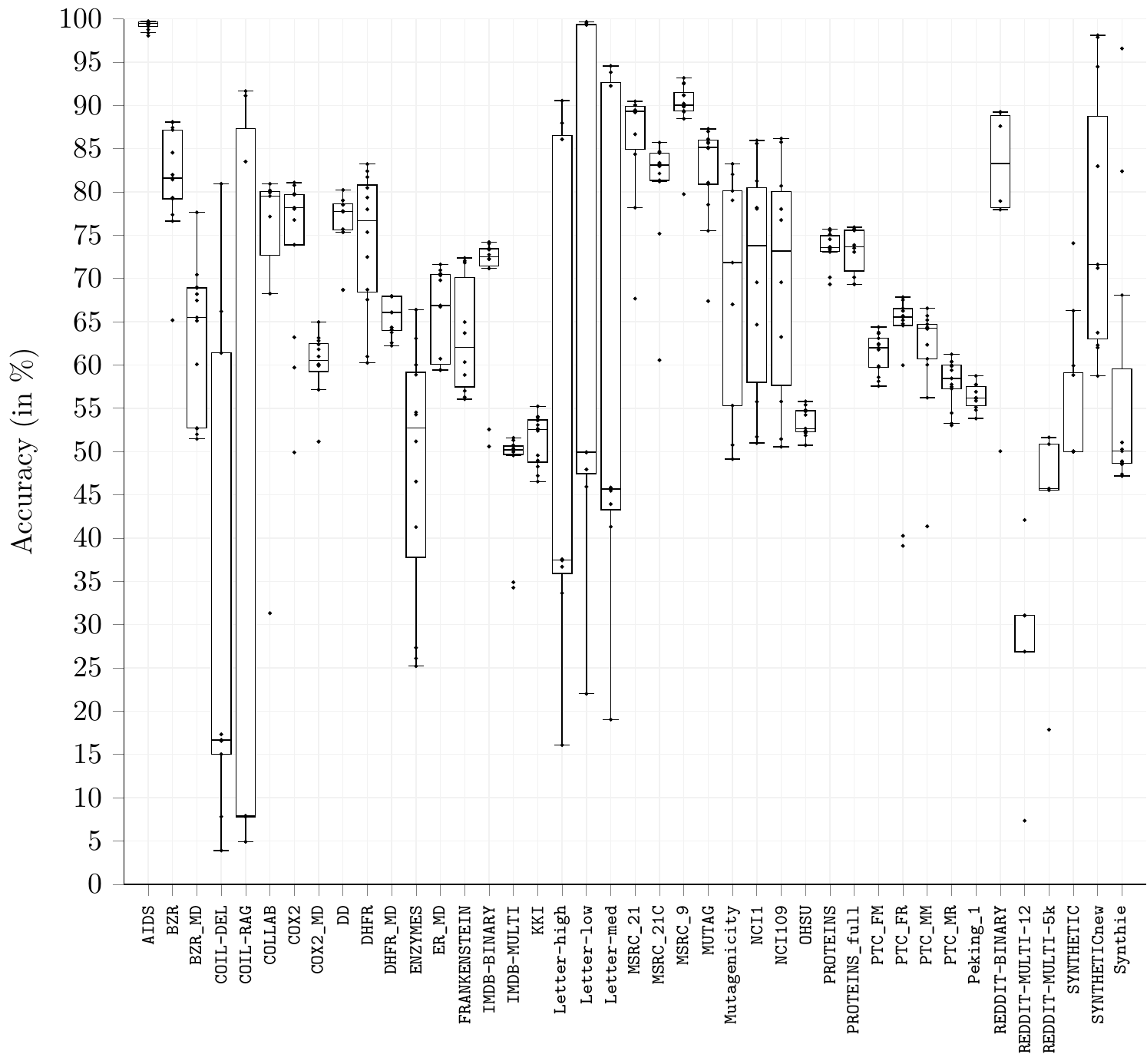}
    \else
  \begin{tikzpicture}
    \pgfplotsset{%
      major grid style = {%
        gray!10,
      }
    }
    \begin{axis}[
      axis x line*                    = bottom,
      axis y line*                    = left,
      xmajorgrids                     = true,
      ymajorgrids                     = true,
      enlarge x limits                = false,
      enlarge y limits                = false,
      width                           = \textwidth,
      boxplot/draw direction          = y,
      boxplot/every box/.append style = {fill=white},
      xtick = {%
        1, 2, 3, 4, 5, 6, 7, 8, 9, 10, 11, 12, 13, 14, 15, 16, 17, 18, 19,
        20, 21, 22, 23, 24, 25, 26, 27, 28, 29, 30, 31, 32, 33, 34, 35, 36,
        37, 38, 39, 40, 41
      },
      tick align = outside,
      xmin  = 0,
      xmax  = 42,
      ymin  = 0,
      ymax  = 100,
      ytick = {%
          0, 5, 10, 15, 20, 25, 30, 35, 40, 45, 50, 55, 60, 65, 70, 75,
         80, 85, 90, 95, 100
      },
      ylabel = {Accuracy~(in \si{\percent})},
      xticklabels = {%
        AIDS,
        BZR,
        BZR\_MD,
        COIL-DEL,
        COIL-RAG,
        COLLAB,
        COX2,
        COX2\_MD,
        DD,
        DHFR,
        DHFR\_MD,
        ENZYMES,
        ER\_MD,
        FRANKENSTEIN,
        IMDB-BINARY,
        IMDB-MULTI,
        KKI,
        Letter-high,
        Letter-low,
        Letter-med,
        MSRC\_21,
        MSRC\_21C,
        MSRC\_9,
        MUTAG,
        Mutagenicity,
        NCI1,
        NCI109,
        OHSU,
        PROTEINS,
        PROTEINS\_full,
        PTC\_FM,
        PTC\_FR,
        PTC\_MM,
        PTC\_MR,
        Peking\_1,
        REDDIT-BINARY,
        REDDIT-MULTI-12,
        REDDIT-MULTI-5k,
        SYNTHETIC,
        SYNTHETICnew,
        Synthie
      },
      xticklabel style = {
        font   = \scriptsize\ttfamily,
        rotate = 90,
      }
    ]
      \addplot[boxplot, mark=none] table[y index=0] {Data/Boxplots/AIDS.txt};
      \addplot[boxplot, mark=none] table[y index=0] {Data/Boxplots/BZR.txt};
      \addplot[boxplot, mark=none] table[y index=0] {Data/Boxplots/BZR_MD.txt};
      \addplot[boxplot, mark=none] table[y index=0] {Data/Boxplots/COIL-DEL.txt};
      \addplot[boxplot, mark=none] table[y index=0] {Data/Boxplots/COIL-RAG.txt};
      \addplot[boxplot, mark=none] table[y index=0] {Data/Boxplots/COLLAB.txt};
      \addplot[boxplot, mark=none] table[y index=0] {Data/Boxplots/COX2.txt};
      \addplot[boxplot, mark=none] table[y index=0] {Data/Boxplots/COX2_MD.txt};
      \addplot[boxplot, mark=none] table[y index=0] {Data/Boxplots/DD.txt};
      \addplot[boxplot, mark=none] table[y index=0] {Data/Boxplots/DHFR.txt};
      \addplot[boxplot, mark=none] table[y index=0] {Data/Boxplots/DHFR_MD.txt};
      \addplot[boxplot, mark=none] table[y index=0] {Data/Boxplots/ENZYMES.txt};
      \addplot[boxplot, mark=none] table[y index=0] {Data/Boxplots/ER_MD.txt};
      \addplot[boxplot, mark=none] table[y index=0] {Data/Boxplots/FRANKENSTEIN.txt};
      \addplot[boxplot, mark=none] table[y index=0] {Data/Boxplots/IMDB-BINARY.txt};
      \addplot[boxplot, mark=none] table[y index=0] {Data/Boxplots/IMDB-MULTI.txt};
      \addplot[boxplot, mark=none] table[y index=0] {Data/Boxplots/KKI.txt};
      \addplot[boxplot, mark=none] table[y index=0] {Data/Boxplots/Letter-high.txt};
      \addplot[boxplot, mark=none] table[y index=0] {Data/Boxplots/Letter-low.txt};
      \addplot[boxplot, mark=none] table[y index=0] {Data/Boxplots/Letter-med.txt};
      \addplot[boxplot, mark=none] table[y index=0] {Data/Boxplots/MSRC_21.txt};
      \addplot[boxplot, mark=none] table[y index=0] {Data/Boxplots/MSRC_21C.txt};
      \addplot[boxplot, mark=none] table[y index=0] {Data/Boxplots/MSRC_9.txt};
      \addplot[boxplot, mark=none] table[y index=0] {Data/Boxplots/MUTAG.txt};
      \addplot[boxplot, mark=none] table[y index=0] {Data/Boxplots/Mutagenicity.txt};
      \addplot[boxplot, mark=none] table[y index=0] {Data/Boxplots/NCI1.txt};
      \addplot[boxplot, mark=none] table[y index=0] {Data/Boxplots/NCI109.txt};
      \addplot[boxplot, mark=none] table[y index=0] {Data/Boxplots/OHSU.txt};
      \addplot[boxplot, mark=none] table[y index=0] {Data/Boxplots/PROTEINS.txt};
      \addplot[boxplot, mark=none] table[y index=0] {Data/Boxplots/PROTEINS_full.txt};
      \addplot[boxplot, mark=none] table[y index=0] {Data/Boxplots/PTC_FM.txt};
      \addplot[boxplot, mark=none] table[y index=0] {Data/Boxplots/PTC_FR.txt};
      \addplot[boxplot, mark=none] table[y index=0] {Data/Boxplots/PTC_MM.txt};
      \addplot[boxplot, mark=none] table[y index=0] {Data/Boxplots/PTC_MR.txt};
      \addplot[boxplot, mark=none] table[y index=0] {Data/Boxplots/Peking_1.txt};
      \addplot[boxplot, mark=none] table[y index=0] {Data/Boxplots/REDDIT-BINARY.txt};
      \addplot[boxplot, mark=none] table[y index=0] {Data/Boxplots/REDDIT-MULTI-12K.txt};
      \addplot[boxplot, mark=none] table[y index=0] {Data/Boxplots/REDDIT-MULTI-5K.txt};
      \addplot[boxplot, mark=none] table[y index=0] {Data/Boxplots/SYNTHETIC.txt};
      \addplot[boxplot, mark=none] table[y index=0] {Data/Boxplots/SYNTHETICnew.txt};
      \addplot[boxplot, mark=none] table[y index=0] {Data/Boxplots/Synthie.txt};

      \addplot[only marks, mark size = 0.5pt, x filter/.expression = { 1}] table[y index=0] {Data/Boxplots/AIDS.txt};
      \addplot[only marks, mark size = 0.5pt, x filter/.expression = { 2}] table[y index=0] {Data/Boxplots/BZR.txt};
      \addplot[only marks, mark size = 0.5pt, x filter/.expression = { 3}] table[y index=0] {Data/Boxplots/BZR_MD.txt};
      \addplot[only marks, mark size = 0.5pt, x filter/.expression = { 4}] table[y index=0] {Data/Boxplots/COIL-DEL.txt};
      \addplot[only marks, mark size = 0.5pt, x filter/.expression = { 5}] table[y index=0] {Data/Boxplots/COIL-RAG.txt};
      \addplot[only marks, mark size = 0.5pt, x filter/.expression = { 6}] table[y index=0] {Data/Boxplots/COLLAB.txt};
      \addplot[only marks, mark size = 0.5pt, x filter/.expression = { 7}] table[y index=0] {Data/Boxplots/COX2.txt};
      \addplot[only marks, mark size = 0.5pt, x filter/.expression = { 8}] table[y index=0] {Data/Boxplots/COX2_MD.txt};
      \addplot[only marks, mark size = 0.5pt, x filter/.expression = { 9}] table[y index=0] {Data/Boxplots/DD.txt};
      \addplot[only marks, mark size = 0.5pt, x filter/.expression = { 9}] table[y index=0] {Data/Boxplots/DD.txt};
      \addplot[only marks, mark size = 0.5pt, x filter/.expression = {10}] table[y index=0] {Data/Boxplots/DHFR.txt};
      \addplot[only marks, mark size = 0.5pt, x filter/.expression = {11}] table[y index=0] {Data/Boxplots/DHFR_MD.txt};
      \addplot[only marks, mark size = 0.5pt, x filter/.expression = {12}] table[y index=0] {Data/Boxplots/ENZYMES.txt};
      \addplot[only marks, mark size = 0.5pt, x filter/.expression = {13}] table[y index=0] {Data/Boxplots/ER_MD.txt};
      \addplot[only marks, mark size = 0.5pt, x filter/.expression = {14}] table[y index=0] {Data/Boxplots/FRANKENSTEIN.txt};
      \addplot[only marks, mark size = 0.5pt, x filter/.expression = {15}] table[y index=0] {Data/Boxplots/IMDB-BINARY.txt};
      \addplot[only marks, mark size = 0.5pt, x filter/.expression = {16}] table[y index=0] {Data/Boxplots/IMDB-MULTI.txt};
      \addplot[only marks, mark size = 0.5pt, x filter/.expression = {17}] table[y index=0] {Data/Boxplots/KKI.txt};
      \addplot[only marks, mark size = 0.5pt, x filter/.expression = {18}] table[y index=0] {Data/Boxplots/Letter-high.txt};
      \addplot[only marks, mark size = 0.5pt, x filter/.expression = {19}] table[y index=0] {Data/Boxplots/Letter-low.txt};
      \addplot[only marks, mark size = 0.5pt, x filter/.expression = {20}] table[y index=0] {Data/Boxplots/Letter-med.txt};
      \addplot[only marks, mark size = 0.5pt, x filter/.expression = {21}] table[y index=0] {Data/Boxplots/MSRC_21.txt};
      \addplot[only marks, mark size = 0.5pt, x filter/.expression = {22}] table[y index=0] {Data/Boxplots/MSRC_21C.txt};
      \addplot[only marks, mark size = 0.5pt, x filter/.expression = {23}] table[y index=0] {Data/Boxplots/MSRC_9.txt};
      \addplot[only marks, mark size = 0.5pt, x filter/.expression = {24}] table[y index=0] {Data/Boxplots/MUTAG.txt};
      \addplot[only marks, mark size = 0.5pt, x filter/.expression = {25}] table[y index=0] {Data/Boxplots/Mutagenicity.txt};
      \addplot[only marks, mark size = 0.5pt, x filter/.expression = {26}] table[y index=0] {Data/Boxplots/NCI1.txt};
      \addplot[only marks, mark size = 0.5pt, x filter/.expression = {27}] table[y index=0] {Data/Boxplots/NCI109.txt};
      \addplot[only marks, mark size = 0.5pt, x filter/.expression = {28}] table[y index=0] {Data/Boxplots/OHSU.txt};
      \addplot[only marks, mark size = 0.5pt, x filter/.expression = {29}] table[y index=0] {Data/Boxplots/PROTEINS.txt};
      \addplot[only marks, mark size = 0.5pt, x filter/.expression = {30}] table[y index=0] {Data/Boxplots/PROTEINS_full.txt};
      \addplot[only marks, mark size = 0.5pt, x filter/.expression = {31}] table[y index=0] {Data/Boxplots/PTC_FM.txt};
      \addplot[only marks, mark size = 0.5pt, x filter/.expression = {32}] table[y index=0] {Data/Boxplots/PTC_FR.txt};
      \addplot[only marks, mark size = 0.5pt, x filter/.expression = {33}] table[y index=0] {Data/Boxplots/PTC_MM.txt};
      \addplot[only marks, mark size = 0.5pt, x filter/.expression = {34}] table[y index=0] {Data/Boxplots/PTC_MR.txt};
      \addplot[only marks, mark size = 0.5pt, x filter/.expression = {35}] table[y index=0] {Data/Boxplots/Peking_1.txt};
      \addplot[only marks, mark size = 0.5pt, x filter/.expression = {36}] table[y index=0] {Data/Boxplots/REDDIT-BINARY.txt};
      \addplot[only marks, mark size = 0.5pt, x filter/.expression = {37}] table[y index=0] {Data/Boxplots/REDDIT-MULTI-12K.txt};
      \addplot[only marks, mark size = 0.5pt, x filter/.expression = {38}] table[y index=0] {Data/Boxplots/REDDIT-MULTI-5K.txt};
      \addplot[only marks, mark size = 0.5pt, x filter/.expression = {39}] table[y index=0] {Data/Boxplots/SYNTHETIC.txt};
      \addplot[only marks, mark size = 0.5pt, x filter/.expression = {40}] table[y index=0] {Data/Boxplots/SYNTHETICnew.txt};
      \addplot[only marks, mark size = 0.5pt, x filter/.expression = {41}] table[y index=0] {Data/Boxplots/Synthie.txt};
    \end{axis}
  \end{tikzpicture}
  \fi
  \caption{%
    Boxplots of the accuracy distribution of all graph kernels,
    separated by data set. A high spread indicates that a data set
    poses difficulties for a certain class of graph kernels.
  }
  \label{fig:Boxplots accuracies}
\end{figure}
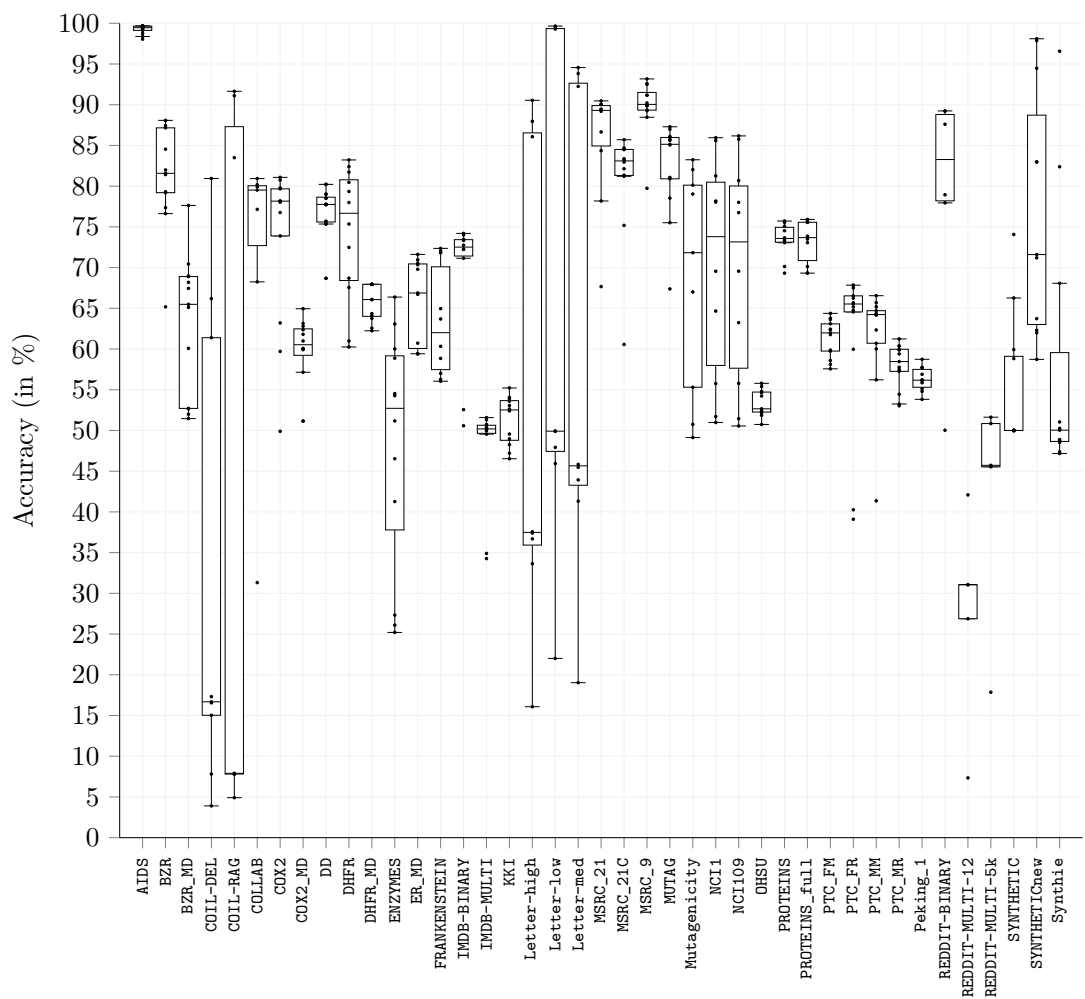
%
\subsection{Boxplot analysis}

The preceding visualisations are very ``dense'' in the sense that they
show all data at the same time. To assess the difficulty of data sets,
a coarse perspective is sufficient.
Figure~\ref{fig:Boxplots accuracies} depicts boxplots of
the accuracy distributions of kernels, grouped by data set~(kernels that
did not finish computing were \emph{not} included; thus, the number of accuracy
values for each data set might vary). The underlying idea of such
a visualisation is to show whether \emph{all} kernels behave similarly
on a specific data set or not.
Even though one might be tempted to compare different data sets via
their boxplots, only the variance of kernels on each data set should be
considered---this visualisation cannot be used to assess whether
a certain graph kernel is suitable for classification because no
baselines for a random classifier are shown.

\subsubsection{High-variance data sets}
%
We first observe that some of the data sets exhibit a large spread in
their accuracies.
The list of high-variance data sets includes
\begin{compactenum}[(1)]
  \item\texttt{COIL-RAG},
  \item\texttt{COIL-DEL},
  \item\texttt{ENZYMES},
  \item\texttt{Letter-high},
  \item\texttt{Letter-low},
  \item\texttt{Letter-med},
  \item\texttt{Mutagenicity},
  \item\texttt{NCI1},
  \item\texttt{NCI109},
  \item\texttt{SYNTHETICnew}.
\end{compactenum}
Out of those, only \texttt{COIL-DEL}, \texttt{ENZYMES},
\texttt{Mutagenicity}, \texttt{NCI1}, and \texttt{NCI109} have node or
edge labels. The remaining data sets either feature node or edge
attributes.
If we link this back to Figure~\ref{fig:Vertex histogram kernel
scatterplot}, which depicted the differences in performance with respect
to the vertex histogram kernel here, we see that this list comprises
data sets in which the vertex histogram kernel did not perform as well
as other types of kernels.
Thus, these data sets can be considered ``hard'' to classify accurately,
but the performance also highlights the need for developing~(more) graph
kernels that
\begin{inparaenum}[(i)]
  \item are capable of handling graphs with node or edge attributes, and
  \item scale well.
\end{inparaenum}
Such graph kernels have the potential to significantly improve
classification performance here.

\subsubsection{Low-variance data sets}
%
By contrast, there are also data sets that can be considered to be
``solved'' in the sense that most graph kernels perform extremely
similarly~(such as for the \texttt{AIDS} data set, which \emph{every}
kernel can classify very well). This does \emph{not} necessarily imply
that all such data sets are solved. For example, for the
\texttt{PTC-$\ast$} data sets, it is unclear whether performance can be
significantly improved with, for example, a new technique that is better
able to exploit structural information, or whether the best performance
on this data set has already been reached. Prior to discussing how to
estimate the difficulty of a data set in a more principled manner, we
first provide a numerical view on the accuracies of all data sets.

\begin{table}[tbp]
  \sisetup{
    detect-weight           = true,
    detect-inline-weight    = math,
    table-format            = 2.2(2),
    separate-uncertainty    = true,
    table-align-uncertainty = true,
    table-text-alignment    = center,
  }
  \newcommand{\w}{\color{red}}
  \footnotesize
  \centering
  \begin{tabular}{>{\ttfamily}lSSS[table-format=2.2(2)]S}
    \toprule
    \normalfont{Data set} & {\normalfont{$\min$}}
                          & {\normalfont{$\max$}} & {\normalfont{Avg.}}
                          & {\normalfont{$\%\Delta$}}\\
    \midrule
      \w AIDS                 & 98.05  & 99.70  & 99.24 \pm 0.53  & 1.65\\
       BZR                    & 65.18  & 88.08  & 81.38 \pm 6.32  & 22.90\\
       BZR\_MD                & 51.46  & 77.63  & 63.16 \pm 8.53  & 26.17\\
       COIL-DEL               & 3.90   & 80.94  & 31.76 \pm 29.11 & 77.04\\
       COIL-RAG               & 4.91   & 91.65  & 30.42 \pm 37.69 & 86.74\\
       COLLAB                 & 31.32  & 80.93  & 71.04 \pm 18.05 & 49.61\\
       COX2                   & 49.90  & 81.08  & 73.64 \pm 9.73  & 31.18\\
       COX2\_MD               & 51.15  & 64.95  & 59.63 \pm 4.42  & 13.80\\
       DD                     & 68.68  & 80.22  & 76.62 \pm 3.59  & 11.54\\
       DHFR                   & 60.25  & 83.22  & 74.20 \pm 8.17  & 22.97\\
       \w DHFR\_MD            & 62.24  & 67.95  & 65.74 \pm 2.24  & 5.71\\
       ENZYMES                & 25.20  & 66.38  & 47.90 \pm 14.75 & 41.18\\
       ER\_MD                 & 59.42  & 71.62  & 65.99 \pm 5.19  & 12.20\\
       FRANKENSTEIN           & 56.05  & 72.36  & 63.34 \pm 6.70  & 16.31\\
       IMDB-BINARY            & 50.58  & 74.20  & 68.66 \pm 9.06  & 23.62\\
       IMDB-MULTI             & 34.27  & 51.58  & 47.30 \pm 6.73  & 17.31\\
       \w KKI                 & 46.53  & 55.22  & 51.28 \pm 2.99  & 8.69\\
       Letter-high            & 16.08  & 90.54  & 51.37 \pm 26.20 & 74.46\\
       Letter-low             & 22.01  & 99.66  & 63.19 \pm 27.49 & 77.65\\
       Letter-med             & 19.03  & 94.56  & 58.21 \pm 26.16 & 75.53\\
       MSRC\_21               & 67.67  & 90.47  & 85.55 \pm 7.34  & 22.80\\
       MSRC\_21C              & 60.56  & 85.70  & 80.78 \pm 6.92  & 25.14\\
       MSRC\_9                & 79.74  & 93.17  & 89.78 \pm 3.49  & 13.43\\
       MUTAG                  & 67.38  & 87.29  & 82.41 \pm 5.77  & 19.91\\
       Mutagenicity           & 49.13  & 83.24  & 68.72 \pm 13.80 & 34.11\\
       NCI1                   & 50.98  & 85.95  & 70.17 \pm 13.68 & 34.97\\
       NCI109                 & 50.55  & 86.17  & 69.80 \pm 13.76 & 35.62\\
       \w OHSU                & 50.74  & 55.79  & 53.36 \pm 1.66  & 5.05\\
       \w PROTEINS            & 69.32  & 75.72  & 73.38 \pm 2.15  & 6.40\\
       \w PROTEINS\_full      & 69.32  & 75.92  & 73.20 \pm 2.68  & 6.60\\
       \w PTC\_FM             & 57.57  & 64.38  & 61.33 \pm 2.29  & 6.81\\
       PTC\_FR                & 39.10  & 67.84  & 61.61 \pm 9.94  & 28.74\\
       PTC\_MM                & 41.36  & 66.55  & 61.52 \pm 6.67  & 25.19\\
       \w PTC\_MR             & 53.03  & 61.24  & 57.92 \pm 2.77  & 8.21\\
       \w Peking\_1           & 53.82  & 58.74  & 56.30 \pm 1.51  & 4.92\\
       REDDIT-BINARY          & 50.03  & 89.24  & 78.83 \pm 14.99 & 39.21\\
       REDDIT-MULTI-12K       & 7.34   & 42.09  & 27.69 \pm 12.70 & 34.75\\
       REDDIT-MULTI-5K        & 17.86  & 51.63  & 42.32 \pm 13.96 & 33.77\\
       SYNTHETIC              & 50.00  & 74.07  & 54.93 \pm 8.14  & 24.07\\
       SYNTHETICnew           & 58.73  & 98.10  & 76.90 \pm 15.07 & 39.37\\
       Synthie                & 47.17  & 96.57  & 58.10 \pm 16.86 & 49.40\\
    \bottomrule
  \end{tabular}
  \caption{%
    An overview of the performance values of all graph kernels on the
    benchmark data sets. Data sets whose performance difference is less
    than \SI{10}{\percent} have been highlighted.
  }
  \label{tab:Difficulties}
\end{table}

\subsubsection{Summarising differences in accuracy}
%
Table~\ref{tab:Difficulties} summarises the performance measures that we
extracted in the previous plots by showing the performance gap between
the worst-performing method and the best-performing method, as well as
the average accuracy obtained on a given data set.
Using an arbitrary cut-off of \SI{10}{\percent} accuracy difference, we
have the following data sets:
\begin{compactenum}[(1)]
  \item \texttt{AIDS},
  \item \texttt{DHFR\_MD},
  \item \texttt{KKI},
  \item \texttt{OHSU},
  \item \texttt{PROTEINS},
  \item \texttt{PROTEINS\_full},
  \item \texttt{PTC\_FM}, 
  \item \texttt{PTC\_MR}, and
  \item \texttt{Peking\_1}.
\end{compactenum}
All of these data sets also exhibit small standard deviations in their
accuracy distributions. This suggests that \emph{all} graph kernels are
performing almost equally and none of them have a clear advantage over
the other.

Again, this does not necessarily imply that these data sets are too
easy: while \texttt{AIDS} can be considered as ``solved'', the
performance in the remaining data sets could be improved. However, the
list shows that these data sets appear to contain a sufficient amount of
structural information that can be exploited to some extent by all graph
kernels.  This makes these data sets beneficial for comparing different
types of graph kernels. Any analysis that claims the superiority of
a specific graph kernel should nonetheless employ other data sets that
provide more information about its generalisation performance.

\subsection{Estimating maximum predictive performance}
%
So far, we have analysed all data sets from different perspectives,
focusing on how ``easy'' they make classification for different graph
kernels. Now we want to assess their difficulty in a more principled
manner. To this end, we estimate the \emph{maximum} predictive
performance that can be achieved on a given data set \re{for the considered graph kernels}. We follow a very
conservative procedure here: we first take the predictions of all graph
kernels over all folds and all repetitions of the training process.
Following our cross-validation procedure, we will thus be able to
collect predictions of all graphs in every data set. For each data set
and each of its graphs, we now count how many kernels exist that are
capable of classifying a graph correctly. For example, assume we observe
graph $i$ at fold $j$ and some kernel $k$ is predicting the correct
label. We then add $k$ to a set $K_{ij}$ that contains all graph kernels
that are capable of classifying graph $i$ in this fold. This information
can now be summarised in multiple ways. We opt for the most conservative
one, which involves counting all matrices for which $K_{ij}
= \emptyset$, \ie\ all graphs over all folds that cannot be classified
by \emph{any} \re{of the graph kernels we considered}. Letting $N$ refer to the number of all
graphs over all folds and $N'$ to the number of matrices with $K_{ij}
= \emptyset$, the fraction $\nicefrac{N'}{N}$ can thus serve as an
indicator of the difficulty of a benchmark data set. We call graphs
that \re{were unable to be correctly classified on any split of the data and by any of the included graph kernels} \emph{\re{generally misclassified}}.

\begin{figure}[tbp]
  \centering
    \iffinal
    \includegraphics[width=\textwidth]{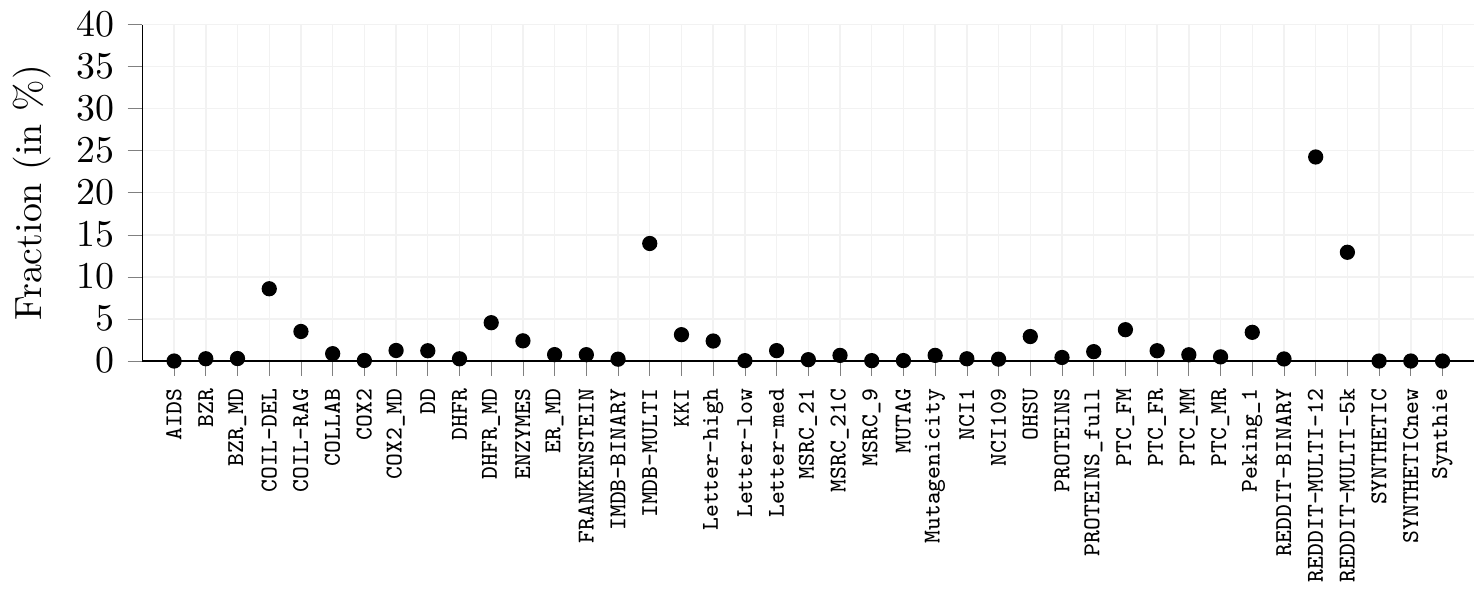}
    \else
  \pgfplotsset{%
    major grid style = {%
      gray!10,
    }
  }
  \begin{tikzpicture}
    \begin{axis}[%
      axis x line*                    = bottom,
      axis y line*                    = left,
      xmajorgrids                     = true,
      ymajorgrids                     = true,
      enlarge x limits                = false,
      enlarge y limits                = false,
      height                          = 5cm,
      width                           = \textwidth,
      xtick = {%
        2, 3, 4, 5, 6, 7, 8, 9, 10, 11, 12, 13, 14, 15, 16, 17, 18, 19,
        20, 21, 22, 23, 24, 25, 26, 27, 28, 29, 30, 31, 32, 33, 34, 35,
        36, 37, 38, 39, 40, 41, 42
      },
      ytick = {
        0, 5, 10, 15, 20, 25, 30, 35, 40
      },
      tick align = outside,
      xmin  = 1,
      xmax  = 43,
      ymin  = 0,
      ymax  = 40,
      ylabel = {Fraction~(in \si{\percent})},
      xticklabels = {%
        AIDS,
        BZR,
        BZR\_MD,
        COIL-DEL,
        COIL-RAG,
        COLLAB,
        COX2,
        COX2\_MD,
        DD,
        DHFR,
        DHFR\_MD,
        ENZYMES,
        ER\_MD,
        FRANKENSTEIN,
        IMDB-BINARY,
        IMDB-MULTI,
        KKI,
        Letter-high,
        Letter-low,
        Letter-med,
        MSRC\_21,
        MSRC\_21C,
        MSRC\_9,
        MUTAG,
        Mutagenicity,
        NCI1,
        NCI109,
        OHSU,
        PROTEINS,
        PROTEINS\_full,
        PTC\_FM,
        PTC\_FR,
        PTC\_MM,
        PTC\_MR,
        Peking\_1,
        REDDIT-BINARY,
        REDDIT-MULTI-12,
        REDDIT-MULTI-5k,
        SYNTHETIC,
        SYNTHETICnew,
        Synthie
      },
      xticklabel style = {
        font   = \scriptsize\ttfamily,
        rotate = 90,
      }
    ]
      \addplot[only marks] table[%
        x       = index,
        y       = unclassifiable,
        col sep = comma,
      ] {Data/Difficulty_unclassifiable.csv};
    \end{axis}
  \end{tikzpicture}
  \fi

   \iffinal
   \includegraphics[width=\textwidth]{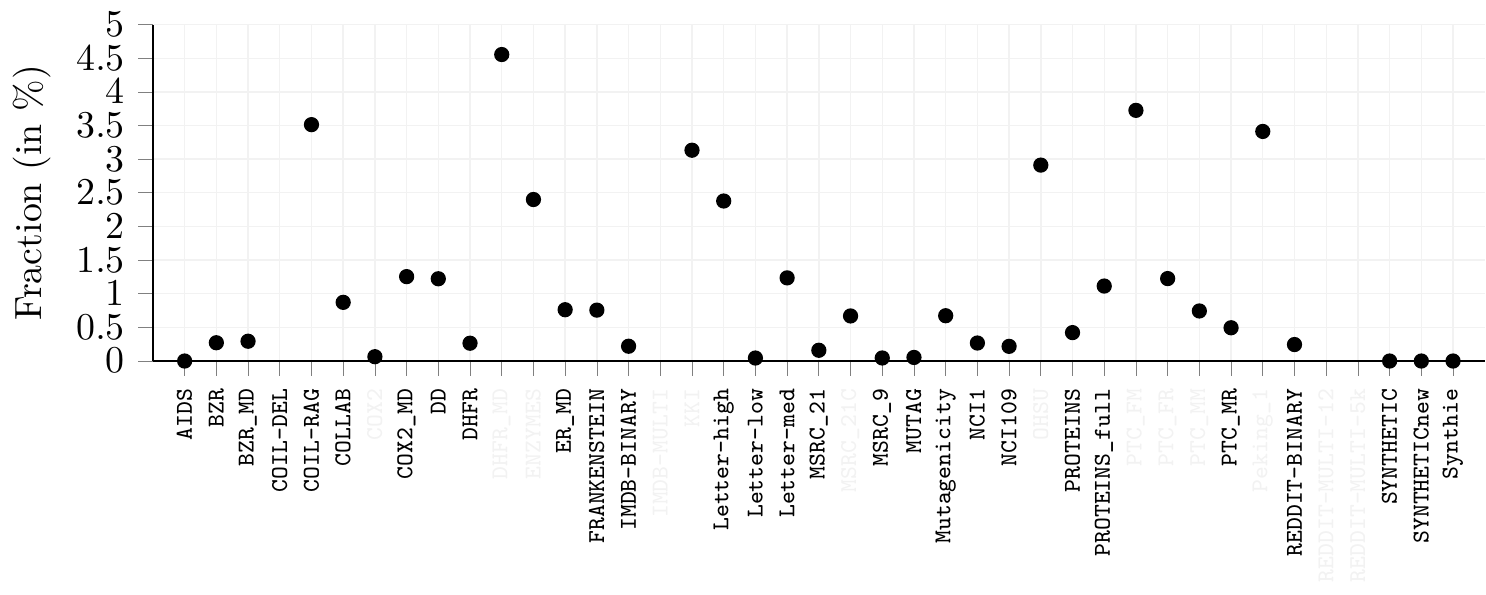}
   \else
  \begin{tikzpicture}
    \begin{axis}[%
      axis x line*                    = bottom,
      axis y line*                    = left,
      xmajorgrids                     = true,
      ymajorgrids                     = true,
      enlarge x limits                = false,
      enlarge y limits                = false,
      height                          = 5cm,
      width                           = \textwidth,
      xtick = {%
        2, 3, 4, 5, 6, 7, 8, 9, 10, 11, 12, 13, 14, 15, 16, 17, 18, 19,
        20, 21, 22, 23, 24, 25, 26, 27, 28, 29, 30, 31, 32, 33, 34, 35,
        36, 37, 38, 39, 40, 41, 42
      },
      ytick = {
        0, 0.5, 1.0, 1.5, 2.0, 2.5, 3.0, 3.5, 4.0, 4.5, 5.0
      },
      tick align = outside,
      xmin  = 1,
      xmax  = 43,
      ymin  = 0,
      ymax  = 5,
      ylabel = {Fraction~(in \si{\percent})},
      xticklabels = {%
        AIDS,
        BZR,
        BZR\_MD,
        COIL-DEL,
        COIL-RAG,
        COLLAB,
        \textcolor{gray!10}{COX2},
        COX2\_MD,
        DD,
        DHFR,
        \textcolor{gray!10}{DHFR\_MD},
        \textcolor{gray!10}{ENZYMES},
        ER\_MD,
        FRANKENSTEIN,
        IMDB-BINARY,
        \textcolor{gray!10}{IMDB-MULTI},
        \textcolor{gray!10}{KKI},
        Letter-high,
        Letter-low,
        Letter-med,
        MSRC\_21,
        \textcolor{gray!10}{MSRC\_21C},
        MSRC\_9,
        MUTAG,
        Mutagenicity,
        NCI1,
        NCI109,
        \textcolor{gray!10}{OHSU},
        PROTEINS,
        PROTEINS\_full,
        \textcolor{gray!10}{PTC\_FM},
        \textcolor{gray!10}{PTC\_FR},
        \textcolor{gray!10}{PTC\_MM},
        PTC\_MR,
        \textcolor{gray!10}{Peking\_1},
        REDDIT-BINARY,
        \textcolor{gray!10}{REDDIT-MULTI-12},
        \textcolor{gray!10}{REDDIT-MULTI-5k},
        SYNTHETIC,
        SYNTHETICnew,
        Synthie
      },
      xticklabel style = {
        font   = \scriptsize\ttfamily,
        rotate = 90,
      }
    ]
      \addplot[only marks] table[%
        x       = index,
        y       = unclassifiable,
        col sep = comma,
      ] {Data/Difficulty_unclassifiable.csv};
    \end{axis}
  \end{tikzpicture}
  \fi
  \caption{%
    A depiction of the fraction of \emph{\re{generally misclassified}} graphs in each
    data set, \ie\ graphs that \re{were not classified correctly on any split of the data for any graph kernel in our experiments.}
    The upper part of the figure shows \emph{all} fractions, whereas in
    the lower part, only those data sets whose fraction of
    \re{generally misclassified} graphs is below \SI{5}{\percent} are shown.
  }
  \label{fig:Unclassifiable}
\end{figure}
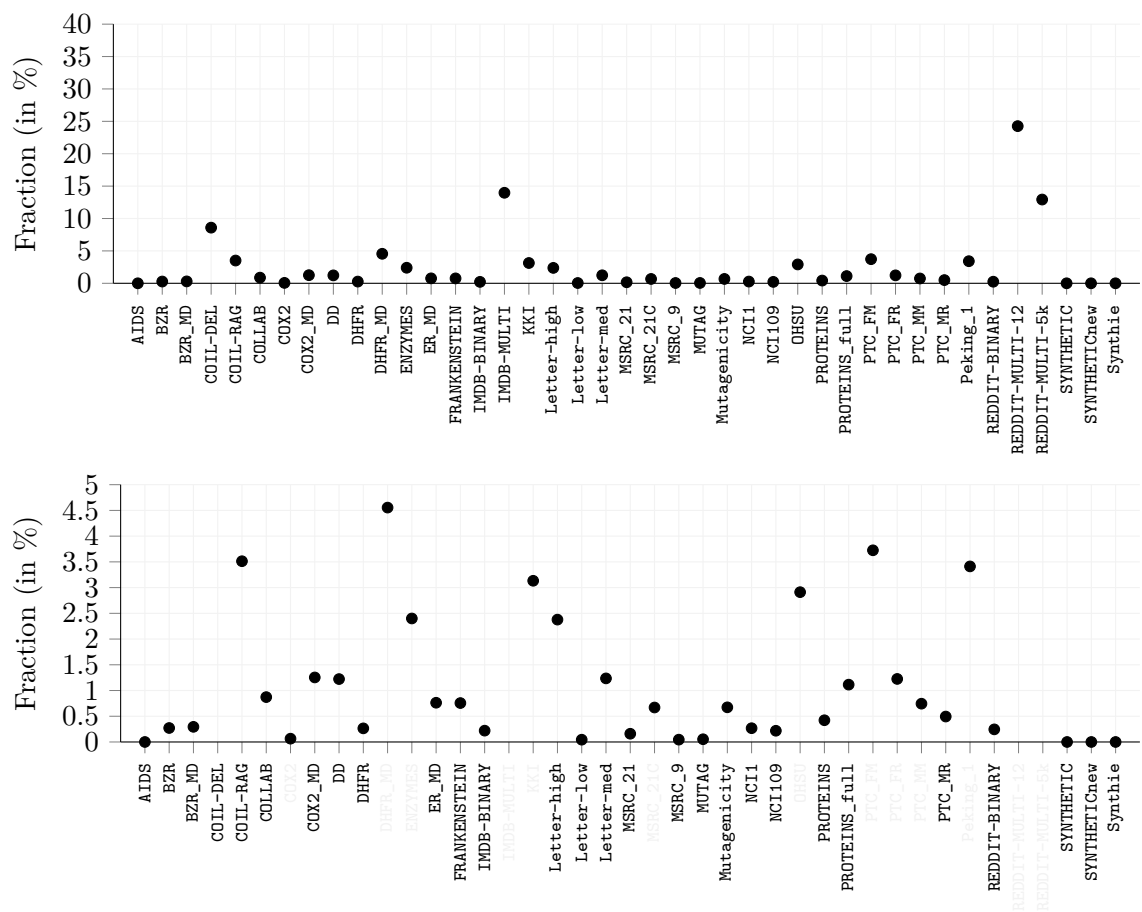

\subsubsection{Depicting generally misclassified graphs}
%
Figure~\ref{fig:Unclassifiable} depicts the fraction of \re{generally misclassified}
graphs over all data sets---once for \emph{all} fractions, and once for
only those data sets whose fraction of \re{generally misclassified} graphs is below
\SI{5}{\percent}. We observe that numerous data sets remain.
The implications of this plot are positive: the fact that there are few
\re{generally misclassified} graphs in all data sets means that the data sets are
\emph{highly consistent} in the sense that they contain few graphs that \re{were always misclassified in our experiments}~(for example, because their label is
inconsistent, or their representation is non-unique and overlaps with
another graph that contains a different label).
The obvious exceptions from this are the unlabelled data sets
\texttt{IMDB-MULTI}, \texttt{REDDIT-MULTI-12}, and
\texttt{REDDIT-MULTI-5k}.
We conjecture that these data sets suffer from graphs that are either
duplicates or quasi-isomorphic but with different labels. Given the
provenance of these data sets, \ie\ their generation based on online
forums without the inclusion of labels~\citep{Yanardag15}, care needs
to be taken in assessing classification results here.
\re{Having generally misclassified graphs in a data set poses two plausible interpretations. On the one hand, it is possible that these graphs are unclassifiable, perhaps due to non-unique representation or mistaken labels, as mentioned above. On the other hand, it is also possible that the graph kernels considered are simply not powerful enough to differentiate these challenging graphs. Since our experiments considered a broad array of graph kernels across nearly all categories defined in Chapter~\ref{chap:Kernels}, we suspect these graphs are indeed quite often generally misclassfied, which therefore would represent an upper limit on the classification performance we would expect any graph kernel to achieve.} 

\begin{figure}[tbp]
  \centering
   \iffinal
   \includegraphics[width=0.7\textwidth]{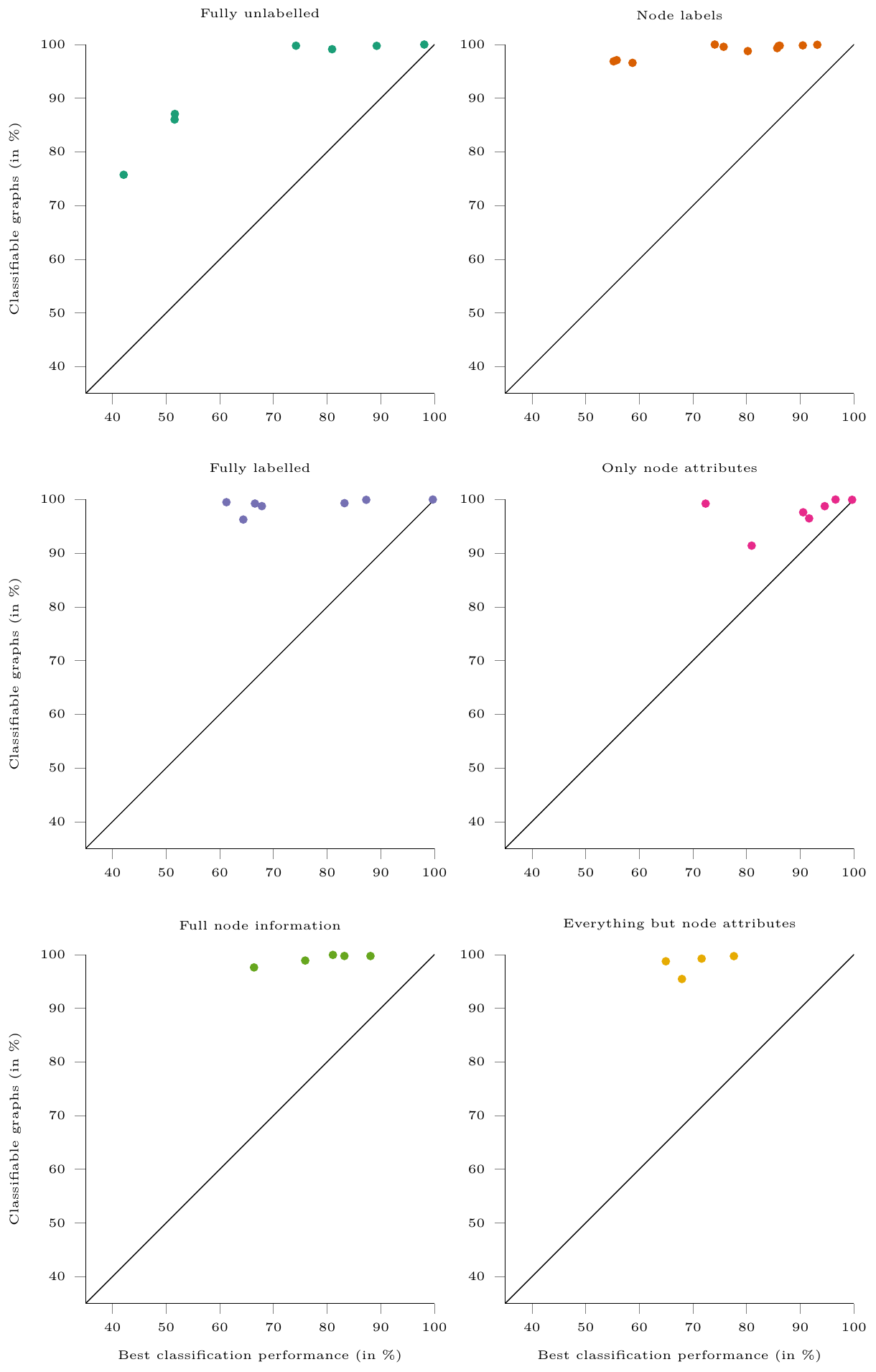}
   \else
  \tikzstyle{every node}=[font=\tiny]
  \pgfplotsset{%
    every axis plot/.append style = {
      mark size = 1.5pt,
    },
  }
  \begin{tikzpicture}
    \begin{groupplot}[%
      group style      = {%
        group size     = 2 by 3,
        vertical sep   = 1.5cm,
        xlabels at     = edge bottom,
        ylabels at     = edge left,
      }, 
      axis x line*     = bottom,
      axis y line*     = left,
      enlarge x limits = false,
      enlarge y limits = false,
      width            = 0.50\textwidth,
      every node near coord/.append style = {%
        font = \tiny\ttfamily,
      },
      tick align = outside,
      xlabel = {Best classification performance~(in \si{\percent})},
      ylabel = {Classifiable graphs~(in \si{\percent})},
      xmin = 35.0, xmax = 100.0,
      ymin = 35.0, ymax = 100.0,
      xtick = {40, 50, 60, 70, 80, 90, 100},
      ytick = {40, 50, 60, 70, 80, 90, 100},
      %
      unit vector ratio*= 1 1 1,
      scatter/classes = {%
        1={Dark2-A},
        2={Dark2-B},
        3={Dark2-C},
        4={Dark2-D},
        5={Dark2-E},
        6={Dark2-F},
        7={Dark2-G} 
      },
      legend pos = south east,
      legend cell align={left} 
    ]

      \nextgroupplot[title = {Fully unlabelled}]

      \addplot[domain={0:200}] {x};

      \addplot[%
         scatter, only marks,
         scatter src = explicit symbolic,
      ] table[%
        x       = best,
        y expr  = 100.0 - \thisrow{unclassifiable},
        col sep = comma,
        meta    = class,
        restrict expr to domain = {\thisrow{class}}{1:1},
      ] {Data/Difficulty_unclassifiable.csv};

      \nextgroupplot[title = {Node labels}]

      \addplot[domain={0:200}] {x};

      \addplot[%
         scatter, only marks,
         scatter src = explicit symbolic,
      ] table[%
        x       = best,
        y expr  = 100.0 - \thisrow{unclassifiable},
        col sep = comma,
        meta    = class,
        restrict expr to domain = {\thisrow{class}}{2:2},
      ] {Data/Difficulty_unclassifiable.csv};

      \nextgroupplot[title = {Fully labelled}]

      \addplot[domain={0:200}] {x};

      \addplot[%
         scatter, only marks,
         scatter src = explicit symbolic,
      ] table[%
        x       = best,
        y expr  = 100.0 - \thisrow{unclassifiable},
        col sep = comma,
        meta    = class,
        restrict expr to domain = {\thisrow{class}}{3:3},
      ] {Data/Difficulty_unclassifiable.csv};

      \nextgroupplot[title = {Only node attributes}]

      \addplot[domain={0:200}] {x};

      \addplot[%
         scatter, only marks,
         scatter src = explicit symbolic,
      ] table[%
        x       = best,
        y expr  = 100.0 - \thisrow{unclassifiable},
        col sep = comma,
        meta    = class,
        restrict expr to domain = {\thisrow{class}}{4:4},
      ] {Data/Difficulty_unclassifiable.csv};

      \nextgroupplot[title = {Full node information}]

      \addplot[domain={0:200}] {x};

      \addplot[%
         scatter, only marks,
         scatter src = explicit symbolic,
      ] table[%
        x       = best,
        y expr  = 100.0 - \thisrow{unclassifiable},
        col sep = comma,
        meta    = class,
        restrict expr to domain = {\thisrow{class}}{5:5},
      ] {Data/Difficulty_unclassifiable.csv};

      \nextgroupplot[title = {Everything but node attributes}]

      \addplot[domain={0:200}] {x};

      \addplot[%
         scatter, only marks,
         scatter src = explicit symbolic,
      ] table[%
        x       = best,
        y expr  = 100.0 - \thisrow{unclassifiable},
        col sep = comma,
        meta    = class,
        restrict expr to domain = {\thisrow{class}}{6:6},
      ] {Data/Difficulty_unclassifiable.csv};

    \end{groupplot}
  \end{tikzpicture}
  \fi
  \caption{%
    A visualisation of the difficulty of each data set, \re{grouped by data set
    type,} by plotting the
    fraction of graphs \re{that were correctly classified on at least one fold of one kernel} against the best performance
    achieved by some kernel. The further away from the diagonal, the
    more difficult a data set is. By contrast, data sets that are
    situated close to the diagonal \re{may have reached their upper limit of performance, since the best performance of the included graph kernels} is roughly similar to
    the fraction of \re{graphs in the data set that} at least
    one graph kernel can identify correctly. \re{On the other hand, it could also indicate the need for a more powerful kernel in order to classify these \emph{generally misclassified} graphs.}
    The colour-coding follows the classes described in
    Section~\ref{sec:Breakdown}.
  }
  \label{fig:Difficulty}
\end{figure}
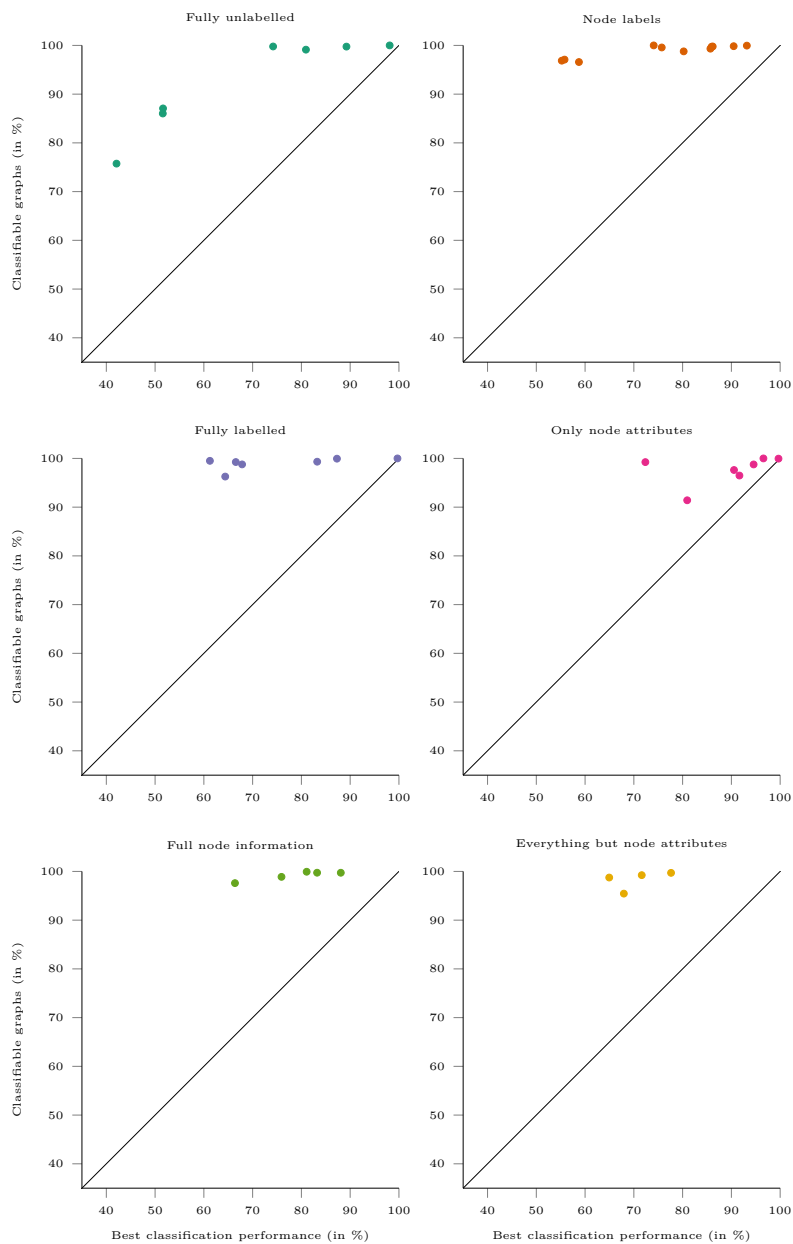

\subsubsection{Gauging the difficulty of all data sets}
%
Having seen that most of the data sets are highly consistent, we now
finally gauge their difficulty. This requires making use of the best
classification accuracy that we obtained for them.
Before we depict the resulting visualisations, we want to motivate the
subsequent assessment by the following observation: if a data set poses a
\emph{simple} classification task, \re{we would expect that} the fraction of graphs that cannot
be classified by \emph{any} \re{considered} graph kernel \re{could be approximately ascertained from the best performance on the data set.} Specifically, if $x$ is the
fraction of \re{generally misclassified} graphs, the performance of the best graph
kernel should be~$\approx 1 - x$, \re{the fraction of graphs that were successfully classified at least once}.
In other words, the best graph kernel
should be able to classify a simple data set \emph{up to} its subset of
\re{generally misclassified} graphs.
This gives us a separate axis, namely the best classification
performance for each data set. Figure~\ref{fig:Difficulty} depicts the
resulting plot, \re{grouped by data set type}. Each dot represents a benchmark data set; the \mbox{$x$-axis} depicts the
best classification performance, while the \mbox{$y$-axis} depicts the
percentage of \re{graphs that were successfully classified at least once in our experiments}. In the interest of readability, we
removed the data set labels from the plot and will only refer to them in
the subsequent analysis.

This figure lends itself to numerous insights: we first observe that
there are several data sets whose fraction of \re{graphs that were correctly classified at least once} is
approached by the best performance of some graph kernel~(within less
than \SI{0.5}{\percent}), namely
%
\begin{inparaenum}[(i)]
  \item \texttt{AIDS},
  \item \texttt{Letter-low},
  \item \texttt{Synthie}, and
  \item \texttt{SYNTHETICnew}.
\end{inparaenum}
\re{We therefore state that these data sets have reached their upper limit of performance with respect to the graph kernels considered. Any additional kernel comparisons should therefore take care to ensure that the performance surpasses the percent of graphs that are currently classified correctly, because otherwise} they might inadvertently conclude that the
performance of a new graph kernel surpasses existing graph kernels,
when in reality, the performance benefits are minuscule~(or in
the worst case, are just caused by fold variations).

%
If we extend the threshold between the fraction of \re{the graphs that were correctly classified at least once} 
versus the best performance of a graph kernel to \SI{5}{\percent}, the
list of simple data sets starts to include \texttt{COIL-RAG} and \texttt{Letter-med}.
Their optimal classification performance, according to these
considerations, should be larger \re{than} \SI{95}{\percent}.
As the values in Table~\ref{tab:Empirical performance} demonstrate,
there is still a gap that cannot be fully explained by the standard
deviation. Hence, it is likely that the classification performance of
these data sets may yet be increased by a few performance points by some
future graph kernel.


By contrast, it is interesting to see that the challenging data
sets---according to this metric---are the \emph{unlabelled} data
sets of class~i, which can be clearly seen as outliers in the
aforementioned plot. Data sets of class~vi, containing everything
\emph{but} node attributes also appear to have a hidden complexity
that is yet to be overcome.
On the other hand, data sets from class~ii~(only node labels) and
class~iii~(node and edge labels), are distributed in the plot. Some of
these data sets hence appear to be more difficult than others of the
same type, which is generally preferable for a collection of benchmark
data sets.

\subsection{Consequences}

We conclude this analysis by discussing the consequences of the
preceding analyses. Our recommendations vary in terms of their epistemic
status between ``authoritative'', implying that we consider our claim to
be strong, and ``exploratory'', which we consider to be
\emph{suggestions} for the community.

\subsubsection{Exclusion of \re{easy} data sets}
%
We consider this to be an authoritative claim that is backed up very
well by the previous analyses: given their simplicity, the data sets
\begin{inparaenum}[(i)]
  \item \texttt{AIDS},
  \item \texttt{Letter-low},
  \item \texttt{Synthie}, and
  \item \texttt{SYNTHETICnew}
\end{inparaenum}
should \emph{not} be considered in any formal comparison of graph
kernels any more.


\subsubsection{Exclusion of node-attributed data sets}

As Figure~\ref{fig:Difficulty} demonstrates, the performance of many
data sets of type~iv, \ie\ containing node attributes and no
labels~(except for \texttt{COIL-DEL}, which we included in this category
for reasons of simplicity), is already \emph{close} to optimal.
As an exploratory suggestion, to guard against picking up a wrong signal
when discussing the merits of a specific graph kernel, we suggest to
consider excluding at least
\begin{inparaenum}[(i)]
  \item \texttt{Letter-\-low},
  \item \texttt{Letter-med}, and
  \item \texttt{Letter-high}
\end{inparaenum}
from an analysis~(please note that \texttt{Synthie}, as suggested above,
should \emph{always} be removed because we consider it to be solved).
The other data sets of this type, \ie\ \texttt{COIL-DEL} and \texttt{FRANKENSTEIN},
can be kept, but we suggest caution when basing any performance claims
on these data sets alone.

\section{Grouping graph kernels}\label{sec:Grouping graph kernels}

Despite classification accuracy being the primary metric of interest, we
nonetheless require tools to select a graph kernel in practice. Thus, we
discuss multiple methods for \emph{grouping} graph kernels before
providing a flowchart to \emph{choose} them.

\subsection{Critical difference analysis}\label{sec:Critical difference analysis}

Prior to employing methods that focus on individual predictions or
kernel matrices, we perform a statistical analysis of the AUROC values
and the ranks of individual graph kernels. Our goal is to assess to what
extent different graph kernels are statistically significantly different
from each other. 
%
%
We then calculate a \emph{critical difference
plot}~\citep{Calvo16, Demsar06}. Originally developed for the comparison
of classification algorithms~\citep{Demsar06}, the critical difference
plot is now commonly employed in large-scale surveys of classifiers~\citep{Bagnall17}.
Briefly put, a critical difference plot employs a Nemenyi test to obtain
a critical difference value. If the performance difference between two
algorithms \emph{exceeds} said value, the algorithms are considered to
be statistically significantly different. This can be visualised in
a corresponding diagram, in which algorithms whose performance is not
statistically significantly different are connected by a line---thus
immediately grouping pairs of available algorithms in terms of
their differences. However, it needs to be stressed that the test
analyses \emph{pairwise} differences, so the plot should only be used to
make claims about \emph{pairs} of classifiers; it is not to be seen as
a ``clustering'' method.

\begin{figure}[tbp]
  \centering
  \includegraphics[width=.7\textwidth]{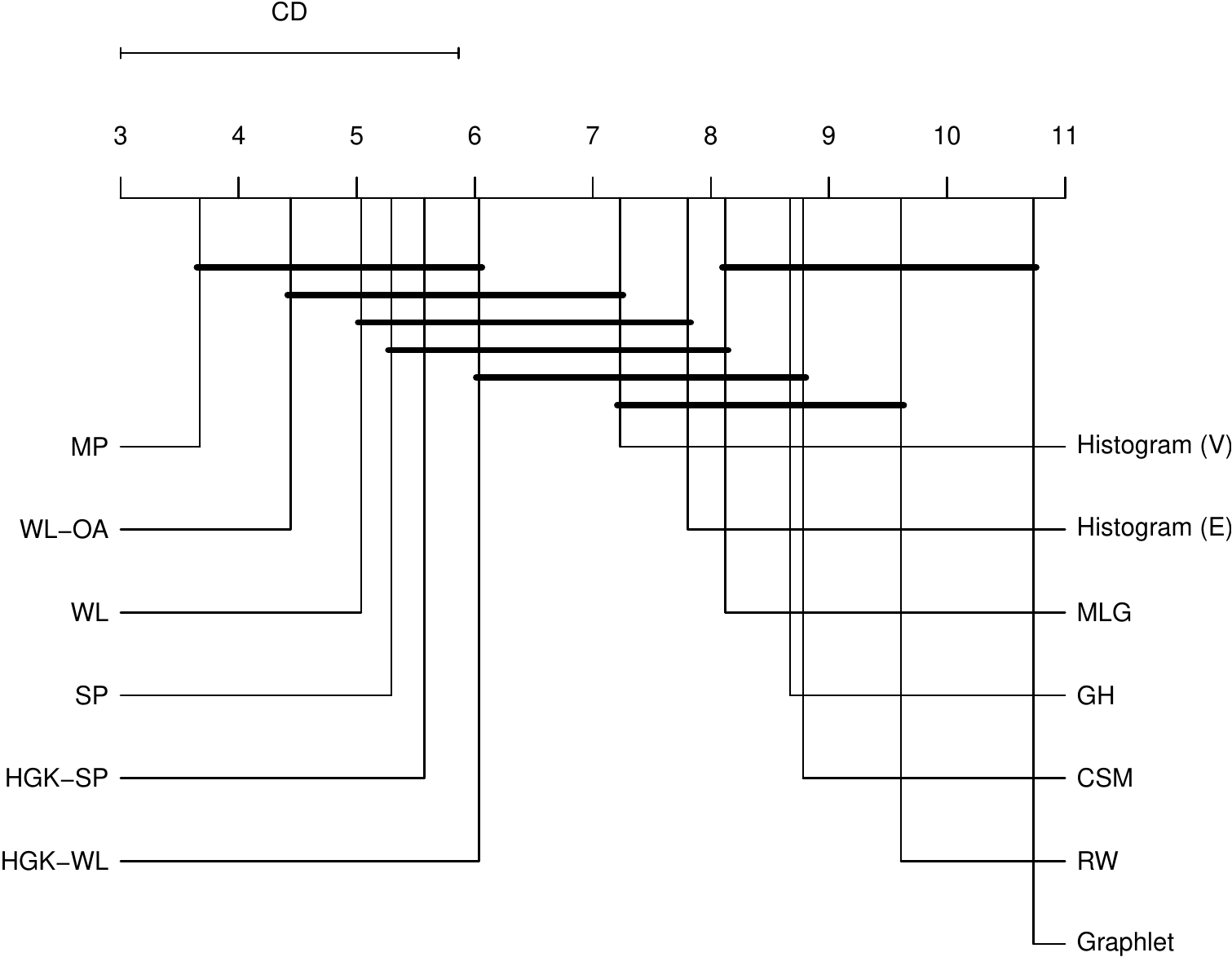}
  \caption{%
    A \emph{critical difference plot} of the graph kernels at the
    $\alpha = 0.05$ level. Any two kernels that are part of the same
    interval do \emph{not} exhibit a statistically significantly
    different classification performance.
  }
  \label{fig:Critical difference plot}
\end{figure}

Figure~\ref{fig:Critical difference plot} depicts the critical
difference plot for different graph kernels, at a significance level of
$\alpha = 0.05$. We can see that, despite
the better average rank of the graph kernels based on the
Weisfeiler--Lehman relabelling scheme, the performance of a group of
graph kernels is statistically not significantly different. These
kernels include
\begin{inparaenum}[(i)]
  \item MP,
  \item \mbox{WL-OA},
  \item WL, and
  \item the two Hash Graph Kernel variants,
\end{inparaenum}
Each pair of these graph kernels is \emph{not} statistically
significantly different from each other in terms of their performance on
the whole benchmark data set.
As the test is very conservative, we can only claim statistical
significance of differences between several pairs of other graph
kernels. For example, from the group of kernels mentioned above, there
is only a statistically significant difference between MP and the
vertex histogram kernel or between \mbox{WL-OA} and the vertex histogram
kernel. For \emph{all} other pairs of graph kernels of the same group,
the test lacks statistical power to claim a statistically significant
difference.
Hence, these results need to be taken with a grain of salt. At the very
least, given the large number of pairwise comparisons, the benchmark
data set repository does not seem to be entirely suitable to make claims
about the statistical significance of large groups of graph kernels.

\subsection{Grouping based on predictions}\label{sec:Grouping by predictions}

Next to the critical difference analysis, a straightforward way of
grouping graph kernels involves their \emph{predictions} on the
benchmark data sets.
To this end, we aggregate all predictions~(on all folds) of a graph
kernel into a high-dimensional label vector $\mathbf{y}$. Given another
such vector $\mathbf{y'}$ created from the predictions of another graph
kernel, we calculate their \emph{Hamming distance} to see how similar
their predictions are. We are \emph{not} interested in knowing whether
these predictions are correct; we are merely interested in knowing to
what extent they agree. Since the Hamming distance is a metric, we can
collect the pairwise dissimilarity scores in a quadratic matrix and use
\emph{metric multidimensional scaling}~\citep[Chapter~9]{Borg05} to
obtain a two-dimensional embedding.

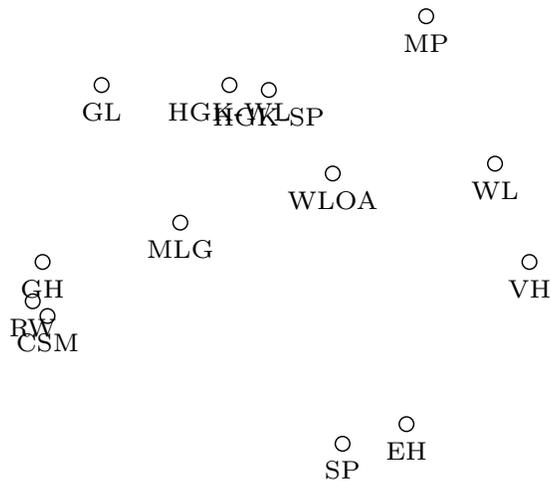
\begin{figure}[tb]
  \centering
  \iffinal
  \includegraphics[width=0.5\textwidth]{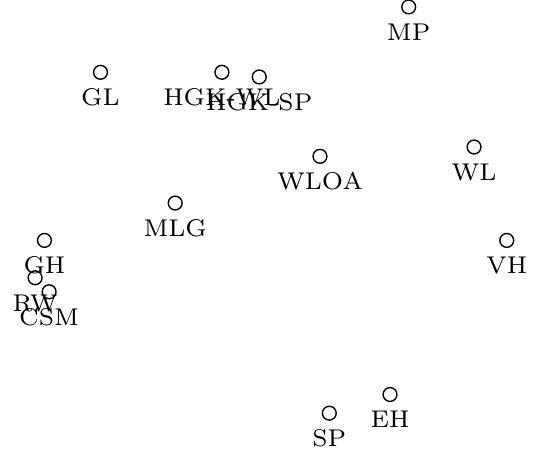}
  \else
    \begin{tikzpicture}
    \pgfplotsset{%
      every node near coord/.append style = {
        anchor = north,
        font   = \scriptsize,
        yshift = -1pt,
      }
    }
    \begin{axis}[
      axis lines  = none,
      xmin = -0.6, ymin = -0.6,
      xmax =  0.6, ymax =  0.6,
      unit vector ratio* = 1 1 1,
    ]
      \addplot[%
        mark = o,
        scatter,
        scatter/use mapped color = {
          draw = black,
          fill = none,
        },
        only marks,
      ] table {Data/Kernel_predictions_MDS.txt};
      \addplot[%
        nodes near coords, only marks, mark = none,
        point meta = explicit symbolic] table[meta=kernel]
        {Data/Kernel_predictions_MDS.txt};
    \end{axis}
  \end{tikzpicture}
  \fi
  \caption{%
    An embedding of the graph kernels of this section in terms of their
    actual predictions. Distances in the plot correspond to how similar
    the prediction profile of two kernels is.
  }
  \label{fig:Kernel predictions embedding}
\end{figure}

Figure~\ref{fig:Kernel predictions embedding} depicts the resulting
embedding. The distances in this plot correspond to the differences
in predictions between the individual graph kernels.
Contrary to our intuition, there are no direct ``obvious'' groups in the
embedding. While \mbox{HGK-WL} and \mbox{HGK-SP} are put relatively
close to each other, there is no group of methods based on
Weisfeiler--Lehman propagations, for example.
Nevertheless, there are some noteworthy aspects in this plot: we
observe that the message passing kernel~(MP), which employs additional
approximation schemes, is predicting labels differently than other
kernels. 

All in all, this plot is \emph{not} sufficient to ``pick'' a graph
kernel to use, though, so we require a more involved method.

\subsection{Grouping based on hierarchical clustering}

In a manner similar to the grouping presented in Section~\ref{sec:Grouping by predictions},
we can also employ hierarchical clustering to obtain different views on
all graph kernels.
Such a hierarchy will be useful because it makes it easier to assess the
differences in graph kernels at multiple levels. In the following, we
will describe two different hierarchies, one based on
the \emph{accuracies}, the other one based on the predicted labels.

\subsubsection{Clustering based on AUROC values}
\begin{figure}[tb]
  \centering
  \subcaptionbox{Clustering based on AUROCs\label{sfig:hclust AUROCs}}{%
    \includegraphics[width=0.50\linewidth]{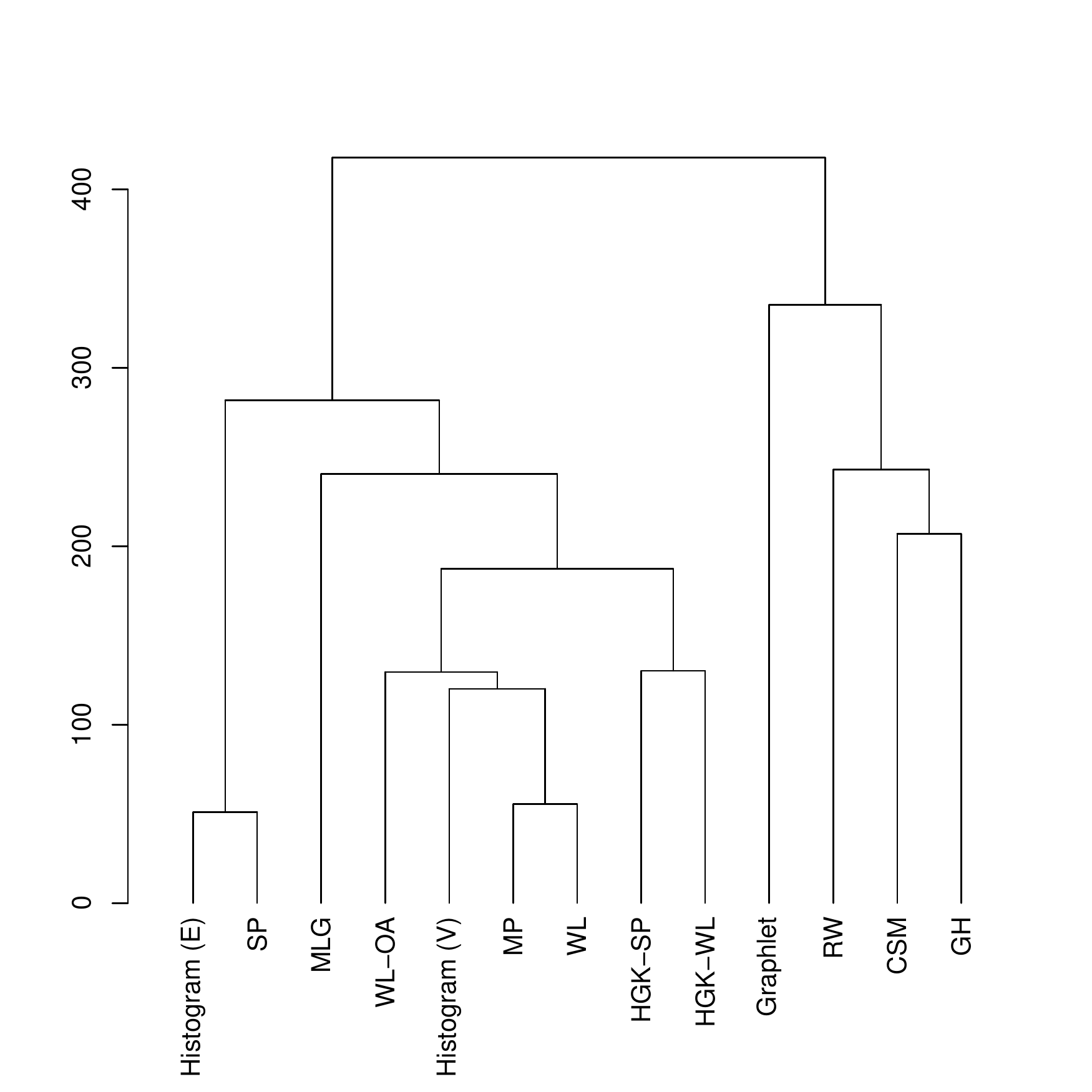}%
  }%
  \subcaptionbox{Clustering based on Hamming distance\label{sfig:hclust Hamming}}{%
    \includegraphics[width=0.50\linewidth]{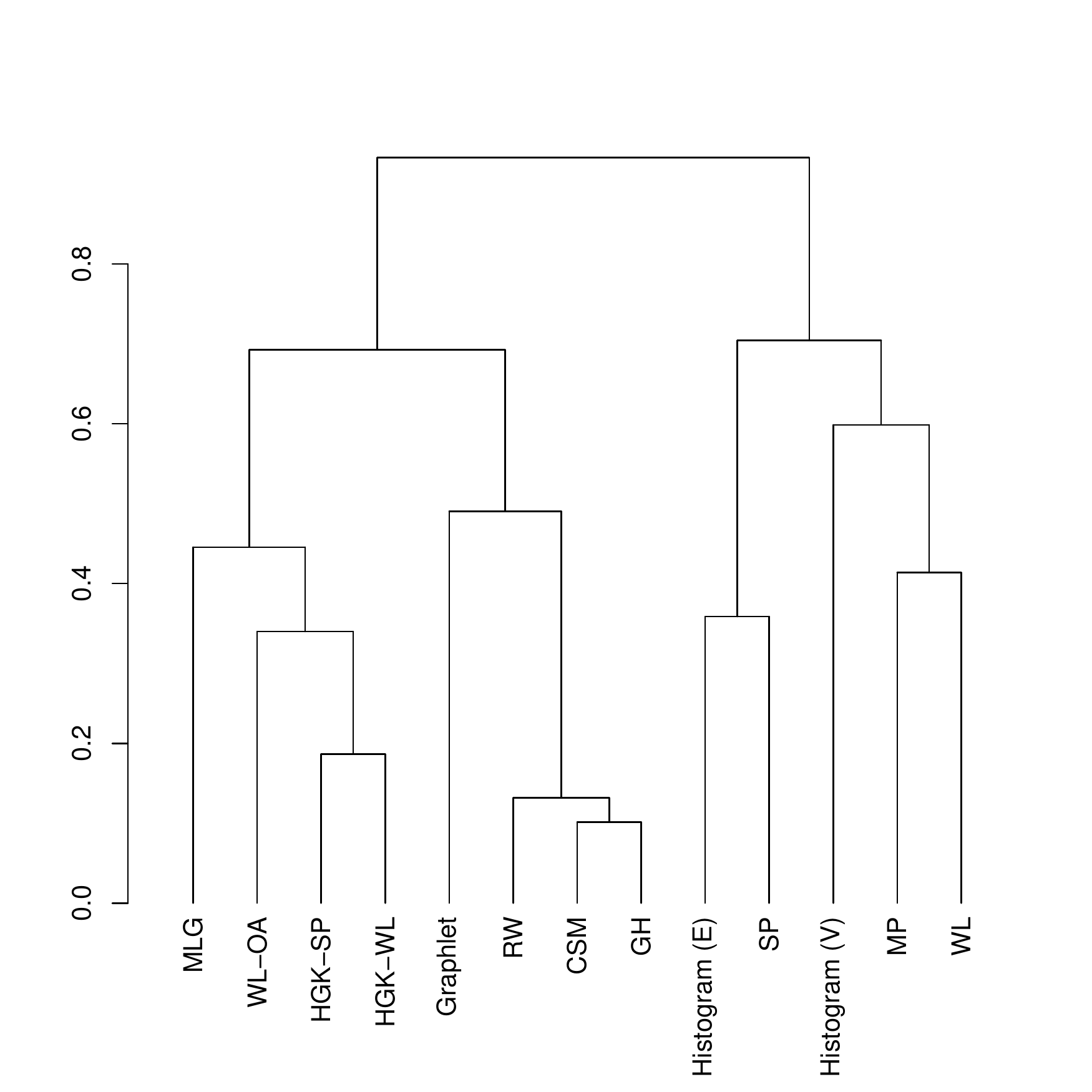}%
  }
  \caption{%
    Dendrograms obtained from performing hierarchical clustering on
    the graph kernels and their predictions.
  }
  \label{fig:Dendrograms}
\end{figure}

%
For this clustering, we treat each graph kernel as a ``sample'' of data
set and each benchmark data set as a ``feature'', yielding an $n \times
m$ matrix with $n$ graph kernels in the rows and $m$ data sets in the
columns. We take each entry of the matrix to be an AUROC such that the
values are comparable across multiple data sets.
Calculating the pairwise Euclidean distance then results in an
$n \times n$ matrix, which we can cluster using \emph{complete linkage
hierarchical clustering}~\citep{Muellner11}.
Figure~\ref{sfig:hclust AUROCs} depicts the resulting dendrogram.
Interestingly, Weisfeiler--Lehman approaches are clustered together
here; there is a cluster containing
\mbox{WL-OA}, \mbox{MP}, and \mbox{WL}, and \mbox{HGK-WL} suggesting that the performance
of these kernels across all data sets is extremely similar.
While this can be used to make a coarse pre-selection of a graph kernel
in practice, a more precise analysis would also include the actual
predictions of each method. Hence, we will now define a more-detailed
variant of this plot.

\subsubsection{Clustering based on Hamming distance}
%
To have a more in-depth analysis of the differences and similarities
between the graph kernels, we once again use the Hamming distance
between their individual predictions, as described in
Section~\ref{sec:Grouping by predictions}. Thus, two algorithms will be
considered \emph{similar} or \emph{related} if their predictions are
similar over all data sets and all folds.

Figure~\ref{sfig:hclust Hamming} shows the corresponding dendrogram~(which in
turn should correspond to Figure~\ref{fig:Kernel predictions embedding},
as this plot shows a two-dimensional embedding based on he same
distances). 
Here, \mbox{WL} and \mbox{MP} form one cluster, as do \mbox{HGK-SP} and
\mbox{HGK-WL}. However, the higher level clusters are not as
informative.

\section{Choosing a graph kernel}\label{sec:Choosing a graph kernel}

Having analysed graph kernels and the benchmark data sets at length and
under different perspectives, we now provide guidelines for choosing
a graph kernel \emph{in practice}.
The recommendations we give in this section are informed mainly by the
per-type ranking shown in Table~\ref{tab:Breakdown mean accuracy
ranking}, p.~\pageref{tab:Breakdown mean accuracy ranking}, as well as
on the different groupings we developed in Section~\ref{sec:Grouping
graph kernels}, p.~\pageref{sec:Grouping graph kernels}.
We distinguish two different scenarios: first, a scenario with unlimited
computational resources; second, a scenario in which computational
resources are more limited. Since the graph kernels are
implemented using different programming languages, we cannot provide
a fair comparison in terms of runtime. We can, however, discuss the
extent to which parameter tuning is required in order to obtain good
predictive performance.

\tikzstyle{IO} = [
  trapezium,
  trapezium left angle  =  70,
  trapezium right angle = 110,
  minimum width         = 2cm,
  minimum height        = 1cm,
  align                 = center,
  draw                  = black,
  text centered,
]

\tikzstyle{process} = [
  rectangle,
  minimum width  = 2.00cm,
  minimum height = 0.75cm,
  draw           = black,
  align          = center,
  text centered,
]

\tikzstyle{decision} = [
  diamond,
  minimum width  = 2cm,
  minimum height = 1cm,
  draw           = black,
  align          = center,
  text centered,
]

\tikzstyle{arrow} = [
  thick,    
  ->,       
  >=stealth 
]

\begin{figure}[tp]
  \centering
  \iffinal
  \includegraphics{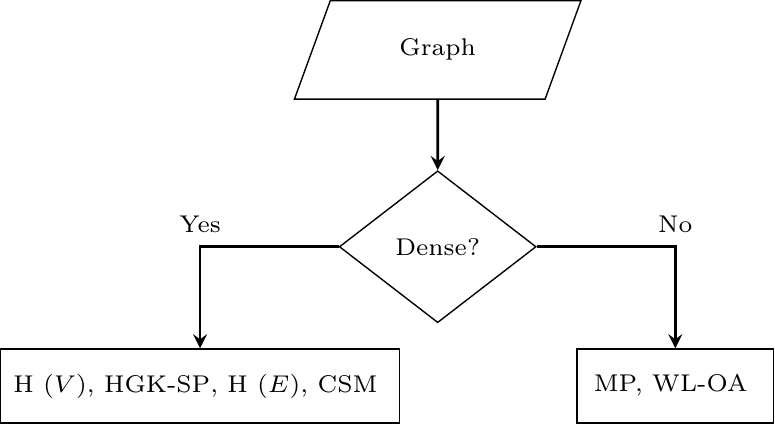}
  \else
  \tikzstyle{every node} = [
    font = \scriptsize
  ]
  \begin{tikzpicture}[node distance = 2cm]
    \node[IO]                         (input)   {Graph};
    \node[decision, below of = input] (density) {Dense?};
    \node[process, below left of  = density, xshift=-1cm] (dense) {%
        H~($\vertices$), \mbox{HGK-SP}, H~($\edges$), \mbox{CSM}
    };
    \node[process, below right of = density, xshift= 1cm] (sparse) {%
      MP, WL-OA
    };

    \draw[arrow] (input) -- (density);
    \draw[arrow] (density.west) -| node[anchor=south] {Yes} (dense.north);
    \draw[arrow] (density.east) -| node[anchor=south] {No} (sparse.north);
  \end{tikzpicture}
  \fi
  \caption{%
    Selection process for a graph kernel in a scenario with no
    restrictions on computational resources.
  }
  \label{fig:Flowchart 1}
\end{figure}
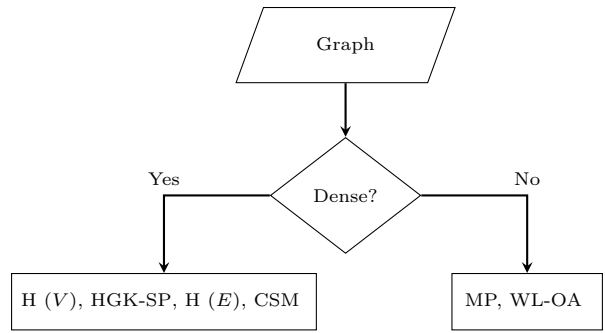
\subsection{Scenario 1: unlimited computational resources}
%
In this scenario, runtime and memory does not matter---we assume that
one is only interested in classification performance. Suitable
candidates are thus those graph kernels that outperform all other
kernels on a specific type of data set.
Using the ranking from Table~\ref{tab:Breakdown mean accuracy ranking} on
p.~\pageref{tab:Breakdown mean accuracy ranking}, we suggest to follow the flowchart
in Figure~\ref{fig:Flowchart 1} in order to choose a suitable graph
kernel. The only decision a user needs to take here is to check the
\emph{density} of the graphs data set beforehand. We purposefully leave
the definition of what constitutes a dense graph open---a
straightforward threshold would be to define graphs with a density of $>
0.5$ to be dense, as opposed to sparse. For complete graphs, \ie\ graphs
with a density of $1$, all schemes based on a Weisfeiler--Lehman propagation of label
information are not applicable any more because all neighbourhoods will
be essentially the same. In such a case, histogram-based kernels or
a Hash graph kernel based on shortest paths can be more suitable. The
former group of kernels has computational advantages---which are not
relevant in this scenario---while the latter graph kernel has the
advantage of being flexible in terms of how to use label or attribute
information~(recall that the shortest-path graph kernel family can
handle arbitrary node/edge information; since in complete graphs, the shortest path
between two vertices typically degenerates to an edge, this remains
computationally feasible). 

%
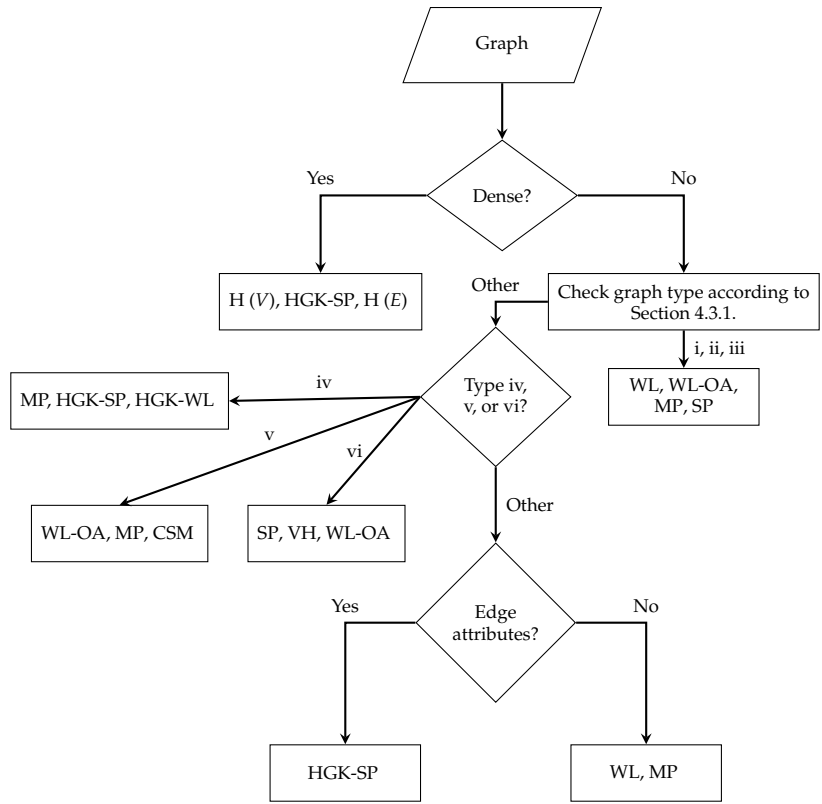
\begin{figure}[t]
  \centering
  \tikzstyle{every node} = [
    font = \scriptsize
  ]
  \begin{tikzpicture}[node distance = 2cm]
    \node[IO]                         (input)   {Graph};
    \node[decision, below of = input] (density) {Dense?};
    \node[process, below left of  = density, xshift=-1cm] (dense) {%
      H~($\vertices$), \mbox{HGK-SP}, H~($\edges$)
    };
    \node[process, below right of = density, xshift= 1cm] (sparse) {%
      Check graph type according to\\
      Section~4.3.1.
    };

    \node[process, below = 0.5cm of sparse] (123) {%
        WL, \mbox{WL-OA},\\ \mbox{MP}, {SP}
    };

    \node[decision, left of = 123, xshift=-0.50cm] (select) {%
      Type~iv,\\
      v, or vi?
    };

    \node[process, below = 0.5cm of select, xshift = -5.0cm] (5) {%
      \mbox{WL-OA}, \mbox{MP}, \mbox{CSM}
    };

    \node[process, above = 1.0cm of 5] (4) {%
      MP, \mbox{HGK-SP}, \mbox{HGK-WL}
    };

    \node[process, right of = 5, xshift = 0.75cm] (6) {%
      \mbox{SP}, \mbox{VH}, \mbox{WL-OA}
    };

    \node[decision, below = 1cm of select] (other) {%
      Edge\\attributes?
    };

    \node[process, below of = other, xshift = -2cm] (edges) {%
      HGK-SP
    };

    \node[process, below of = other, xshift =  2cm] (noedges) {%
      WL, MP
    };

    \draw[arrow] (input) -- (density);
    \draw[arrow] (density.west) -| node[anchor=south] {Yes} (dense.north);
    \draw[arrow] (density.east) -| node[anchor=south] {No} (sparse.north);

    \draw[arrow] (sparse.south) -- node[anchor=west]  {i, ii, iii} (123.north);
    \draw[arrow] (sparse.west)  -| node[anchor=south] {Other} (select.north);

    \draw[arrow] (select.west) -- node[anchor=south]  {iv} (4.east);
    \draw[arrow] (select.west) -- node[anchor=south]  {v}  (5.north);
    \draw[arrow] (select.west) -- node[anchor=east ]  {vi} (6.north);

    \draw[arrow] (select.south) -- node[anchor=west] {Other} (other.north);

    \draw[arrow] (other.west) -| node[anchor=south] {Yes} (edges.north);
    \draw[arrow] (other.east) -| node[anchor=south] {No}  (noedges.north);

  \end{tikzpicture}
  \caption{%
    Selection process for a graph kernel in a scenario where only
    restricted computational resources are available.
  }
  \label{fig:Flowchart 2}
\end{figure}
%
\subsection{Scenario 2: limited computational resources}
%
In this scenario, computational resources are \emph{limited}, either in
the sense of having large-scale data sets, limited storage, or limited
CPU time. In this case, the selection process has to be a little bit
more measured and also depend on the data set type.
Figure~\ref{fig:Flowchart 2} presents the flowchart that we recommend to
follow in this situation. It is informed by Table~\ref{tab:Breakdown
mean accuracy ranking} on
p.~\pageref{tab:Breakdown mean accuracy ranking}, but also by the clustering of graph
kernels in Figure~\ref{sfig:hclust Hamming}.

This flowchart is more detailed because it takes into the different data
set types as well as potential runtime requirements and parameter
choices. Whenever multiple kernels are listed in a field of the chart,
our ordering goes from kernels that feature few hyperparameters to
kernels that feature more. This is motivated by the observation that
more hyperparameters require more complex search strategies, which may
quickly become infeasible in case resources are restricted.

\section{Conclusion}

This chapter presented a comprehensive empirical analysis of a variety of graph
kernels. We discussed the difficulty and suitability of the benchmark
data sets and gave recommendations about their usage. A comprehensive
comparison of predictive performance showed that graph kernels that are
based on some form of the Weisfeiler--Lehman algorithm are among the
best-performing graph kernels. After discussing several strategies for
grouping them, either based on statistical measures or based on
clustering, we closed this chapter with suggestions on how to choose
a graph kernel for handling new data sets. In the next and final
chapter, we will discuss novel research directions for this field but
also some necessary actions that result from our analyses.

\chapter{Discussion \& future directions}\label{chap:Future}

In this final section, we will discuss actionable items that
arise from the preceding analyses. We will also describe future
directions, extensions, and emerging topics in the field of graph
kernels.

\section{Current limitations in graph kernels research}

As our discussion in the preceding sections demonstrated, the current
benchmark data sets and methods suffer from several limitations. We consider the
crucial issues to be
\begin{inparaenum}[(i)]
  \item limitations of benchmark data sets, 
  \item challenges in kernel usage,
  \item reproducibility and software availability, and
  \item scalability.
\end{inparaenum}
In the following, we will briefly comment on each of these issues and
give our recommendations on how to address them.

\subsection{Limitations of benchmark data sets} 

Overall, as emerged from our experimental evaluation, we consider many of the
current benchmark data sets to be insufficient to assess whether graph kernels
possess expressivity.
First, the lack of information in their topological structure is a major
issue. We already observed that the topological structure of graphs only
partially contributes to the information gain, as we can infer from the
histogram kernels being good predictors~(see Section~\ref{sec:Histogram
kernels}).
This aspect has been previously discussed by \citet{Sugiyama15}, who showed
that histograms of node and edge labels, combined with a Gaussian kernel, can
be extremely competitive.

Second, tightly linked to the first point, our analysis also uncovered the
issue of missing \emph{provenance information}, \ie\ information about the
construction process, for the data sets: some of the graph data sets
contain ``derived'' or ``constructed'' graphs---graphs that involved
user-defined choices in their creation. The repository does not provide
sufficient provenance information to understand or reproduce these graphs. For
example, if a data set of sparse graphs has been created by thresholding a set
of dense graphs, information about this thresholding should be added to the
data set or, even better, the \emph{original} data set should be provided as
well. This will make it possible to develop graph kernels that take structural
information at multiple scales into account---but it will also make the
creation process of the data sets more transparent. In light of the performance
of histogram kernels, which we analysed in Section~\ref{sec:Histogram kernels}
on p.~\ref{sec:Histogram kernels}, we conjecture that the creation process of
the data sets contributes to their performance, which is sometimes surprisingly
competitive, as we would expect graph data sets to require structural
information for correct classification.

Third, another limitation of the current benchmark data sets, analysed in
Section~\ref{sec:Difficulty}, is the general lack of ``difficulty''. Our
discussion suggested that the current graph benchmark repository contains data
sets that can and should be excluded due to being \emph{too easy}, \ie\
classifiable by a simple vertex histogram kernel, or \emph{already solvable}.
We recall that a data set was considered as already solvable if the overlap of
correctly-classified graphs among the different methods was sufficiently large.
Intuitively, this implies that the achievable performance has already been
reached, and the remaining non-classifiable graphs are either noisy or
outliers, and in conclusion too ``tricky'' to classify.

Fourth, current benchmark data sets are not sufficiently \emph{diverse}. In
Section~\ref{sec:Breakdown} on p.~\pageref{sec:Breakdown}, we partitioned the
benchmark repository into six different types. Considering the presence or
absence of the four individual types of information~(node labels, edge labels,
node attributes, and edge attributes) to be binary variables, there are 16
possible data set types---with the repository containing no examples for some
of them. While it is theoretically possible to remove or mask certain types of
features, this is not necessarily the same as lacking a given type of feature,
such as a node label; this is particularly problematic because some types of
graph features can be derived from another type, thus potentially leaking
information.  This lack of diversity is also expressed in other ways, such as
the density of data sets---see Figure~\ref{fig:Density distribution} on
p.~\pageref{fig:Density distribution}---and their size.


\paragraph{Recommendation}


Clearly, there is a strong need for new graph benchmark data sets, given the
lack of topological information, diversity and difficulty in the current ones.
Two strategies appear to be paramount to achieve this: first, to develop new methods
for generating graph classification benchmark data sets that overcome these
shortcomings. Second, to define and explore new application domains of graph
kernels that result in a larger variety of real-world data sets for graph
classification~(see also Section~\ref{sec:New application domains}).
When including new---either synthetic or real-world---data sets into the
collection of benchmark data sets, provenance information should be provided,
including
\begin{inparaenum}[(i)]
\item information about parameters~(if any) that were used to create
    the data set, as well as
  \item choices in pre-selecting edges or vertices.
\end{inparaenum}
This information will help to decide whether one can expect the topology of
a graph to positively influence classification in a data set, and whether
a particular graph kernel has an advantage because its features mirror
important parameters in the data set generation.

While a priori it is not possible to define the difficulty of a \emph{new}
graph data set, being able to assess it is crucial in order to choose the most
suitable graph kernel approach. It is particularly relevant to evaluate and
compare the performance of a vertex histogram kernel to other---more
complex---methods. On the one hand, if the vertex histogram kernel outperforms
the others, one might conclude that the data set is \emph{too easy} and no information
about graph topology is required to classify it. On the other hand, if most of
the methods perform the same or similarly, one should precisely analyse
\emph{which} graphs are classified incorrectly by each kernel.
Following the logical flow analysis conducted in Section~\ref{sec:Difficulty},
one can then conclude whether to consider the data as already ``solved'', and
therefore not requiring the development of additional more advanced methods for
it.

Once these issues are addressed, a future wealth of diverse graph benchmark
data sets can then be utilised to conduct a fairer empirical comparison of
graph kernels, and for a targeted design of novel graph kernels for particular
types of graphs, \eg\ graphs with high-dimensional node attributes. 
We support the recent efforts of \citet{hu2020ogb} and \citet{Morris2020}, who
 each created repositories of new benchmark data sets. The latter is
in fact an updated version of the repository of data sets we used in this review, and now
encompasses over 120 different benchmark data sets.
%

%
%




\subsection{Challenges in graph kernels usage}

Another challenge in graph kernel research arises from how graph kernels
are compared. With more and more kernels being defined in the literature, more and more
empirical comparisons between them will be conducted. It is important to be
aware of stumbling blocks in these comparisons.

First, one should be aware that most graph kernels do not define \emph{one}
single way of comparing graphs to each other, but rather a \emph{family} of
methods. This property is often under-utilised in comparisons or applications,
leading to an unnecessarily low predictive performance that is not competitive.
For instance, there is a whole family of random walk kernels, which differ by
the way they weigh steps in the walks: geometric random walk kernels, for
example, use exponentially decaying weights for subsequent steps, whereas
random walks of a fixed length $k$ give the \emph{same} weight to all $k$
steps. This difference can have drastic effects: in fact, \citet{Sugiyama15} showed
that the decaying factor in the geometric random walk kernel often has to be
chosen so small that it degenerates to a simple edge comparison between two
graphs, resulting in poor classification accuracy. Still, the fixed-length
random walk is reported to achieve results that are competitive with the state
of the art. It is therefore important to not falsely generalise empirical
findings across all instances of family of graph kernels, but to select
a competitive instance.

Second, our empirical results indicate that node and edge label histograms
information is extremely beneficial for good classification. Several graph
kernels can capture this type of information: for example, Weisfeiler--Lehman
kernels with $h = 0$ iterations are already equivalent to a comparison of node
histograms. Nevertheless, some publications inadvertently use a parameter
grid that excludes $h = 0$, thereby preventing only original label information from
being used. 
%
As another example, graphlet kernels that consider graphlets of size~1 and~2
count nodes and edges, respectively, whereas fixed-length random walk kernels with
$k=1$ count edges in a graph.
A way to severely hurt the performance of a graph kernel is to \emph{exclude}
these simple graph properties from kernel computation by not considering such
simple kernel instances in the hyperparameter search of the kernels.

\paragraph{Recommendation} Kernel choice and hyperparameter tuning should be
performed for all the competitive methods, in order to guarantee the best
performance of the state-of-the-art methods. Hyperparameters should be chosen
such that they also allow for inclusion of simple graph statistics such as node
and edge histograms. Moreover, the choice of kernel as well as all
hyperparameters should be clearly reported, at least in the appendix of
published manuscripts, and ideally be reproducible with published code.
Furthermore, as \citet{Sugiyama15} pointed out, we again emphasise the crucial
importance of using histogram kernels as baselines when developing and
benchmarking new graph kernels. 

\subsection{Reproducibility and software availability}

Aside from an appropriate choice of graph kernel comparison partners, the growing number of empirical graph kernel comparisons 
above also creates an enormous 
need for reproducibility and, in particular, necessitates the availability of
open source software to reproduce results.

\paragraph{Reproducibility} The lack of reproducibility in the graph kernel
community is largely due to the lack of a ``common agreement'' concerning the
experimental setup and its parameters, such as the number of folds, or the
number of splits to employ for a given data set. A lack of code and information
about the experimental setup may cause the accuracy on a certain data set to
differ from one publication to another one, thereby leading to inconclusive
results and, in the worst case, incorrect claims about kernel performance.
The comparability of graph kernels is further exacerbated by a non-uniform
selection of benchmark data sets when it comes to evaluating prediction
performance.

\paragraph{Software availability} There are also cases in which the code for
kernel computation is not published; even if it is available, there is still
the issue of heterogeneity between programming languages. To address this
problem, public software packages that facilitate the application and
implementation of graph kernels in popular and uniform coding languages have
recently been developed.

The \texttt{graphkernels}~\citep{Sugiyama17} package is a Python and R wrapper
that relies on a C++ backend implementation. The advantage of C++ can certainly
be found in the high speed and the efficiency of the code. Furthermore, the
user-friendly interface permits computing all the individual graph kernel
matrices with similar steps. The analogous interface between Python and R contributes
to the versatility of the two languages.

The \texttt{GraKeL}~\citep{grakel2018} package was entirely developed in
Python, is compatible with \texttt{scikit-learn}, and exploits the
\texttt{Cython} extension to benefit from a fast implementation in~C. At
present, \texttt{GraKel} supports a larger spectrum of graph kernel methods
than \texttt{graphkernels}. Its compatibility with the popular
\texttt{scikit-learn} library simplifies the integration into a classification
pipeline. 

\paragraph{Recommendation} We strongly encourage researchers to always provide
code as well as pre-compiled data set splits when publishing a new graph
kernel. It is crucial to also report experimental setup information for the
competitor methods in order to guarantee a fair and complete assessment.
Furthermore, it would be extremely beneficial for the community to define
standard splits on the benchmark data sets, provide results with the existing
methods, and always use them when a new approach is developed. 
We welcome the recent efforts of \citet{hu2020ogb, Morris2020,
dwivedi2020benchmarkgnns} in this direction. 

\subsection{Scalability}

Lastly, scalability remains a key challenge in graph kernel computation. 
As Table~\ref{tab:Empirical performance} on p.~\pageref{tab:Empirical
performance} demonstrates, some graph kernels cannot be trained efficiently
even on a high-performance computing cluster architecture. While a lot of past
graph kernel research was motivated by the need to develop faster graph
kernels, there may still be room to find strategies how to speed up existing
kernels.

\paragraph{Recommendation}
We think that the community should continue to focus on computational
efficiency. In addition to parallelising some calculations, we suggest
investigating \emph{approximation strategies} to speed up the computation of
kernel matrices. Classical examples of this are the Nystr\"om
method~\citep{Nystroem30}, which was successfully used to speed up calculations
of the message passing graph kernel~\citep{Nikolentzos18}. Similarly, the use
of less restrictive, \ie\ \emph{non-perfect}, hashing functions was
instrumental in speeding up the family of hash graph kernels~\citep{Morris16}.
We also envision that progress could be made by employing probabilistic data
structures such as \emph{bloom filters}~\citep{Bloom70}. These data structures
could be used to replace traditional data structures such as \emph{sets} to
improve computational efficiency, at the expense of correctness in certain
queries, making the method approximative.


\section{Emerging topics and future challenges}

Despite these limitations, there are many different exciting new lines
of research. In this section, we elaborate on emerging topics from which the field of graph
kernels could benefit. Our discussion is structured as follows:
\begin{inparaenum}[(i)]
  \item we discuss the idea of building more complex graph kernels,  
  \item we outline the link between graph kernels and graph neural networks
    while paying particular attention to the Weisfeiler--Lehman
    framework and the theoretical link to the graph isomorphism problem, and
  \item we describe new application domains for graph kernels.
\end{inparaenum}

\subsection{Increased complexity for graph kernels}

\re{A natural path forward is to consider more complex
  graph kernels.  We will focus on a few initial
  efforts in this area, beginning with the idea of building
  \emph{hybrid graph kernels}, \ie\
    ensembles of graph kernels. We then describe the limitations of the popular \mbox{$\mathcal{R}$-convolution}
  framework and present ideas how to overcome them. 
  }

\subsubsection{Ensembles of graph kernels}

Given the wealth of graph kernels in the literature, an exciting question to explore is whether superior kernels could be built by combining existing ones. This could result in \emph{graph kernel ensembles} that are capable of exploiting different sets of structural elements of graphs, thereby surpassing any individual graph kernel on certain data sets.

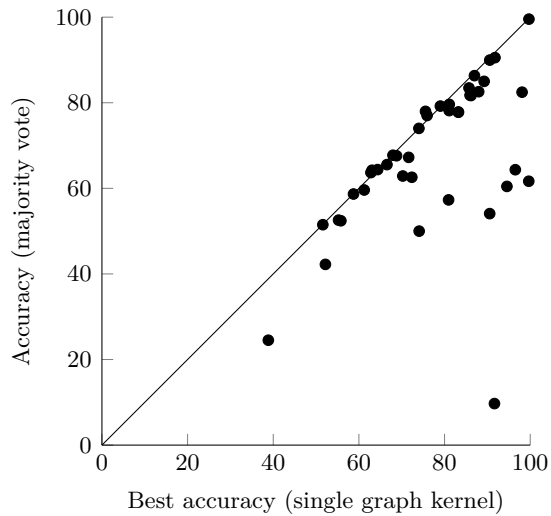
\begin{figure}[tbp]
  \centering
  \iffinal
    \includegraphics{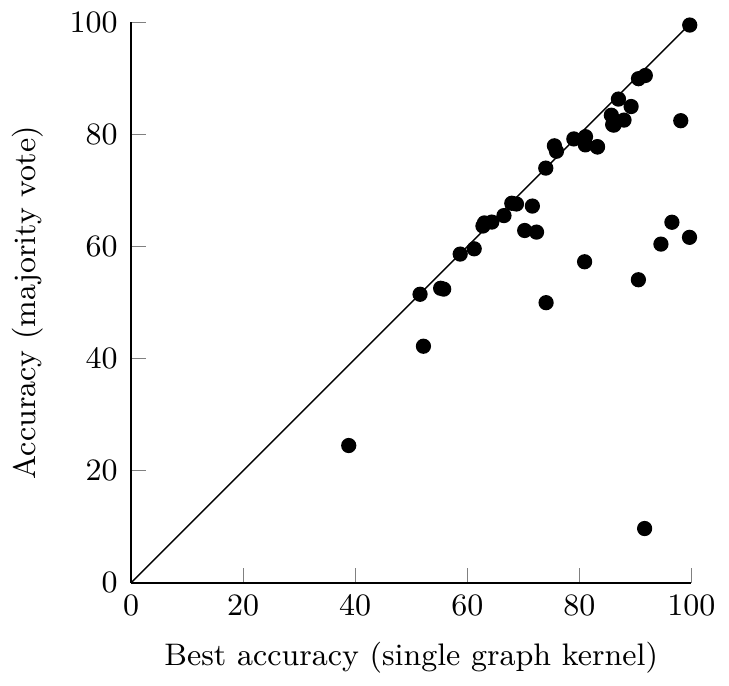}
  \else
  \tikzstyle{every node}=[font=\footnotesize]
  \begin{tikzpicture}
    \begin{axis}[%
      xmin               =   0.0,
      xmax               = 100.0,
      ymin               =   0.0,
      ymax               = 100.0,
      axis x line*       = bottom,
      axis y line*       = left,
      unit vector ratio* = 1 1 1,
      xlabel             = {Best accuracy~(single graph kernel)},
      ylabel             = {Accuracy~(majority vote)},
    ]
      \addplot[domain = {0:100}] {x};
      \addplot[only marks] table[col sep = comma, x = best, y = mean] {Data/Ensemble.csv};
    \end{axis}
  \end{tikzpicture}
  \fi
  \caption{%
    A comparison of the best performance of a single graph kernel and
    the ensemble performance obtained via a majority vote.
  }
  \label{fig:Ensemble performance}
\end{figure}

Such an endeavour is fraught with obstacles, though. As a simple
experiment, we use the results from the preceding chapter to create a simple graph kernel ensemble. Specifically, we collate all predictions of all graph kernels and use a majority vote to predict the label. Figure~\ref{fig:Ensemble performance} depicts the performance of this simplistic combination and compares its performance to that of the best graph kernel on the corresponding data set.
We observe that in most of the cases, the predictive performance of the ensemble graph kernel is lower than that of the best-performing individual one. In fact, there are only five data sets for which this ensemble kernel improves predictive performance, namely
\begin{inparaenum}[(i)]
  \item \texttt{COX2\_MD},
  \item \texttt{DD},
  \item \texttt{ENZYMES},
  \item \texttt{PROTEINS}, and
  \item \texttt{PROTEINS\_full}.
\end{inparaenum}
Next to the computational challenges inherent in any ensemble method, this experiment also demonstrates the difficulty of creating useful ensembles---more involved methods are required; in our simple experiment, the predictor does not benefit from the fact that different graph kernels are capable of capturing different characteristics of a data set. By only following the majority vote, the resulting ensemble cannot exploit the specialisation of individual graph kernels, as different features are not weighted according to their relevance to the task at hand.

One promising direction for future research is therefore the integration of confidence information; in the simplest case, such information could be used to predict according to the most \emph{confident} graph kernel.
However, it would also be possible to create a hierarchy of graph kernel predictors that are activated ``on demand'' whenever the confidence drops below a certain threshold. We also suggest that, according to the priority of the user, one can restrict the ensemble to only a subclass of kernels, such as propagation-based or Weisfeiler--Lehman based schemes.
Additionally, more complex ensemble approaches than a simple majority vote could be employed. Furthermore, we speculate that multiple kernel learning~\citep{lanckriet2004learning, sonnenburg2006large} could be used to effectively learn an optimal way to combine different substructures into a more powerful graph kernel, an avenue that has already begun to be explored~\citep{aiolli15}. Nevertheless, the major limitation of ensemble approaches is obviously their high computational complexity.

\subsubsection{Alternatives to the $\mathcal{R}$-convolution framework}

Another direction of graph kernel research reconsiders the foundation of
how most graph kernels are designed, with the goal of finding alternatives
that can improve performance.  Most graph kernels
have been developed based on a simple instance of the
$\mathcal{R}$-convolution framework~(see Section~\ref{sec:R-convolution
kernels}), which decomposes two structured objects into their sets of
substructures, to then aggregate the pairwise similarities of these
substructures via a na{\"i}ve sum or average. Recent efforts have been made
to overcome the limitations arising from this aggregation step, which
might possibly disregard complex interactions between substructures.
Originally, a kernel based on an optimal assignment of node labels was
proposed~\citet{Froehlich05}; while being empirically successful, this
kernel has been later shown to be non positive definite~\citep{Vert08}.
Recently,~\citet{Kriege16} extended these ideas by developing
a Weisfeiler--Lehman based optimal assignment kernel~(\mbox{WL-OA}).
Successively, \citet{Togninalli19} also proposed an extension of the
original Weisfeiler--Lehman kernel, which is especially designed for
continuously attributed graphs~(WWL). This method employs the
Wasserstein distance~\citep{villani08} to capture more complex
similarities between substructures, computing a graph level
representation from node features obtained via a multi iteration
Weisfeiler--Lehman inspired scheme. The \mbox{WL-OA} and WWL kernels for
graphs with categorical node labels have been shown to be positive
semi-definite; assessing the positive definiteness of WWL on continuous
node attributed graphs is still an open problem. These approaches open
the door to a new line of research, connecting the emerging field of
\emph{optimal transport}~\citep{villani08}, which has recently gained
considerable interest in the community, to graph kernels. Challenging
theoretical problems, such as assessing positive definiteness of
existing methods, and theoretical contributions, including the design of
new kernels based on optimal transport theory, will be of interest in
the future.

The field of topological data analysis, focusing on connectivity
properties of structured objects in general, has recently started to
demonstrate its capabilities in graph classification, constituting
a somewhat complementary view to existing methods. Topological
features, such as connected components and cycles, have shown their
capabilities for improving existing graph kernels~\citep{Rieck19}, but
they can also ``hold their own'' upon being combined with appropriate
machine learning architectures~\citep{Hofer17, Zhao19}. With recent work
establishing a framework for \emph{learning} topological descriptors in
an end-to-end fashion to improve classification performance~\citep{Hofer20},
we envision that this topic will be of increasing relevance in the
future.

We conclude this section by pointing out that the last step in kernel
computation on structured objects, \ie\ the aggregation of node
representations, is also a limitation in the field of graph neural
networks~\citep{xu18}, where it is commonly referred to as a \texttt{READOUT}
layer. Most GNNs approaches use a \texttt{mean}, \texttt{sum},
\texttt{max} or a combination of these functions to generate the
graph-level representations from the node features~(as obtained via
Equations~\ref{eq:GCN neighborhood} and~\ref{eq:GCN update}). Extensions
of the current scheme, based for instance on attention
mechanisms~\citep{Gilmer17}, pooling strategies~\citep{Ying18}, or
network architectures~\citep{schutt17}, have yet to be fully explored
and certainly represent an interesting direction to pursue in graph
neural networks and graph kernel research. We proceed now to a larger
discussion of graph neural networks, and explore their connection to
graph kernels.





\subsection{Link between graph kernels and graph neural networks}

Graph Neural Networks~(GNNs) have emerged in recent years and
established a successful line of research, achieving state-of-the-art
performance in both graph classification and regression tasks~(see
\citet{Wu19} for a recent survey). We will first provide a definition and brief
overview of GNNs, and will later discuss how this procedure is related
to the Weisfeiler--Lehman labelling scheme (see
Section~\ref{sec:Weisfeiler--Lehman kernel}). 

The main idea underlying GNNs is to propagate the initial feature
representation of the nodes and edges across the graph, thereby
exploiting a multi-iterative scheme that at each step updates the
current state by looking at the neighbourhood information of a node.
In the following, the term \emph{node feature} refers to the node
representation at a given iteration, which can be either the original
node label or attribute, or an update of it obtained after one or more
steps.

Graph Neural Networks employ an affine transformation followed by
a pointwise non-linear activation function to update
the node or edge information, which encourages \emph{smooth}
information propagation on the graph. We will follow the structure and
notation introduced in~\citet{xu18}. Given a graph $G = (V,E)$ and a set
of node features $X_{G}\in \real^{|V|\times p}$, most GNNs employ
a neighbourhood aggregation strategy. Such a strategy updates the node
feature of the current iteration by \emph{aggregating} the
representations of the neighbours using, for example, the calculation of
a mean. The aggregation function is crucial because it makes all learned
representations invariant with respect to permutations; a GNN is
therefore impervious to changing the indices of nodes.
Let $x_{v}^{0}\in\real^{p}$ be the initial node feature~(\ie\ either
attributes or a label) of node $v$ in graph $G$. We recursively define 
\begin{align}
  z_{v}^{(h)} &:= \texttt{AGGREGATE}\left\{
    x_{v'}^{(h-1)} \mid \vertex' \in \neighbourhood\left(v\right)
  \right\},\label{eq:GCN neighborhood}
  \shortintertext{and update the node feature of $v$ as}
    x_{v}^{(h)} &:= \texttt{COMBINE} \Big(x_{v}^{(h-1)},z_{v}^{(h)}. \Big)\label{eq:GCN update}
\end{align}
for up to $h_{\text{max}}$ rounds of propagation.
Finally, the collection of all node feature vectors obtained during the
iteration process can be summarised into a \emph{single} vectorial
representation for the graph by
\begin{equation}
    \phi(G) = \texttt{READOUT}\left(\left\{x_{v}^{(h)} \mid v \in V, h=\{0,\hdots,h_{\text{max}}\} \right\} \right).
\end{equation}
Multiple possibilities exist for defining the \texttt{COMBINE},
\texttt{AGGREGATE}, and \texttt{READOUT} functions, leading to
different approaches as described by \citet{duvenaud15}, \citet{kipf17},
and \citet{hamilton17}, for example.

\paragraph{Weisfeiler--Lehman versus GNNs}
We note that the idea behind GNNs follows the Weisfeiler--Lehman
propagation scheme, as highlighted by~\citet{xu18}. 
Recalling the terminology and notation introduced in
Section~\ref{sec:Weisfeiler--Lehman kernel}, we defined the
Weisfeiler--Lehman node feature update as
\begin{equation}
  \Vlabel\left(\vertex\right)_{\text{WL}}^{(h+1)} := f\left( \left(\Vlabel\left(\vertex\right)_{\text{WL}}^{(h)}, \left\{
    \Vlabel\left(\vertex'\right)_{\text{WL}}^{(h)} \mid \vertex' \in \neighbourhood\left(v\right)
  \right\} \right)\right),
  \label{eq:WL relabelling2}
\end{equation}
where $f(\cdot)$ represents a \emph{perfect hashing} scheme that
uniquely maps tuples formed by the node label of the current vertex,
$\Vlabel\left(\vertex\right)_{\text{WL}}^{(h)}$, and the multiset of
node labels of all neighbours of the current vertex, $\left\{
\Vlabel\left(\vertex'\right)_{\text{WL}}^{(h)} \mid \vertex' \in
\neighbourhood\left(v\right) \right\}$, to a new categorical label.
Denoting by $\phi_{\text{V}}(v)_{\text{WL}}^{(h)}$ the one-hot vector
corresponding to $\Vlabel\left(\vertex\right)_{\text{WL}}^{(h)}$, the
simplest instance of the Weisfeiler-Lehman framework would then
aggregate these node representations across all propagation steps
$h=0,1,\hdots,h_{\text{max}}$ as
\begin{equation} \phi_{\text{WL}}(G) = \mathrm{concat}\left(\sum_{v \in
  V}\Vlabel\left(\vertex\right)_{\text{WL}}^{(0)}, \hdots, \sum_{v \in
  V}\Vlabel\left(\vertex\right)_{\text{WL}}^{(h_{\text{max}})}\right),
\end{equation}
to obtain a single vectorial embedding for the graph.

Thus, the main difference between the Weisfeiler--Lehman scheme and GNN
approaches boils down to the definition of the \texttt{COMBINE},
\texttt{AGGREGATE}, and \texttt{READOUT} functions. Perhaps the most crucial
distinction is that while graph neural networks typically implement the
\linebreak \texttt{AGGREGATE} and \texttt{COMBINE} steps as smooth functions with
learnable parameters, the Weisfeiler--Lehman uses instead a perfect
hash. Theoretically, despite being virtually parameter-free, the
Weisfeiler--Lehman implementation of information propagation is at least
as expressive as that of graph neural networks due to the use of perfect
hashing. However, it lacks the ability to be ``tuned'' end-to-end to
a specific task if a sufficiently large data set is available.
Similarly, the \texttt{READOUT} phase of the simplest instance of the
Weisfeiler-Lehman framework can be understood as forming
an~(unnormalised) node label histogram for each step $h$ and then
concatenating these histograms. In contrast, graph neural networks use
a wide range of alternatives, some being virtually equivalent to that of
the Weisfeiler-Lehman framework, while others implement complex smooth
functions with a large number of learnable parameters. 
%

\subsubsection{Link to the graph isomorphism problem}

Having established the connection between the Wesifeiler--Lehman
\linebreak scheme
and GNNs, we now want to understand how expressive GNNs can be. To do
so, we will use the return to the graph isomorphism problem introduced
in Section~\ref{sec:Graph isomorphism problem}. 
In general, determining whether two graphs are isomorphic is so far
a problem that is not known to be solvable in polynomial time. In
practice, the WL test works for the majority of all cases, though it is
possible to find---or construct---pairs and families of non-isomorphic graphs
that the WL test cannot distinguish. Recently, it has been investigated whether graph neural
networks are more powerful than the WL test and can succeed in
distinguishing these graphs~\citep{xu18}.
\citet{xu18} concluded that GNNs are \emph{at most} as powerful as the
WL test in distinguishing graph structures. The authors further
commented that certain requirements in the scheme of GNNs need to be
satisfied in order to \emph{achieve} such power, and they propose
a novel Graph Isomorphism Network~(GIN) architecture, which is capable
of reaching a discriminative power comparable to that of the
Weisfeiler--Lehman isomorphism test for distinguishing graph
structures. 
\citet{xu18} argue that the \texttt{READOUT} function in a GNN needs to be
injective, to guarantee that two \emph{non-isomorphic} graphs will be
mapped to different graph embeddings, thereby being correctly identified
as \emph{non-isomorphic}. 

\subsubsection{Learning node representations with GNNs \& graph kernels}
%
Finally, we would like to highlight an additional use case for GNNs and
graph kernels which is of increasing interest to the community: learning
representations of nodes. While this review has been centered primarily around graph
classification, many graph neural networks are often used for the purpose 
of node classification~\citep{kipf17} or the related task of learning representations of
nodes in a graph~\citep{dai2016discriminative, hamilton2017representation}. 
While the graph kernels we described have been primarily used for classification or
regression tasks, for most of the existing approaches, an
explicit feature vector representation of a graph can be derived, or
an approximation of it can be computed~\citep{kriege2019}.
Due to most current graph kernels being instances of the
$\mathcal{R}$-convolution framework, in which a graph is represented as
a set of nodes, graph kernels indirectly also provide a vectorial
representation of each node in the graph. While there is long-standing
interest in kernels between nodes in one
graph~\citep{kondor2002diffusion,SmoKon03},
these efficient-to-compute node representations based on graph kernels have
not been studied in any detail. At the very least, they will offer
a baseline that node kernel and deep learning approaches need to improve over in order
to prove the merits of their representation learning.   
\paragraph{} As we've seen in this section, graph neural networks have emerged alongside graph kernels as
a state-of-the-art approach to solve graph classification tasks. We
conclude by highlighting an exciting line of future research that
explores the relative
benefits of graph kernels and GNNs and exploits them in hybrid
approaches.  Initial research~\citep{Nikolentzos18a, Du19} demonstrated
that such \emph{hybrid} approaches, which combine the ``best of both
worlds'', can indeed achieve good predictive performance. \re{\citet{Du19},
for instance, introduce the \emph{graph neural tangent kernel}, which
under certain assumptions can be shown to be equivalent to an infinitely wide neural
network trained by gradient descent. This therefore promises to have the expansive expressivity of a neural network,
while still maintaining the benefits of a convex optimization objective.
Given the distinct strengths of graph kernels and graph neural
networks, developing methods that can fuse components of the two
presents an promising new direction in the field of graph
classification.
}

\subsection{New application domains}\label{sec:New application domains}

While GNNs represent a new suite of methods applicable to the task of
graph classification, another promising direction for future research is the exploration of
new application domains. Structural biology and chemoinformatics
will remain a primary application domain of graph kernels and graph
learning, but we predict that emerging application domains will increase the variety of the
data sets and the number and type of learning tasks on graphs.
In light of the issues that we discussed in the preceding sections, we
consider multiple domains to be promising for the future, particularly
within new medical and mathematical applications.
%
%
Each of these fields, which we will subsequently discuss, has their own
idiosyncratic data set types, which will enrich future research.

\subsubsection{Medical applications}

One of the most promising new application domains is within medical
applications. We will now introduce two specific areas where graph-based
approaches are starting to take hold, namely with 
\begin{inparaenum}[(i)]
  \item electronic health records and
  \item and brain connectivity networks.
\end{inparaenum}
%
\paragraph{Electronic health records}
%
Electronic health records refer to the ensemble of all records of
a patient in a hospital. Their multi-modal nature makes them hard
to use in classification scenarios.
Recent advances in machine learning show that the inclusion of
structural, \ie\ \emph{graphical}, information can be used for medical
purposes, such as mortality or medications prediction~\citep{Choi19}. Similarly, there are
ambitious projects to learn knowledge graphs from such
records~\citep{Rotmensch17}.
The future of graph kernels research should embrace this domain because
of its challenges~(large, multi-modal, noisy graphs) and its potential
to improve patient welfare.

\paragraph{Brain connectivity networks}
%
In a similar fashion, magnetic resonance imaging~(MRI) data have started
to become ubiquitous over recent years and various analysis methods have
been proposed.  MRI data can be subject to thresholding~(representing an
uncertainty, or a certain noise level) to yield functional connectivity
networks of the human brain. There are numerous publications discussing
network extraction and network analysis techniques~(see \citet{Ktena18,
Wang10} for two randomly-selected examples), making them prime examples
for the development of graph kernels that can handle heterogeneous data
sets at multiple scales or different ``resolution'' levels.
The relevance of the topology, \ie\ the definition of connectivity to
extract such a graph, is known to be one of the recurring problems of
the field~\citep{Expert19}, and we are convinced that graph kernels
could provide solutions. Pioneering studies by \citet{vega2013} and
\citet{gkirtzou2013} exploited Weisfeiler--Lehman based techniques to
analyse the fMRI graphs, and their results are encouraging for the
further development of the field. 

\subsubsection{Mathematical applications}

A second area where there is potential for graph-based methods is within
more theoretical mathematical applications. We now highlight two such
areas, namely
\begin{inparaenum}[(i)]
  \item geometric graphs and
  \item the planted clique problem,
\end{inparaenum}
where we anticipate that graph-based methods can progress the field. 

\paragraph{Geometric graphs}
%
Moving to a somewhat more unorthodox domain, geometric graphs refer to
graphs that are constructed on point cloud data---that is to say, sets
of unstructured points in a \mbox{$d$-dimensional} real-valued vector
space---by a \emph{proximity operator} that uses different geometric
properties to define edges between individual points.
The \emph{Gabriel graph}~\citep{Gabriel69}, for example, creates an edge
between two points $p$ and $q$ if and only if their diameter circle~(or
diameter sphere in higher dimensions) contains no other points. Several
variants of such graphs exist~\citep{Correa11, Jaromczyk92} and their
construction can be shown to have interesting geometric properties.
Since point cloud data occur in different domains, we consider them to
be an interesting example for further research in graph kernels. Given
the existence of some previous work~\citep{Bach08}, we are convinced
that the principled construction process of these graphs yields an
interesting starting point for the development of new ``geometric''
graph kernels.

\paragraph{The planted clique problem}
%
Adopting a more theoretical perspective, we briefly discuss how graph
kernels can be of interest to solve optimisation problems on graphs,
thereby helping discover new theorems in complexity theory.
While many of these problems exist~\citep{Arora09}, we focus on a specific
instance, namely the \emph{planted clique problem}~\citep{Alon98}.
A clique is defined as a subset of vertices in an undirected graph whose
induced subgraph is complete, \ie\ every pair of vertices is
connected by an edge. A planted clique in a graph is a clique that has
been added to the graph by selecting a subset of vertices at random and
turning them into a clique. In combinatorial optimisation, the planted
clique problem consists of distinguishing random graphs from graphs with
a planted clique; the probability of adding such a planted clique is
typically taken to be $p = 0.5$.
This problem can be solved in polynomial time only for
specific sufficiently large values of $k$, with $k$ being the size of
a clique. In practice, this problem can be turned into
a binary classification classification on graphs, thus permitting the
use of graph kernels~(since $p = 0.5$, the classification problem does
not suffer from class imbalance).
We see a great potential in this application, which to the best of our
knowledge is yet unexplored. In this domain---and potentially for
related tasks---graph kernels could provide major benefits and speed-ups
for solving the planted clique problem from a classification perspective
with high accuracy. 

\section{Conclusion}

This survey showed that the field of graph kernels is a vibrant and rich
domain of machine learning research. Being well-grounded in the theory
of reproducing kernel Hilbert ppaces, the field permits contributions on
various levels, ranging from the theoretical assessment of kernel
properties to the empirical assessment of the integration of new graph
features.
Given the numerous challenges, open problems and emerging topics around
graph kernels, we are convinced that there are plenty of opportunities
for future work.  We therefore hope that this review provides a stimulus
for novel graph kernel research.

\newglossaryentry{Kernel function}
{%
  name        = {$\kernel$},
  description = {%
    A generic kernel function. Subscripts will be used to denote
    specific graph kernels or base kernels in feature vector spaces.
  },
  sort        = {Kernel function},
}

\newglossaryentry{Kernel function, basic}
{%
  name        = {$\basekernel$},
  description = {%
    A generic base kernel function for kernels that are composed of
    smaller parts permitting the definition of their own kernel.
  },
  sort        = {Kernel function, base},
}

\newglossaryentry{Real numbers}
{%
  name        = {$\real$},
  description = {The field of real numbers},
  sort        = {Real numbers},
}

\newglossaryentry{Complex numbers}
{%
  name        = {$\real$},
  description = {The field of complex numbers},
  sort        = {Complex numbers},
}

\newglossaryentry{Natural numbers}
{%
  name        = {$\natural$},
  description = {The set of natural numbers},
  sort        = {Natural numbers},
}

\newglossaryentry{Graph}
{%
  name        = {$\graph = \left(\vertices, \edges\right)$},
  description = {A graph with a set of vertices $\vertices$ and a set of edges $\edges$},
  sort        = {Graph},
}

\newglossaryentry{Vertex}
{%
  name        = {$\vertex$, $\vertices$},
  description = {%
    The set of vertices $\vertices$ of a graph. An individual vertex is
    denoted by $\vertex$.
  },
  sort        = {Vertex},
}

\newglossaryentry{Edge}
{%
  name        = {$\edge$, $\edges$},
  description = {%
    The set of edges $\edges$ of a graph. An individual edge is denoted
    by $\edge$. If not mentioned otherwise, edges are undirected.
  },
  sort        = {Edge},
}

\newglossaryentry{Hilbert space}
{%
  name        = {$\hilbertspace$},
  description = {A Hilbert space},
  sort        = {Hilbert space},
}

\newglossaryentry{Adjacency matrix}
{%
  name        = {$\adjacency$},
  description = {%
    An adjacency matrix of a graph. If not mentioned otherwise, the adjacency
    matrix is symmetrical because it describes an undirected graph.
  },
  sort = {Adjacency matrix}
}

\newglossaryentry{Inner product}
{%
  name = {$\innerproduct{\cdot}{\cdot}$},
  description = {%
    An inner product on a normed vector space. The concrete definition of the
    inner product will become clear from the context.
  },
  sort = {Inner product}
}

\newglossaryentry{Alphabets}
{%
  name        = {$\Vlabels$, $\Elabels$},
  description = {%
    The domain of vertex labels~($\Vlabels$) or edge labels~($\Elabels$)
    for a graph that contains additional label information.
  },
  sort = {Sigma} 
}

\newglossaryentry{Paths}
{%
  name        = {$\paths$},
  description = {%
    The set of all paths in a graph
  },
  sort = {Paths}
}

\newglossaryentry{Ones}
{%
  name        = {$\ones$},
  description = {%
    A vector of unspecified dimensionality that contains only ones
  },
  sort = {1}
}

\newglossaryentry{Laplacian}
{%
  name        = {$\laplacian$},
  description = {%
    The Laplacian matrix of a graph. The precise definition~(normalized,
    weighted, etc.) will become clear from the context.
  },
  sort = {Laplacian}
}

\newglossaryentry{Trace}
{%
  name        = {$\trace$},
  description = {%
    The trace operator of a matrix
  },
  sort = {Trace}
}

\newglossaryentry{Distance}
{%
  name        = {$\distance$},
  description = {%
    A generic distance function whose precise definition depends on the context
  },
  sort = {Distance, generic}
}

\newglossaryentry{Graph edit distance}
{%
  name        = {$\gdistance$},
  description = {%
    The graph edit distance function
  },
  sort = {Distance, graph edit}
}

\newglossaryentry{Label function}
{%
  name        = {$\flabel$},
  description = {%
    A generic label function for objects
  },
  sort = {Label function, basic}
}

\newglossaryentry{Vertex label function}
{%
  name        = {$\Vlabel$},
  description = {%
    A function that assigns each vertex a label from a vertex label
    alphabet $\Vlabels$
  },
  sort = {Label function, vertices}
}

\newglossaryentry{Edge label function}
{%
  name        = {$\Elabel$},
  description = {%
    A function that assigns each edge a label from an edge label
    alphabet $\Elabels$
  },
  sort = {Label function, edges}
}

\newglossaryentry{weight function}
{%
  name        = {$\fweight$},
  description = {%
    A generic weight function for objects that assigns real-valued
    weights
  },
  sort = {Weight function, basic}
}

\newglossaryentry{Vertex weight function}
{%
  name        = {$\Vweight$},
  description = {%
    A function that assigns each vertex a real-valued weight
  },
  sort = {Weight function, vertices}
}

\newglossaryentry{Edge weight function}
{%
  name        = {$\Eweight$},
  description = {%
    A function that assigns each edge a real-valued weight
  },
  sort = {Weight function, edges}
}

\newglossaryentry{Neighbourhood}
{%
  name        = {$\neighbourhood$},
  description = {%
    Operator for calculating the neighbourhood of a vertex in a graph,
    \ie\ all vertices that are \emph{adjacent} to a given source vertex
  },
  sort = {Neighbourhood function},
}

\newglossaryentry{Feature vector}
{%
  name        = {$\featurevector\left(x\right)$},
  description = {%
    A feature vector representation of an object $x$~(such as a graph,
    or one of its substructures).
    This notation is typically used in the context of a Hilbert
    space~$\hilbertspace$.
  },
  sort = {Feature vector},
}

\newglossaryentry{Hash family}
{%
  name        = {$\hashfamily$},
  description = {%
    A family of hash functions. This is typically used in the context of
    graph kernel frameworks that require the selection of hash
    functions.
  },
  sort        = {Hash family},
}

\backmatter
\glsaddall
\printglossaries
\printbibliography

@Article{Vert08,
  author        = {Vert, Jean-Philippe},
  title         = {The optimal assignment kernel is not positive definite},
  journal       = {arXiv e-prints},
  year          = {2008},
  archiveprefix = {arXiv},
  eprint        = {0801.4061},
  primaryclass  = {cs.LG},
}

@article{Wilson08,
	Author = {Wilson, Richard C. and Zhu, Ping},
	Journal = {Pattern Recognition},
	Number = {9},
	Pages = {2833--2841},
	Title = {A study of graph spectra for comparing graphs and trees},
	Volume = {41},
	Year = {2008}}

@Article{Hofmann08,
  author    = {Hofmann, Thomas and Sch{\"o}lkopf, Bernhard and Smola, Alexander J.},
  title     = {Kernel Methods in Machine Learning},
  journal   = {The Annals of Statistics},
  year      = {2008},
  volume    = {36},
  number    = {3},
  pages     = {1171--1220},
  publisher = {The Institute of Mathematical Statistics},
}

@Book{Brouwer12,
  title     = {Spectra of Graphs},
  publisher = {Springer},
  year      = {2012},
  author    = {Andries E. Brouwer and Willem H. Haemers},
  address   = {New York, NY, USA},
}

@Book{Chung97,
  title     = {Spectral Graph Theory},
  publisher = {American Mathematical Society},
  year      = {1997},
  author    = {Fan R. K. Chung},
  address   = {Providence, RI, USA},
}

@Article{Babai15,
  author        = {L{\'{a}}szl{\'{o}} Babai},
  title         = {Graph Isomorphism in Quasipolynomial Time},
  journal       = {arXiv e-prints},
  year          = {2015},
  archiveprefix = {arXiv},
  eprint        = {1512.03547},
  primaryclass  = {cs.DS},
}

@techreport{Haussler99,
	Author = {Haussler, David},
	Institution = {University of California at Santa Cruz},
	Title = {Convolution kernels on discrete structures},
	Year = {1999}}

@Article{Read77,
  author  = {Read, Ronald C. and Corneil, Derek G.},
  title   = {The Graph Isomorphism Disease},
  journal = {Journal of Graph Theory},
  year    = {1977},
  volume  = {1},
  number  = {4},
  pages   = {339-363},
}

@Article{Shervashidze11,
  author  = {Shervashidze, N. and Schweitzer, P. and van Leeuwen, E. Jan and Mehlhorn, K. and Borgwardt, K. M.},
  title   = {Weisfeiler--Lehman Graph Kernels},
  journal = {Journal of Machine Learning Research},
  year    = {2011},
  number  = {12},
  pages   = {2539--2561},
}

@InProceedings{Kashima03,
  author    = {Kashima, Hisashi and Tsuda, Koji and Inokuchi, Akihiro},
  title     = {Marginalized Kernels between Labeled Graphs},
  booktitle = {Proceedings of the 20th International Conference on Machine Learning},
  year      = {2003},
  pages     = {321--328},
}

@InProceedings{Borgwardt05,
  author    = {Borgwardt, Karsten and Kriegel, Hans-Peter},
  title     = {Shortest-path kernels on graphs},
  booktitle = {Proceedings of the Fifth IEEE International Conference on Data Mining},
  year      = {2005},
  pages     = {74--81},
  address   = {Washington, DC, USA},
  publisher = {IEEE Computer Society},
}

@InProceedings{Vishwanathan06,
  author    = {Vishwanathan, S. V. N. and Borgwardt, Karsten M. and Schraudolph, Nicol N.},
  title     = {Fast Computation of Graph Kernels},
  booktitle = {Advances in Neural Information Processing Systems~19},
  year      = {2006},
  pages     = {1449--1456},
}

@inproceedings{Zhang18,
  title     = {{RetGK}: Graph kernels based on return probabilities of random walks},
  author    = {Zhang, Zhen and Wang, Mianzhi and Xiang, Yijian and Huang, Yan and Nehorai, Arye},
  booktitle = {Advances in Neural Information Processing Systems~32},
  pages     = {3964--3974},
  year      = {2018}
}

@InProceedings{Shervashidze09a,
  author    = {Shervashidze, Nino and Vishwanathan, SVN and Petri, Tobias and Mehlhorn, Kurt and Borgwardt, Karsten},
  title     = {Efficient graphlet kernels for large graph comparison},
  booktitle = {Proceedings of the 12th International Conference on Artificial Intelligence and Statistics},
  year      = {2009},
  pages     = {488--495},
}

@InCollection{Gaertner03,
  author    = {G{\"a}rtner, Thomas and Flach, Peter and Wrobel, Stefan},
  title     = {On Graph Kernels: {H}ardness Results and Efficient Alternatives},
  booktitle = {Learning Theory and Kernel Machines},
  publisher = {Springer},
  year      = {2003},
  editor    = {Sch{\"o}lkopf, Bernhard and Warmuth, Manfred K.},
  pages     = {129--143},
  address   = {Heidelberg, Germany},
}

@InProceedings{Shervashidze09b,
  author    = {Shervashidze, Nino and Borgwardt, Karsten},
  title     = {Fast subtree kernels on graphs},
  booktitle = {Advances in Neural Information Processing Systems~22},
  year      = {2009},
  pages     = {1660--1668},
}

@Article{Neumann16,
  author  = {Neumann, Marion and Garnett, Roman and Bauckhage, Christian and Kersting, Kristian},
  title   = {Propagation kernels: efficient graph kernels from propagated information},
  journal = {Machine Learning},
  year    = {2016},
  volume  = {102},
  number  = {2},
  pages   = {209--245},
}

@InProceedings{Orsini15,
  author    = {Orsini, Francesco and Frasconi, Paolo and De Raedt, Luc},
  title     = {Graph Invariant Kernels},
  booktitle = {Proceedings of the 24th International Conference on Artificial Intelligence},
  year      = {2015},
  pages     = {3756--3762},
  address   = {Palo Alto, CA, USA},
  publisher = {AAAI Press},
}

@InProceedings{Katoka16,
  author    = {Tetsuya Kataoka and Akihiro Inokuchi},
  title     = {Hadamard Code Graph Kernels for Classifying Graphs},
  booktitle = {Proceedings of the 5th International Conference on Pattern Recognition Applications and Methods~(ICPRAM)},
  year      = {2016},
  pages     = {24--32},
}

@InProceedings{Kondor16,
  author    = {Kondor, Risi and Pan, Horace},
  title     = {The Multiscale Laplacian Graph Kernel},
  booktitle = {Advances in Neural Information Processing Systems~29},
  year      = {2016},
  editor    = {D. D. Lee and M. Sugiyama and U. V. Luxburg and I. Guyon and R. Garnett},
  pages     = {2990--2998},
}

@InProceedings{Yanardag15,
  author    = {Yanardag, Pinar and Vishwanathan, S.V.N.},
  title     = {Deep Graph Kernels},
  booktitle = {Proceedings of the 21th ACM SIGKDD International Conference on Knowledge Discovery and Data Mining},
  year      = {2015},
  pages     = {1365--1374},
  address   = {New York, NY, USA},
  publisher = {ACM},
}

@misc{KKMMN16,
	Author = {Kristian Kersting and Nils M. Kriege and Christopher Morris and Petra Mutzel and Marion Neumann},
	Title = {Benchmark Data Sets for Graph Kernels},
	Url = {http://graphkernels.cs.tu-dortmund.de},
	Year = {2016}}

@Article{Sugiyama17,
  author  = {Sugiyama, Mahito and Ghisu, M. Elisabetta and Llinares-L{\'o}pez, Felipe and Borgwardt, Karsten},
  title   = {\texttt{graphkernels}: {R} and {P}ython packages for graph comparison},
  journal = {Bioinformatics},
  year    = {2017},
  volume  = {34},
  number  = {3},
  pages   = {530--532},
}

@InCollection{Sugiyama15,
  author    = {Sugiyama, Mahito and Borgwardt, Karsten},
  title     = {Halting in Random Walk Kernels},
  booktitle = {Advances in Neural Information Processing Systems~28},
  year      = {2015},
  editor    = {C. Cortes and N. D. Lawrence and D. D. Lee and M. Sugiyama and R. Garnett},
  pages     = {1639--1647},
}

@InProceedings{Feragen13,
  author    = {Feragen, Aasa and Kasenburg, Niklas and Petersen, Jens and de Bruijne, Marleen and Borgwardt, Karsten},
  title     = {Scalable kernels for graphs with continuous attributes},
  booktitle = {Advances in Neural Information Processing Systems~26},
  year      = {2013},
  editor    = {C. J. C. Burges and L. Bottou and M. Welling and Z. Ghahramani and K. Q. Weinberger},
  pages     = {216--224},
}

@InProceedings{Gilmer17,
  author    = {Justin Gilmer and Samuel S. Schoenholz and Patrick F. Riley and Oriol Vinyals and George E. Dahl},
  title     = {Neural Message Passing for Quantum Chemistry},
  booktitle = {Proceedings of the 34th International Conference on Machine Learning},
  year      = {2017},
  editor    = {Doina Precup and Yee Whye Teh},
  series    = {Proceedings of Machine Learning Research},
  pages     = {1263--1272},
  publisher = {PMLR},
}

@InProceedings{Hido09,
  author    = {Hido, Shohei and Kashima, Hisashi},
  title     = {A Linear-Time Graph Kernel},
  booktitle = {Proceedings of the Ninth IEEE International Conference on Data Mining},
  year      = {2009},
  pages     = {179--188},
  address   = {Washington, DC, USA},
  publisher = {IEEE Computer Society},
}

@InProceedings{Datar04,
  author    = {Datar, Mayur and Immorlica, Nicole and Indyk, Piotr and Mirrokni, Vahab S.},
  title     = {Locality-Sensitive Hashing Scheme Based on $p$-stable Distributions},
  booktitle = {Proceedings of the 20th Annual Symposium on Computational Geometry},
  year      = {2004},
  pages     = {253--262},
  address   = {New York, NY, USA},
  publisher = {ACM},
}

@Book{Arora09,
  title     = {Computational complexity: {A} Modern Approach},
  publisher = {Cambridge University Press},
  year      = {2009},
  author    = {Sanjeev Arora and Boaz Barak},
  address   = {Cambridge, United Kingdom},
}

@article{Weisfeiler68,
	Author = {Boris Weisfeiler and Andrei A. Lehman},
	Journal = {Nauchno-Technicheskaya Informatsia},
	Number = {9},
	Pages = {12--16},
	Title = {A reduction of a graph to a canonical form and an algebra arising during this reduction},
	Volume = {2},
	Year = {1968}}

@Book{Riesen15,
  title     = {Structural Pattern Recognition with Graph Edit Distance},
  publisher = {Springer},
  year      = {2015},
  author    = {Riesen, Kaspar},
  address   = {Cham, Switzerland},
}

@InProceedings{Bunke07,
  author    = {Bunke, Horst and Riesen, Kaspar},
  title     = {A Family of Novel Graph Kernels for Structural Pattern Recognition},
  booktitle = {Progress in Pattern Recognition, Image Analysis and Applications},
  year      = {2007},
  editor    = {Rueda, Luis and Mery, Domingo and Kittler, Josef},
  pages     = {20--31},
  address   = {Heidelberg, Germany},
  publisher = {Springer},
}

@InProceedings{Bai13,
  author    = {Bai, Lu and Hancock, Edwin R. and Torsello, Andrea and Rossi, Luca},
  title     = {A Quantum {J}ensen--{S}hannon Graph Kernel Using the Continuous-time Quantum Walk},
  booktitle = {Graph-Based Representations in Pattern Recognition},
  year      = {2013},
  editor    = {Kropatsch, Walter G. and Artner, Nicole M. and Haxhimusa, Yll and Jiang, Xiaoyi},
  pages     = {121--131},
  address   = {Heidelberg, Germany},
  publisher = {Springer},
}

@InProceedings{Froehlich05,
  author       = {Fr{\"o}hlich, Holger and Wegner, J{\"o}rg K. and Sieker, Florian and Zell, Andreas},
  title        = {Optimal Assignment Kernels for Attributed Molecular Graphs},
  booktitle    = {Proceedings of the 22nd International Conference on Machine Learning},
  year         = {2005},
  pages        = {225--232},
  address      = {New York, NY, USA},
  organization = {ACM},
}

@InProceedings{Menchetti05,
  author    = {Menchetti, Sauro and Costa, Fabrizio and Frasconi, Paolo},
  title     = {Weighted Decomposition Kernels},
  booktitle = {Proceedings of the 22nd International Conference on Machine Learning},
  year      = {2005},
  pages     = {585--592},
  address   = {New York, NY, USA},
  publisher = {ACM},
}

@Article{Ceroni07,
  author  = {Ceroni, Alessio and Costa, Fabrizio and Frasconi, Paolo},
  title   = {Classification of small molecules by two- and three-dimensional decomposition kernels},
  journal = {Bioinformatics},
  year    = {2007},
  volume  = {23},
  number  = {16},
  pages   = {2038--2045},
}

@Article{Jebara04,
  author  = {Jebara, Tony and Kondor, Risi and Howard, Andrew},
  title   = {Probability Product Kernels},
  journal = {Journal of Machine Learning Research},
  year    = {2004},
  volume  = {5},
  pages   = {819--844},
}

@Article{Dijkstra59,
  author  = {Edsger Wybe Dijkstra},
  title   = {A Note on Two Problems in Connexion with Graphs},
  journal = {Numerische Mathematik},
  year    = {1959},
  volume  = {1},
  number  = {1},
  pages   = {269--271},
}

@Book{Tan19,
  title     = {Introduction to Data Mining},
  publisher = {Pearson},
  year      = {2019},
  author    = {Pang-Ning Tan and Michael Steinbach and Anuj Karpatne and Vipin Kumar},
  address   = {London, United Kingdom},
  edition   = {2},
}

@InCollection{Karp72,
  author    = {Karp, Richard M.},
  title     = {Reducibility among Combinatorial Problems},
  booktitle = {Complexity of Computer Computations},
  publisher = {Springer},
  year      = {1972},
  editor    = {Miller, Raymond E. and Thatcher, James W. and Bohlinger, Jean D.},
  pages     = {85--103},
  address   = {Boston, MA, USA},
}

@InProceedings{Berger09,
  author    = {Berger, Franziska and Gritzmann, Peter and de Vries, Sven},
  title     = {Minimum Cycle Bases and Their Applications},
  booktitle = {Algorithmics of Large and Complex Networks: {D}esign, Analysis, and Simulation},
  year      = {2009},
  editor    = {Lerner, J{\"u}rgen and Wagner, Dorothea and Zweig, Katharina A.},
  pages     = {34--49},
  address   = {Heidelberg, Germany},
  publisher = {Springer},
}

@techreport{Bernstein00,
	Author = {Bernstein, Mira and {D}e Silva, Vin and Langford, John C. and Tenenbaum, Joshua B.},
	Institution = {Stanford University},
	Title = {Graph approximations to geodesics on embedded manifolds},
	Year = {2000}}

@inproceedings{Vert04,
	Address = {Cambridge, MA, USA},
	Author = {Vert, Jean-Philippe and Tsuda, Koji and Sch{\"o}lkopf, Bernhard},
	Booktitle = {Kernel Methods in Computational Biology},
	Editor = {Sch{\"o}lkopf, Bernhard and Tsuda, Koji and Vert, Jean-Philippe},
	Pages = {35--70},
	Publisher = {MIT Press},
	Title = {A primer on kernel methods},
	Year = {2004}}

@Article{Mikolov13,
  author        = {Mikolov, Tomas and Chen, Kai and Corrado, Greg and Dean, Jeffrey},
  title         = {Efficient Estimation of Word Representations in Vector Space},
  journal       = {arXiv e-prints},
  year          = {2013},
  archiveprefix = {arXiv},
  eprint        = {1301.3781},
  primaryclass  = {cs.CL},
}

@InProceedings{Kriege16,
  author    = {Kriege, Nils M. and Giscard, Pierre-Louis and Wilson, Richard C.},
  title     = {On Valid Optimal Assignment Kernels and Applications to Graph Classification},
  booktitle = {Advances in Neural Processing Systems~29},
  year      = {2016},
  pages     = {1623--1631},
}

@Article{Floyd62,
  author  = {Floyd, Robert W.},
  title   = {Algorithm 97: {S}hortest Path},
  journal = {Communications of the ACM},
  year    = {1962},
  volume  = {5},
  number  = {6},
  pages   = {345},
}

@Article{Warshall62,
  author  = {Warshall, Stephen},
  title   = {A Theorem on {B}oolean Matrices},
  journal = {Journal of the ACM},
  year    = {1962},
  volume  = {9},
  number  = {1},
  pages   = {11--12},
}

@Article{Przulj04,
  author  = {Pr{\v{z}}ulj, N. and Corneil, D. G. and Jurisica, I.},
  title   = {Modeling interactome: scale-free or geometric?},
  journal = {Bioinformatics},
  year    = {2004},
  volume  = {20},
  number  = {18},
  pages   = {3508--3515},
}

@Article{Przulj06,
  author  = {Pr{\v{z}}ulj, N. and Corneil, D. G. and Jurisica, I.},
  title   = {Efficient estimation of graphlet frequency distributions in protein–protein interaction networks},
  journal = {Bioinformatics},
  year    = {2006},
  volume  = {22},
  number  = {8},
  pages   = {974--980},
}

@InProceedings{Curado15,
  author    = {Curado, Manuel and Escolano, Francisco and Hancock, Edwin R. and Nourbakhsh, Farshad and Pelillo, Marcello},
  title     = {Similarity Analysis from Limiting Quantum Walks},
  booktitle = {Similarity-Based Pattern Recognition -- SIMBAD},
  year      = {2015},
  editor    = {Feragen, Aasa and Pelillo, Marcello and Loog, Marco},
  pages     = {38--53},
  address   = {Cham, Switzerland},
  publisher = {Springer},
}

@InProceedings{Bai15,
  author    = {Bai, Lu and Zhang, Zhihong and Ren, Peng and Rossi, Luca and Hancock, Edwin R.},
  title     = {An Edge-based Matching Kernel through Discrete-time Quantum Walks},
  booktitle = {Image Analysis and Processing -- ICIAP},
  year      = {2015},
  editor    = {Murino, Vittorio and Puppo, Enrico},
  pages     = {27--38},
  address   = {Cham, Switzerland},
  publisher = {Springer},
}

@InProceedings{Bai14,
  author    = {Bai, Lu and Rossi, Luca and Bunke, Horst and Hancock, Edwin R.},
  title     = {Attributed Graph Kernels Using the {J}ensen--{T}sallis $q$-Differences},
  booktitle = {Machine Learning and Knowledge Discovery in Databases},
  year      = {2014},
  editor    = {Calders, Toon and Esposito, Floriana and H{\"u}llermeier, Eyke and Meo, Rosa},
  pages     = {99--114},
  address   = {Heidelberg, Germany},
  publisher = {Springer},
}

@Book{Wasserman94,
  title      = {Social Network Analysis: {M}ethods and Applications},
  publisher  = {Cambridge University Press},
  year       = {1994},
  author     = {Wasserman, Stanley and Faust, Katherine},
  address    = {Cambridge, UK},
  collection = {Structural Analysis in the Social Sciences},
}

@Article{Scott11,
  author  = {Scott, John},
  title   = {Social network analysis: developments, advances, and prospects},
  journal = {Social Network Analysis and Mining},
  year    = {2011},
  volume  = {1},
  number  = {1},
  pages   = {21--26},
}

@InProceedings{Morris16,
  author    = {Morris, Christopher and Kriege, Nils M. and Kersting, Kristian and Mutzel, Petra},
  title     = {Faster Kernels for Graphs with Continuous Attributes via Hashing},
  booktitle = {Proceedings of the 16th IEEE International Conference on Data Mining},
  year      = {2016},
  pages     = {1095--1100},
}

@Article{Pauleve10,
  author  = {Paulevé, Loïc and Jégou, Hervé and Amsaleg, Laurent},
  title   = {Locality sensitive hashing: {A} comparison of hash function types and querying mechanisms},
  journal = {Pattern Recognition Letters},
  year    = {2010},
  volume  = {31},
  number  = {11},
  pages   = {1348--1358},
}

@Book{Borg05,
  title     = {Modern Multidimensional Scaling},
  publisher = {Springer},
  year      = {2005},
  author    = {Ingwer Borg and Patrick J. F. Groenen},
  address   = {New York, NY, USA},
  edition   = {2},
  subtitle  = {Theory and Applications},
}

@Article{Demsar06,
  author  = {Dem{\v{s}}ar, Janez},
  title   = {Statistical Comparisons of Classifiers Over Multiple Data Sets},
  journal = {Journal of Machine Learning Research},
  year    = {2006},
  volume  = {7},
  pages   = {1--30},
}

@Article{Calvo16,
  author  = {Calvo, Borja and Santafé, Guzmán},
  title   = {\texttt{scmamp}: {S}tatistical Comparison of Multiple Algorithms in Multiple Problems},
  journal = {The R Journal},
  year    = {2016},
  volume  = {8},
  number  = {1},
  pages   = {248--256},
}

@Article{Bagnall17,
  author  = {Bagnall, Anthony and Lines, Jason and Bostrom, Aaron and Large, James and Keogh, Eamonn},
  title   = {The great time series classification bake off: a review and experimental evaluation of recent algorithmic advances},
  journal = {Data Mining and Knowledge Discovery},
  year    = {2017},
  volume  = {31},
  number  = {3},
  pages   = {606--660},
}

@Article{Muellner11,
  author        = {M{\"u}llner, Daniel},
  title         = {Modern hierarchical, agglomerative clustering algorithms},
  journal       = {arXiv e-prints},
  year          = {2011},
  archiveprefix = {arXiv},
  eprint        = {1109.2378},
  primaryclass  = {stat.ML},
}

@Article{Nikolentzos18,
  author        = {Nikolentzos, Giannis and Vazirgiannis, Michalis},
  title         = {Message Passing Graph Kernels},
  journal       = {arXiv e-prints},
  year          = {2018},
  archiveprefix = {arXiv},
  eprint        = {1808.02510},
  primaryclass  = {stat.ML},
}

@Article{Zhou18,
  author        = {Zhou, Jie and Cui, Ganqu and Zhang, Zhengyan and Yang, Cheng and Liu, Zhiyuan and Wang, Lifeng and Li, Changcheng and Sun, Maosong},
  title         = {Graph Neural Networks: {A} Review of Methods and Applications},
  journal       = {arXiv e-prints},
  year          = {2018},
  archiveprefix = {arXiv},
  eprint        = {1812.08434},
  primaryclass  = {cs.LG},
}

@Article{Kanungo02,
  author  = {Kanungo, T. and Mount, D. M. and Netanyahu, N. S. and Piatko, C. D. and Silverman, R. and Wu, A. Y.},
  title   = {An Efficient {$k$-Means} Clustering Algorithm: {A}nalysis and Implementation},
  journal = {IEEE Transactions on Pattern Analysis and Machine Intelligence},
  year    = {2002},
  volume  = {24},
  number  = {7},
  pages   = {881--892},
}

@InCollection{Williams01,
  author    = {Williams, Christopher K. I. and Seeger, Matthias},
  title     = {Using the Nystr\"{o}m Method to Speed Up Kernel Machines},
  booktitle = {Advances in Neural Information Processing Systems~13},
  publisher = {MIT Press},
  year      = {2001},
  editor    = {T. K. Leen and T. G. Dietterich and V. Tresp},
  pages     = {682--688},
}

@Article{Togninalli19,
  author        = {Togninalli, Matteo and Ghisu, Elisabetta and Llinares-L{\'o}pez, Felipe and Rieck, Bastian and Borgwardt, Karsten},
  title         = {Wasserstein {W}eisfeiler--{L}ehman Graph Kernels},
  journal       = {arXiv e-prints},
  year          = {2019},
  archiveprefix = {arXiv},
  eprint        = {1906.01277},
  primaryclass  = {cs.LG},
}

@InProceedings{Rieck19,
  author    = {Rieck, Bastian and Bock, Christian and Borgwardt, Karsten},
  title     = {A Persistent {W}eisfeiler--{L}ehman Procedure for Graph Classification},
  booktitle = {Proceedings of the 36th International Conference on Machine Learning},
  year      = {2019},
  editor    = {Chaudhuri, Kamalika and Salakhutdinov, Ruslan},
  volume    = {97},
  series    = {Proceedings of Machine Learning Research},
  pages     = {5448--5458},
  publisher = {PMLR},
}

@Article{Bai18,
  author        = {Bai, Yunsheng and Ding, Hao and Bian, Song and Chen, Ting and Sun, Yizhou and Wang, Wei},
  title         = {Graph Edit Distance Computation via Graph Neural Networks},
  journal       = {arXiv e-prints},
  year          = {2018},
  archiveprefix = {arXiv},
  eprint        = {1808.05689},
  primaryclass  = {cs.LG},
}

@Article{Glem06,
  author  = {Glem, Robert and Bender, Andreas and Hasselgren, Catrin and Carlsson, Lars and Boyer, Scott and Smith, James},
  title   = {Circular fingerprints: Flexible molecular descriptors with applications from physical chemistry to ADME},
  journal = {IDrugs : the investigational drugs journal},
  year    = {2006},
  volume  = {9},
  pages   = {199--204},
}

@Article{Aronszajn50,
  author    = {Nachman Aronszajn},
  title     = {Theory of Reproducing Kernels},
  journal   = {Transactions of the American Mathematical Society},
  year      = {1950},
  volume    = {68},
  number    = {3},
  pages     = {337--404},
  publisher = {American Mathematical Society},
}

@Book{OSearcoid07,
  title     = {Metric Spaces},
  publisher = {Springer},
  year      = {2007},
  author    = {{\'O} Searc{\'o}id, M{\'}iche{\'}al},
  address   = {London, England},
}

@Article{Blizard88,
  author    = {Blizard, Wayne D.},
  title     = {Multiset Theory},
  journal   = {Notre Dame Journal of Formal Logic},
  year      = {1988},
  volume    = {30},
  number    = {1},
  pages     = {36--66},
  publisher = {Duke University Press},
}

@InProceedings{Babai79,
  author    = {L. Babai and L. Kucera},
  title     = {Canonical labelling of graphs in linear average time},
  booktitle = {20th Annual Symposium on Foundations of Computer Science},
  year      = {1979},
  pages     = {39--46},
  issn      = {0272-5428},
}

@InProceedings{Silva04,
  author    = {de Silva, Vin and Carlsson, Gunnar},
  title     = {Topological estimation using witness complexes},
  booktitle = {Symposium on Point-Based Graphics},
  year      = {2004},
  editor    = {Markus Gross and Hanspeter Pfister and Marc Alexa and Szymon Rusinkiewicz},
  publisher = {The Eurographics Association},
  comment   = {FIXME: Consistent names of the symposium},
}

@InProceedings{Graf01,
  author    = {Graf, Arnulf B. A. and Borer, Silvio},
  title     = {Normalization in Support Vector Machines},
  booktitle = {Pattern Recognition},
  year      = {2001},
  editor    = {Radig, Bernd and Florczyk, Stefan},
  pages     = {277--282},
  address   = {Heidelberg, Germany},
  publisher = {Springer},
  isbn      = {978-3-540-45404-5},
}

@Article{Wang10,
  author  = {Wang, Jinhui and Zuo, Xinian and He, Yong},
  title   = {Graph-based network analysis of resting-state functional {MRI}},
  journal = {Frontiers in Systems Neuroscience},
  year    = {2010},
  volume  = {4},
  pages   = {16:1--16:14},
}

@Article{Expert19,
  author  = {Expert, Paul and Lord, Louis-David and Kringelbach, Morten L. and Petri, Giovanni},
  title   = {Editorial: {T}opological Neuroscience},
  journal = {Network Neuroscience},
  year    = {2019},
  volume  = {3},
  number  = {3},
  pages   = {653--655},
}

@Article{Ktena18,
  author  = {Ktena, Sofia Ira and Parisot, Sarah and Ferrante, Enzo and Rajchl, Martin and Lee, Matthew and Glocker, Ben and Rueckert, Daniel},
  title   = {Metric learning with spectral graph convolutions on brain connectivity networks},
  journal = {NeuroImage},
  year    = {2018},
  volume  = {169},
  pages   = {431--442},
}

@InProceedings{Choi19,
  author    = {Choi, Edward and Dusenberry, Michael W. and Flores, Gerardo and Xu, Zhen and Li, Yujia and Xue, Yuan and Dai, Andrew M.},
  title     = {Learning Graphical Structure of Electronic Health Records with Transformer for Predictive Healthcare},
  booktitle = {ICML Workshop on Learning and Reasoning with Graph-Structured Data},
  year      = {2019},
}

@Article{Rotmensch17,
  author  = {Rotmensch, Maya and Halpern, Yoni and Tlimat, Abdulhakim and Horng, Steven and Sontag, David},
  title   = {Learning a Health Knowledge Graph from Electronic Medical Records},
  journal = {Scientific Reports},
  year    = {2017},
  volume  = {7},
  number  = {1},
  pages   = {5994},
}

@Article{Jaromczyk92,
  author  = {Jerzy W. Jaromczyk and Godfried T. Toussaint},
  title   = {Relative Neighborhood Graphs and Their Relatives},
  journal = {Proceedings of the IEEE},
  year    = {1992},
  volume  = {80},
  number  = {9},
  pages   = {1502--1517},
}

@Article{Gabriel69,
  author  = {Kuno R. Gabriel and Robert R. Sokal},
  title   = {A new statistical approach to geographic variation analysis},
  journal = {Systematic Biology},
  year    = {1969},
  volume  = {18},
  number  = {3},
  pages   = {259--278},
}

@Article{Correa11,
  author        = {Correa, Carlos D. and Lindstrom, Peter},
  title         = {Towards Robust Topology of Sparsely Sampled Data},
  journal       = {IEEE Transactions on Visualization and Computer Graphics},
  year          = {2011},
  volume        = {17},
  number        = {12},
  pages         = {1852--1861},
}

@InProceedings{Bach08,
  author    = {Bach, Francis R.},
  title     = {Graph Kernels between Point Clouds},
  booktitle = {Proceedings of the 25th International Conference on Machine Learning},
  year      = {2008},
  pages     = {25--32},
}

@Article{Nystroem30,
  author  = {Nystr{\"o}m, E. J.},
  title   = {{\"U}ber die praktische {A}ufl{\"o}sung von Integralgleichungen mit Anwendungen auf Randwertaufgaben},
  journal = {Acta Mathematica},
  year    = {1930},
  volume  = {54},
  pages   = {185--204},
}

@Article{Bloom70,
  author  = {Bloom, Burton H.},
  title   = {Space/Time Trade-offs in Hash Coding with Allowable Errors},
  journal = {Communications of the ACM},
  year    = {1970},
  volume  = {13},
  number  = {7},
  pages   = {422--426},
}

@InProceedings{Nikolentzos18a,
  author    = {Nikolentzos, Giannis and Meladianos, Polykarpos and Tixier, Antoine Jean-Pierre and Skianis, Konstantinos and Vazirgiannis, Michalis},
  title     = {Kernel Graph Convolutional Neural Networks},
  booktitle = {Artificial Neural Networks and Machine Learning -- ICANN 2018},
  year      = {2018},
  editor    = {K{\r{u}}rkov{\'a}, V{\v{e}}ra and Manolopoulos, Yannis and Hammer, Barbara and Iliadis, Lazaros and Maglogiannis, Ilias},
  pages     = {22--32},
  address   = {Cham, Switzerland},
  publisher = {Springer},
}

@Article{Wu19,
  author        = {Wu, Zonghan and Pan, Shirui and Chen, Fengwen and Long, Guodong and Zhang, Chengqi and Yu, Philip S.},
  title         = {A Comprehensive Survey on Graph Neural Networks},
  journal       = {arXiv e-prints},
  year          = {2019},
  archiveprefix = {arXiv},
  eprint        = {1901.00596},
  primaryclass  = {cs.LG},
}

@InProceedings{xu18,
  author    = {Xu, Keyulu and Hu, Weihua and Leskovec, Jure and Jegelka, Stefanie},
  title     = {How Powerful are Graph Neural Networks?},
  booktitle = {International Conference on Learning Representations~(ICLR)},
  year      = {2019},
}

@Article{Du19,
  author        = {Simon S. Du and Kangcheng Hou and Barnab{\'{a}}s P{\'{o}}czos and Ruslan Salakhutdinov and Ruosong Wang and Keyulu Xu},
  title         = {Graph Neural Tangent Kernel: Fusing Graph Neural Networks with Graph Kernels},
  journal       = {arXiv e-prints},
  year          = {2019},
  pages         = {2224--2232},
  archiveprefix = {arXiv},
  eprint        = {1905.13192},
  primaryclass  = {cs.LG},
}

@InProceedings{hamilton17,
  author    = {Hamilton, William L. and Ying, Rex and Leskovec, Jure},
  title     = {Inductive Representation Learning on Large Graphs},
  booktitle = {Advances in Neural Information Processing Systems~30},
  year      = {2017},
  pages     = {1024--1034},
}

@InProceedings{duvenaud15,
  author    = {Duvenaud, David K and Maclaurin, Dougal and Iparraguirre, Jorge and Bombarell, Rafael and Hirzel, Timothy and Aspuru-Guzik, Alan and Adams, Ryan P},
  title     = {Convolutional Networks on Graphs for Learning Molecular Fingerprints},
  booktitle = {Advances in Neural Information Processing Systems~28},
  year      = {2015},
  pages     = {2224--2232},
}

@InProceedings{kipf17,
  author    = {Kipf, Thomas N. and Welling, Max},
  title     = {Semi-Supervised Classification with Graph Convolutional Networks},
  booktitle = {International Conference on Learning Representations~(ICLR)},
  year      = {2017},
}

@Book{villani08,
  title     = {Optimal Transport: Old and New},
  publisher = {Springer},
  year      = {2008},
  author    = {Villani, C{\'e}dric},
  address   = {Heidelberg, Germany},
}

@InProceedings{Ying18,
  author    = {Ying, Rex and You, Jiaxuan and Morris, Christopher and Ren, Xiang and Hamilton, William L. and Leskovec, Jure},
  title     = {Hierarchical Graph Representation Learning with Differentiable Pooling},
  booktitle = {Advances in Neural Information Processing Systems~32},
  year      = {2018},
  pages     = {4805--4815},
}

@Article{schutt17,
  author  = {Schütt, Kristof T. and Arbabzadah, Farhad and Chmiela, Stefan and Müller, Klaus R. and Tkatchenko, Alexandre},
  title   = {Quantum-chemical insights from deep tensor neural networks},
  journal = {Nature Communications},
  year    = {2017},
  volume  = {8},
}

@Article{kriege2019,
  author  = {Kriege, Nils M. and Neumann, Marion and Morris, Christopher and Kersting, Kristian and Mutzel, Petra},
  title   = {A unifying view of explicit and implicit feature maps of graph kernels},
  journal = {Data Mining and Knowledge Discovery},
  year    = {2019},
  volume  = {33},
  number  = {6},
  pages   = {1505--1547},
}

@Article{hamilton2017representation,
  author        = {Hamilton, William L. and Ying, Rex and Leskovec, Jure},
  title         = {Representation Learning on Graphs: Methods and Applications},
  journal       = {arXiv e-prints},
  year          = {2017},
  archiveprefix = {arXiv},
  eprint        = {1709.05584},
  primaryclass  = {cs.SI},
}

@InProceedings{aiolli15,
  author    = {Aiolli, F. and Donini, M. and Navarin, N. and Sperduti, A.},
  title     = {Multiple Graph-Kernel Learning},
  booktitle = {IEEE Symposium Series on Computational Intelligence},
  year      = {2015},
  pages     = {1607--1614},
}

@Book{trinajstic2018chemical,
  title     = {Chemical Graph Theory},
  publisher = {CRC Press},
  year      = {2018},
  author    = {Trinajstic, Nenad},
  address   = {Boca Raton, FL, USA},
  edition   = {2},
}

@Article{wu2018moleculenet,
  author    = {Wu, Zhenqin and Ramsundar, Bharath and Feinberg, Evan N and Gomes, Joseph and Geniesse, Caleb and Pappu, Aneesh S and Leswing, Karl and Pande, Vijay},
  title     = {MoleculeNet: a benchmark for molecular machine learning},
  journal   = {Chemical Science},
  year      = {2018},
  volume    = {9},
  number    = {2},
  pages     = {513--530},
  publisher = {Royal Society of Chemistry},
}

@Article{vamathevan2019applications,
  author    = {Vamathevan, Jessica and Clark, Dominic and Czodrowski, Paul and Dunham, Ian and Ferran, Edgardo and Lee, George and Li, Bin and Madabhushi, Anant and Shah, Parantu and Spitzer, Michaela and others},
  title     = {Applications of machine learning in drug discovery and development},
  journal   = {Nature Reviews Drug Discovery},
  year      = {2019},
  volume    = {18},
  pages     = {463--477},
  publisher = {Nature Publishing Group},
}

@Article{szklarczyk2018string,
  author    = {Szklarczyk, Damian and Gable, Annika L and Lyon, David and Junge, Alexander and Wyder, Stefan and Huerta-Cepas, Jaime and Simonovic, Milan and Doncheva, Nadezhda T and Morris, John H and Bork, Peer and others},
  title     = {STRING v11: protein--protein association networks with increased coverage, supporting functional discovery in genome-wide experimental datasets},
  journal   = {Nucleic Acids Research},
  year      = {2018},
  volume    = {47},
  number    = {D1},
  pages     = {D607--D613},
  publisher = {Oxford University Press},
}

@Article{zhang2005general,
  author    = {Zhang, Bin and Horvath, Steve},
  title     = {A General Framework for Weighted Gene Co-expression Network Analysis},
  journal   = {Statistical Applications in Genetics and Molecular Biology},
  year      = {2005},
  volume    = {4},
  number    = {1},
  pages     = {17:1--17:43},
  publisher = {De Gruyter},
}

@Article{lonsdale2013genotype,
  author    = {Lonsdale, John and Thomas, Jeffrey and Salvatore, Mike and Phillips, Rebecca and Lo, Edmund and Shad, Saboor and Hasz, Richard and Walters, Gary and Garcia, Fernando and Young, Nancy and others},
  title     = {The Genotype-Tissue Expression (GTEx) project},
  journal   = {Nature Genetics},
  year      = {2013},
  volume    = {45},
  number    = {6},
  pages     = {580--585},
  publisher = {Nature Publishing Group},
}

@Article{kanehisa2000kegg,
  author    = {Kanehisa, Minoru and Goto, Susumu},
  title     = {{KEGG}: Kyoto Encyclopedia of Genes and Genomes},
  journal   = {Nucleic Acids Research},
  year      = {2000},
  volume    = {28},
  number    = {1},
  pages     = {27--30},
  publisher = {Oxford University Press},
}

@article{karlebach2008modelling,
  title={Modelling and analysis of gene regulatory networks},
  author={Karlebach, Guy and Shamir, Ron},
  journal={Nature Reviews Molecular Cell Biology},
  volume={9},
  number={10},
  pages={770},
  year={2008},
  publisher={Nature Publishing Group}
}

@article{goh2007human,
  title={The human disease network},
  author={Goh, Kwang-Il and Cusick, Michael E and Valle, David and Childs, Barton and Vidal, Marc and Barab{\'a}si, Albert-L{\'a}szl{\'o}},
  journal={Proceedings of the National Academy of Sciences},
  volume={104},
  number={21},
  pages={8685--8690},
  year={2007},
  publisher={National Acad Sciences}
}

@article{borgwardt2005protein,
  title={Protein function prediction via graph kernels},
  author={Borgwardt, Karsten M and Ong, Cheng Soon and Sch{\"o}nauer, Stefan and Vishwanathan, SVN and Smola, Alex J and Kriegel, Hans-Peter},
  journal={Bioinformatics},
  volume={21},
  number={suppl\_1},
  pages={i47--i56},
  year={2005},
  publisher={Oxford University Press}
}

@Article{huson2005application,
  author    = {Huson, Daniel H. and Bryant, David},
  title     = {Application of Phylogenetic Networks in Evolutionary Studies},
  journal   = {Molecular Biology and Evolution},
  year      = {2005},
  volume    = {23},
  number    = {2},
  pages     = {254--267},
  publisher = {Oxford University Press},
}

@Article{he2010graph,
  author    = {He, Yong and Evans, Alan},
  title     = {Graph theoretical modeling of brain connectivity},
  journal   = {Current Opinion in Neurology},
  year      = {2010},
  volume    = {23},
  number    = {4},
  pages     = {341--350},
  publisher = {LWW},
}

@InProceedings{choi2017gram,
  author       = {Choi, Edward and Bahadori, Mohammad Taha and Song, Le and Stewart, Walter F and Sun, Jimeng},
  title        = {\texttt{GRAM}: Graph-based Attention Model for Healthcare Representation Learning},
  booktitle    = {Proceedings of the 23rd ACM SIGKDD International Conference on Knowledge Discovery and Data Mining},
  year         = {2017},
  pages        = {787--795},
  address      = {New York, NY, USA},
  organization = {ACM},
}

@Book{todeschini2008handbook,
  title     = {Handbook of Molecular Descriptors},
  publisher = {Wiley-VCH},
  year      = {2008},
  author    = {Todeschini, Roberto and Consonni, Viviana},
  volume    = {11},
  address   = {Weinheim, Germany},
}

@Article{leskovec2007dynamics,
  author    = {Leskovec, Jure and Adamic, Lada A. and Huberman, Bernardo A.},
  title     = {The Dynamics of Viral Marketing},
  journal   = {ACM Transactions on the Web},
  year      = {2007},
  volume    = {1},
  number    = {1},
  pages     = {5:1--5:39},
  publisher = {ACM},
}

@InProceedings{du2007community,
  author       = {Du, Nan and Wu, Bin and Pei, Xin and Wang, Bai and Xu, Liutong},
  title        = {Community Detection in Large-scale Social Networks},
  booktitle    = {Proceedings of the 9th WebKDD and 1st SNA-KDD 2007 Workshop on Web Mining and Social Network Analysis},
  year         = {2007},
  pages        = {16--25},
  address      = {New York, NY, USA},
  organization = {ACM},
}

@InProceedings{du2013scalable,
  author    = {Du, Nan and Song, Le and Rodriguez, Manuel Gomez and Zha, Hongyuan},
  title     = {Scalable Influence Estimation in Continuous-time Diffusion Networks},
  booktitle = {Advances in Neural information Processing Systems~26},
  year      = {2013},
  editor    = {C. J. C. Burges and L. Bottou and M. Welling and Z. Ghahramani and K. Q. Weinberger},
  pages     = {3147--3155},
}

@InProceedings{tschiatschek2018fake,
  author    = {Tschiatschek, Sebastian and Singla, Adish and Gomez Rodriguez, Manuel and Merchant, Arpit and Krause, Andreas},
  title     = {Fake News Detection in Social Networks via Crowd Signals},
  booktitle = {Companion Proceedings of the The Web Conference 2018},
  year      = {2018},
  pages     = {517--524},
}

@Book{scholkopf2002learning,
  title     = {Learning with Kernels: Support Vector Machines, Regularization, Optimization, and Beyond},
  publisher = {MIT Press},
  year      = {2002},
  author    = {Sch{\"o}lkopf, Bernhard and Smola, Alexander J.},
  address   = {Cambridge, MA, USA},
}

@inproceedings{vega2013,
 author = {Vega-Pons, Sandro and Avesani, Paolo},
 title = {Brain Decoding via Graph Kernels},
 booktitle = {Proceedings of the 2013 International Workshop on Pattern Recognition in Neuroimaging},
 series = {PRNI '13},
 year = {2013},
 pages = {136--139},
 numpages = {4},
 publisher = {IEEE Computer Society}
}

@InProceedings{gkirtzou2013,
  author    = {Gkirtzou, Katerina and Honorio, Jean and Samaras, Dimitris and Goldstein, Rita and Blaschko, Matthew},
  title     = {fMRI Analysis with Sparse Weisfeiler--Lehman Graph Statistics},
  booktitle = {Machine Learning in Medical Imaging},
  year      = {2013},
  editor    = {Wu, Guorong and Zhang, Daoqiang and Shen, Dinggang and Yan, Pingkun and Suzuki, Kenji and Wang, Fei},
  pages     = {90--97},
  address   = {Cham, Switzerland},
  publisher = {Springer},
}

@Article{sonnenburg2006large,
  author  = {Sonnenburg, S{\"o}ren and R{\"a}tsch, Gunnar and Sch{\"a}fer, Christin and Sch{\"o}lkopf, Bernhard},
  title   = {Large Scale Multiple Kernel Learning},
  journal = {Journal of Machine Learning Research},
  year    = {2006},
  volume  = {7},
  pages   = {1531--1565},
}

@InProceedings{mahe2004extensions,
  author    = {Mah{\'e}, Pierre and Ueda, Nobuhisa and Akutsu, Tatsuya and Perret, Jean-Luc and Vert, Jean-Philippe},
  title     = {Extensions of Marginalized Graph Kernels},
  booktitle = {Proceedings of the 21st Conference on Machine learning},
  year      = {2004},
  pages     = {70--78},
  address   = {New York, NY, USA},
  publisher = {ACM},
}

@InProceedings{kondor2009graphlet,
  author       = {Kondor, Risi and Shervashidze, Nino and Borgwardt, Karsten M.},
  title        = {The Graphlet Spectrum},
  booktitle    = {Proceedings of the 26th Annual International Conference on Machine Learning},
  year         = {2009},
  pages        = {529--536},
  address      = {New York, NY, USA},
  organization = {ACM},
}

@InProceedings{kondor2002diffusion,
  author    = {Kondor, Risi Imre and Lafferty, John},
  title     = {Diffusion Kernels on Graphs and Other Discrete Structures},
  booktitle = {Proceedings of the 19th International Conference on Machine Learning},
  year      = {2002},
  volume    = {2002},
  pages     = {315--322},
}

@InCollection{jebara2003bhattacharyya,
  author    = {Jebara, Tony and Kondor, Risi},
  title     = {Bhattacharyya and Expected Likelihood Kernels},
  booktitle = {Learning Theory and Kernel Machines},
  publisher = {Springer},
  year      = {2003},
  editor    = {Sch{\"o}lkopf, Bernhard and Warmuth, Manfred K.},
  pages     = {57--71},
  address   = {Heidelberg, Germany},
}

@Book{koller2009probabilistic,
  title     = {Probabilistic Graphical Models: Principles and Techniques},
  publisher = {MIT press},
  year      = {2009},
  author    = {Koller, Daphne and Friedman, Nir},
  address   = {Cambridge, MA, USA},
}

@Article{Yang99,
  author  = {Yang, Yiming},
  title   = {An Evaluation of Statistical Approaches to Text Categorization},
  journal = {Information Retrieval},
  year    = {1999},
  volume  = {1},
  number  = {1},
  pages   = {69--90},
}

@Article{pearson1901liii,
  author    = {Pearson, Karl},
  title     = {LIII. On Lines and Planes of Closest Fit to Systems of Points in Space},
  journal   = {The London, Edinburgh, and Dublin Philosophical Magazine and Journal of Science},
  year      = {1901},
  volume    = {2},
  number    = {11},
  pages     = {559--572},
  publisher = {Taylor \& Francis},
}

@Article{hotelling1933analysis,
  author    = {Hotelling, Harold},
  title     = {Analysis of a Complex of Statistical Variables into Principal Components},
  journal   = {Journal of Educational Psychology},
  year      = {1933},
  volume    = {24},
  number    = {6},
  pages     = {417},
  publisher = {Warwick \& York},
}

@Article{lloyd1982least,
  author    = {Lloyd, Stuart},
  title     = {Least Squares Quantization in PCM},
  journal   = {IEEE Transactions on Information Theory},
  year      = {1982},
  volume    = {28},
  number    = {2},
  pages     = {129--137},
  publisher = {IEEE},
}

@Article{hoerl1970ridge,
  author    = {Hoerl, Arthur E. and Kennard, Robert W.},
  title     = {Ridge Regression: Biased Estimation for Nonorthogonal Problems},
  journal   = {Technometrics},
  year      = {1970},
  volume    = {12},
  number    = {1},
  pages     = {55--67},
  publisher = {Taylor \& Francis Group},
}

@InProceedings{boser1992training,
  author       = {Boser, Bernhard E and Guyon, Isabelle M and Vapnik, Vladimir N},
  title        = {A Training Algorithm for Optimal Margin Classifiers},
  booktitle    = {Proceedings of the Fifth Annual Workshop on Computational Learning Theory},
  year         = {1992},
  pages        = {144--152},
  address      = {New York, NY, USA},
  organization = {ACM},
}

@Book{shawe2004kernel,
  title     = {Kernel Methods for Pattern Analysis},
  publisher = {Cambridge University Press},
  year      = {2004},
  author    = {Shawe-Taylor, John and Cristianini, Nello},
  address   = {Cambridge, United Kingdom},
}

@Article{Wiener47,
  author    = {Wiener, Harry},
  title     = {Structural Determination of Paraffin Boiling Points},
  journal   = {Journal of the American Chemical Society},
  year      = {1947},
  volume    = {69},
  number    = {1},
  pages     = {17--20},
  issn      = {0002-7863},
  publisher = {American Chemical Society},
}

@InCollection{Koebler08,
  author    = {K{\"o}bler, Johannes and Verbitsky, Oleg},
  title     = {From Invariants to Canonization in Parallel},
  booktitle = {Computer Science -- Theory and Applications},
  publisher = {Springer},
  year      = {2008},
  editor    = {Hirsch, Edward A. and Razborov, Alexander A. and Semenov, Alexei and Slissenko, Anatol},
  pages     = {216--227},
  address   = {Heidelberg, Germany},
  isbn      = {978-3-540-79709-8},
}

@Article{Alon98,
  author  = {Alon, Noga and Krivelevich, Michael and Sudakov, Benny},
  title   = {Finding a large hidden clique in a random graph},
  journal = {Random Structures {\&} Algorithms},
  year    = {1998},
  volume  = {13},
  number  = {3--4},
  pages   = {457--466},
}

@InCollection{SmoKon03,
  author    = {A. J. Smola and I.R. Kondor},
  title     = {Kernels and Regularization on Graphs},
  booktitle = {Learning Theory and Kernel Machines},
  publisher = {Springer},
  year      = {2003},
  editor    = {B. Sch\"olkopf and M. K. Warmuth},
  pages     = {144--158},
  address   = {Heidelberg, Germany},
}

@Article{lanckriet2004learning,
  author  = {Lanckriet, Gert R. G. and Cristianini, Nello and Bartlett, Peter and Ghaoui, Laurent El and Jordan, Michael I.},
  title   = {Learning the Kernel Matrix with Semidefinite Programming},
  journal = {Journal of Machine Learning Research},
  year    = {2004},
  volume  = {5},
  pages   = {27--72},
}

@Article{kriege2012subgraph,
  author  = {Kriege, Nils and Mutzel, Petra},
  journal = {Proceedings of the 29th International Conference on Machine Learning},
  title   = {Subgraph matching kernels for attributed graphs},
  year    = {2012},
}

@Article{levi1973,
  author    = {Levi, Giorgio},
  journal   = {Calcolo},
  title     = {A note on the derivation of maximal common subgraphs of two directed or undirected graphs},
  year      = {1973},
  number    = {4},
  pages     = {341},
  volume    = {9},
  publisher = {Springer},
}

@inproceedings{Horvath2004cyclic,
    author = {Horv\'{a}th, Tam\'{a}s and G\"{a}rtner, Thomas and Wrobel, Stefan},
    title = {Cyclic Pattern Kernels for Predictive Graph Mining},
    year = {2004},
    publisher = {Association for Computing Machinery},
    address = {New York, NY, USA},
    booktitle = {Proceedings of the Tenth ACM SIGKDD International Conference on Knowledge Discovery and Data Mining},
    pages = {158–167},
}

@inproceedings{ramon2003subtree,
    title={Expressivity versus efficiency of graph kernels},
    author={Ramon, Jan and G{\"a}rtner, Thomas},
    booktitle={Proceedings of the 1st International Workshop on Mining
               Graphs, Trees and Sequences},
    pages={65--74},
    year={2003}
}

@inproceedings{costa2010fast,
  title={Fast neighborhood subgraph pairwise distance kernel},
  author={Costa, Fabrizio and De Grave, Kurt},
  booktitle={Proceedings of the 26th International Conference on Machine Learning},
  pages={255--262},
  year={2010},
}

@inproceedings{nikolentzos2018kcore,
  title     = {A Degeneracy Framework for Graph Similarity},
  author    = {Giannis Nikolentzos and Polykarpos Meladianos and Stratis Limnios and Michalis Vazirgiannis},
  booktitle = {Proceedings of the Twenty-Seventh International Joint Conference on
               Artificial Intelligence, {IJCAI-18}},
  publisher = {International Joint Conferences on Artificial Intelligence Organization},             
  pages     = {2595--2601},
  year      = {2018},
  month     = {7},
}

@article{grakel2018,
  author  = {Giannis Siglidis and Giannis Nikolentzos and Stratis Limnios and Christos Giatsidis and Konstantinos Skianis and Michalis Vazirgiannis},
  title   = {GraKeL: A Graph Kernel Library in Python},
  journal = {Journal of Machine Learning Research},
  year    = {2020},
  volume  = {21},
  number  = {54},
  pages   = {1--5}
}

@article{mahe2009graph,
  title={Graph kernels based on tree patterns for molecules},
  author={Mah{\'e}, Pierre and Vert, Jean-Philippe},
  journal={Machine learning},
  volume={75},
  number={1},
  pages={3--35},
  year={2009},
  publisher={Springer}
}

@inproceedings{dai2016discriminative,
  title={Discriminative embeddings of latent variable models for structured data},
  author={Dai, Hanjun and Dai, Bo and Song, Le},
  booktitle={International Conference on Machine Learning},
  pages={2702--2711},
  year={2016}
}

@article{Kriege2020GKsurvey,
	Author = {Kriege, Nils M. and Johansson, Fredrik D. and Morris, Christopher},
	Journal = {Applied Network Science},
	Number = {1},
	Pages = {6},
	Title = {A survey on graph kernels},
	Volume = {5},
	Year = {2020},
}

@InCollection{lee2012graph,
  author    = {Lee, John A and Verleysen, Michel},
  booktitle = {Image Processing and Analysis with Graphs: Theory and Practice},
  title     = {Graph-based dimensionality reduction},
  year      = {2012},
  pages     = {351--382},
}

@incollection{Aggarwal2010,
author = {Aggarwal, Charu and Wang, Haixun},
year = {2010},
month = {02},
pages = {275--301},
title = {A Survey of Clustering Algorithms for Graph Data},
booktitle = {Managing and Mining Graph Data},
volume = {40},
publisher = {Springer},
}

@misc{nikolentzos2019graph,
      title={Graph Kernels: A Survey}, 
      author={Nikolentzos, Giannis and Siglidis, Giannis and
              Vazirgiannis, Michalis},
      year={2019},
      eprint={1904.12218},
      archivePrefix={arXiv},
      primaryClass={stat.ML}
}

@inproceedings{Green06,
author = {Greene, Derek and Cunningham, P\'{a}draig},
title = {Practical Solutions to the Problem of Diagonal Dominance in Kernel Document Clustering},
year = {2006},
publisher = {Association for Computing Machinery},
address = {New York, NY, USA},
doi = {10.1145/1143844.1143892},
booktitle = {Proceedings of the 23rd International Conference on Machine Learning~(ICML)},
pages = {377–-384},
}

@misc{hu2020ogb,
      title={Open Graph Benchmark: Datasets for Machine Learning on Graphs}, 
      author={Weihua Hu and Matthias Fey and Marinka Zitnik and Yuxiao
              Dong and Hongyu Ren and Bowen Liu and Michele Catasta and Jure Leskovec},
      year={2020},
      eprint={2005.00687},
      archivePrefix={arXiv},
}

@inproceedings{Morris2020,
    title={TUDataset: A collection of benchmark datasets for learning with graphs},
    author={Christopher Morris and Nils M. Kriege and Franka Bause and Kristian Kersting and Petra Mutzel and Marion Neumann},
    booktitle={ICML 2020 Workshop on Graph Representation Learning and Beyond (GRL+ 2020)},
    archivePrefix={arXiv},
    eprint={2007.08663},
    url={www.graphlearning.io},
    year={2020}
}

@article{dwivedi2020benchmarkgnns,
  title={Benchmarking Graph Neural Networks},
  author={Dwivedi, Vijay Prakash and Joshi, Chaitanya K and Laurent, Thomas and Bengio, Yoshua and Bresson, Xavier},
  archivPrefix={arXiv},
  eprint={2003.00982},
  year={2020}
}

@incollection{Hsieh14,
title = {Fast Prediction for Large-Scale Kernel Machines},
author = {Hsieh, Cho-Jui and Si, Si and Dhillon, Inderjit S},
booktitle = {Advances in Neural Information Processing Systems 27},
editor = {Z. Ghahramani and M. Welling and C. Cortes and N. D. Lawrence and K. Q. Weinberger},
pages = {3689--3697},
year = {2014},
publisher = {Curran Associates, Inc.},
}

@InCollection{Hofer17,
  author    = {Hofer, Christoph and Kwitt, Roland and Niethammer, Marc and Uhl, Andreas},
  title     = {Deep Learning with Topological Signatures},
  booktitle = {Advances in Neural Information Processing Systems~30~(NeurIPS)},
  year      = {2017},
  editor    = {I. Guyon and U. V. Luxburg and S. Bengio and H. Wallach and R. Fergus and S. Vishwanathan and R. Garnett},
  publisher = {Curran Associates, Inc.},
  pages     = {1634--1644},
  address   = {Red Hook, NY, USA},
}

@InProceedings{Hofer20,
  author        = {Hofer, Christoph D. and Graf, Florian and Rieck, Bastian and Niethammer, Marc and Kwitt, Roland},
  booktitle     = {Proceedings of the 37th International Conference on Machine Learning~(ICML)},
  title         = {Graph Filtration Learning},
  eprint        = {1905.10996},
  pubstate      = {inpress},
  series        = {Proceedings of Machine Learning Research},
  archiveprefix = {arXiv},
  author+an     = {3=highlight},
  primaryclass  = {cs.LG},
  year          = {2020},
}

@InCollection{Zhao19,
  author        = {Zhao, Qi and Wang, Yusu},
  title         = {Learning metrics for persistence-based summaries and applications for graph classification},
  year          = {2019},
  booktitle     = {Advances in Neural Information Processing Systems~32~(NeurIPS)},
  publisher     = {Curran Associates, Inc.},
  editor        = {Wallach, H. and Larochelle, H. and Beygelzimer, A. and d'Alché-Buc, F. and Fox, E. and Garnett, R.},
  pages         = {9855--9866},
}

\end{document}